\documentclass{article}
\PassOptionsToPackage{table}{xcolor}
\PassOptionsToPackage{sort&compress}{natbib}
\usepackage{iclr2027_conference,times}
\setcitestyle{numbers,square,comma}

\usepackage{amsmath,amssymb,amsfonts,bm,mathtools,amsthm}
\IfFileExists{bbm.sty}{\usepackage{bbm}}{}

\usepackage[colorlinks=true, citecolor=blue, linkcolor=blue, urlcolor=blue]{hyperref}
\usepackage{url}
\usepackage{microtype}
\usepackage{graphicx}
\usepackage{subcaption}
\usepackage{booktabs}
\usepackage{array,tabularx,longtable}
\usepackage{multirow}
\usepackage{enumitem}
\usepackage[table]{xcolor}
\usepackage{float}
\usepackage{mdframed}
\usepackage[capitalize,noabbrev]{cleveref}

\floatstyle{ruled}
\newfloat{algorithm}{tb}{loa}
\floatname{algorithm}{Algorithm}
\crefname{algorithm}{Algorithm}{Algorithms}

\theoremstyle{definition}
\newtheorem{definition}{Definition}[section]

\theoremstyle{plain}
\newtheorem{theorem}[definition]{Theorem}
\newtheorem{proposition}[definition]{Proposition}

\mdfdefinestyle{statementbox}{
  linewidth=0.8pt,
  linecolor=gray!45,
  backgroundcolor=gray!6,
  roundcorner=3pt,
  innertopmargin=6pt,
  innerbottommargin=6pt,
  innerleftmargin=8pt,
  innerrightmargin=8pt,
  skipabove=8pt,
  skipbelow=8pt
}
\surroundwithmdframed[style=statementbox]{definition}
\surroundwithmdframed[style=statementbox]{assumption}
\surroundwithmdframed[style=statementbox]{proposition}
\surroundwithmdframed[style=statementbox]{remark}
\surroundwithmdframed[style=statementbox]{lemma}
\surroundwithmdframed[style=statementbox]{theorem}
\surroundwithmdframed[style=statementbox]{corollary}

\def\rvz{{\mathbf{z}}}

\newcommand{\method}{\textsc{POMA}}
\newcommand{\umethod}{\textsc{LogiCA-POMA}}
\newsavebox{\pomatablebox}
\newcommand{\pomatabbox}[1]{{\small\setlength{\tabcolsep}{3.8pt}\sbox{\pomatablebox}{#1}\ifdim\wd\pomatablebox>\linewidth\resizebox{\linewidth}{!}{\usebox{\pomatablebox}}\else\usebox{\pomatablebox}\fi}}

\newcounter{suppsummary}

\crefname{suppsummary}{summary}{summaries}
\Crefname{suppsummary}{Summary}{Summaries}
\title{Probabilistic Output-Space Multimodal Alignment for Biological Foundation Models}
\author{Aurelien Pelissier \\
\texttt{aurelien.pelissier.38@gmail.com}}
\iclrfinalcopy
\begin{document}
\maketitle
\lhead{Preprint}
\begin{abstract}
Masked language modeling (MLM) has transformed biological sequence
modeling by learning conditional token distributions. However,
continuous biological measurements are often modeled through
pointwise regression or numerical tokenization, limiting structured
probabilistic inference. We introduce \textbf{Probabilistic
Output-Space Multimodal Alignment} (\method{}), a framework that
unifies categorical MLM and mean squared error as special cases
of a generalized masked likelihood objective over native observation
spaces. By retaining predictive distributions, \method{} connects
heterogeneous pretrained foundation models through unit-invariant,
reference-centered likelihood ratios. The statistical structure of
these distributions naturally enables correlation-aware scoring,
analytical conditioning, and Bayesian evidence propagation across
biological candidates, supporting concurrent prediction,
experimental acquisition, and property-conditioned generation.
Extensive evaluations across molecules, spectroscopy, protein
engineering, antibody design, single-cell multi-omics, and regulatory
genomics demonstrate the versatility of this framework.
\method{} thus establishes native predictive distributions as a
reusable interface for \textbf{multimodal probabilistic reasoning}
and uncertainty-aware biological discovery.
\end{abstract}

\vspace{-0.4cm}

\begin{figure}[H]
\centering
\refstepcounter{figure}\label{fig:poma-architecture}
\includegraphics[width=\linewidth,height=2.9in,keepaspectratio]{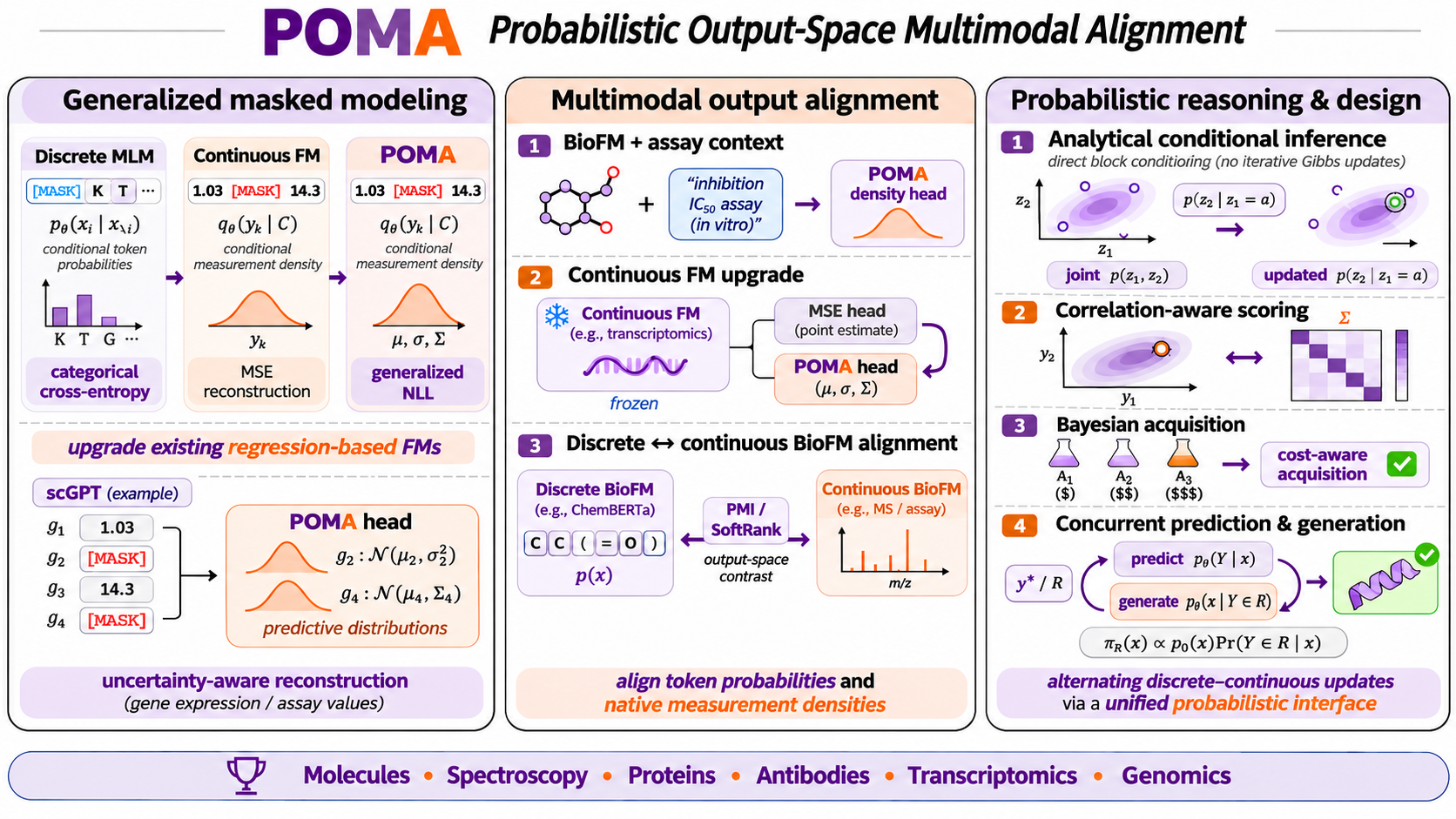}
\end{figure}

\section{Introduction}
\label{sec:intro}

\vspace{-0.1cm}

Biological foundation models have transformed sequence modeling,
providing a general-purpose interface for prediction and
design~\citep{lin2023evolutionary,dalla2025nucleotide,madani2023large,brixi2026genome}.
Central to many of these advances is masked language modeling
(MLM), which learns conditional token distributions through
categorical cross-entropy~\citep{devlin2019bert}.
The probabilistic information encoded in token logits extends
their utility beyond reconstruction, enabling mutation scoring,
functional dependency analysis, candidate ranking, and sequence
generation~\citep{meier2021language,hie2024efficient,tomaz2024nucleotide}.
Yet biological information extends beyond discrete sequences:
gene expression, molecular spectra, binding affinities, and
biophysical properties are quantitative observations characterized
by uncertainty and statistical dependencies. Developing an analogous
probabilistic output interface for continuous measurements could
extend these capabilities across heterogeneous biological modalities.

Existing approaches to quantitative masked modeling commonly
rely on pointwise regression, value discretization, or
expression-ranked tokens~\citep{cui2024scgpt,hao2024large,yang2022scbert,theodoris2023transfer,bushuiev2025self}.
Mean squared error (MSE) corresponds to a fixed-variance Gaussian
negative log-likelihood, targeting the conditional mean without
explicitly learning input-dependent uncertainty or residual
dependence. This creates a fundamental asymmetry between categorical
decoders, which expose predictive distributions, and conventional
regression decoders, which expose point estimates.
Regression Transformer enables concurrent prediction and generation
through numerical tokenization~\citep{born2023regression}, while
regression-aware token objectives improve numerical
accuracy~\citep{zausinger2024regress}.
Neither formulation, however, directly parameterizes structured
joint densities in native measurement spaces.

Categorical cross-entropy and MSE can instead be understood as
special cases of likelihood-based reconstruction.
Probabilistic reconstruction has established foundations in
distributional regression~\citep{bishop1994mixture,kendall2017uncertainties,seitzer2022pitfalls,stirn2023faithful},
neural processes~\citep{bruinsma2021gaussian}, biological models
such as scVI, totalVI and
scPRINT~\citep{lopez2018deep,gayoso2021joint,kalfon2025scprint},
and probabilistic forecasting models such as
Moirai~\citep{woo2024unified}.
Complementary approaches combine categorical language modeling
with continuous diffusion~\citep{zhou2025transfusion}.
Beyond uncertainty quantification, native predictive distributions
provide a statistical interface for comparing observations,
incorporating evidence, and reasoning about dependencies.
Their integration into a common framework for masked modeling,
multimodal alignment, and biological discovery remains
comparatively underexplored.

Multimodal alignment typically operates in shared representation
spaces~\citep{radford2021clip,kalia2025jestr,chen2026flare,xu2026scclip},
including probabilistic embeddings that capture ambiguity in
cross-modal correspondences~\citep{chun2021probabilistic,chun2024improved}.
While effective for retrieval, these representations do not
directly specify predictive distributions over native observations.
Complementary approaches combine latent contrastive and generative
objectives~\citep{yu2022coca}, while
LogiCA~\citep{shao2026contextualizing} contrasts contextualized
token likelihoods while preserving native model outputs.
Extending output-space alignment to continuous modalities
introduces distinct challenges: densities depend on measurement
coordinates, quantitative observations exhibit graded similarity,
and contextual likelihood changes may reflect background shifts
rather than pair-specific information. Related challenges motivate
domain-conditional likelihood scoring, conditional contrastive
learning, and shortcut-aware multimodal
alignment~\citep{holtzman2021surface,tsai2022conditional,bleeker2024demonstrating}.
Addressing these challenges requires compatibility measures that
preserve the statistical meaning of heterogeneous outputs.

The value of native predictive distributions extends beyond
alignment. In biological discovery, observations are acquired
progressively, measurements have unequal experimental costs,
and desired properties may exhibit competing dependencies.
Partial spectra can inform missing spectral regions, while
inexpensive screening assays can constrain predictions of costly
antibody properties. Accurate marginal predictions alone are
insufficient to reason about joint feasibility or determine how
a new observation affects competing candidates. Structured
predictive distributions enable analytical conditioning, while
shared parameter uncertainty allows experimental evidence to
propagate across candidates. Although established in conditional
density estimation and Bayesian experimental
design~\citep{ivanov2019variational,strauss2021arbitrary,daxberger2021laplace,daulton2021parallel},
these principles remain poorly integrated with the learning and
alignment objectives of biological foundation models.

\vspace{-0.2cm}

\paragraph{Our contribution.}
We introduce \textbf{Probabilistic Output-Space Multimodal
Alignment} (\method{}), establishing native predictive
distributions as a common interface for learning, connecting,
and reasoning with biological foundation models.
Through \textbf{generalized masked modeling}, categorical
cross-entropy and MSE are recovered as special cases of a
likelihood-based objective that extends reconstruction to
structured measurement distributions. These distributions
naturally enable \textbf{multimodal output-space alignment}
through unit-invariant, reference-centered likelihood ratios,
with contrastive supervision that accounts for quantitative
measurement similarity. Crucially, preserving the predictive
law enables capabilities beyond reconstruction and retrieval.
Measurement dependencies and shared Bayesian uncertainty
provide the basis for correlation-aware scoring, analytical
conditioning without repeated neural evaluations, and
decision-aware experimental acquisition.
The same probabilistic interface supports concurrent prediction
and property-conditioned generation, allowing biological design
to account for joint feasibility rather than individual property
estimates alone. We establish the mathematical foundations of
these operations and evaluate them across diverse biological
systems, demonstrating how \textbf{probabilistic reasoning and
design} emerge from the statistical structure of native model
outputs.

\section{Probabilistic Output-Space Alignment}
\label{sec:poma}

\method{} connects learning, alignment and probabilistic reasoning through native predictive distributions (\cref{fig:poma-architecture}). These operations share a statistical interface rather than a universal architecture.

\subsection{Generalized probabilistic masked modeling}
\label{sec:poma-output-distributions}\label{sec:poma-mlm-loss}

Let $V=(X,Y)$ comprise discrete coordinates $X$ (including counts) and quantitative measurements $Y$. For a nonempty masked set $M=M_D\cup M_C$, the generalized masked likelihood is
\begin{equation}
 \mathcal L_{\rm MLM}=-\mathbb E_{V,M}\!\left[\frac{\sum_{i\in M_D}\log p_\theta(x_i\mid C_i)+\log q_\theta(y_{M_C}\mid C)}{|M|}\right],
 \label{eq:main-mixed-mlm}
\end{equation}
where $C$ contains visible observations, partner modalities and, where applicable, assay descriptions. Evaluated coordinates are withheld before encoding; $C_i=C$ for common masking and is position-specific for leave-one-out scoring. Empty blocks contribute zero. Here $p_\theta$ denotes token probabilities or densities in reconstruction coordinates, and $q_\theta$ the native measurement law. The continuous block scores a normalized joint observation law, whereas the composite token score need not define a joint sequence likelihood.

\paragraph{Distributional reconstruction.} Categorical cross-entropy and fixed-variance Gaussian regression are special cases of this objective. A heteroscedastic Gaussian readout predicts $\mu(C)$ and $\sigma^2(C)>0$, replacing squared error by the negative log-likelihood $\frac12[(z-\mu)^2/\sigma^2+\log\sigma^2+\log(2\pi)]$. Thus reconstruction learns predictive precision as well as location, even with a frozen backbone. For train-fitted invertible transformations $z_k=T_k(y_k)$ with nonzero derivative, a normalized density on the transformed support induces $q_\theta(y_S\mid C)=p_\theta(T_S(y_S)\mid C)\prod_{k\in S}|T'_k(y_k)|$. The observation law must respect the measurement support; missing coordinates are marginalized under the specified observation assumptions, and censored values contribute event probabilities. Alternative heavy-tailed, mixture, bounded and count laws are developed in \cref{app:poma-mixed-decomposition,app:poma-extended-heads}. While likelihood learning retains the native statistical meaning of observations, absolute densities depend on measurement coordinates.

\subsection{Multimodal alignment in native output spaces}
\label{sec:poma-unified-score}

An observation can have high likelihood even without its partner modality. We therefore measure the improvement relative to a reference population. Let $\ell_X,\ell_Y$ denote the per-coordinate conditional log-scores from \cref{eq:main-mixed-mlm} on fixed nonempty blocks $A_X,A_Y$. For a partner reference law $Q_m$ specified by metadata $m$, define
\begin{align}
 r_{\theta,Y}(y_{A_Y}\mid m,C)&=\int q_\theta(y_{A_Y}\mid x',C)\,dQ_m(x'),\nonumber\\
 \ell_Y^{\rm ref}(y\mid m,C)&=|A_Y|^{-1}\log r_{\theta,Y}(y_{A_Y}\mid m,C).
 \label{eq:main-reference-law}
\end{align}
The reverse reference is analogous, mixing probabilities coordinate-wise for composite token scores. Holding the evaluated block and other context fixed, we define
\begin{align}
 \Delta\ell_X&=\ell_X(x\mid y,C)-\ell_X^{\rm ref}(x\mid m,C),\nonumber\\
 \Delta\ell_Y&=\ell_Y(y\mid x,C)-\ell_Y^{\rm ref}(y\mid m,C),
 \label{eq:main-pmi}\\
 s_\alpha(x,y)&=\alpha\Delta\ell_X+(1-\alpha)\Delta\ell_Y,\quad \alpha\in[0,1].
 \label{eq:main-bidir}
\end{align}
An absent-partner reference measures context gain; population or metadata-matched references additionally control background attributes such as molecular formula. A domain-matched null embedding is a distinct reference prediction, not necessarily a mixture. References do not use unknown test pairings (\cref{app:matched-reference}).

\paragraph{Coordinate invariance and reliability.} Under a common smooth invertible coordinate change, numerator and reference densities acquire the same Jacobian; their positive likelihood ratio is therefore invariant. The unnormalized log-ratio is conditional PMI for compatible full conditional and marginal laws, but composite scores or arbitrary references need not have this interpretation. A common target and reference preserve the conditional-likelihood ranking. These properties do not guarantee pairing specificity (Propositions~\ref{prop:poma-alpha-pmi} and~\ref{prop:poma-likelihood-ratio}). Directional weights can reflect unequal modality information; the covariance-aware minimum-variance rule and repeated-mask reliability heuristic are detailed in \cref{app:supp-e7-adaptive,eq:app-cov-aware-alpha}.

\paragraph{Contextualization and contrastive learning.} Following LogiCA~\citep{shao2026contextualizing}, gated cross-attention contextualizes native readouts, e.g., $a_{X\mid Y}=a_X+\tanh(3\gamma_X)g_X(H_X,H_Y)$, with a zero-initialized gate and an analogous reverse adapter. Categorical heads are retained; quantitative heads parameterize measurement laws, including newly fitted readouts on pretrained features. Assay descriptions provide additional context through a language-conditioned readout (\cref{app:full-method,app:two-biofm-alignment}).

Unlike one-hot contrastive supervision, graded targets can respect the similarity of measured quantities. For paired examples let $s_{bj}=s_\alpha(x_b,y_j)$ and $w_{bj}\propto\exp(-d_{bj}/\tau_{\rm soft})$, normalized over $j$, with $d_{bj}$ a measurement-space distance. We train with
\begin{equation}
 \begin{aligned}
 \mathcal L_{\rm rank}&=-\frac1{N_b}\sum_{b,j}w_{bj}\log\frac{\exp(s_{bj}/\tau)}{\sum_c\exp(s_{bc}/\tau)},\\
 \mathcal L&=\mathcal L_{\rm MLM}+\lambda_{\rm rank}\mathcal L_{\rm rank}.
 \end{aligned}
 \label{eq:main-ranking}
\end{equation}
One-hot targets recover InfoNCE~\citep{oord2018cpc}; symmetric supervision averages both directions. The soft-target geometry is coordinate-dependent, unlike the ratio itself, and the experiments use fixed similarity scales (\cref{app:contrast-geometry,app:reproducibility}). The contrastive term need not preserve proper scoring of the measurement law. Masked likelihood alone or latent contrast can also learn densities for inference-time output scoring, enabling controlled comparisons of reconstruction and alignment.

\subsection{Structured probabilistic inference and biological design}
\label{sec:main-coupling}\label{sec:logica-poma}

Prediction summarizes conditional output distributions; generation samples from them. Their dependence structure enables evidence propagation within and across candidates. A structured Gaussian readout $z=\mu+U\eta+\epsilon$, with independent $\eta\sim\mathcal N(0,I_r)$ and $\epsilon\sim\mathcal N(0,D)$ ($D$ positive diagonal), represents measurement covariance $\Sigma=D+UU^\top$. Observing a subset $z_B$ updates the latent factor through $C_B=I_r+U_B^\top D_B^{-1}U_B$ and $m_B=C_B^{-1}U_B^\top D_B^{-1}(z_B-\mu_B)$. The remaining measurements follow
\begin{equation}
 z_A\mid z_B,C\sim\mathcal N(\mu_A+U_A m_B,\ D_A+U_A C_B^{-1}U_A^\top).
 \label{eq:main-schur}
\end{equation}
This conditional law permits direct block prediction and sampling without coordinate-wise Gibbs iterations. Cached factor-system construction and solution cost $O(|B|r^2+r^3)$, excluding output covariance materialization. Heavy-tailed and mixture laws extend this principle (Propositions~\ref{prop:app-woodbury-schur} and~\ref{prop:app-mvt-schur}). The learned law also supports model-based dependency analysis through token-weighted counterfactual KL and precision-matrix partial correlations, quantifying sequence-to-property sensitivity and conditional measurement associations, respectively, rather than causality (\cref{app:joint-gibbs-dep-maps}).

\paragraph{Shared Bayesian uncertainty.} To propagate evidence across candidates, let $z_i=\Phi_i\mathbf w+\epsilon_i$ for fixed features $\Phi_i$ and shared posterior $\mathbf w\mid\mathcal D\sim\mathcal N(\widehat{\mathbf w},\Sigma_W)$. Assume conditionally centered residuals with covariance $D_i+U_iU_i^\top$ and zero conditional cross-covariance between distinct candidates. Then
\begin{equation}
 \operatorname{Cov}(z_i,z_j\mid\mathcal D)=\underbrace{\Phi_i\Sigma_W\Phi_j^\top}_{\rm epistemic}+\underbrace{\mathbb I_{i=j}(D_i+U_iU_i^\top)}_{\rm aleatoric}.
 \label{eq:main-cross-cov}
\end{equation}
For conditionally jointly Gaussian residuals with weight-independent covariance, integrating the posterior preserves the low-rank-plus-diagonal family. A new observation can therefore update predictions across candidates through the shared posterior while retaining analytical conditioning. Conjugate updates, spectral truncation and the scope of approximate Laplace inference are established in Theorem~\ref{thm:joint-candidate-assay-closure} and Proposition~\ref{prop:app-adaptive-mixed-tv}.

\paragraph{Joint feasibility and decision-aware acquisition.} For a target region $\mathcal R=\prod_k[a_k,b_k]$, define $r(x)=\Pr(z\in\mathcal R\mid x,\mathcal D)$. Conditional independence given a latent factor gives $r(x)=\mathbb E_{\eta\mid x,\mathcal D}[h_x(\eta)]$ with $h_x(\eta)=\prod_k\Pr(z_k\in[a_k,b_k]\mid x,\eta,\mathcal D)$. This accounts for dependencies that can reverse rankings based on products of marginal feasibility probabilities (Theorem~\ref{thm:correlation-acquisition}).

Information shared by competing candidates need not improve their relative ranking. For a finite candidate set under a coherent joint law, let $p_i$ and $p_i(Y_a)$ denote feasibility before and after observing assay $Y_a$, and $i_0\in\arg\max_i p_i$. The expected one-step improvement in candidate selection~\citep{frazier2008kg,frazier2009kg} is
\begin{equation}
 V(a)=\mathbb E_{Y_a\mid\mathcal D}[\max_i p_i(Y_a)]-p_{i_0}.
 \label{eq:main-voi}
\end{equation}
With feasibility gap $g_i=p_{i_0}-p_i$ and differential variance $s_i^2=\operatorname{Var}(p_i(Y_a)-p_{i_0}(Y_a)\mid\mathcal D)$, we establish the distribution-independent bound
\begin{equation}
 0\le V(a)\le\min\!\left\{1-p_{i_0},\ \frac12\sum_{i\ne i_0}\left(\sqrt{g_i^2+s_i^2}-g_i\right)\right\}.
 \label{eq:main-voi-bound}
\end{equation}
At fixed differential variance, each summand decreases with the feasibility gap and vanishes when differential variance is zero. Conditional replicas express the required covariance of updated feasibility probabilities via joint events, with tractable Gaussian cross-covariances (Theorem~\ref{thm:v1-decision-voi}). The bound thus supports pruning with valid lower and upper bounds on value per positive assay cost, rather than prioritizing marginal uncertainty alone. It concerns one-step, model-relative decision value, not globally optimal multistep allocation; model-relative selection-regret bounds are developed in \cref{app:decision-regret}.

\paragraph{Property-conditioned generation.} Alternating conditional measurement draws with masked-token edits yields a Gibbs-style mechanism. Since masked token conditionals need not define a compatible joint law, we specify explicit targets. For a proper positive prior $p_0$ on a finite editing space and nonzero prior-weighted feasibility, the event-conditioned sequence law and its factor augmentation are
\begin{equation}
 \begin{aligned}
 \pi_{\mathcal R}(x)&\propto p_0(x)r(x),\\
 \bar\pi_{\mathcal R}(x,\eta)&\propto p_0(x)p(\eta\mid x,\mathcal D)h_x(\eta).
 \end{aligned}
 \label{eq:main-aug-target}
\end{equation}
The augmented law retains the sequence marginal without repeated integration of $r(x)$ during proposals. For joint sequence--measurement generation, Schur draws form an exact continuous Gibbs block, complemented by Hastings-corrected token proposals. Theorem~\ref{thm:mh-joint-invariance} establishes these joint updates and exact-target event samplers; adaptive proposals and finite-budget guarantees are developed in \cref{app:supp-design,app:poma-certified-adaptive}.

\section{Experiments}
\label{sec:experiments}

We evaluate the three methodological pillars through controlled comparisons of learning objectives, cross-modal scoring and inference operations. \Cref{tab:main-unified-results,fig:global-experiments} summarize the representative results and core methodological evidence; \cref{fig:supp-biological-applications} illustrates complementary biological applications. Dataset provenance and evaluation protocols are provided in \cref{app:full-method}.

\begin{table}[!t]
\centering
\caption{\textbf{Representative experimental results across the three
methodological pillars of \method{}.}
Comparisons use matched evaluation protocols within each row.
Bold indicates the numerically better result, without implying
statistical significance. The protein-engineering results use the
capped protocol. Acquisition and design experiments are retrospective
or oracle-based, while sampling fidelity is evaluated against an
enumerated model-defined target. Full comparisons, uncertainty
estimates and experimental protocols are provided in
\cref{app:supp-masked,app:supp-alignment,app:supp-inference,%
app:supp-bayesian,app:supp-decisions,app:supp-design}.}
\label{tab:main-unified-results}
\small
\setlength{\tabcolsep}{3pt}
\renewcommand{\arraystretch}{1.08}
\pomatabbox{
\begin{tabular}{llllrr}
\toprule
\textbf{Benchmark} &
\textbf{Metric} &
\textbf{Comparator} &
\textbf{\method{} variant} &
\textbf{Ref.} &
\textbf{Ours} \\
\midrule

\multicolumn{6}{l}{\textbf{Generalized probabilistic masked modeling}}\\
\addlinespace[2pt]

BMMC / scGPT~\citep{luecken2021sandbox,cui2024scgpt}
& CRPS $\downarrow$
& Native MSE
& MVN + SoftRank
& .266 & \textbf{.235} \\

BMMC / scGPT
& Joint NLL $\downarrow$
& Post-hoc heterosc.
& Coupled NLL
& .128 & \textbf{.088} \\

AIntibody~\citep{erasmus2026aintibody}
& Precision@15 $\uparrow$
& Fixed variance
& Heterosc.\ NLL
& 13.3\% & \textbf{53.3\%} \\

TDC ADMET~\citep{huang2021therapeutics}
& Mean rank $\downarrow$
& Morgan GBDT~\citep{rogers2010ecfp}
& POMA-Rank
& 4.91 & \textbf{3.73} \\

\midrule
\multicolumn{6}{l}{\textbf{Multimodal output-space alignment}}\\
\addlinespace[2pt]

MassSpecGym~\citep{bushuiev2024massspecgym}
& Hit@1 $\uparrow$
& CLIP-style~\citep{radford2021clip}
& Adaptive bidir.
& 10.4\% & \textbf{30.3\%} \\

ChemBERTa--DreaMS~\citep{ahmad2022chemberta,bushuiev2025self}
& Hit@1 $\uparrow$
& Unmatched contrast
& Matched contrast
& 12.2\% & \textbf{36.6\%} \\

ChemBERTa--DreaMS
& Spectral NLL $\downarrow$
& Latent contrast
& Matched contrast
& .591 & \textbf{.572} \\

& Hit@1 $\uparrow$
& Latent contrast
& Matched contrast
& \textbf{35.2\%} & 26.7\% \\

CITE-seq~\citep{stoeckius2017citeseq,luecken2021sandbox}
& FOSCTTM $\downarrow$
& totalVI-style~\citep{gayoso2021joint}
& Coupled likelihood
& .0557 & \textbf{.0430} \\

\midrule
\multicolumn{6}{l}{\textbf{Probabilistic reasoning and biological design}}\\
\addlinespace[2pt]

BMMC / scGPT
& Cond.\ MAE $\downarrow$
& Neural recond.
& Schur conditioning
& .166 & \textbf{.160} \\

QM9S~\citep{zou2023qm9s}
& IR MAE $\downarrow$
& Unconditioned
& Banded+LR Schur
& .713 & \textbf{.353} \\

lentiMPRA~\citep{agarwal2025massively}
& HepG2 MAE $\downarrow$
& Unconditioned
& MVN Schur
& .711 & \textbf{.365} \\

AbRank~\citep{liu2026abrank}
& Impact AUROC $\uparrow$
& PLM entropy
& Counterfactual KL
& .512 & \textbf{.634} \\

GDPa1 (4-shot)~\citep{arsiwala2025high}
& NLL $\downarrow$
& NTK-BLL~\citep{calvoordonez2026richer}
& Laplace
& 1.606 & \textbf{1.596} \\

Protein eng.\ (27)~\citep{herrerarocha2026multitask}
& Joint NLL $\downarrow$
& Ensemble~\citep{lakshminarayanan2017simple}
& Coupled Laplace
& \textbf{1.150} & 1.254 \\

& Joint Brier $\downarrow$
& Ensemble
& Coupled Laplace
& .166 & \textbf{.159} \\

AIntibody (unseen assay)
& Spearman $\rho$ $\uparrow$
& One-hot task ID
& Qwen3 context~\citep{yang2025qwen3}
& .005 & \textbf{.184} \\

AIntibody (acquisition)
& Leads/100u $\uparrow$
& GP-KG~\citep{frazier2009kg}
& Laplace EVSI
& .546 & \textbf{.692} \\

FLAb (5 libraries)~\citep{chungyoun2024flab}
& Co-opt.\ hit $\uparrow$
& BOAT~\citep{rao2026boat}
& LogiCA-POMA~\citep{shao2026contextualizing}
& 25.1\% & \textbf{30.2\%} \\

PMO (23 tasks)~\citep{gao2022sample}
& Top-10 AUC $\uparrow$
& Graph GA~\citep{jensen2019graphga}
& GA + POMA
& .628 & \textbf{.681} \\

AbLang2~\citep{olsen2024ablang2}
& Sampling TV $\downarrow$
& Heuristic Gibbs~\citep{johnson2021generating}
& Factor-aug.\ MH
& .492 & \textbf{.307} \\

\bottomrule
\end{tabular}
}
\end{table}

\subsection{Probabilistic adaptation of pretrained foundation models}
\label{sec:main-prediction}

We first assess whether pretrained representations support useful predictive distributions without changing the backbone. On held-out BMMC sequencing Site~4~\citep{luecken2021sandbox}, adapting frozen scGPT~\citep{cui2024scgpt} reduces masked-expression CRPS by $11.7\%$ and improves empirical $95\%$ coverage from $88.9\%$ to $96.2\%$, although the original point decoder retains better MAE (\cref{fig:global-experiments}A). To distinguish distributional learning from post-hoc calibration, we compare heads retaining the pretrained mean with jointly learned mean and uncertainty. Post-hoc heteroscedastic modeling recovers much of the marginal calibration improvement, but the jointly learned coupled head achieves better joint NLL and conditional prediction than post-hoc covariance estimation. Additional contrast does not improve reconstruction in this setting (\cref{tab:main-objective,tab:app-e4-scgpt-calib,tab:app-biofm-scgpt-ladder}).

The distinction between point accuracy and distributional quality extends to other biological systems, particularly where predictions guide candidate selection. On AIntibody~\citep{erasmus2026aintibody}, heteroscedastic learning improves co-optimal top-15 precision from $13.3\%$ to $53.3\%$, despite the fixed-variance reference retaining the best joint NLL. Across TDC ADMET~\citep{huang2021therapeutics}, ranking-augmented densities outperform several neural and numerical-token baselines, while the advantage over fingerprint-based gradient boosting is not statistically established (\cref{tab:app-mse-vs-nll,app:admet-results}). Thus the generalized objective can improve uncertainty-aware decisions without uniformly improving point predictions.

\subsection{Multimodal alignment of heterogeneous foundation models}
\label{sec:main-alignment}

\begin{figure}[!t]
\centering
\makebox[\linewidth][c]{%
  \begin{minipage}{1.3\linewidth}
    \centering
    \includegraphics[width=\linewidth,keepaspectratio]{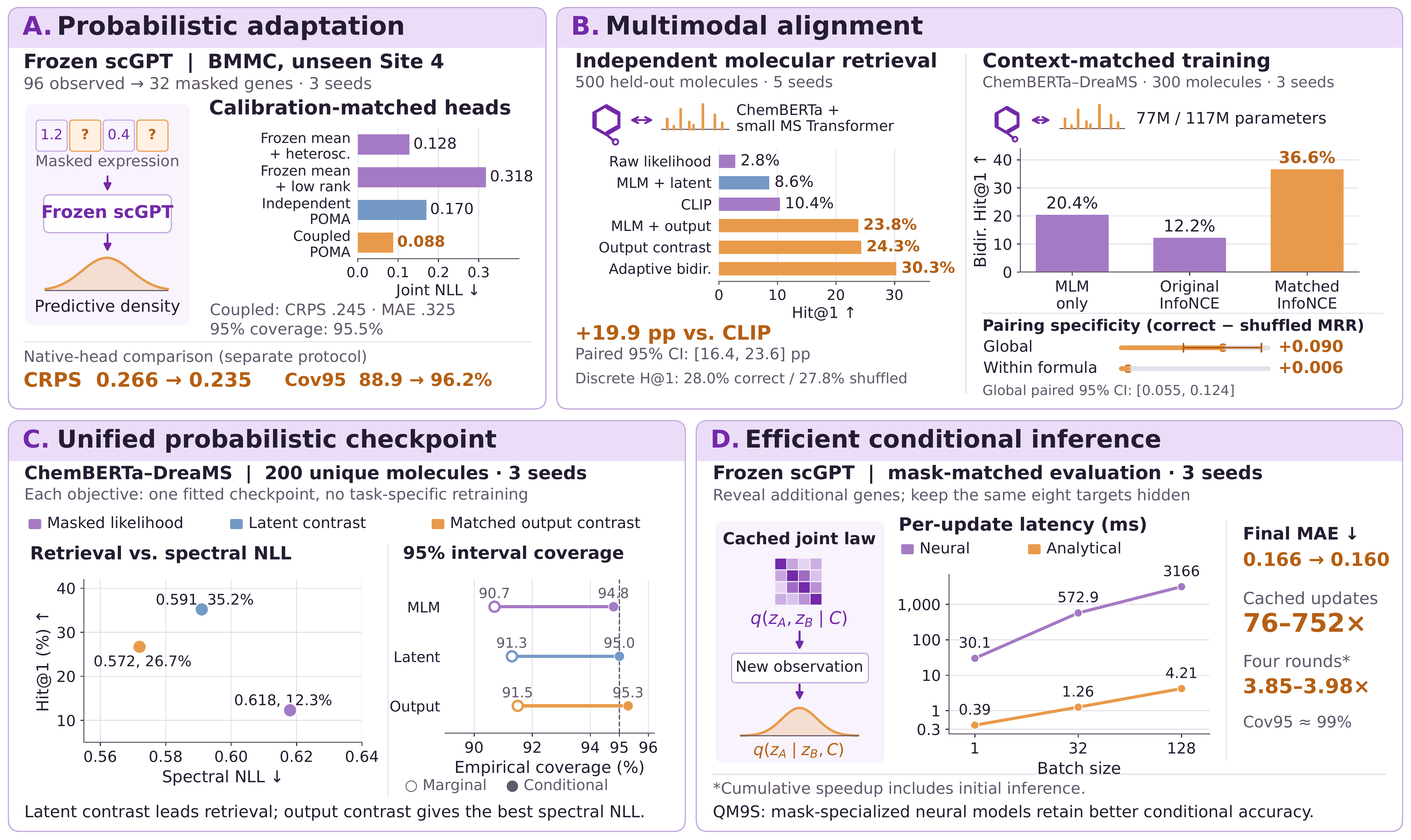}
    \caption{\textbf{Core experimental validation of \method{}.}
    \textbf{(A)} Probabilistic adaptation of frozen scGPT on held-out BMMC Site~4: calibration-matched heads (three seeds) and a native-decoder comparison (bottom), evaluated under separate protocols.
    \textbf{(B)} Independent MassSpecGym retrieval using ChemBERTa and the in-pipeline pretrained spectral Transformer (500 unique molecules, five seeds), and context-matched training using ChemBERTa--DreaMS (300 previously unused unique molecules, three seeds). The correct-minus-shuffled MRR differences distinguish global pairing sensitivity from within-formula discrimination; paired bootstrap intervals are shown where reported.
    \textbf{(C)} Unified ChemBERTa--DreaMS checkpoint evaluation (200 unique molecules, three seeds): retrieval, spectral NLL and coverage from the same checkpoint for each training objective, without task-specific retraining. Open and filled markers denote marginal and conditional coverage, respectively.
    \textbf{(D)} Mask-matched scGPT conditioning (three seeds): cached analytical updates versus neural reconditioning. Cumulative speedups over four rounds include initial inference; mask-specialized neural predictors remain more accurate on QM9S. Task illustrations are schematic.}
    \label{fig:global-experiments}
  \end{minipage}%
}
\end{figure}

\paragraph{Native output-space compatibility.} Molecular identification from tandem mass spectra requires connecting a chemical structure to its fragmentation pattern. We contextualize ChemBERTa~\citep{ahmad2022chemberta} and an independently pretrained masked-peak spectral Transformer, retaining their respective reconstruction tasks. On 500 held-out unique MassSpecGym molecules~\citep{bushuiev2024massspecgym}, adaptive bidirectional likelihood scoring achieves $30.3\%$ Hit@1 versus $10.4\%$ for CLIP-style alignment (paired improvement $19.9$ points, $95\%$ CI $[16.4,23.6]$). Reference-centered scoring also outperforms raw likelihood contrast, highlighting the importance of background observation plausibility. In a separate controlled comparison, replacing latent similarity with bidirectional likelihood scoring also improves retrieval on the same latent-trained checkpoint without retraining. These results establish predictive likelihoods as an alternative alignment interface, independently of whether the model is trained with output-space contrast (\cref{fig:global-experiments}B; \cref{tab:app-e10-definitive,tab:app-e1-e3}).

\paragraph{Reference design and pairing specificity.} High retrieval accuracy need not imply correct pairing~\citep{bleeker2024demonstrating}: shuffling the spectral context leaves discrete-direction Hit@1 nearly unchanged ($28.0\%$ versus $27.8\%$), consistent with a modality-induced prior shift.

We address this issue with context-matched references using ChemBERTa and DreaMS, the latter pretrained on 24 million spectra~\citep{bushuiev2025self}. Whereas ChemBERTa retains its token head, DreaMS supplies pretrained features for a newly fitted density readout. Incorporating matched references into contrastive training, rather than merely changing them at inference, improves bidirectional Hit@1 from $12.2\%$ to $36.6\%$ on a separate held-out slice. The correct-versus-globally-shuffled MRR difference reaches $0.090$ (paired $95\%$ CI $[0.055,0.124]$), although within-formula gain remains only $0.006$. Reference design therefore improves global pairing specificity without resolving fine-grained isomer discrimination (\cref{tab:app-e10-falsification,tab:app-e12-shortcut}). Target-coordinate withholding audits distinguish contextual prediction from direct target leakage (\cref{tab:app-e13-leakage}).

\paragraph{Limited and asymmetric observations.} Masked likelihood with discrete output scoring attains the strongest integrated retrieval across nested paired-data subsets under matched-step and matched-epoch budgets in the context-free protocol. Excessive contrast can compromise reconstruction. Reliability-aware fusion improves MRR from $0.235$ to $0.307$ under $50\%$ peak dropout; in scGPT--ESM-2 cellular contextualization~\citep{lin2023evolutionary}, the forward RNA-to-protein direction remains preferable. 

These effects also depend on the observation law and the candidate pool. Hurdle laws better accommodate sparse spectral observations, while contrastive benefits vary across likelihood families. On held-out CITE-seq cells~\citep{luecken2021sandbox,gayoso2021joint,xu2026scclip}, feature-based coupled likelihood scoring reduces FOSCTTM from $0.0557$ to $0.0430$ relative to a totalVI-style conditional VAE (\cref{fig:supp-biological-applications}C). Conversely, latent alignment is preferable for open-scaffold vibrational-spectral retrieval. These results motivate retaining predictive distributions as an additional interface rather than universally replacing latent similarity (\cref{app:supp-alignment}).

\subsection{Probabilistic reasoning with learned output distributions}
\label{sec:main-conditional}

\paragraph{A unified probabilistic checkpoint.} We test whether retrieval, reconstruction and conditional inference can coexist within the same learned distribution. For each ChemBERTa--DreaMS training objective, all operations use the same checkpoint without task-specific retraining (\cref{fig:global-experiments}C). Context-matched output contrast improves Hit@1 from $12.3\%$ to $26.7\%$ and gives the lowest spectral NLL, whereas latent contrast achieves the strongest retrieval ($35.2\%$ Hit@1) and pairing specificity. Both retain the joint law needed to incorporate measurements. Revealing spectral bands narrows predictive intervals and brings coverage toward its nominal level, although conditional NLL need not improve on sparse spectra (\cref{tab:main-unified-checkpoint,tab:app-e15-unified}). Thus the learned distribution supports multiple operations irrespective of the chosen alignment objective.

\paragraph{Incorporating experimental evidence.} On frozen scGPT, cached analytical conditioning achieves comparable or slightly better reconstruction than neural reconditioning under matched masking protocols, despite conservative intervals. Individual updates are $76$--$752\times$ faster across evaluated batch sizes, corresponding to an approximately fourfold cumulative speedup over four observation rounds including initial inference (\cref{fig:global-experiments}D). Sequential conditioning agrees numerically with joint conditioning, but mask-specialized neural predictors remain more accurate on QM9S~\citep{zou2023qm9s}. Analytical updates are computationally advantageous when the fitted dependence structure is reliable (\cref{app:supp-e14-mask-matched,tab:app-conditioning-latency}).

The same operation addresses incomplete biological measurements. Conditioning on observed IR bands reduces reconstruction MAE by $50.5\%$ in a simulated water-window occlusion task; conditioning on measured K562 activity in lentiMPRA~\citep{agarwal2025massively} reduces HepG2 prediction MAE by $48.7\%$ on the conditional evaluation subset (\cref{fig:supp-biological-applications}A,E). Gains extend to partially measured cellular and proteomic panels, but not to the smaller paired IR-to-NMR/MS dataset. Heavy-tailed conditioning improves likelihood for outlying GDPa1 antibody measurements~\citep{arsiwala2025high}, illustrating the importance of the observation law (\cref{app:spectroscopy,app:multiomics,tab:app-mvt-softrank}).

\paragraph{Biological dependencies.} Distributional sensitivity extends the dependency analysis of categorical biological language models~\citep{tomaz2024nucleotide,zhang2024interactingmotifs}. On AbRank~\citep{liu2026abrank}, counterfactual KL improves discrimination between large- and small-effect mutations (AUROC $0.634$ on 520 mutation pairs); in AIntibody, sequence-conditioned partial correlations reveal different affinity--hydrophobicity associations in aromatic-rich and polar paratopes. These are model-based rather than causal dependencies (\cref{app:joint-gibbs-dep-maps}).

\paragraph{Bayesian adaptation.} While the preceding experiments update properties within an individual candidate, shared parameter uncertainty connects predictions across candidates. Across 27 protein-engineering systems~\citep{herrerarocha2026multitask}, coupled Laplace improves joint NLL over independent prediction in 23 systems. Its joint-event Brier score improves on a tuned multi-output GP and is comparable to a five-member ensemble with $4.1\times$ longer inference time; the ensemble retains better joint NLL, and their Brier difference is not statistically established. Uncapped experiments over 38,000 variants further illustrate the computational trade-off (\cref{tab:app-mtlmpo-27datasets,tab:app-mtlmpo-bll-pareto,tab:app-gpu-mtlmpo}).

In few-shot GDPa1 adaptation, the Bayesian head approaches nominal coverage at approximately an order of magnitude lower reported latency than the NTK-based comparator~\citep{calvoordonez2026richer}; covariance regularization improves feasibility estimation under lineage shift.

Assay descriptions provide a complementary form of generalization~\citep{giancardo2026camap}: on AIntibody and GDPa1, semantic conditioning raises zero-shot unseen-assay rank correlation from near zero to approximately $0.18$, with most gains explained by semantic pooling. Together, these studies illustrate how native distributions incorporate evidence from new observations, candidates and experimental contexts (\cref{app:supp-bayesian,tab:main-breadth}).

\subsection{From probabilistic inference to experimental discovery}
\label{sec:main-design-results}

\paragraph{Decision-aware acquisition.} We evaluate whether experiments should be selected for their effect on joint candidate feasibility rather than information gain alone. On 354 fully characterized AIntibody clones, decision-specific Bayesian acquisition improves mean confirmed leads per cost by $26.8\%$ relative to cost-aware GP knowledge gradient~\citep{frazier2008kg,frazier2009kg}, although this difference is not statistically established. Paired comparisons favor it over non-Bayesian coupled prediction and hypervolume-based GP acquisition. In the sparse library it also exceeds global information gain in observable confirmed leads, illustrating that uncertainty reduction and decision improvement can favor different experiments. The replica-covariance bound prunes $45.6\%$ of evaluated actions without reported numerical violations, while refitted GPs remain preferable on smaller GDPa1 and under some selection-dependent missingness mechanisms (\cref{app:supp-decisions}). Decision-local certification audits are reported in \cref{app:decision-certification-audit}.

\paragraph{Multiproperty optimization.} Joint feasibility guides selection among competing property requirements. Across five measured FLAb libraries~\citep{chungyoun2024flab}, LogiCA-POMA achieves $30.2\%$ mean co-optimal hit rate at matched budgets, versus $26.1\%$ for a tuned multi-output GP and $25.1\%$ for BOAT~\citep{rao2026boat}. Probabilistic guidance is also compatible with existing proposal mechanisms: on the 23 PMO tasks~\citep{gao2022sample}, POMA-guided SELFIES search improves upon Regression Transformer~\citep{born2023regression}, while probabilistic prescreening of Graph GA~\citep{jensen2019graphga} improves mean top-10 AUC from $0.628$ to $0.681$, with gains on 22 tasks. These are retrospective and oracle-based evaluations, not prospective synthesis (\cref{tab:app-independent-flab-boat,tab:pmo-generation-results}).

\paragraph{Property-conditioned generation.} To distinguish optimization from sampling a specified target law, we evaluate corrected transitions on an enumerable AbLang2-based sequence space~\citep{olsen2024ablang2}. Factor-augmented Metropolis--Hastings reduces total-variation error from $0.492$ for uncorrected Gibbs sampling to $0.307$, avoiding repeated outer Monte Carlo integration. Exact and pseudo-marginal transitions give similar results, and pilot-gated global proposals improve finite-budget sampling in favorable epistatic regimes. This supports coherent generation while distinguishing stationary invariance from finite-time fidelity and prospective biological validity (\cref{app:supp-design,app:pilot-gated-sampling}).

\section{Outlook}
\label{sec:outlook}

Biological foundation models can provide reusable probabilistic interfaces alongside reusable representations. By connecting native measurement distributions, \method{} integrates masked reconstruction, multimodal alignment and experimental reasoning. The experiments reveal complementary roles for likelihood learning, reference design, latent and output-space contrast, and statistical coupling. Analytical conditioning offers efficient updates, while specialized neural models can retain greater predictive flexibility. Reliable inference requires accurate dependence and partner-specific information, particularly under distribution shift. External pretraining overlap remains incompletely auditable, and retrospective or oracle-based design requires prospective validation.

More broadly, independently pretrained models can exchange evidence through their native outputs rather than a single universal representation. Such an interface could support iterative discovery systems in which measurements refine predictions, uncertainty propagates across candidates, and experiments are selected for their contribution to scientific objectives. This perspective extends biological foundation models toward integrated probabilistic reasoning and design.

\section*{Reproducibility Statement}
Source code, experimental configurations, and evaluation scripts
are available at \url{https://github.com/AI-SysBio/POMA}.
The supplementary material provides mathematical proofs,
experimental protocols, and extended results.
Lean~4 formalizations of selected theoretical statements
are provided in the \texttt{lean/} directory.

\section*{AI Use Statement}
Generative AI tools assisted with literature synthesis,
theoretical development, proof generation, experimental
implementation, and manuscript preparation. Accompanying
Lean formalizations are provided in the \texttt{lean/}
directory of the source repository. The author retains
full responsibility for the correctness and integrity
of all theoretical and empirical results.

\clearpage
\appendix
\setcounter{table}{0}
\renewcommand{\thetable}{S\arabic{table}}
\setcounter{figure}{0}
\renewcommand{\thefigure}{S\arabic{figure}}
\renewcommand{\theHfigure}{supp.\arabic{figure}}
\section*{Supplementary material}

The supplement follows the statistical operations enabled by \method{}. We first establish the likelihood interface and its mathematical properties, then describe the experimental protocols used to evaluate it. The empirical development proceeds from learning predictive distributions to alignment, conditioning on measurements, and propagating uncertainty across candidates. These operations provide the basis for experimental acquisition and property-conditioned design. Decision-regret theory is developed with the mathematical foundations and assessed alongside the acquisition experiments. Structure-adaptive sampling theory accompanies the generation studies. Stronger finite-budget generation guarantees are collected separately at the end, where their additional assumptions and numerical limitations are made explicit.

\begin{figure}[H]
\centering
\makebox[\linewidth][c]{%
  \begin{minipage}{1.3\linewidth}
    \centering
    \includegraphics[width=\linewidth]{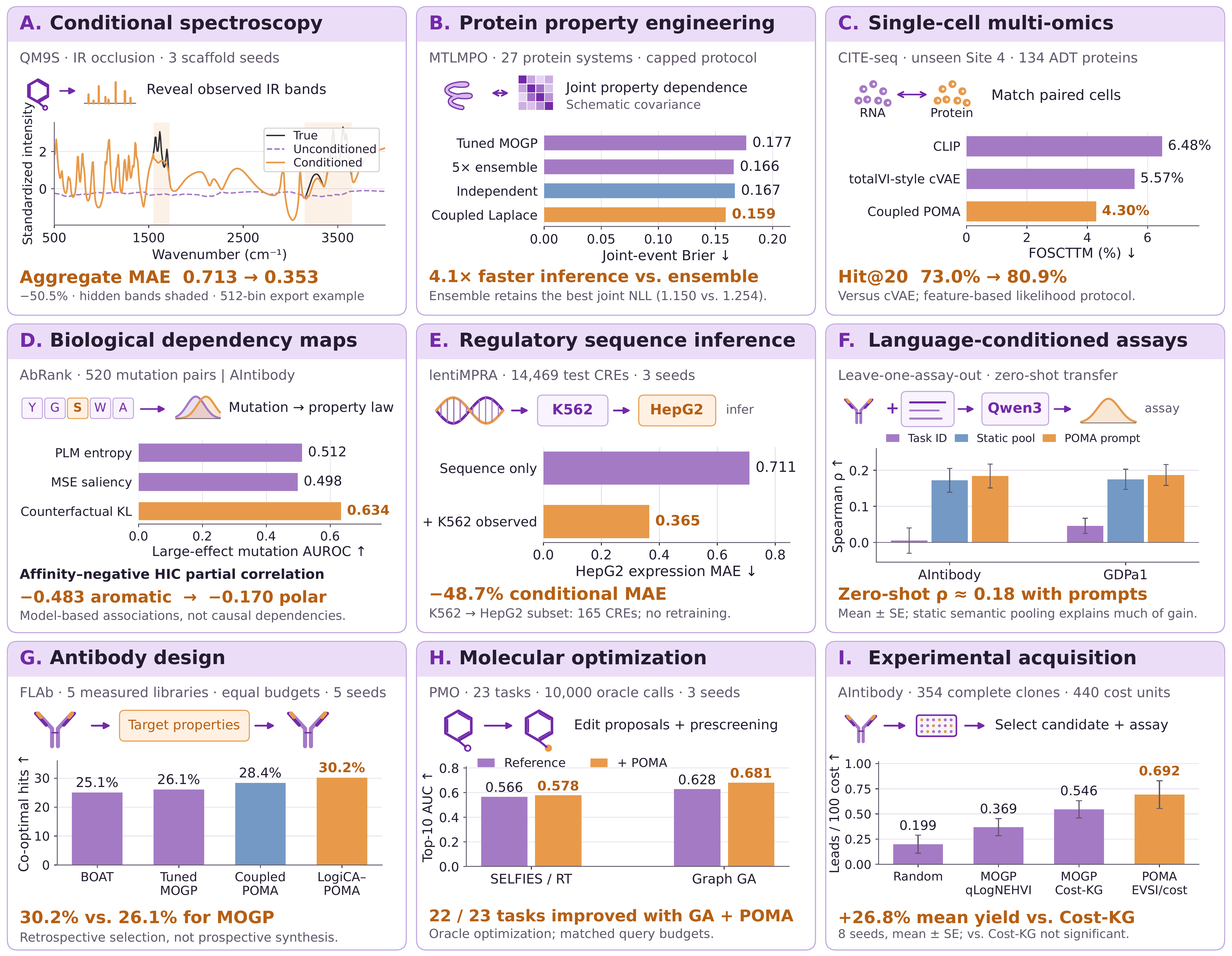}
    \caption{\textbf{Biological applications of the probabilistic output interface.}
    \textbf{(A)} Conditional IR reconstruction on QM9S: an illustrative 512-channel spectrum with shaded occluded bands; the aggregate MAE is from the separate 1024-channel, three-seed evaluation.
    \textbf{(B)} Joint-event Brier scores on 27 protein-engineering systems (capped protocol); the covariance illustration is schematic, and the ensemble retains lower joint NLL.
    \textbf{(C)} Feature-based RNA--protein cell matching on held-out CITE-seq Site~4, distinct from the scGPT--ESM-2 protocol.
    \textbf{(D)} Counterfactual KL sensitivity for AbRank mutation-effect discrimination and sequence-conditioned partial correlations for AIntibody; these are model-based, not causal, dependencies.
    \textbf{(E)} HepG2 expression prediction after observing K562 expression (lentiMPRA, 165-element conditional subset).
    \textbf{(F)} Zero-shot leave-one-assay-out transfer (mean $\pm$ SE): Qwen3 supplies contextual representations to a probabilistic measurement head, not a bidirectional masked-token decoder.
    \textbf{(G)} Retrospective antibody selection across five measured FLAb libraries.
    \textbf{(H)} PMO molecular optimization on 23 oracle tasks at matched budgets; the SELFIES reference uses Regression Transformer.
    \textbf{(I)} Complete-label AIntibody candidate--assay acquisition (eight seeds, mean $\pm$ SE); the mean advantage over cost-aware GP knowledge gradient is not statistically established. Task illustrations are schematic. No new prospective assays were performed.}
    \label{fig:supp-biological-applications}
  \end{minipage}%
}
\end{figure}

Each experimental section distinguishes the scientific question, the controlled comparison, and the limits of the evidence. Results obtained on different splits, candidate pools or training budgets are not pooled. \Cref{fig:supp-biological-applications} provides a visual overview of representative applications, with full protocols and results in the corresponding sections.

\begin{center}
\small
\begin{tabularx}{\linewidth}{@{}lXr@{}}
\toprule Section & Contents & Page\\
\midrule
\ref{app:supp-reference} & \hyperref[app:supp-reference]{Extended related work} & \pageref{app:supp-reference}\\
\ref{app:poma-theory} & \hyperref[app:poma-theory]{Mathematical foundations} & \pageref{app:poma-theory}\\
\ref{app:full-method} & \hyperref[app:full-method]{Experimental methodology and reproducibility} & \pageref{app:full-method}\\
\ref{app:supp-masked} & \hyperref[app:supp-masked]{Generalized probabilistic masked modeling} & \pageref{app:supp-masked}\\
\ref{app:supp-alignment} & \hyperref[app:supp-alignment]{Multimodal alignment in native output spaces} & \pageref{app:supp-alignment}\\
\ref{app:supp-inference} & \hyperref[app:supp-inference]{Structured conditional inference} & \pageref{app:supp-inference}\\
\ref{app:supp-bayesian} & \hyperref[app:supp-bayesian]{Bayesian adaptation and generalization} & \pageref{app:supp-bayesian}\\
\ref{app:supp-decisions} & \hyperref[app:supp-decisions]{Decision-aware experimental discovery and certification} & \pageref{app:supp-decisions}\\
\ref{app:supp-design} & \hyperref[app:supp-design]{Property-conditioned optimization and generation} & \pageref{app:supp-design}\\
\ref{app:poma-certified-adaptive} & \hyperref[app:poma-certified-adaptive]{Finite-budget guarantees for property-conditioned generation} & \pageref{app:poma-certified-adaptive}\\
\bottomrule
\end{tabularx}
\end{center}

\clearpage

\clearpage
\section{Extended related work}\label{app:supp-reference}

The output interface of a foundation model determines which statistical questions can be addressed directly after training. We review the relevant literature along this distinction, beginning with continuous measurements and then considering multimodal alignment and experimental design. POMA combines native predictive likelihoods and output-space alignment with established principles of distributional reconstruction, conditional inference, and Bayesian decision-making.

\subsection{Continuous and discrete probabilistic modeling}\label{sec:related-work}
\paragraph{Continuous variables in foundation models.} Expression-ranked tokens, value discretization and point-valued masked reconstruction have been widely used in single-cell foundation models~\citep{yang2022scbert,theodoris2023transfer,hao2024large,cui2024scgpt}. Related approaches appear in time series, weather, materials, mass spectroscopy and clinical laboratory modeling~\citep{cheng2026timemae,man2023w,kim2024masked,chen2024mattergpt,shen2025denoising,bushuiev2025self,bellamy2023labrador,restrepo2025representation}. These approaches learn informative representations but do not necessarily expose native joint measurement distributions. Distributional reconstruction itself is established: scPRINT uses a count likelihood, scVI and totalVI use probabilistic latent-variable models, and Moirai uses a mixture-distribution forecasting head~\citep{kalfon2025scprint,lopez2018deep,gayoso2021joint,woo2024unified}. Other recent work combines discrete language modeling with continuous diffusion or flow-based modeling~\citep{zhou2025transfusion,nguyen2026omnicast,morehead2026zatom}. POMA addresses a complementary question, namely how the output distributions of modality-specific models can participate directly in an alignment objective and remain available for subsequent probabilistic inference.

\paragraph{Numerical tokenization and native densities.} Regression Transformer~\citep{born2023regression} uses a numerical vocabulary to support concurrent prediction and generation, and regression-aware token objectives improve numerical prediction~\citep{zausinger2024regress}. Numerical token probabilities can induce a distribution over the represented support, but they do not directly parameterize the continuous joint density used here. The distinction concerns measurement geometry, the induced probability law, and the tractability of conditioning rather than an absence of all uncertainty in tokenized models.

\subsection{Probabilistic reconstruction and arbitrary conditioning}

\paragraph{Distributional regression and conditioning.} Mixture density networks, probabilistic boosting, calibrated regression, deep ensembles and conformal prediction offer established approaches to uncertainty estimation~\citep{bishop1994mixture,duan2020ngboost,kuleshov2018accurate,lakshminarayanan2017simple,lei2018distribution}. Proper scoring rules provide a principled means of comparing predictive distributions~\citep{gneiting2007strictly,scheuerer2015variogram,landsgesell2026distributional}. Domain-constrained models such as MultiPUFFIN embed property-specific physical structure~\citep{nogueira2026multipuffin}. Arbitrary conditional VAEs, energy-based conditional distributions and diffusion-based imputation already support inference from partial observations~\citep{ivanov2019variational,strauss2021arbitrary,tashiro2021csdi}. POMA uses a more restricted, computationally tractable joint measurement family, integrating these probabilistic principles with the masked likelihood and alignment objectives. It is compatible with additional variance scaling and conformal calibration.

\subsection{Biological contextualization and output-space alignment}

\paragraph{Modality and language context.} Structure-conditioned and inverse-folding language models use continuous structural information to condition categorical sequence prediction~\citep{hsu2022learning,yang2023masked}. ESM3 and SaProt incorporate structure through discrete structural tokens~\citep{hayes2025simulating,su2024saprot}. These models expand the context and outputs of sequence models; POMA addresses the different problem of assigning densities to continuous measurement values. Natural-language interfaces offer another means of specifying tasks and experimental contexts. BioT5 combines molecular, protein and natural-language sequences, while BioReason projects genomic representations into a language model~\citep{pei2023biot5,fallahpour2026bioreason}. The prompt-conditioned POMA variant uses a related encoder--projector--language-model pattern, but retains a native probabilistic output for the measured quantity.

\paragraph{Representation-space and output-space alignment.} Latent contrastive alignment, exemplified by CLIP, has been adapted to molecular--spectral and cellular multimodal data~\citep{radford2021clip,kalia2025jestr,chen2026flare,xu2026scclip}. In biological language models, native per-position probabilities are also useful for sequence plausibility, mutation effects, structure and design~\citep{lin2023evolutionary,riesselman2018deep,meier2021language,hie2024efficient,hie2022evolutionary,yuksel2023selformer,dalla2025nucleotide,cui2025towards,brixi2026genome,tomaz2024nucleotide,gong2024evorank,zhang2024interactingmotifs,madani2023large,johnson2021generating,mccarter2026make}. LogiCA contextualizes these categorical outputs directly~\citep{shao2026contextualizing}. Contrastive predictive coding estimates density ratios in learned representations, and Rank-N-Contrast introduces target-dependent similarity for regression~\citep{oord2018cpc,zha2023rank}. POMA extends native output-space contrast to continuous likelihoods; its scores are defined on the observations that the model reconstructs. This preserves the output distribution for conditional inference while allowing compatibility to be assessed through relative likelihoods.

Representation-space contrast and probabilistic reconstruction can coexist in a single model. A latent similarity does not itself specify a likelihood over observations; conversely, a model with a native predictive density need not use it as the sole retrieval score. This distinction motivates the controlled comparisons in \cref{app:supp-alignment,app:supp-e15-unified}, including regimes where latent contrast is preferable.

\subsection{Bayesian adaptation and experimental design}

\paragraph{Bayesian adaptation and design.} Multi-output GPs, constrained optimization and noisy hypervolume improvement provide established alternatives for correlated experimental design~\citep{bonilla2007multi,gardner2014bayesian,daulton2021parallel,ament2023unexpected}. Bayesian last layers and Laplace approximations provide efficient ways to attach parameter uncertainty to neural representations~\citep{daxberger2021laplace,calvoordonez2026richer}. Our use of a shared parameter posterior is motivated by evidence propagation across candidates and its interaction with a structured observation law. The acquisition and sampling results depend on the specified probabilistic model, not on a claim that Bayesian optimization or Metropolis--Hastings is novel in isolation.

\subsection{Antibody assay context and domain-specific baselines}\label{app:related}

Therapeutic antibody development requires more than strong affinity; expression, stability, solubility, viscosity, specificity and manufacturability may determine whether a candidate progresses~\citep{gordon2025therapeutic,erasmus2024aintibody,mason2021optimization}. Reduced affinity can even be desirable for certain biological mechanisms~\citep{yu2023reducing}. These trade-offs motivate the joint feasibility studies. Common triage measurements include HIC (hydrophobic interaction chromatography), AC-SINS (self-interaction), BVP and OVA/CHO polyreactivity assays, thermal stability and aggregation temperatures. SPR and KinExA provide affinity measurements. The thresholds and nominal assay costs used in this work are protocol-specific and do not constitute universal therapeutic development criteria.

Sequence and structure-based profilers such as TAP, CamSol and Protein-Sol assess developability liabilities~\citep{raybould2019five,sormanni2015camsol,hebditch2017protein}. Humanness or nativeness provides another useful design score, with tools including Hu-mAb, BioPhi and AbNatiV~\citep{marks2021humanization,prihoda2022biophi,ramon2024assessing}. Such scores do not by themselves establish clinical immunogenicity, which can also depend on peptide presentation, immune recognition and patient context. MHC-II presentation and antibody-specific immunogenicity predictors are complementary approaches~\citep{reynisson2020netmhcpan,chen2019predicting,wang2024abimmpred,o2024modular,huang2025antibody}. Heavy--light compatibility and repertoire assembly further constrain realistic sequence design~\citep{burbach2024improving,guo2025immunomatch}. Multi-objective design frameworks explicitly balance such measurements and risk scores~\citep{parkinson2025resp2,rao2026boat}. In our evaluations, measured affinity and developability are distinguished from surrogate sequence-level validation; no clinical safety or prospective experimental claim follows from the latter.

These studies motivate the formal development that follows. A likelihood-based interface must first define valid observation distributions and comparable cross-modal scores. Its value for experimental reasoning then depends on how those distributions represent dependence, propagate evidence and define decision utilities. We address these requirements in \cref{app:poma-theory}, and treat stronger guarantees for finite-budget generation separately in \cref{app:poma-certified-adaptive}.
\clearpage
\section{Mathematical foundations of POMA}\label{app:poma-theory}

We develop the mathematical formulation independently of any particular biological encoder or benchmark. The starting point is a likelihood over heterogeneous observations, with the appropriate probability measure for each type of output. Centering these likelihoods provides a compatibility score for alignment. A structured measurement law then makes conditioning and marginalization tractable, while a shared parameter posterior propagates evidence across candidates. We use these properties to formulate decision-specific acquisition, characterize regret and optimality under explicit uncertainty assumptions, and define coherent targets for biological design.

The change-of-variables and Gaussian conditioning identities are standard statistical tools; their role here is to establish the validity and computational form of the interface. The decision-specific bounds and sampling constructions build on these tools. Throughout, exactness refers to the stated model and assumptions. In particular, nonlinear encoder adaptation uses an approximate posterior, and correctness relative to a fitted measurement law does not imply correctness relative to the physical data-generating process. The empirical tests of these distinctions are presented in \cref{app:supp-masked,app:supp-alignment,app:supp-inference,app:supp-decisions,app:supp-design}. Specialized proposal constructions are developed alongside their generation experiments; conditional finite-budget guarantees are treated in \cref{app:poma-certified-adaptive}.

\subsection{Mixed-variable likelihoods and changes of measurement coordinates}\label{app:poma-mixed-decomposition}

Let $V$ collect discrete tokens or labels, integer counts, and continuous measurements, and let $W$ denote the available context. We score observed coordinates without serializing continuous values as tokens. For each coordinate, a contextual readout $h_i$ specifies either a mass function or a density in a transformed coordinate. The single-coordinate output score has the following forms; the Tobit row denotes the boundary-mass contribution, with the uncensored interior term given in \cref{eq:app-tobit-head}.

\begin{equation}
  q_{\theta,i}(v_i \mid h_i) =
  \begin{cases}
    \mathrm{Cat}(\pi_i) \;\text{or}\; \mathrm{ZINB}(\lambda_i,\theta_i,\pi_{0,i}), & \text{disc.},\\[2pt]
    p_{\eta_{\theta,i}(h_i)}\!\big(T_i(v_i)\big)\,\big|T_i'(v_i)\big|, & \text{cont.},\\[2pt]
    \Phi\!\big(\tfrac{z_{L,i}-\mu_i}{\sigma_i}\big)^{\mathbb{I}_{L}}\Phi\!\big(\tfrac{\mu_i-z_{U,i}}{\sigma_i}\big)^{\mathbb{I}_{U}}, & \text{\textsc{Tobit}},
  \end{cases}
  \label{eq:poma-mixed-output}
\end{equation}

The categorical and count heads are probability masses; a continuous head must account for the Jacobian to define a normalized density in the original measurement units. In the censored case, the observation law combines masses at the limits of quantitation with a density in the interior. The following derivation makes the continuous change of variables explicit.

Let $A = A_D \cup A_C$ partition the evaluated coordinates of biological object
$V$ into discrete coordinates $A_D$ and continuous coordinates $A_C$. For each
$k \in A_C$, let $T_k : \mathcal{I}_k \to \mathcal{J}_k$ be a $C^1$ strictly
monotone bijection with nonzero derivative, fitted strictly on training data (e.g.,
$T_k(v) = (\log_{10}(v - c_k) - m_k)/s_k$ for positive skewed quantities,
$T_k(v) = (\mathrm{logit}((v - a_k)/(b_k - a_k)) - m_k)/s_k$ for bounded
intervals $(a_k, b_k)$, or standardized Yeo--Johnson transformations). Under
change of variables $\mathbf{z}_{A_C} = (T_k(v_k))_{k \in A_C}$, any normalized
joint density $p_\eta(\mathbf{z}_{A_C} \mid W)$ supported and normalized on $\prod_{k\in A_C}\mathcal J_k$ induces a
valid probability density in original measurement units:
\begin{equation}
    \log q_\theta(\mathbf{v}_{A_C} \mid W)
    =
    \log p_\eta(\mathbf{z}_{A_C} \mid W)
    +
    \sum_{k \in A_C} \log \big|T_k'(v_k)\big|.
    \label{eq:app-poma-jacobian-joint}
\end{equation}
Here $\mathcal I_k$ is the specified modeling support, which may differ from the physically admissible measurement range. The normalization statement assumes that the transformed law places no mass outside $T(\prod_k\mathcal I_k)$. An unrestricted Gaussian after a variance-stabilizing transformation does not automatically satisfy this condition on a restricted physical domain: the Jacobian alone supplies neither truncation nor a boundary atom. The explicit bounded and censored heads in \cref{app:poma-extended-heads} specify those observation laws separately.

Let $C_i$ denote the visible discrete context for token $i$: the common masked input for masked reconstruction, or $V_{\setminus i}$ for leave-one-out scoring. Combining discrete categorical terms with Eq.~(\ref{eq:app-poma-jacobian-joint})
yields the unified per-coordinate masked log-score
\begin{equation}
    \ell_A(V \mid W)
    =
    \frac{1}{|A|}
    \left[
        \sum_{i \in A_D} \log \pi_{\theta,i}(v_i \mid C_i, W)
        +
        \log p_\eta(\mathbf{z}_{A_C} \mid W)
        +
        \sum_{k \in A_C} \log \big|T_k'(v_k)\big|
    \right].
    \label{eq:app-poma-mixed-decomposition}
\end{equation}

In a minibatch of size $B$, let $A_{b,D}$ and $A_{b,C}$ be the evaluated discrete and continuous coordinates, respectively, and $A_b=A_{b,D}\cup A_{b,C}$. The training loss corresponding to this score is

\begin{equation}
  \begin{aligned}
    \mathcal{L}_{\mathrm{MLM}} = -\frac{1}{B}\sum_{b=1}^B \frac{1}{|A_b|}\!\Biggl[
      &\sum_{i\in A_{b,D}}\!\log\pi_{\theta,i}(v_{b,i}\mid C_{b,i},W_b) \\[-3pt]
      &+ \log q_\theta(\mathbf{v}_{b,A_{b,C}}\mid W_b)
    \Biggr].
  \end{aligned}
  \label{eq:poma-mlm-loss}
\end{equation}

For a categorical head this is cross-entropy; for an independent fixed-variance Gaussian head it reduces to scaled MSE in the reconstruction coordinates, up to terms independent of the predictor (including the train-fitted Jacobian). Empty discrete or continuous blocks contribute zero, and the evaluated set itself must be nonempty. For counts, the discrete contribution is evaluated as a probability mass; for censored continuous measurements, the quantitative likelihood is taken with respect to the mixed observation measure described below. A sum of position-specific masked scores is not necessarily a normalized joint sequence likelihood. The quantitative block evaluates the joint observation law on the selected coordinates. Missing coordinates are integrated out rather than assigned numerical placeholder targets, under the stated observation-model assumptions. These distinctions are important when using the scores for inverse identification or sampling.

\paragraph{Distributional reconstruction and the native-coordinate law.}
A contextual representation specifies parameters $\boldsymbol\xi(C)$ of a normalized predictive density $p_\theta(z\mid C)=p(z;\boldsymbol\xi(C))$. For a scalar Gaussian head, predicting both the location and the positive variance yields
\begin{equation}
    -\log p_\theta(z\mid C)
    =\frac12\left[
    \frac{(z-\mu(C))^2}{\sigma^2(C)}
    +\log\sigma^2(C)+\log(2\pi)
    \right].
    \label{eq:main-gaussian-nll}
\end{equation}
Fixing variance gives scaled MSE up to predictor-independent terms; learning it instead weights residuals by predictive precision and penalizes excessive dispersion through the log-variance term. The negative log-score is proper for the specified observation law, but calibration remains an empirical property. For an observed continuous subset $S$, the corresponding native-coordinate density is
\begin{equation}
    q_\theta(y_S\mid C)
    =p_\theta(T_S(y_S)\mid C)
    \prod_{k\in S}|T'_k(y_k)|.
    \label{eq:main-jac}
\end{equation}
under the support and differentiability conditions above. This density is normalized in native measurement coordinates and changes with the units; it is the ratio of two such densities that admits the invariance result below.

\subsection{Reference-centered output-space compatibility}\label{app:supp-theory-alignment}

Likelihoods are affected by the background plausibility of an observation and by the coordinates in which a density is expressed. The context-free reference below is a special case of \cref{eq:main-pmi}. We then define the more general population and metadata-matched references used to distinguish partner-specific information from the effect of the presence of a modality.

\begin{equation}
  \begin{aligned}
    \Delta\ell_X(X\mid Y) &= \ell_{A_X}(X\mid Y) - \ell_{A_X}(X\mid\varnothing), \\
    \Delta\ell_Y(Y\mid X) &= \ell_{A_Y}(Y\mid X) - \ell_{A_Y}(Y\mid\varnothing),
  \end{aligned}
  \label{eq:poma-pmi-centered}
\end{equation}

\subsection{Reference laws and invariance}\label{app:matched-reference}

Let $q_\theta(y\mid x,C)$ be a normalized predictive density or mass function, and let $Q_m$ be a reference law on partner contexts, where $m$ denotes available metadata. With the evaluated coordinates and other context $C$ held fixed, define
\begin{equation}
 r_\theta(y\mid m,C)=\int q_\theta(y\mid x',C)\,dQ_m(x'),\qquad
 s_Y(x,y\mid m,C)=\log\frac{q_\theta(y\mid x,C)}{r_\theta(y\mid m,C)}.
 \label{eq:app-matched-reference}
\end{equation}
The reverse direction is defined analogously. Choosing a point mass on the absent context recovers the original reference; a population or metadata-matched distribution defines a mixture. An average embedding or learned null context instead defines a single reference prediction and need not equal the mixture. Reference selection must not depend on the unknown test pairing. For a compatible joint law conditioned on $(m,C)$, choosing $Q_m=p(x'\mid m,C)$ and the conditional component $q_\theta(y\mid x',m,C)=p(y\mid x',m,C)$ gives $r_\theta=p(y\mid m,C)$. The corresponding unnormalized ratio is conditional pointwise mutual information; the metadata may be implicit in $x$ or explicitly supplied to the head. Different coordinate normalizations need not yield a common normalized PMI.

For composite masked scores, two implementations must be distinguished. Averaging the normalized coordinate-wise predictive probabilities before taking logarithms defines coordinate-wise mixture references. Alternatively, centering a composite log-score by $\log \mathbb E_{Q_m}\exp(\ell)$ or by $\mathbb E_{Q_m}\ell$ defines a relative score; the exponentiated composite score need not itself be a joint density. These choices have different statistical interpretations even when they yield similar empirical rankings. The comparison in \cref{app:supp-e12-shortcut} examines their practical behavior.

\begin{proposition}[Reference invariance and covariance-aware directional weighting]
\label{prop:poma-alpha-pmi}
For fixed evaluated coordinates and positive numerator and reference density at the observation, a common smooth invertible coordinate transformation leaves the log likelihood ratio invariant, provided the reference law and metadata are transformed consistently. The same cancellation applies to composite score centering if the additive Jacobian term is independent of the partner context. If the two directional scores are conditionally unbiased estimators of a common normalized PMI with variances $\nu_X^2,\nu_Y^2$ and error covariance $c_{XY}$, then the minimum-variance convex combination uses
\begin{equation}
 \alpha^\star=\operatorname{clip}_{[0,1]}\!\left(\frac{\nu_Y^2-c_{XY}}{\nu_X^2+\nu_Y^2-2c_{XY}}\right).
 \label{eq:app-cov-aware-alpha}
\end{equation}
This expression applies when the denominator is positive; otherwise the conditional error difference is constant and every convex weight has the same variance.
\end{proposition}
\paragraph{Proof.}\label{app:poma-alpha-proof}
Write $J(y)=|\det D\psi(y)|$ for the absolute Jacobian determinant of a smooth invertible change of coordinates $\widetilde y=\psi(y)$ on the evaluated continuous block, with positive densities and nonzero Jacobian at the observation. Both the numerator and every component of the mixture reference acquire the factor $J(y)^{-1}$. Since this factor does not depend on the integration context, the reference acquires exactly the same factor:
\begin{equation}
 \begin{aligned}
 \widetilde r_\theta(\widetilde y\mid m,C)
 &=\int q_\theta(y\mid x',C)J(y)^{-1}\,dQ_m(x')\\
 &=J(y)^{-1}r_\theta(y\mid m,C),\\
 \log\frac{\widetilde q_\theta(\widetilde y\mid x,C)}
 {\widetilde r_\theta(\widetilde y\mid m,C)}
 &=\log\frac{q_\theta(y\mid x,C)J(y)^{-1}}
 {r_\theta(y\mid m,C)J(y)^{-1}}
 =\log\frac{q_\theta(y\mid x,C)}{r_\theta(y\mid m,C)}.
 \end{aligned}
 \label{eq:app-reference-jacobian-cancellation}
\end{equation}
For mixed observation laws the analogous statement follows from the Radon--Nikodym derivative under a common push-forward; the corresponding atom probabilities are unchanged. For composite log-scores, the same additive offset cancels against either reference convention described above. The variance of the fused estimator is
\[
 \alpha^2\nu_X^2+(1-\alpha)^2\nu_Y^2+2\alpha(1-\alpha)c_{XY}.
\]
Its quadratic coefficient is the variance of the directional error difference. If positive, differentiating and projecting onto $[0,1]$ gives the result. Otherwise the errors coincide almost surely under the stated unbiasedness assumption. Setting $c_{XY}=0$ recovers inverse-variance weighting. \(\square\)

The empirical weighting procedure estimates mask-induced score variability rather than error against an observed PMI. It is a reliability heuristic unless the common-target and unbiasedness assumptions are independently justified. Similarly, invariance of a score does not by itself imply that it isolates pair-specific information. Matched references and shuffling controls address that empirical question.

For a fixed observed target, reference centering also cancels when comparing different candidate biological objects. This permits direct ranking by a conditional likelihood ratio without evaluating a separate probability of the observed target.

For a nonempty evaluated block $A$ with positive predictive and reference densities, write $s_\theta(a,u)=|A|^{-1}\log[q_\theta(u_A\mid a)/q_0(u_A)]$. We optionally define a pairwise preference model using a logistic link,
\[
    P_\theta(u^A\succ u^B\mid a)
    \coloneqq \sigma\!\left(\frac{s_\theta(a,u^A)-s_\theta(a,u^B)}{\tau}\right),
    \qquad \sigma(t)=(1+e^{-t})^{-1},\quad \tau>0.
\]
This is a modeling choice, not a derived probability of an experimentally observed preference. The useful algebraic property is the reduction of score differences to likelihood ratios under a common reference.

\begin{proposition}[Output-likelihood ratio cancellation]
\label{prop:poma-likelihood-ratio}
For candidates $u^A,u^B$ evaluated on the same nonempty block $A$ under anchor $a$, the difference of reference-centered scores is
\[
    s_\theta(a,u^A)-s_\theta(a,u^B)
    =\frac{1}{|A|}\log\frac{q_\theta(u^A_A\mid a)q_0(u^B_A)}
    {q_\theta(u^B_A\mid a)q_0(u^A_A)}.
\]
For a fixed target assay assignment $u$ and two biological candidates $a^A,a^B$, a common reference and common measurement transformation give
\[
    s_\theta(a^A,u)-s_\theta(a^B,u)
    =\frac{1}{|A|}\log\frac{q_\theta(u_A\mid a^A)}{q_\theta(u_A\mid a^B)}.
\]
The latter difference is independent of the reference density and invariant under a consistent change of measurement coordinates.
\end{proposition}

\paragraph{Proof.}\label{app:poma-likelihood-ratio-proof}
Subtracting the two reference-centered log-scores and collecting logarithms gives the first identity. For the second identity, both scores evaluate the same $u_A$, so the common $\log q_0(u_A)$ cancels. A change of measurement coordinates adds the same log-Jacobian to both conditional log-densities, which also cancels. \(\square\)

The logistic link is monotone for $\tau>0$ and therefore preserves the ordering induced by these score differences. Its calibration as a preference probability is a separate empirical question. The cancellation requires a common target, evaluated coordinate set and reference law; a candidate-dependent reference need not cancel.

A related construction scores local perturbations against a reference. The following result concerns cancellation of the shared reference terms; if two assignments have different observed continuous values, their raw density differences need not be invariant under changes of coordinates.

\begin{proposition}[Reference-local density-ratio cancellation]
\label{prop:app-poma-reference-local-cancellation}
Let $V^A$ and $V^B$ be two continuous assignments conditioned on $W$ that perturb
the same nonempty coordinate subset $S$ relative to reference $V^0$, and define
$r_S(V \mid W; V^0) = \frac{1}{|S|}\sum_{k \in S} \log \frac{q_{\theta,k}(v_k \mid V_{\setminus k}, W)}{q_{\theta,k}(v^0_k \mid V^0_{\setminus k}, W)}$.
Then $r_S(V^A \mid W; V^0) - r_S(V^B \mid W; V^0) = \frac{1}{|S|}\sum_{k \in S}[\log q_{\theta,k}(v^A_k \mid V^A_{\setminus k}, W) - \log q_{\theta,k}(v^B_k \mid V^B_{\setminus k}, W)]$.
\end{proposition}

\paragraph{Proof.}
Subtracting $r_S(V^B \mid W; V^0)$ from $r_S(V^A \mid W; V^0)$ cancels the
shared reference terms $\log q_{\theta,k}(v^0_k \mid V^0_{\setminus k}, W)$ for
each $k \in S$. \(\square\)

\subsection{Contrastive supervision and measurement geometry}\label{app:contrast-geometry}

When contrastive supervision is used, we augment masked likelihood with an auxiliary alignment objective. Matched pairs provide the positive targets, and other observed in-batch values provide support-matched alternatives. Under independent sampling of paired examples, off-diagonal pairs are drawn from the product of the sampling marginals; constrained hard-negative sampling changes this distribution. The contrast is defined on the reference-centered scores, rather than on the absolute likelihoods. Using observed values rather than off-support perturbations discourages trivial support discrimination. The row-anchored objective is

\begin{equation}
  \begin{aligned}
    \mathcal{L}_{\textsc{POMA}}^{\mathrm{soft}} &= -\frac{1}{B}\sum_{b=1}^B \sum_{j=1}^B w_{b,j}\log\frac{\exp(s_\theta(a_b,u_j)/\tau)}{\sum_{c=1}^B\exp(s_\theta(a_b,u_c)/\tau)}, \\
    \mathcal{L}_{\mathrm{total}} &= \mathcal{L}_{\mathrm{MLM}} + \lambda_{\mathrm{rank}}\mathcal{L}_{\textsc{POMA}}^{\mathrm{soft}},
  \end{aligned}
  \label{eq:poma-poma-loss}
\end{equation}

The one-hot InfoNCE target assigns zero target mass to every non-self observation, irrespective of measurement similarity (although the score-dependent gradients need not weight all negatives equally). For scalar measurements $z_b,z_j$ in reconstruction coordinates, a noise-standardized distance and its associated Gibbs weights are
\begin{equation}
 \begin{aligned}
 d_{bj}&=\frac{|z_b-z_j|}{\sqrt{\sigma_{\mathrm{exp},b}^2+\sigma_{\mathrm{exp},j}^2+\epsilon_{\mathrm{noise}}^2}},\\
 w_{bj}&=\frac{\exp(-d_{bj}/\tau_{\mathrm{soft}})}{\sum_c\exp(-d_{bc}/\tau_{\mathrm{soft}})},
 \end{aligned}
 \label{eq:app-noise-standardized-geometry}
\end{equation}
where $\sigma_{\mathrm{exp},b}$ and $\sigma_{\mathrm{exp},j}$ denote replicate-derived measurement scales when available, $\epsilon_{\mathrm{noise}}>0$ regularizes the distance, and $\tau_{\mathrm{soft}}>0$. This is a general formulation: the reported SoftRank implementations use fixed variance-stabilized Gaussian/Tanimoto similarity scales rather than experimentally estimated replicate noise (\cref{app:reproducibility}). For the symmetric distance above, column-normalizing the same unnormalized similarities gives the weights $w_{j,b}$ for the reverse direction. Writing $S_{b,j}$ for the reference-centered compatibility score, for example $\log q_\theta(y_j\mid X_b,P_e)-\log q_0(y_j\mid P_e)$ in the assay-conditioned continuous direction, the symmetric soft cross-entropy is
\begin{equation}
  \mathcal{L}_{\textsc{POMA}}^{\mathrm{soft}} = -\frac{1}{2B}\sum_{b=1}^B \sum_{j=1}^B \left[w_{b,j}\log\frac{\exp(S_{b,j}/\tau)}{\sum_{c=1}^B \exp(S_{b,c}/\tau)} + w_{j,b}\log\frac{\exp(S_{b,j}/\tau)}{\sum_{a=1}^B \exp(S_{a,j}/\tau)}\right],
  \label{eq:app-soft-ranking}
\end{equation}
which assigns nonzero target mass to similar observations rather than treating every nonmatching pair as a strict negative. The reference-centered score retains the Jacobian cancellation of \cref{prop:poma-alpha-pmi}.

The invariance established above concerns the centered scores under a common change of variables. The soft-label geometry is separately specified in the reconstruction coordinates. The contrastive term is not itself a proper scoring rule, so a retrieval improvement need not imply improved calibration (\cref{app:supp-alignment}). In addition to the nominal $\lambda_{\mathrm{rank}}$, the contrastive temperature and the relative gradient magnitudes determine the effective contribution of the auxiliary objective. The gradient-ratio and temperature experiments in \cref{app:supp-e8-lambda} characterize this trade-off. The case $\lambda_{\mathrm{rank}}=0$ retains the likelihood objective and the inference-time output score; latent contrast is another compatible supervision choice.
\subsection{Observation-specific predictive distributions}\label{app:poma-extended-heads}

The mixed-variable objective permits different observation supports without changing its role as a log-score interface. In particular, a censored outcome contributes a probability mass rather than a density at the threshold. Bounded and overdispersed count measurements can also be represented directly. We give the observation-specific heads and the conditional Student-$t$ result used in \cref{app:extended-experiments}.

\paragraph{Two-sided Tobit censored density head.}
Many biological assays report lower-censored values ($v \le L_k$) or upper-censored values ($v \ge U_k$), including detection floors and saturation ceilings. Censoring direction depends on the reported coordinate: a non-binder reported as $K_D \ge 1\,\mu\mathrm{M}$ is upper-censored on the $K_D$ scale and lower-censored after a decreasing affinity transformation. For a strictly increasing $T_k$, modeling the latent transformed readout $Z_k^\star \sim \mathcal{N}(\mu_k(h_k), \sigma_k^2(h_k))$ observed through $Z_k = \mathrm{clip}(Z_k^\star, \tilde{L}_k, \tilde{U}_k)$ with $\tilde{L}_k = T_k(L_k)$ and $\tilde{U}_k = T_k(U_k)$ yields the exact hybrid log-likelihood~\citep{tobin1958estimation}
\begin{equation}
  \log q_{\theta,k}^{\mathrm{Tobit}}(v_k \mid h_k) =
  \begin{cases}
    \log \Phi\!\left(\dfrac{\tilde{L}_k - \mu_k}{\sigma_k}\right), & v_k \le L_k \;\text{(LLOQ)},\\[6pt]
    \log \phi\!\left(\dfrac{T_k(v_k) - \mu_k}{\sigma_k}\right) - \log\sigma_k + \log\big|T_k'(v_k)\big|, & L_k < v_k < U_k,\\[6pt]
    \log \Phi\!\left(\dfrac{\mu_k - \tilde{U}_k}{\sigma_k}\right), & v_k \ge U_k \;\text{(ULOQ)},
  \end{cases}
  \label{eq:app-tobit-head}
\end{equation}
For a decreasing transformation, the transformed limits must be reordered and the lower- and upper-tail events exchanged. The expectation below is in transformed coordinates; a nonlinear inverse transformation does not commute with expectation. For increasing $T_k$, the closed-form censored expectation is $\mathbb{E}[\mathrm{clip}(Z_k^\star, \tilde{L}_k, \tilde{U}_k) \mid h_k] = \tilde{L}_k \Phi(\alpha_L) + \tilde{U}_k (1 - \Phi(\alpha_U)) + \mu_k(\Phi(\alpha_U) - \Phi(\alpha_L)) + \sigma_k(\phi(\alpha_L) - \phi(\alpha_U))$, where $\alpha_L = (\tilde{L}_k - \mu_k)/\sigma_k$ and $\alpha_U = (\tilde{U}_k - \mu_k)/\sigma_k$.

\paragraph{Bounded Kumaraswamy density and Zero-Inflated Negative Binomial (\textsc{ZINB}) count heads.}
For double-bounded percentages $u_k = (v_k - a_k)/(b_k - a_k) \in (0, 1)$ (such as plasma protein binding in $[0,100]\%$), \method{} optionally parameterizes a \textsc{Kumaraswamy} distribution~\citep{kumaraswamy1980generalized} with shape parameters $a_\theta(h_k), b_\theta(h_k) > 0$:
\begin{equation}
  \log q_{\theta,k}^{\mathrm{Kuma}}(v_k \mid h_k)
  = \log a_\theta + \log b_\theta + (a_\theta - 1)\log u_k + (b_\theta - 1)\log(1 - u_k^{a_\theta}) - \log(b_k - a_k),
  \label{eq:app-kumaraswamy}
\end{equation}
which places strictly zero probability mass outside $[a_k, b_k]$ and admits closed-form CDF $F(u_k) = 1 - (1 - u_k^{a_\theta})^{b_\theta}$ and quantile function $F^{-1}(p) = (1 - (1 - p)^{1/b_\theta})^{1/a_\theta}$. For integer sequencing read counts $c_k \in \mathbb{N}_0$ from phage/yeast display selections or single-cell perturbation screens~\citep{risso2018general}, \method{} emits mean rate $\lambda_k > 0$, inverse dispersion $\theta_k > 0$, and dropout probability $\pi_{0,k} \in (0, 1)$ defining the \textsc{ZINB} PMF:
\begin{equation}
  q_{\theta,k}^{\mathrm{ZINB}}(c_k \mid h_k) =
  \begin{cases}
    \pi_{0,k} + (1 - \pi_{0,k})\!\left(\dfrac{\theta_k}{\theta_k + \lambda_k}\right)^{\!\theta_k}, & c_k = 0,\\[8pt]
    (1 - \pi_{0,k})\dfrac{\Gamma(c_k + \theta_k)}{\Gamma(\theta_k)\,\Gamma(c_k + 1)}\!\left(\dfrac{\theta_k}{\theta_k + \lambda_k}\right)^{\!\theta_k}\!\left(\dfrac{\lambda_k}{\theta_k + \lambda_k}\right)^{\!c_k}, & c_k \ge 1.
  \end{cases}
  \label{eq:app-zinb}
\end{equation}

A mixture head represents a density as $q(z\mid C)=\sum_{j=1}^J\omega_j(C)q_j(z\mid C)$, where the nonnegative weights sum to one. A hurdle head separately models nondetection and a normalized positive-value density: on the observation space $\{\mathrm{ND}\}\cup(\epsilon,\infty)$, write $P(\mathrm{ND}\mid C)=\pi(C)$ and $q(y\mid C)=(1-\pi(C))f_+(y\mid C)$ for $y>\epsilon$. Here $f_+$ integrates to one on the detected support. For a censored rather than hurdle model, the atom probability is the latent CDF at the detection threshold and the interior term is the corresponding latent density. These distinctions are relevant to the common observed-data scoring convention in \cref{app:metrics-protocol}.

\subsection{Structured measurement distributions and Gaussian conditional inference}\label{app:poma-woodbury-schur}

For a panel of $K$ continuous measurements, the readout predicts the mean, positive diagonal scale and rank-$r$ loading matrix. In the language-conditioned implementation, a zero-initialized cross-assay self-attention block precedes these readouts. The shared family can be written as a Gaussian or a multivariate Student-$t$ with degrees of freedom $\nu>2$:

\begin{equation}
  \begin{aligned}
    q_\theta(\mathbf{z}_C \mid X,\{P_k\}_{k=1}^K) &= \mathcal{T}_{\nu_\theta(X)}\!\big(\mathbf{z}_C;\,\boldsymbol{\mu}_\theta(X),\,\Sigma_\theta(X)\big), \\
    \Sigma_\theta(X) &= D_\theta(X) + U_\theta(X)U_\theta(X)^\top.
  \end{aligned}
  \label{eq:poma-lowrank-mvn}
\end{equation}

For clarity, $\mathcal T_\nu(\mu,\Sigma)$ denotes the $d$-dimensional Student-$t$ law with location $\mu$, positive-definite scale $\Sigma$, and density
\begin{equation}
 t_{\nu,d}(z;\mu,\Sigma)
 =\frac{\Gamma((\nu+d)/2)}{\Gamma(\nu/2)(\nu\pi)^{d/2}|\Sigma|^{1/2}}
 \left(1+\frac{(z-\mu)^\top\Sigma^{-1}(z-\mu)}{\nu}\right)^{-(\nu+d)/2}.
 \label{eq:app-mvt-density}
\end{equation}
For $\nu>2$, its covariance is $\nu\Sigma/(\nu-2)$; in the Gaussian limit the scale is the covariance. The factor representation is useful because it avoids dense linear algebra on the full assay panel, even when only a subset is observed. We establish the Gaussian marginalization and conditional identities first.

Suppose $K$ continuous coordinates have transformed representation
$\mathbf{z} = \boldsymbol{\mu} + U \boldsymbol{\eta} + \boldsymbol{\varepsilon}$
given conditioning object $X$, where $\boldsymbol{\eta} \sim \mathcal{N}(\mathbf{0}, I_r)$
and $\boldsymbol{\varepsilon} \sim \mathcal{N}(\mathbf{0}, D)$ independently with
$D = \mathrm{diag}(\sigma_1^2, \ldots, \sigma_K^2)$ and
$U = [\mathbf{u}_1, \ldots, \mathbf{u}_K]^\top \in \mathbb{R}^{K \times r}$.

The factor model has measurement covariance
\begin{equation}
    \Sigma=D+UU^\top,
    \label{eq:main-factor}
\end{equation}
with context-dependent parameters. The same cached factors can be used for subset likelihoods and conditional updates, rather than recomputing the neural representation after each observation.

\begin{proposition}[Exact sparse marginalization and Schur conditioning]
\label{prop:app-woodbury-schur}
Let $S \subseteq \{1, \ldots, K\}$ be any nonempty subset of observed assays for
object $X$, and define the $r \times r$ capacitance matrix
$C_S = I_r + U_S^\top D_S^{-1} U_S$. Then:
\begin{enumerate}[leftmargin=1.5em,itemsep=2pt]
    \item The exact marginal distribution of $\mathbf{z}_S$ is
    $\mathcal{N}(\boldsymbol{\mu}_S, D_S + U_S U_S^\top)$, whose log-density can
    be evaluated in $\mathcal{O}(|S|r^2 + r^3)$ operations via
    \begin{align}
        \log \det(D_S + U_S U_S^\top)
        &= \sum_{k \in S} \log \sigma_k^2 + \log \det C_S, \\
        (\mathbf{z}_S - \boldsymbol{\mu}_S)^\top (D_S + U_S U_S^\top)^{-1} (\mathbf{z}_S - \boldsymbol{\mu}_S)
        &= \|\mathbf{d}_S\|_2^2 - \mathbf{w}_S^\top C_S^{-1} \mathbf{w}_S,
    \end{align}
    where $d_k = (z_k - \mu_k)/\sigma_k$ and $\mathbf{w}_S = U_S^\top D_S^{-1}(\mathbf{z}_S - \boldsymbol{\mu}_S)$.
    \item Given observed assays $\mathbf{z}_B$ ($B \subset \{1, \ldots, K\}$), the
    posterior over the latent coupling factor is
    $\boldsymbol{\eta} \mid \mathbf{z}_B, X \sim \mathcal{N}(\mathbf{m}_{\eta|B}, C_B^{-1})$
    with $\mathbf{m}_{\eta|B} = C_B^{-1} U_B^\top D_B^{-1}(\mathbf{z}_B - \boldsymbol{\mu}_B)$.
    Consequently, for every unmeasured coordinate $k \notin B$,
    \begin{equation}
        z_k \mid \mathbf{z}_B, X
        \sim
        \mathcal{N}\!\left(
            \mu_k + \mathbf{u}_k^\top \mathbf{m}_{\eta|B},\;
            \sigma_k^2 + \mathbf{u}_k^\top C_B^{-1} \mathbf{u}_k
        \right).
        \label{eq:app-schur-posterior}
    \end{equation}
\end{enumerate}
\end{proposition}

\paragraph{Proof.}
Because marginalization of a multivariate Gaussian simply extracts the principal
submatrix indexed by $S$, $\mathrm{Cov}(\mathbf{z}_S \mid X) = D_S + U_S U_S^\top$.
Applying Sylvester's determinant identity gives
$\det(D_S + U_S U_S^\top) = \det(D_S)\det(I_r + U_S^\top D_S^{-1} U_S) = \det(D_S)\det(C_S)$,
and the Woodbury matrix identity gives
$(D_S + U_S U_S^\top)^{-1} = D_S^{-1} - D_S^{-1} U_S C_S^{-1} U_S^\top D_S^{-1}$,
proving (1). For (2), by linear-Gaussian conjugacy of the likelihood
$\mathbf{z}_B \mid \boldsymbol{\eta} \sim \mathcal{N}(\boldsymbol{\mu}_B + U_B \boldsymbol{\eta}, D_B)$
with prior $\boldsymbol{\eta} \sim \mathcal{N}(\mathbf{0}, I_r)$, the precision of
$\boldsymbol{\eta} \mid \mathbf{z}_B$ is $I_r + U_B^\top D_B^{-1} U_B = C_B$ and
its mean is $C_B^{-1} U_B^\top D_B^{-1}(\mathbf{z}_B - \boldsymbol{\mu}_B)$. Since
$\varepsilon_k \perp (\boldsymbol{\eta}, \mathbf{z}_B)$ for $k \notin B$,
$z_k = \mu_k + \mathbf{u}_k^\top \boldsymbol{\eta} + \varepsilon_k$ is Gaussian
with mean $\mu_k + \mathbf{u}_k^\top \mathbf{m}_{\eta|B}$ and variance
$\sigma_k^2 + \mathbf{u}_k^\top C_B^{-1} \mathbf{u}_k$. Setting $B = \{1,\ldots,K\}\setminus\{k\}$
immediately yields the exact leave-one-out conditional $q_{\theta,k}(z_k \mid \mathbf{z}_{\setminus k}, X)$.
\(\square\)

These identities are evaluated for arbitrary observed subsets of the fitted joint law. Their use for missing-data inference assumes the observation mechanism permits the specified likelihood or that it is explicitly modeled. Marginalization does not by itself correct selection bias from unobserved outcomes.

\subsection{Heavy-tailed and mixture conditional distributions}

A multivariate heavy-tailed model differs from a collection of independent heavy-tailed marginals because a shared scale factor couples uncertainty across coordinates. Conditioning on unusual observations updates this scale, producing a conditional distribution whose dispersion depends on the realized screening outcome.

\begin{proposition}[Coupled Low-Rank Multivariate Student-$t$ Woodbury marginalization and adaptive Schur conditioning]
\label{prop:app-mvt-schur}
Suppose $\mathbf{z} \mid X \sim \mathcal{T}_\nu(\boldsymbol{\mu}, D + U U^\top)$ follows a multivariate Student-$t$ distribution~\citep{kotz2004multivariate,shah2014student} with degrees of freedom $\nu > 2$, diagonal scale $D = \mathrm{diag}(s_1^2, \ldots, s_K^2)$, and rank-$r$ factor $U \in \mathbb{R}^{K \times r}$, constructed as the Gamma scale mixture $\mathbf{z} = \boldsymbol{\mu} + \tau^{-1/2}(U\boldsymbol{\eta} + \boldsymbol{\varepsilon})$ with mutually independent $\tau \sim \mathrm{Gamma}(\nu/2, \nu/2)$ (shape--rate convention), $\boldsymbol\eta\sim\mathcal N(0,I_r)$, and $\boldsymbol\varepsilon\sim\mathcal N(0,D)$, where $s_k>0$. The density convention is given in \cref{eq:app-mvt-density}. Then:
\begin{enumerate}[leftmargin=1.5em,itemsep=2pt]
  \item For any observed assay subset $S \subseteq \{1,\ldots,K\}$, the exact marginal is $\mathbf{z}_S \mid X \sim \mathcal{T}_\nu(\boldsymbol{\mu}_S, D_S + U_S U_S^\top)$, whose log-density evaluates in $\mathcal{O}(|S|r^2 + r^3)$ operations via $C_S = I_r + U_S^\top D_S^{-1} U_S$ and Woodbury quadratic form $\delta_S^2 = (\mathbf{z}_S - \boldsymbol{\mu}_S)^\top (D_S + U_S U_S^\top)^{-1}(\mathbf{z}_S - \boldsymbol{\mu}_S)$:
  \begin{equation}
    \log p(\mathbf{z}_S \mid X) = \log\frac{\Gamma(\frac{\nu + |S|}{2})}{\Gamma(\frac{\nu}{2})(\nu\pi)^{|S|/2}} - \frac{1}{2}\!\left(\sum_{k \in S}\!\log s_k^2 + \log\det C_S\!\right)\!-\!\frac{\nu\!+\!|S|}{2}\log\!\left(\!1 + \frac{\delta_S^2}{\nu}\!\right)\!.
  \end{equation}
  \item Given screening assays $\mathbf{z}_B$, define the Mahalanobis screening surprise
\[
 \delta_B^2 = (\mathbf{z}_B - \boldsymbol{\mu}_B)^\top
 (D_B + U_B U_B^\top)^{-1}(\mathbf{z}_B - \boldsymbol{\mu}_B).
\]
The exact conditional posterior over unmeasured assays $A = \{1,\ldots,K\}\setminus B$ is Student-$t$:
  \begin{equation}
    \mathbf{z}_A \mid \mathbf{z}_B, X \;\sim\; \mathcal{T}_{\nu + |B|}\!\left(\tilde{\boldsymbol{\mu}}_A,\; \frac{\nu + \delta_B^2}{\nu + |B|}\big(D_A + U_A C_B^{-1} U_A^\top\big)\right),
    \label{eq:app-mvt-schur}
  \end{equation}
  where $\tilde{\boldsymbol{\mu}}_A = \boldsymbol{\mu}_A + U_A C_B^{-1} U_B^\top D_B^{-1}(\mathbf{z}_B - \boldsymbol{\mu}_B)$. Unlike Gaussian Schur conditioning, whose covariance depends only on which assays $B$ were measured, Eq.~(\ref{eq:app-mvt-schur}) has an observation-dependent scale. The multiplier $(\nu + \delta_B^2)/(\nu + |B|)$ exceeds one exactly when $\delta_B^2>|B|$, and is below one when $\delta_B^2<|B|$. This is a scale multiplier, not a covariance multiplier; the conditional covariance additionally includes the Student-$t$ factor $(\nu+|B|)/(\nu+|B|-2)$.
\end{enumerate}
\end{proposition}

\paragraph{Proof.}
Marginalizing the Gaussian block in the normal--Gamma representation and integrating the common scale gives the Student-$t$ marginal with the same degrees of freedom. The determinant and quadratic-form identities in \cref{prop:app-woodbury-schur} give the stated evaluation cost and log-density. For the conditional law, set $\Sigma=D+UU^\top$ and $S_{A\mid B}=D_A+U_A C_B^{-1}U_A^\top$. The Gaussian block decomposition of the quadratic form yields
\[
 \delta_{A\cup B}^2=\delta_B^2+\delta_{A\mid B}^2,
 \qquad
 \delta_{A\mid B}^2=(\mathbf z_A-\tilde{\boldsymbol\mu}_A)^\top
 S_{A\mid B}^{-1}(\mathbf z_A-\tilde{\boldsymbol\mu}_A).
\]
The elementary factorization underlying the adaptive scale is
\begin{equation}
 1+\frac{\delta_B^2+\delta_{A\mid B}^2}{\nu}
 =\left(1+\frac{\delta_B^2}{\nu}\right)
  \left(1+\frac{\delta_{A\mid B}^2}{\nu+\delta_B^2}\right).
 \label{eq:app-mvt-scale-factorization}
\end{equation}
At fixed $\mathbf z_B$, the first factor is constant in $\mathbf z_A$. Inserting the second factor into \cref{eq:app-mvt-density} identifies the conditional density kernel as that of a Student-$t$ law with degrees of freedom $\nu+|B|$, location $\tilde{\boldsymbol\mu}_A$, and scale $(\nu+\delta_B^2)S_{A\mid B}/(\nu+|B|)$. Normalization gives \cref{eq:app-mvt-schur}. Equivalently, the normal--Gamma update has shape $(\nu+|B|)/2$ and rate $(\nu+\delta_B^2)/2$, with Gaussian conditional covariance $S_{A\mid B}/\tau$. \(\square\)

This conditional closure should not be conflated with the result of integrating an additional Gaussian parameter posterior, which in general gives a location mixture rather than an exact Student-$t$ law.

A finite mixture of structured laws can likewise be conditioned without retraining. If the component marginal densities on $B$ are $q_j(z_B\mid C)$, Bayes' rule gives updated weights $\omega_j(z_B,C)\propto\omega_j(C)q_j(z_B\mid C)$ and the conditional mixture $\sum_j\omega_j(z_B,C)q_j(z_A\mid z_B,C)$. Gaussian components retain their Schur conditionals. This identity does not imply that mixtures of independent scalar heads model cross-coordinate dependence unless they share a component variable. The coupled-mixture experiments are given in \cref{app:supp-coupled-gmm}.

\subsection{Bayesian posterior closure and propagation across candidates}\label{app:supp-theory-bayes}\label{app:poma-laplace-proof}

The low-rank factor in a measurement law describes dependence within a candidate. A shared posterior over mean-readout parameters introduces a second source of dependence, between candidates. In the fixed-feature Gaussian case, take a common weight vector $\mathbf w_\mu\in\mathbb R^p$ with prior $\mathcal N(\mathbf w_{\mathrm{pre}},\tau_0^{-1}I_p)$ and design matrix $\Phi(X)\in\mathbb R^{K\times p}$. For the cross-candidate covariance identity, it suffices that residuals have conditional mean zero, conditional covariance $D_\theta(X)+U_\theta(X)U_\theta(X)^\top$, and zero conditional cross-covariance between distinct candidates, given the weights and observed data. The law of total covariance then gives \cref{eq:main-cross-cov} without a Gaussianity or independence assumption. The exact posterior and conditioning results below additionally assume a jointly Gaussian residual vector with weight-independent block-diagonal covariance; in that regime uncorrelated residual blocks are independent. The next result combines measurement and parameter uncertainties in this Gaussian representation and bounds the effect of truncating the parameter covariance.

\begin{theorem}[Structured joint posterior closure, cross-candidate propagation, and spectral truncation]
\label{thm:joint-candidate-assay-closure}
\label{prop:poma-laplace-closure}
In the fixed-feature Gaussian regime, suppose the residual vector across candidates $X_1,\ldots,X_n$ (and any newly observed candidate) is conditionally jointly Gaussian with zero mean and weight-independent block-diagonal covariance. Then:
(i) the joint posterior predictive vector $(\mathbf{Z}_1,\ldots,\mathbf{Z}_n)\mid\mathcal{D}$ is Gaussian with block covariance $\mathrm{Cov}(\mathbf{Z}_i,\mathbf{Z}_j\mid\mathcal{D})=\Phi(X_i)\Sigma_W\Phi(X_j)^\top + \mathbb{I}_{\{i=j\}}\Sigma_{\mathrm{ale}}(X_i)$, and factoring $\Sigma_W=L_W L_W^\top$ ($L_W\in\mathbb{R}^{p\times r'}$) preserves single-candidate Woodbury--Schur closure $\Sigma_{\mathrm{pred}}(X)=D_\theta(X)+\tilde{U}(X)\tilde{U}(X)^\top$ with $\tilde{U}(X)=[U_\theta(X),\Phi(X)L_W]\in\mathbb{R}^{K\times\tilde{r}}$ ($\tilde{r}=r+r'$);
(ii) observing partial assays $S\subseteq\{1,\ldots,K\}$ on $X_{n+1}$ updates $(\hat{\mathbf{w}}_{\mathrm{post}},\Sigma_W)$ and shifts the predictive mean of unmeasured assay $a$ for any previously indexed candidate $X_j$ ($j\le n$) by
\[
\boldsymbol{\phi}_{j,a}^\top\Sigma_W\Phi_{n+1,S}^\top
\Sigma_{\mathrm{pred},S}(X_{n+1})^{-1}
(\mathbf{z}_{n+1,S}-\hat{\boldsymbol{\mu}}_{n+1,S}).
\]
With a cached posterior factor, the mean-update contraction costs $\mathcal{O}(|S|\tilde{r}^2+\tilde{r}^3+|S|pr')$ operations (excluding dense covariance materialization); and
(iii) truncating $\Sigma_W$ to its top-$r'$ eigenpairs satisfies $D_{\mathrm{KL}}(\mathcal{N}(\hat{\boldsymbol{\mu}}_S,\tilde{\Sigma}_{S,r'})\,\|\,\mathcal{N}(\hat{\boldsymbol{\mu}}_S,\Sigma_S))\le \frac{\|\Phi_S(X)\|_F^2}{2\sigma_{\min}(D_S)}\sum_{\ell=r'+1}^p\lambda_\ell(\Sigma_W)$.
\end{theorem}

\paragraph{Proof of Theorem~\ref{thm:joint-candidate-assay-closure} (Joint candidate--assay closure, sequential updates, and spectral truncation).}
Let $\Phi(X) \in \mathbb{R}^{K \times p}$ ($p = d/2 + 1$) denote the homogeneous penultimate readout matrix across $K$ prompts so that $\boldsymbol{\mu}_{\mathbf{w}}(X) = \Phi(X)\mathbf{w}_\mu$ (or let $\Phi(X) = \Psi_{\mathrm{NTK}}(X) \in \mathbb{R}^{K \times m}$ in the subsampled NTK-enriched parameterization of \citealt{calvoordonez2026richer}). Under fixed features, fixed Gaussian observation covariance $\Sigma_{\mathrm{ale},S_n}(X_n) = D_{n,S_n} + U_{n,S_n}U_{n,S_n}^\top$, independent experimental residuals across candidates, and prior $\mathbf{w}_\mu \sim \mathcal{N}(\mathbf{w}_{\mathrm{pre}}, \tau_0^{-1}I_p)$, linear-Gaussian conjugacy yields an exact Gaussian posterior $\mathbf{w}_\mu \mid \mathcal{D} \sim \mathcal{N}(\hat{\mathbf{w}}_{\mathrm{post}}, \Sigma_W)$ with
\begin{align}
    \Lambda_{\mathrm{post}} &= \tau_0 I_p + \sum_{n=1}^N \Phi_{n,S_n}^\top \Sigma_{\mathrm{ale},S_n}^{-1} \Phi_{n,S_n}, \qquad \Sigma_W = \Lambda_{\mathrm{post}}^{-1}, \\
    \hat{\mathbf{w}}_{\mathrm{post}} &= \Sigma_W\!\left(\tau_0 \mathbf{w}_{\mathrm{pre}} + \sum_{n=1}^N \Phi_{n,S_n}^\top \Sigma_{\mathrm{ale},S_n}^{-1} \mathbf{z}_{n,S_n}\right).
\end{align}
Under nonlinear joint fine-tuning, a generalized Gauss--Newton precision can instead define a local Gaussian approximation; it is not the conjugate posterior above. With independent Gaussian uncertainty in the location of a Student-$t$ observation model ($\nu>2$), the predictive law is a Gaussian--Student-$t$ location mixture, with covariance equal to the Gaussian location covariance plus $\nu/(\nu-2)$ times the Student-$t$ scale matrix.

\emph{Part (i):} For any collection of candidates $X_1,\ldots,X_n$, write $\mathbf{Z}_i = \Phi(X_i)\mathbf{w}_\mu + \boldsymbol{\epsilon}_i$ where $\mathbf{w}_\mu \mid \mathcal{D} \sim \mathcal{N}(\hat{\mathbf{w}}_{\mathrm{post}}, \Sigma_W)$ is shared across all $i$ and $\boldsymbol{\epsilon}_i \sim \mathcal{N}(\mathbf{0}, \Sigma_{\mathrm{ale}}(X_i))$ are mutually independent across $i$ and independent of $\mathbf{w}_\mu$. Joint linearity implies $(\mathbf{Z}_1,\ldots,\mathbf{Z}_n)\mid\mathcal{D}$ is jointly Gaussian. By the law of total covariance, the conditional residual cross-covariances vanish for distinct candidates, and the covariance of the conditional means is $\Phi(X_i)\Sigma_W\Phi(X_j)^\top$. Thus the joint law has $\mathrm{Cov}(\mathbf{Z}_i, \mathbf{Z}_j \mid \mathcal{D}) = \Phi(X_i)\Sigma_W\Phi(X_j)^\top + \mathbb{I}_{\{i=j\}}\Sigma_{\mathrm{ale}}(X_i)$. Setting $i=j$ and factoring $\Sigma_W = L_W L_W^\top$ ($L_W \in \mathbb{R}^{p\times r'}$) gives $\Sigma_{\mathrm{pred}}(X) = D_\theta(X) + \tilde{U}(X)\tilde{U}(X)^\top$ with $\tilde{U}(X) = [U_\theta(X), \Phi(X)L_W] \in \mathbb{R}^{K \times \tilde{r}}$ ($\tilde{r}=r+r'$), so Proposition~\ref{prop:app-woodbury-schur} applies verbatim. If $L_W$ retains only a truncated eigenspace, the same identities apply to that approximate Gaussian posterior; part~(iii) bounds its predictive discrepancy from the full posterior.

\emph{Part (ii):} Consider observing partial panel $\mathbf{z}_{n+1,S}$ on $X_{n+1}$. Because $(\mathbf{Z}_j, \mathbf{Z}_{n+1,S}) \mid \mathcal{D}$ is jointly Gaussian with cross-covariance $\mathrm{Cov}(\mathbf{Z}_j, \mathbf{Z}_{n+1,S} \mid \mathcal{D}) = \Phi(X_j)\Sigma_W \Phi_{n+1,S}^\top$ and marginal covariance $\Sigma_{\mathrm{pred},S}(X_{n+1}) = D_{n+1,S} + \tilde{U}_{n+1,S}\tilde{U}_{n+1,S}^\top$, standard Gaussian conditioning updates $\Sigma_W \mapsto \Sigma_W - \Sigma_W \Phi_{n+1,S}^\top \Sigma_{\mathrm{pred},S}(X_{n+1})^{-1} \Phi_{n+1,S} \Sigma_W$ and shifts the posterior predictive mean of every assay $a$ of candidate $X_j$ by
\begin{align*}
    &\mathbb{E}\big[Z_{j,a} \mid \mathcal{D} \cup \{(X_{n+1}, \mathbf{z}_{n+1,S})\}\big] - \mathbb{E}[Z_{j,a} \mid \mathcal{D}] \\
    &\quad = \boldsymbol{\phi}_{j,a}^\top \Sigma_W \Phi_{n+1,S}^\top \Sigma_{\mathrm{pred},S}(X_{n+1})^{-1}(\mathbf{z}_{n+1,S} - \hat{\boldsymbol{\mu}}_{n+1,S}).
\end{align*}
Evaluating $\Sigma_{\mathrm{pred},S}(X_{n+1})^{-1}(\mathbf{z}_{n+1,S} - \hat{\boldsymbol{\mu}}_{n+1,S})$ via the $\tilde{r}\times\tilde{r}$ Woodbury capacitance Cholesky factor costs $\mathcal{O}(|S|\tilde{r}^2 + \tilde{r}^3)$, and contracting $\Sigma_W \Phi_{n+1,S}^\top$ via $L_W L_W^\top$ costs $\mathcal{O}(|S|pr')$, yielding the stated cost for the mean-update contraction with a cached low-rank factor. Explicitly materializing a dense updated parameter covariance requires additional work.

\emph{Part (iii):} Let $\Sigma_W = \sum_{\ell=1}^p \lambda_\ell \mathbf{v}_\ell \mathbf{v}_\ell^\top$ ($\lambda_1 \ge \dots \ge \lambda_p > 0$) and $\Sigma_{W,r'} = \sum_{\ell=1}^{r'} \lambda_\ell \mathbf{v}_\ell \mathbf{v}_\ell^\top$, with residual $\Delta_W = \Sigma_W - \Sigma_{W,r'} \succeq 0$. For any subset $S$, let $\Sigma_S = \Sigma_{\mathrm{ale},S} + \Phi_S \Sigma_W \Phi_S^\top$ and $\tilde{\Sigma}_{S,r'} = \Sigma_{\mathrm{ale},S} + \Phi_S \Sigma_{W,r'} \Phi_S^\top$, so $\Sigma_S - \tilde{\Sigma}_{S,r'} = \Phi_S \Delta_W \Phi_S^\top \succeq 0$. Let $\mu_1,\ldots,\mu_{|S|} \ge 0$ denote the eigenvalues of $\tilde{\Sigma}_{S,r'}^{-1/2}(\Phi_S \Delta_W \Phi_S^\top)\tilde{\Sigma}_{S,r'}^{-1/2}$, so the eigenvalues of $\tilde{\Sigma}_{S,r'}^{-1/2}\Sigma_S \tilde{\Sigma}_{S,r'}^{-1/2}$ are $1 + \mu_k \ge 1$. Using $\log(1+\mu) - \frac{\mu}{1+\mu} = \int_0^\mu \frac{t}{(1+t)^2}\,dt \le \mu$ for all $\mu \ge 0$, we have
\begin{align*}
    D_{\mathrm{KL}}\!\big(\mathcal{N}(\hat{\boldsymbol{\mu}}_S, \tilde{\Sigma}_{S,r'}) \,\|\, \mathcal{N}(\hat{\boldsymbol{\mu}}_S, \Sigma_S)\big)
    &= \frac{1}{2}\sum_{k=1}^{|S|} \left[\log(1 + \mu_k) - \frac{\mu_k}{1 + \mu_k}\right] \\
    &\le \frac{1}{2}\sum_{k=1}^{|S|} \mu_k
    = \frac{1}{2}\mathrm{tr}\!\left(\tilde{\Sigma}_{S,r'}^{-1} \Phi_S \Delta_W \Phi_S^\top\right).
\end{align*}
Since $\tilde{\Sigma}_{S,r'} \succeq D_S \succeq \sigma_{\min}(D_S) I_{|S|}$, Von Neumann's trace inequality gives $\mathrm{tr}(\tilde{\Sigma}_{S,r'}^{-1} \Phi_S \Delta_W \Phi_S^\top) \le \sigma_{\min}(D_S)^{-1} \mathrm{tr}(\Phi_S^\top \Phi_S \Delta_W) \le \sigma_{\min}(D_S)^{-1} \|\Phi_S\|_F^2 \mathrm{tr}(\Delta_W) = \sigma_{\min}(D_S)^{-1}\|\Phi_S\|_F^2 \sum_{\ell=r'+1}^p \lambda_\ell$, establishing Part~(iii). \(\square\)

One consequence of the shared posterior is that the variance of a local mutation difference includes the cross-candidate covariance; positive shared covariance reduces this variance relative to the sum of the two marginal epistemic variances. Another is that information about expensive measurements can be evaluated using low-dimensional matrix determinants. These consequences are useful for the decision studies in \cref{app:supp-decisions}.

\begin{proposition}[Shared-posterior mutation scaling and cost-aware assay information gain]
\label{thm:mutation-and-info-gain}
(i) If $X\mapsto\boldsymbol{\phi}(X)$ is locally $L$-Lipschitz under sequence/graph metric $d$, then under a shared draw $\mathbf{w}_\mu\sim\mathcal{N}(\hat{\mathbf{w}}_{\mathrm{post}},\Sigma_W)$, \[
\begin{aligned}
\mathrm{Var}_{\mathrm{epi}}(Z_1-Z_2\mid\mathcal{D})
 &=\|(\boldsymbol{\phi}(X_1)-\boldsymbol{\phi}(X_2))^\top L_W\|_2^2\\
 &\le \|\Sigma_W\|_{\mathrm{op}}L^2 d(X_1,X_2)^2,
\end{aligned}
\]
which bounds the differential epistemic variance through the edit distance and accounts for shared antigen uncertainty.
(ii) Partitioning assays of candidate $X$ into screening subset $B$ ($c(B)\le C$) and expensive target assays $A$ under $\Sigma_{\mathrm{pred}}(X)=D(X)+\tilde{U}(X)\tilde{U}(X)^\top$ ($\tilde{U}\in\mathbb{R}^{K\times\tilde{r}}$), the exact mutual information reduces to $\tilde{r}\times\tilde{r}$ determinants:
$I(\mathbf{Z}_A;\mathbf{Z}_B\mid X,\mathcal{D})=\frac{1}{2}[\log\det(I_{\tilde{r}}+M_A)+\log\det(I_{\tilde{r}}+M_B)-\log\det(I_{\tilde{r}}+M_A+M_B)]$
with $M_S=\tilde{U}_S^\top D_S^{-1}\tilde{U}_S$. After caching $M_A$ and $\log\det(I_{\tilde r}+M_A)$ for the fixed target panel, each screening set $B$ can be scored in $\mathcal{O}(|B|\tilde{r}^2+\tilde{r}^3)$ time, independent of $|A|$; forming the target cache costs $\mathcal{O}(|A|\tilde r^2+\tilde r^3)$.
\end{proposition}

\paragraph{Proof of Proposition~\ref{thm:mutation-and-info-gain} (Lipschitz mutation scaling and Woodbury assay information gain).}
\emph{Part (i):} Under a shared draw $\mathbf{w}_\mu = \hat{\mathbf{w}}_{\mathrm{post}} + L_W \boldsymbol{\zeta}$ ($\boldsymbol{\zeta} \sim \mathcal{N}(\mathbf{0}, I_{r'})$, $L_W L_W^\top = \Sigma_W$) with independent experimental noises $\varepsilon_1, \varepsilon_2$, $Z_1 - Z_2 = (\boldsymbol{\phi}(X_1) - \boldsymbol{\phi}(X_2))^\top \hat{\mathbf{w}}_{\mathrm{post}} + (\boldsymbol{\phi}(X_1) - \boldsymbol{\phi}(X_2))^\top L_W \boldsymbol{\zeta} + (\varepsilon_1 - \varepsilon_2)$, so \[
\begin{aligned}
\mathrm{Var}_{\mathrm{epi}}(Z_1 - Z_2\mid\mathcal D)
 &= (\boldsymbol\phi(X_1)-\boldsymbol\phi(X_2))^\top L_WL_W^\top
 (\boldsymbol\phi(X_1)-\boldsymbol\phi(X_2))\\
 &\le \|L_WL_W^\top\|_{\mathrm{op}}\|\boldsymbol\phi(X_1)-\boldsymbol\phi(X_2)\|_2^2\\
 &\le \|\Sigma_W\|_{\mathrm{op}}L^2d(X_1,X_2)^2,
\end{aligned}
\]
whenever $\|\boldsymbol{\phi}(X_1) - \boldsymbol{\phi}(X_2)\|_2 \le L\,d(X_1, X_2)$.

\emph{Part (ii):} For disjoint assay subsets $A, B \subseteq \{1,\ldots,K\}$ with $S = A \cup B$, Gaussian mutual information satisfies $I(\mathbf{Z}_A; \mathbf{Z}_B \mid X, \mathcal{D}) = \frac{1}{2}[\log\det\Sigma_{AA} + \log\det\Sigma_{BB} - \log\det\Sigma_{SS}]$. Under $\Sigma_{\mathrm{pred}}(X) = D + \tilde{U}\tilde{U}^\top$ ($\tilde{U} \in \mathbb{R}^{K\times\tilde{r}}$), the matrix determinant lemma gives $\log\det\Sigma_{TT} = \log\det D_T + \log\det(I_{\tilde{r}} + \tilde{U}_T^\top D_T^{-1}\tilde{U}_T)$ for any subset $T$. Since $D_S = \mathrm{diag}(D_A, D_B)$, $\log\det D_A + \log\det D_B - \log\det D_S = 0$ cancels identically, while $\tilde{U}_S^\top D_S^{-1}\tilde{U}_S = \tilde{U}_A^\top D_A^{-1}\tilde{U}_A + \tilde{U}_B^\top D_B^{-1}\tilde{U}_B = M_A + M_B$. Substituting these identities yields $I(\mathbf{Z}_A; \mathbf{Z}_B \mid X, \mathcal{D}) = \frac{1}{2}[\log\det(I_{\tilde{r}} + M_A) + \log\det(I_{\tilde{r}} + M_B) - \log\det(I_{\tilde{r}} + M_A + M_B)]$. Precomputing $M_A$ and $\log\det(I_{\tilde{r}} + M_A)$ once per candidate allows any candidate screening set $B$ to be scored by forming $M_B = \tilde{U}_B^\top D_B^{-1}\tilde{U}_B$ in $\mathcal{O}(|B|\tilde{r}^2)$ time and computing two $\tilde{r}\times\tilde{r}$ Cholesky log-determinants in $\mathcal{O}(\tilde{r}^3)$ time, independent of $|A|$. \(\square\)

For mixed likelihoods or nonlinear parameter adaptation, the Gaussian posterior must instead be treated as an approximation. The following observed-path identity isolates its error under adaptive data collection. It does not assume that the experimental policy is fixed in advance; it requires that the chosen action depend only on the recorded past and not additionally on the latent parameter.

\begin{proposition}[Observed-path identity and conditional TV bound for adaptive mixed likelihoods]
\label{prop:app-adaptive-mixed-tv}
Let $\mathcal{D}_T = \{(x_t, y_t, c_t)\}_{t=1}^T$ be an adaptively collected trajectory where $x_t \sim \pi_t(\cdot\mid \mathcal{D}_{t-1})$ depends only on past observations and $(y_t, c_t)\mid (x_t,\mathbf{w})$ follows any mixed continuous (e.g., heavy-tailed Student-$t_\nu$) and categorical (e.g., logistic gate) likelihood $p(y_t,c_t\mid x_t,\mathbf{w})$ under prior $p_0(\mathbf{w})$. Let $q_{\mathrm{Lap}}(\mathbf{w}) = \mathcal{N}(\hat{\mathbf{w}}_{\mathrm{MAP}}, H_{\mathrm{MAP}}^{-1})$ denote the Laplace-Gaussian posterior and define the observed-path non-Gaussian log-residual $\Delta_L(\mathbf{w}) \coloneqq [\log p_0(\mathbf{w}) + \sum_{t=1}^T \log p(y_t,c_t\mid x_t,\mathbf{w}) - \log p_{\mathrm{MAP}}] + \frac{1}{2}(\mathbf{w}-\hat{\mathbf{w}}_{\mathrm{MAP}})^\top H_{\mathrm{MAP}}(\mathbf{w}-\hat{\mathbf{w}}_{\mathrm{MAP}})$. Then:
(i) the adaptive policy factors $\prod_{t=1}^T \pi_t(x_t\mid\mathcal{D}_{t-1})$ cancel identically in the exact posterior $p_{\mathrm{exact}}(\mathbf{w}\mid\mathcal{D}_T)$;
(ii) the exact reverse KL divergence is $D_{\mathrm{KL}}(q_{\mathrm{Lap}}\,\|\,p_{\mathrm{exact}}) = \log \mathbb{E}_{q_{\mathrm{Lap}}}[\exp(\Delta_L(\mathbf{w}))] - \mathbb{E}_{q_{\mathrm{Lap}}}[\Delta_L(\mathbf{w})]$; and
(iii) by Pinsker's inequality, for every downstream candidate decision event $H_x \in \{0,1\}$,
\begin{align*}
    \big|P_{q_{\mathrm{Lap}}}(H_x=1\mid\mathcal{D}_T) - P_{p_{\mathrm{exact}}}(H_x=1\mid\mathcal{D}_T)\big|
    &\le d_{\mathrm{TV}}(q_{\mathrm{Lap}}, p_{\mathrm{exact}}) \\
    &\le \sqrt{\tfrac{1}{2}\big(\log \mathbb{E}_{q_{\mathrm{Lap}}}[e^{\Delta_L(\mathbf{w})}] - \mathbb{E}_{q_{\mathrm{Lap}}}[\Delta_L(\mathbf{w})]\big)}.
\end{align*}
\end{proposition}

\paragraph{Proof.}
\emph{Part (i):} By the chain rule, $p(\mathbf{w},\mathcal{D}_T) = p_0(\mathbf{w})\prod_{t=1}^T \pi_t(x_t\mid\mathcal{D}_{t-1}) p(y_t,c_t\mid x_t,\mathbf{w})$. Because $\pi_t(x_t\mid\mathcal{D}_{t-1})$ does not depend on $\mathbf{w}$, it factors out of the marginal integral $p(\mathcal{D}_T)=\int p(\mathbf{w},\mathcal{D}_T)\,d\mathbf{w}$ and cancels identically in $p_{\mathrm{exact}}(\mathbf{w}\mid\mathcal{D}_T) \propto p_0(\mathbf{w})\prod_{t=1}^T p(y_t,c_t\mid x_t,\mathbf{w})$.
\emph{Part (ii):} Writing $p_{\mathrm{exact}}(\mathbf{w}\mid\mathcal{D}_T) = Z^{-1} q_{\mathrm{Lap}}(\mathbf{w})\exp(\Delta_L(\mathbf{w}))$ with $Z = \mathbb{E}_{q_{\mathrm{Lap}}}[\exp(\Delta_L(\mathbf{w}))]$, we obtain $D_{\mathrm{KL}}(q_{\mathrm{Lap}}\,\|\,p_{\mathrm{exact}}) = \mathbb{E}_{q_{\mathrm{Lap}}}[\log(Z e^{-\Delta_L(\mathbf{w})})] = \log \mathbb{E}_{q_{\mathrm{Lap}}}[\exp(\Delta_L(\mathbf{w}))] - \mathbb{E}_{q_{\mathrm{Lap}}}[\Delta_L(\mathbf{w})] \ge 0$ (strictly zero iff $\Delta_L$ is constant, i.e., exact Gaussianity).
\emph{Part (iii):} Since $P_q(H_x=1) = \mathbb{E}_{q}[P(H_x=1\mid\mathbf{w})]$ with integrand $P(H_x=1\mid\mathbf{w})\in[0,1]$, variational characterization of total variation plus Pinsker's inequality gives $|P_{q_{\mathrm{Lap}}}(H_x=1) - P_{p_{\mathrm{exact}}}(H_x=1)| \le d_{\mathrm{TV}}(q_{\mathrm{Lap}}, p_{\mathrm{exact}}) \le \sqrt{\frac{1}{2} D_{\mathrm{KL}}(q_{\mathrm{Lap}}\,\|\,p_{\mathrm{exact}})}$. The displayed identity is exact if the log moment-generating function exists; Monte Carlo estimates of its expectation are not automatically upper confidence bounds without additional concentration or tail control. The identity is audited numerically in \cref{app:decisive-evsi-multiprotein}. \(\square\)

\subsection{Target-region probabilities and correlation-sensitive decisions}\label{app:poma-reference-local-ranking}

Let $\mathcal R=\prod_{k\in S}[a_k,b_k]$ specify a joint property requirement. Under a Gaussian factor model, conditioning on the latent factor renders the measurement coordinates independent, so the probability of feasibility can be evaluated with an integral over the factor rather than a full-dimensional Gaussian indicator. This is useful both for acquisition and for generation.

In the general latent-factor representation, write $r(x)=\mathbb E_{\eta\mid x,\mathcal D}[h_x(\eta)]$, where conditional independence gives
\begin{equation}
    h_x(\eta)
    =\prod_k\Pr(z_k\in[a_k,b_k]\mid x,\eta,\mathcal D).
    \label{eq:main-event}
\end{equation}
Gaussian factors permit evaluation with scalar normal CDFs; other families may require additional latent variables. In the Gaussian case with a standard normal factor this yields the following explicit expression.

\begin{equation}
  \begin{aligned}
    &\log P_\theta(\mathbf{z}_S\in\mathcal{R}\mid X) = \log \mathbb{E}_{\boldsymbol{\eta}\sim\mathcal{N}(\mathbf{0},I_r)}\!\Biggl[\prod_{k\in S} \biggl( \\[-4pt]
    &\quad \Phi\!\Big(\tfrac{b_k-\mu_k-\mathbf{u}_k^\top\boldsymbol{\eta}}{\sigma_k}\Big) - \Phi\!\Big(\tfrac{a_k-\mu_k-\mathbf{u}_k^\top\boldsymbol{\eta}}{\sigma_k}\Big)\biggr)\Biggr].
  \end{aligned}
  \label{eq:poma-coupled-region}
\end{equation}
Equation~\eqref{eq:poma-coupled-region} is the \emph{Gaussian} event probability. For a multivariate Student-$t$ head with shared scale mixture $\tau\sim\mathrm{Gamma}(\nu/2,\nu/2)$ and $\boldsymbol{\eta}\sim\mathcal{N}(0,I_r)$, the event probability requires an outer expectation over $\tau$ with CDF arguments $(\sqrt{\tau}(b_k-\mu_k)-\mathbf{u}_k^\top\boldsymbol{\eta})/\sigma_k$ and $(\sqrt{\tau}(a_k-\mu_k)-\mathbf{u}_k^\top\boldsymbol{\eta})/\sigma_k$. All Gaussian-only decision and finite-sample theorems below assume the Gaussian head or a separately justified Gaussian moment approximation, whereas conditional Student-$t$ Schur updates remain exact (\cref{prop:app-mvt-schur}).

\paragraph{Derivation of Eq.~(\ref{eq:poma-coupled-region}).}
Under $\mathbf{z}_S = \boldsymbol{\mu}_S + U_S \boldsymbol{\eta} + \boldsymbol{\varepsilon}_S$
with $\boldsymbol{\eta} \sim \mathcal{N}(\mathbf{0}, I_r)$ and
$\boldsymbol{\varepsilon}_S \sim \mathcal{N}(\mathbf{0}, D_S)$, conditioning on
$\boldsymbol{\eta}$ renders $\{z_k\}_{k \in S}$ mutually independent with
$z_k \mid \boldsymbol{\eta} \sim \mathcal{N}(\mu_k + \mathbf{u}_k^\top \boldsymbol{\eta}, \sigma_k^2)$.
By the law of total expectation,
\[
    P_\theta\!\left(\mathbf{z}_S \in \prod_{k \in S}[a_k, b_k] \,\middle|\, X\right)
    =
    \mathbb{E}_{\boldsymbol{\eta} \sim \mathcal{N}(\mathbf{0}, I_r)}
    \!\left[
        \prod_{k \in S} P_\theta(a_k \le z_k \le b_k \mid \boldsymbol{\eta}, X)
    \right],
\]
which yields Eq.~(\ref{eq:poma-coupled-region}) via standard normal CDF differences
and has no greater Monte Carlo variance than direct indicator sampling in
$\mathbb{R}^{|S|}$ by the Rao--Blackwell theorem.

Thompson sampling~\citep{thompson1933sampling} draws a shared parameter vector $\mathbf w^{(s)}\sim\mathcal N(\widehat{\mathbf w}_{\mathrm{post}},\Sigma_W)$ and evaluates the feasibility probability at $\boldsymbol\mu^{(s)}(X)=\Phi(X)\mathbf w^{(s)}$. It is thus possible to explore parameter uncertainty while integrating measurement noise in the feasibility calculation. The next result formalizes why a product of marginal feasibility probabilities can induce a different ordering of candidates from their joint probability.

\begin{theorem}[Correlation-sensitive target-region acquisition and ranking reversal]
\label{thm:correlation-acquisition}
For $\mathbf{Z}(X)\mid\mathcal{D}\sim\mathcal{N}(\hat{\boldsymbol{\mu}}(X),\Sigma(X))$ with marginal-matching diagonal $\Sigma_{\mathrm{diag}}(X)$ and correlation matrix $R(X)=\Sigma_{\mathrm{diag}}(X)^{-1/2}\Sigma(X)\Sigma_{\mathrm{diag}}(X)^{-1/2}$, compare coupled feasibility $p_{\mathcal{R}}(X)=\Pr(\mathbf{Z}(X)\in\mathcal{R}\mid\mathcal{D})$ against $p_{\mathrm{ind}}(X)=\prod_{k=1}^K\Pr(Z_k(X)\in[a_k,b_k]\mid\mathcal{D})$:
(i) for every measurable $\mathcal{R}\subseteq\mathbb{R}^K$, writing $p_{\mathrm{ind}}(X)$ for the probability of the same event under the marginal-matched diagonal Gaussian (a product for rectangular events), $|p_{\mathcal{R}}(X)-p_{\mathrm{ind}}(X)|\le \sqrt{-\frac{1}{4}\log\det R(X)}$; and
(ii) on $\mathcal{R}=[\tau_1,\infty)\times[\tau_2,\infty)$ with standardized margins $z_k(X)=(\hat{\mu}_k(X)-\tau_k)/\Sigma_{kk}(X)^{1/2}$, whenever candidate $A$ has higher marginals $m_A=z_1(A)=z_2(A)<0$ than candidate $B$ ($m_B<m_A<0$) but stronger negative trade-off correlation $\rho_A<\rho_{\mathrm{crit}}(m_A,m_B,\rho_B)\le\rho_B$, diagonal acquisition strictly ranks $p_{\mathrm{ind}}(A)>p_{\mathrm{ind}}(B)$ while true feasibility reverses order: $p_{\mathcal{R}}(B)>p_{\mathcal{R}}(A)$.
\end{theorem}

\paragraph{Proof of Theorem~\ref{thm:correlation-acquisition} (Correlation-sensitive feasibility bound and decision-ranking reversal).}
\emph{Part (i):} Let $P = \mathcal{N}(\hat{\boldsymbol{\mu}}, \Sigma)$ and $Q = \mathcal{N}(\hat{\boldsymbol{\mu}}, \Sigma_{\mathrm{diag}})$ share identical marginals, so $p_{\mathcal{R}}(X) = P(\mathcal{R})$ and $p_{\mathrm{ind}}(X) = Q(\mathcal{R})$ for any product box $\mathcal{R} = \prod_{k=1}^K [a_k, b_k]$ (and for arbitrary measurable $\mathcal{R}\subseteq\mathbb{R}^K$, $|P(\mathcal{R}) - Q(\mathcal{R})| \le d_{\mathrm{TV}}(P, Q)$ by definition of total variation distance). Because $\mathrm{tr}(\Sigma_{\mathrm{diag}}^{-1}\Sigma) = \sum_{k=1}^K \Sigma_{kk}/\Sigma_{kk} = K$, the Gaussian KL divergence simplifies to the multivariate total correlation:
\begin{align*}
    D_{\mathrm{KL}}(P \,\|\, Q)
    &= \frac{1}{2}\!\left[\mathrm{tr}(\Sigma_{\mathrm{diag}}^{-1}\Sigma) - K + \log\frac{\det\Sigma_{\mathrm{diag}}}{\det\Sigma}\right] \\
    &= -\frac{1}{2}\log\det\!\big(\Sigma_{\mathrm{diag}}^{-1/2}\Sigma\Sigma_{\mathrm{diag}}^{-1/2}\big)
    = -\frac{1}{2}\log\det R(X).
\end{align*}
Applying Pinsker's inequality $d_{\mathrm{TV}}(P, Q) \le \sqrt{\frac{1}{2} D_{\mathrm{KL}}(P \,\|\, Q)}$ yields \[
 |p_{\mathcal{R}}(X) - p_{\mathrm{ind}}(X)| \le \sqrt{-\tfrac14\log\det R(X)}.
\]

\emph{Part (ii):} Standardize each coordinate via $W_k = (Z_k - \hat{\mu}_k)/\Sigma_{kk}^{1/2}$, so that $(W_1, W_2)$ is standard bivariate normal with correlation $\rho \in (-1, 1)$ and $p_{\mathcal{R}}(X) = \Psi(m_1, m_2; \rho) \coloneqq \Pr(W_1 \ge -m_1, W_2 \ge -m_2) = \Phi_2(m_1, m_2; \rho)$. Under diagonal acquisition ($\rho=0$) with symmetric margins $m_X = z_1(X) = z_2(X) < 0$, $p_{\mathrm{ind}}(X) = \Phi(m_X)^2$. Since $\Phi$ is strictly increasing and positive, $m_A > m_B$ implies $p_{\mathrm{ind}}(A) = \Phi(m_A)^2 > \Phi(m_B)^2 = p_{\mathrm{ind}}(B)$ for \emph{all} $(\rho_A, \rho_B)$. Conversely, by Plackett's identity~\citep{kotz2004multivariate}, $\frac{\partial}{\partial\rho}\Phi_2(m, m; \rho) = \frac{1}{2\pi\sqrt{1-\rho^2}}\exp\!\big(-\frac{m^2}{1+\rho}\big) > 0$, so $\rho \mapsto \Phi_2(m, m; \rho)$ is continuous and strictly increasing on $(-1, 1)$, with $\lim_{\rho \downarrow -1} \Phi_2(m_A, m_A; \rho) = \max(0, 2\Phi(m_A) - 1) = 0$ whenever $m_A < 0$ (since $\Phi(m_A) < 1/2$). Meanwhile, for any $m_B \in (-\infty, m_A)$ and $\rho_B \in (-1, 1)$, $\Phi_2(m_B, m_B; \rho_B) > 0$; in particular, for $\rho_B \ge 0$, $\Phi_2(m_B, m_B; \rho_B) \ge \Phi(m_B)^2 > 0$. By the intermediate value theorem and strict monotonicity in $\rho_A$, whenever $\Phi_2(m_A, m_A; \rho_B) > \Phi_2(m_B, m_B; \rho_B)$ there exists a unique threshold $\rho_{\mathrm{crit}}(m_A, m_B, \rho_B) \in (-1, \rho_B)$ such that for all $\rho_A \in (-1, \rho_{\mathrm{crit}})$, $p_{\mathcal{R}}(A) = \Phi_2(m_A, m_A; \rho_A) < \Phi_2(m_B, m_B; \rho_B) = p_{\mathcal{R}}(B)$, reversing the candidate ranking. \(\square\)

\subsection{Decision-specific value of information and replica covariance}

Feasibility probabilities can change after a new assay is observed. A policy that values the decision rather than the total information in an observation should account for how this change affects the best candidate. The following theorem uses standard conditional-expectation and Brier-risk identities to express the covariance of updated feasibility probabilities through conditionally independent replicas. Its decision-specific content is the bound on the expected improvement of the best posterior candidate in part~(ii), with the Gaussian replica construction providing a tractable specialization.

\begin{theorem}[Decision-specific Value of Information and replica covariance bound]
\label{thm:v1-decision-voi}
Conditional on current observations $\mathcal{D}$, let $\mathcal C$ be a finite candidate set with targets $\mathbf{Z}_i$ and a proposed assay measurement $Y_a$ under a coherent joint law with $p_i=\Pr(\mathbf{Z}_i\in\mathcal{R}\mid\mathcal{D})$, $p_i(Y_a)=\Pr(\mathbf{Z}_i\in\mathcal{R}\mid\mathcal{D},Y_a)$, $i_0\in\arg\max_i p_i$, $p_0=p_{i_0}$, and gaps $g_i=p_0-p_i\ge 0$.
(i) \emph{Standard conditional-expectation identities and Gaussian specialization.} The expected Brier-risk reduction for candidate $i$ equals the conditional-replica identity $\mathcal{V}_{\mathrm{Br}}(Y_a)=\mathrm{Var}(p_i(Y_a))=\Pr(\mathbf{Z}_i^{(1)}\!\in\!\mathcal{R},\mathbf{Z}_i^{(2)}\!\in\!\mathcal{R})-p_i^2\le p_i(1-p_i)$, and the posterior-probability covariance matrix $B=\mathrm{Cov}((p_i(Y_a))_i)$ has entries $B_{ij}=\Pr(\widehat{\mathbf{Z}}_i\!\in\!\mathcal{R},\widehat{\mathbf{Z}}_j\!\in\!\mathcal{R})-p_ip_j$ for two distinct replica draws from the respective conditional target laws given the same assay observation, including when $i=j$. In the jointly Gaussian case with nonsingular assay covariance $S=\mathrm{Cov}(Y_a)$ and $C_i=\mathrm{Cov}(\mathbf Z_i,Y_a)$, their cross-covariance is $K_{ij}=C_iS^{-1}C_j^\top$.
(ii) Setting $s_i^2=B_{ii}+B_{i_0 i_0}-2B_{i i_0}$, the decision-specific value of observing assay $a$ obeys
\begin{equation}
  \begin{aligned}
    0 \;\le\; \mathrm{VOI}(a) &\coloneqq \mathbb{E}_{Y_a\mid\mathcal{D}}\!\Big[\max_{i\in\mathcal{C}} p_i(Y_a)\Big] - p_0 \\
    &\le \min\!\Big\{1-p_0,\; \tfrac{1}{2}\sum_{i\neq i_0}\big(\sqrt{g_i^2+s_i^2}-g_i\big)\Big\},
  \end{aligned}
  \label{eq:v1-voi-bound}
\end{equation}
yielding both an exact decision-targeted acquisition score $\mathrm{VOI}(a)/\mathrm{cost}(a)$ and a computable branch-and-bound certificate pruning any assay whose upper bound per cost cannot beat the incumbent action.
\end{theorem}

\paragraph{Proof of Theorem~\ref{thm:v1-decision-voi} (Decision-specific VOI and replica covariance upper bound).}
\emph{Part (i):} These are standard identities. Suppressing the fixed $\mathcal D$, conditional independence of two replica draws and the tower property give
\[
 \Pr(\widehat{\mathbf Z}_i\in\mathcal R,\widehat{\mathbf Z}_j\in\mathcal R)
 =\mathbb E[p_i(Y_a)p_j(Y_a)],\qquad \mathbb E[p_i(Y_a)]=p_i.
\]
Subtracting $p_ip_j$ gives $B_{ij}$, including the variance when $i=j$ and the replicas are distinct draws. The Brier-risk reduction is the law of total variance for the feasibility indicator. In the Gaussian specialization, $(Z_i,Y_a)$ is jointly Gaussian with $\mathrm{Cov}(Z_i)=V_i$, $\mathrm{Cov}(Y_a)=S\succ 0$, and $\mathrm{Cov}(Z_i, Y_a)=C_i$. Construct conditionally independent replicas $Z_i^{(j)}=\mu_i + C_i S^{-1}(Y_a^\star - \nu_a) + \epsilon_j$ ($j=1,2$) with independent $\epsilon_1,\epsilon_2\sim\mathcal{N}(0, V_i - C_i S^{-1} C_i^\top)$ and fresh $Y_a^\star\sim\mathcal{N}(\nu_a, S)$. Their joint covariance is $\begin{psmallmatrix}V_i & K_{ii}\\ K_{ii} & V_i\end{psmallmatrix}$ ($K_{ii}=C_i S^{-1} C_i^\top$), and conditional independence given $Y_a^\star$ yields $\Pr(Z_i^{(1)}\in\mathcal{R}, Z_i^{(2)}\in\mathcal{R})=\mathbb{E}[p_i(Y_a)^2]$. Since pre-assay optimal Brier risk is $p_i(1-p_i)$ and expected post-assay Brier risk is $\mathbb{E}[p_i(Y_a)(1-p_i(Y_a))]=p_i-\mathbb{E}[p_i(Y_a)^2]$, their difference is $\mathcal{V}_{\mathrm{Br}}(Y_a)=\mathrm{Var}(p_i(Y_a))\le p_i(1-p_i)$. The same common-$Y_a^\star$ construction for $(i,j)$ yields $B_{ij}=\mathrm{Cov}(p_i(Y_a), p_j(Y_a))=\Pr(\widehat{Z}_i\in\mathcal{R},\widehat{Z}_j\in\mathcal{R})-p_i p_j$ with replica cross-covariance $K_{ij}=C_i S^{-1} C_j^\top$.
\emph{Part (ii):} Define $X_i = (p_i(Y_a)-p_i) - (p_{i_0}(Y_a)-p_0)$. By the tower property, $\mathbb{E}[X_i]=0$ and $\mathbb{E}[X_i^2]=B_{ii}+B_{i_0 i_0}-2B_{i i_0}=s_i^2$. Rewriting $\mathrm{VOI}(a)=\mathbb{E}[\max\{0,\max_{i\neq i_0}(X_i-g_i)\}]\le \sum_{i\neq i_0}\mathbb{E}[(X_i-g_i)_+]$ and applying $t_+=(|t|+t)/2$ with Cauchy--Schwarz gives $\mathbb{E}[(X_i-g_i)_+]=\frac{1}{2}(\mathbb{E}|X_i-g_i|-g_i)\le\frac{1}{2}(\sqrt{g_i^2+s_i^2}-g_i)$. Combining with $\max_i p_i(Y_a)\le 1$ yields Eq.~\eqref{eq:v1-voi-bound}. \(\square\)

The bound is an upper bound on one-step model-relative expected value of information. Its differential variance $s_i^2$ measures changes in relative posterior feasibility under the same prospective assay outcome. If all feasibility changes are shared, then $s_i^2=0$ for every alternative and the one-step value is zero; information about absolute feasibility can still be useful for other decision objectives. For positive assay costs, an upper bound per cost below a valid incumbent lower bound suffices for pruning. If the quantities are evaluated approximately, additional numerical error control is required to turn the implemented comparison into a certificate. The resulting criterion is not an optimality guarantee for a multistep experimental policy. The experimental pruning audit in \cref{app:supp-decisions} addresses the reported numerical setting.

\subsection{Decision regret and correlation-aware optimality certification}\label{app:decision-regret}

The value-of-information bound concerns the improvement attainable by another experiment. A complementary question arises when experimental acquisition ends: how much utility might be lost by selecting the candidate currently judged best? Two distinct sources of uncertainty matter. The predictive posterior may differ from the physical data-generating process, and its event probabilities may be approximated computationally. The following result separates these effects for a fixed candidate set.

\begin{proposition}[Model-relative and physical decision-regret decomposition]
\label{thm:finite-sample-regret}
Let $\mathcal C$ be a \emph{fixed} finite candidate set of size $N_{\mathcal C}$. For each $x$ define a real physical success probability $p^*(x)$, the full Gaussian posterior-predictive success probability $p_{\mathrm{post}}(x)$ with covariance $\Sigma_{\mathrm{ale}}(x)+\Phi(x)\Sigma_W\Phi(x)^\top$, and its rank-$r'$ truncated counterpart $p_{r'}(x)$. Let $\widehat p_{r',M}$ be the average of $M$ independent Rao--Blackwellized $[0,1]$ factors per candidate, and $\widehat x\in\arg\max_{x\in\mathcal C}\widehat p_{r',M}(x)$. Set $\varepsilon_{\mathrm{model}}=\max_x|p^*(x)-p_{\mathrm{post}}(x)|$ and $D_{\max}=\max_xD_{\mathrm{KL}}(\mathcal N(\widehat\mu_x,\widetilde\Sigma_{x,r'})\,\|\,\mathcal N(\widehat\mu_x,\Sigma_{x,\mathrm{post}}))$. With probability at least $1-\delta$ over the independent integration samples,
\begin{equation}
 \max_{x\in\mathcal C}p^*(x)-p^*(\widehat x)
 \le 2\varepsilon_{\mathrm{model}}+\sqrt{2D_{\max}}
   +2\sqrt{\frac{\log(2N_{\mathcal C}/\delta)}{2M}}.
 \label{eq:finite-sample-regret}
\end{equation}
The KL term is bounded by the spectral-tail bound in \cref{thm:joint-candidate-assay-closure}(iii). The first term is \emph{not} determined by the fitted posterior; it requires a correctly specified data-generating model with an independent statistical guarantee or external calibration/validation. Dropping it yields a model-relative regret bound with $p_{\mathrm{post}}$ as the decision target. The theorem does not promise a small error for unobserved physical experiments.
\end{proposition}

\paragraph{Proof of Proposition~\ref{thm:finite-sample-regret} (model-relative and physical regret).}
Let $p_{\mathrm{post}}$, $p_{r'}$, and $\widehat p_{r',M}$ be the three success-probability functions in the statement. Because $\widehat x$ maximizes $\widehat p_{r',M}$ on the fixed finite set, the usual two-sided uniform-error inequality gives
\[
 \max_xp^*(x)-p^*(\widehat x)\le 2\max_x|p^*(x)-\widehat p_{r',M}(x)|.
\]
The triangle inequality bounds its right-hand side by twice the sum of the maximum model discrepancy, truncation discrepancy, and Monte Carlo discrepancy. For each $x$, the event $\mathcal R$ is measurable, so Pinsker's inequality gives $|p_{\mathrm{post}}(x)-p_{r'}(x)|\le\sqrt{D_{\mathrm{KL}}(\widetilde P_x\,\|\,P_{\mathrm{post},x})/2}$. Twice this error is at most $\sqrt{2D_{\max}}$. Theorem~\ref{thm:joint-candidate-assay-closure}(iii) bounds $D_{\max}$ through the discarded posterior covariance eigenvalues. The independent $[0,1]$ Rao--Blackwellized factors obey Hoeffding's inequality; a union bound over the $N_{\mathcal C}$ candidates gives, with probability $1-\delta$, $\max_x|\widehat p_{r',M}(x)-p_{r'}(x)|\le\sqrt{\log(2N_{\mathcal C}/\delta)/(2M)}$. Combining proves the displayed result. No assertion about the magnitude of $\varepsilon_{\mathrm{model}}$ follows without additional physical or correctly specified statistical assumptions. \(\square\)

Uniform approximation bounds can be conservative even when the estimated utilities separate the leading candidate from its competitors. The shared-posterior covariance developed above suggests a more local analysis: uncertainty common to two candidates need not affect their relative ordering. Under a jointly Gaussian utility model, the variance of each utility difference is determined by the two marginal variances and their cross-covariance. This is analogous to the differential uncertainty in the value-of-information bound, but here the random quantities are candidate utilities, not posterior feasibility changes under a prospective observation. The Gaussian utility and deterministic approximation-slack assumptions are explicit; feasibility probabilities themselves are not generally Gaussian.

\begin{proposition}[Decision-local post-selection regret and pairwise optimality bound]
\label{prop:app-v47-decision-local}
Conditional on observed data $\mathcal{D}$, let candidate action utilities $\mathbf{U}=(U_1,\ldots,U_M)^\top$ have jointly Gaussian posterior $\mathcal{N}(\hat{\mathbf{u}},B)$ plus deterministic local approximation slack $|U_j^{\mathrm{true}}-U_j|\le \delta_j$ on the top-$M$ contenders, let $i^\star\in\arg\max_j \hat{u}_j$, and define empirical margins $\Delta_j = \hat{u}_{i^\star}-\hat{u}_j \ge 0$, slack-adjusted gaps $\tilde{\Delta}_j = \Delta_j - (\delta_{i^\star}+\delta_j)$, and exact pairwise differential standard deviations $\sigma_{\Delta,j} = \sqrt{\max(B_{i^\star i^\star}+B_{jj}-2B_{i^\star j},0)}$. Then:
\begin{enumerate}[leftmargin=1.5em,itemsep=2pt]
  \item \textbf{Decision-local expected-regret certificate:}
  \begin{equation}
    \mathbb{E}\!\Big[\max_{j\in[M]} U_j^{\mathrm{true}} - U_{i^\star}^{\mathrm{true}} \;\Big|\; \mathcal{D}\Big]
    \;\le\; \sum_{j\neq i^\star} \Big[\sigma_{\Delta,j}\,\phi\!\Big(\frac{\tilde{\Delta}_j}{\sigma_{\Delta,j}}\Big) - \tilde{\Delta}_j\,\Phi\!\Big(-\frac{\tilde{\Delta}_j}{\sigma_{\Delta,j}}\Big)\Big],
    \label{eq:app-v47-local-regret}
  \end{equation}
  where the summand is defined as $(-\tilde\Delta_j)_+$ when $\sigma_{\Delta,j}=0$. For fixed positive $\sigma_{\Delta,j}$, it decays as the positive adjusted gap grows. Positive shared covariance $B_{i^\star j}$ reduces the differential variance through the term $-2B_{i^\star j}$.
  \item \textbf{Pairwise and union optimality certification:} Define $\beta_j=\Phi(-\tilde\Delta_j/\sigma_{\Delta,j})$ when $\sigma_{\Delta,j}>0$, and $\beta_j=\mathbb I\{\tilde\Delta_j<0\}$ when $\sigma_{\Delta,j}=0$. Action $i^\star$ is no worse than rival $j$ with posterior probability at least $1-\beta_j$, and optimal (allowing ties) over all top-$M$ contenders with probability at least $\max\{0,1-\sum_{j\ne i^\star}\beta_j\}$.
\end{enumerate}
\end{proposition}
\begin{proof}
Since $\max_{j} U_j^{\mathrm{true}} - U_{i^\star}^{\mathrm{true}} = \max_{j\neq i^\star}(U_j^{\mathrm{true}}-U_{i^\star}^{\mathrm{true}})_+ \le \sum_{j\neq i^\star}(U_j - U_{i^\star} + \delta_j + \delta_{i^\star})_+$, let $Z_j = U_j - U_{i^\star} + \delta_j + \delta_{i^\star} \sim \mathcal{N}(-\tilde{\Delta}_j,\sigma_{\Delta,j}^2)$. For $\sigma_{\Delta,j}>0$, direct Gaussian integration gives $\mathbb{E}[(Z_j)_+] = \sigma_{\Delta,j}\phi(\tilde{\Delta}_j/\sigma_{\Delta,j}) - \tilde{\Delta}_j\Phi(-\tilde{\Delta}_j/\sigma_{\Delta,j})$. For $\sigma_{\Delta,j}=0$, $Z_j=-\tilde\Delta_j$ deterministically and the expectation is $(-\tilde\Delta_j)_+$. This establishes \cref{eq:app-v47-local-regret}. Part (2) follows from $\Pr(U_j^{\mathrm{true}}>U_{i^\star}^{\mathrm{true}}\mid\mathcal{D})\le \Pr(Z_j>0\mid\mathcal{D})=\beta_j$ and a union bound over $j\neq i^\star$. \(\square\)
\end{proof}

The two bounds address complementary aspects of selection. The uniform decomposition isolates posterior approximation and numerical integration errors, while the local bound exploits utility separation and shared covariance. The latter applies only to the declared contenders and approximation slacks; it does not exclude an unexamined candidate or establish the accuracy of the physical measurement model. Numerical diagnostics are presented in \cref{app:decision-certification-audit}.

The same joint feasibility probabilities used for selection can also define a distribution over sequences. We next establish how conditional prediction and generation can use the same native output laws, with an explicit joint target for coherent sampling.

\subsection{Discrete--continuous composition and coherent sampling}

The unified discrete--continuous model retains the masked categorical token head. For pretrained residue logits $\mathbf z^{\mathrm{PLM}}_\ell(X_{\setminus\ell})\in\mathbb R^{20}$, partner antigen features $E_W$, and assay prompt $P_k$, the contextual residue score is

\begin{equation}
  \begin{aligned}
    &\log\pi_\theta(x_\ell=a\mid X_{\setminus\ell},W,P_k) = \log\mathrm{softmax}_a\!\Big( \\[-2pt]
    &\quad \mathbf{z}^{\mathrm{PLM}}_\ell(X_{\setminus\ell}) + \tanh(\gamma_k)\,g_\psi(H_X,E_W,E(P_k))_{\ell,a}\Big).
  \end{aligned}
  \label{eq:logica-poma-residue}
\end{equation}
Initializing $\gamma_k=0$ preserves exact pretrained language-model likelihoods at step $0$ while learning background-cancelled CDR log-likelihood ratios during fine-tuning.

For the AbRank pairwise prediction task, a continuous difference distribution provides probabilities of directional change and neutrality. Let $\Delta Z=Z_1-Z_2\sim\mathcal N(\Delta\mu,\sigma_\Delta^2)$, with dead zone $[-m_0,m_0]$ and impact threshold $m_+\ge m_0>0$. Conditional on the evaluated event $\mathcal E_{m_0,m_+}=\{|\Delta Z|\le m_0\}\cup\{|\Delta Z|\ge m_+\}$, the three class probabilities are normalized as $P_\theta(c)=P_0(c)/\sum_{c'}P_0(c')$, where

\begin{equation}
  \begin{aligned}
    P_0(+1) &= 1-\Phi\!\Big(\tfrac{m_+-\Delta\mu}{\sigma_\Delta}\Big), \\
    P_0(-1) &= \Phi\!\Big(\tfrac{-m_+-\Delta\mu}{\sigma_\Delta}\Big), \\
    P_0(0) &= \Phi\!\Big(\tfrac{m_0-\Delta\mu}{\sigma_\Delta}\Big)-\Phi\!\Big(\tfrac{-m_0-\Delta\mu}{\sigma_\Delta}\Big).
  \end{aligned}
  \label{eq:abrank-3state-integral}
\end{equation}

These probabilities inherit the symmetry of the difference distribution when the pair is swapped. More generally, prediction and generation differ in how the conditional law is used: its summaries give predictions, whereas its draws give possible realizations. The structured Gaussian head allows all unobserved measurements to be drawn jointly using \cref{eq:main-schur}, rather than by iterating scalar Gibbs updates. Alternating such draws with masked-token updates gives a mixed discrete--continuous Gibbs-style procedure (\cref{eq:app-joint-xy-gibbs}). However, masked token conditionals need not be compatible with a joint sequence law. We therefore fix a proper strictly positive prior $p_0$ over a finite supported editing space, or normalize an explicitly evaluable positive weight $w(X)$. The conditional design law is $\pi_\theta(X,\mathbf z_A\mid\mathbf z_B^*,W)\propto p_0(X\mid W)q_\theta(\mathbf z_A,\mathbf z_B^*\mid X,W)$. Under this joint law, the Schur measurement draw is an exact Gibbs block; contextualized token distributions instead supply proposals for a Hastings-corrected sequence update. The distinct event-conditioned sequence law is $\pi_{\mathcal R}(X)\propto w(X)r(X)$. The following theorem establishes target-invariant kernels for both formulations.

\begin{theorem}[Coherent MH, pseudo-marginal and collapsed factor-augmented sampling]
\label{thm:mh-joint-invariance}
Let $p_0(X\mid W)$ be a proper prior over finite sequence space $\mathcal{X}$ and $q_\theta$ a coupled low-rank \method{} \emph{Gaussian} density with Rao--Blackwellized factor integrand $h_X(\boldsymbol{\eta})=\prod_{k\in S}[\Phi(\frac{b_k-\mu_k-\mathbf{u}_k^\top\boldsymbol{\eta}}{\sigma_k})-\Phi(\frac{a_k-\mu_k-\mathbf{u}_k^\top\boldsymbol{\eta}}{\sigma_k})]$.
(i) \emph{Blocked joint $(X,\mathbf{z}_A)$ MH:} alternating an exact Schur draw $\mathbf{z}_A^{(t+1)}\sim q_\theta(\mathbf{z}_A\mid X^{(t)},\mathbf{z}_B^*,W)$ with any strictly positive (possibly asymmetric or pseudo-likelihood-derived) proposal $X'\sim Q_{\mathrm{prop}}(\cdot\mid X^{(t)},\mathbf{z}_A^{(t+1)})$ accepted with ratio
\[
 1\wedge\frac{p_0(X')q_\theta(\mathbf{z}_A^{(t+1)},\mathbf{z}_B^*\mid X')Q_{\mathrm{prop}}(X^{(t)}\mid X',\mathbf{z}_A^{(t+1)})}{p_0(X^{(t)})q_\theta(\mathbf{z}_A^{(t+1)},\mathbf{z}_B^*\mid X^{(t)})Q_{\mathrm{prop}}(X'\mid X^{(t)},\mathbf{z}_A^{(t+1)})}
\]
preserves $\pi_\theta(X,\mathbf{z}_A\mid\mathbf{z}_B^*,W)$ exactly.

(ii) \emph{Pseudo-marginal MH:} for $\pi_{\mathcal{R}}(X)\propto w(X)r(X)$, evaluating
\[
 \widehat{r}_m(X',u')=\frac{1}{m}\sum_{j=1}^m h_{X'}(\boldsymbol{\eta}_j')
\], accepting with
\[
 1\wedge\frac{w(X')\widehat{r}_m(X',u')Q(X\mid X')}{w(X)\widehat{r}_m(X,u)Q(X'\mid X)}
\], and \emph{retaining $(X,u)$ on rejection} preserves exact marginal $\pi_{\mathcal{R}}$ for every $m\ge 1$ (whereas independently refreshing $u$ on rejection generally changes the stationary law).

(iii) \emph{Collapsed factor augmentation:} alternating an independence refresh $\boldsymbol{\eta}'\sim\mathcal{N}(0,I_r)$ accepted with $1\wedge h_X(\boldsymbol{\eta}')/h_X(\boldsymbol{\eta})$ and a fixed-$\boldsymbol{\eta}$ sequence MH move accepted with
\[
 1\wedge\frac{w(X')h_{X'}(\boldsymbol{\eta})Q(X\mid X',\boldsymbol{\eta})}{w(X)h_X(\boldsymbol{\eta})Q(X'\mid X,\boldsymbol{\eta})}
\] preserves $\pi_{\mathcal{R}}$ using only scalar CDFs with no outer Monte Carlo integral.

(iv) \emph{Immediate push-forward consequence:} for a bimeasurable coordinate bijection $T$, pushing the observation law forward and replacing $\mathcal{R}$ by $T(\mathcal{R})$ leaves $r(X)$, $\pi_{\mathcal{R}}(X)$, and exact sequence-space MH acceptance ratios (for the same sequence proposal) invariant.
\end{theorem}

\paragraph{Proof of Theorem~\ref{thm:mh-joint-invariance} (Stationary invariance of MH-corrected joint $(X,\mathbf{z}_A)$ sampling).}
Let $\pi(X,\mathbf{z}_A) \coloneqq \pi_\theta(X,\mathbf{z}_A \mid \mathbf{z}_B^*, W) = Z_\pi^{-1} p_0(X\mid W)\,q_\theta(\mathbf{z}_A,\mathbf{z}_B^*\mid X,W)$, where the normalizing constant $Z_\pi = \sum_{X\in\mathcal{X}} p_0(X\mid W)\,q_\theta(\mathbf{z}_B^*\mid X,W) \in (0,\infty)$ is finite and positive because $\mathcal{X}$ is finite, $p_0$ is a proper mass function, and the nonsingular Gaussian observation densities are finite and positive at the fixed observation.
\emph{Block 1 (Schur continuous kernel $K_1$):} By Bayes' rule, the conditional density of $\mathbf{z}_A$ given $(X,\mathbf{z}_B^*)$ under $\pi$ is $\pi(\mathbf{z}_A\mid X) = q_\theta(\mathbf{z}_A,\mathbf{z}_B^*\mid X,W)/q_\theta(\mathbf{z}_B^*\mid X,W) = q_\theta(\mathbf{z}_A\mid X,\mathbf{z}_B^*,W)$, which is the exact Schur-complement Gaussian drawn by $K_1$. Hence for any $(X,\mathbf{z}_A,\mathbf{z}_A')$, $\pi(X,\mathbf{z}_A)K_1(\mathbf{z}_A'\mid X,\mathbf{z}_A) = \pi(X)\pi(\mathbf{z}_A\mid X)\pi(\mathbf{z}_A'\mid X) = \pi(X,\mathbf{z}_A')K_1(\mathbf{z}_A\mid X,\mathbf{z}_A')$, establishing detailed balance and $\pi$-invariance of $K_1$.
\emph{Block 2 (MH discrete kernel $K_2$):} Holding $\mathbf{z}_A$ fixed, for any $X \neq X'$ in $\mathrm{supp}(p_0)$, the transition density is $K_2(X'\mid X,\mathbf{z}_A) = Q_{\mathrm{prop}}(X'\mid X,\mathbf{z}_A,\mathbf{z}_B^*,W)\,\alpha(X\to X')$. Since $Z_\pi$ cancels identically in the acceptance ratio $\alpha(X\to X') = \min\!\big(1, \frac{\pi(X',\mathbf{z}_A)Q_{\mathrm{prop}}(X\mid X',\cdot)}{\pi(X,\mathbf{z}_A)Q_{\mathrm{prop}}(X'\mid X,\cdot)}\big)$, we obtain
\begin{align*}
    \pi(X,\mathbf{z}_A)\,K_2(X'\mid X,\mathbf{z}_A)
    &= \min\!\big(\pi(X,\mathbf{z}_A)Q_{\mathrm{prop}}(X'\mid X,\cdot),\; \pi(X',\mathbf{z}_A)Q_{\mathrm{prop}}(X\mid X',\cdot)\big) \\
    &= \pi(X',\mathbf{z}_A)\,K_2(X\mid X',\mathbf{z}_A).
\end{align*}
Thus $K_2$ satisfies detailed balance with respect to $\pi(X,\mathbf{z}_A)$ regardless of whether $Q_{\mathrm{prop}}$ is asymmetric or derived from an unnormalized masked-PLM pseudo-likelihood (\textsc{LogiCA}). Consequently, the composite transition kernel $K = K_2 \circ K_1$ satisfies $\pi K = (\pi K_1) K_2 = \pi K_2 = \pi$. The strictly positive proposal in part~(i), together with positive Gaussian conditionals, gives an irreducible joint kernel. For proposals with restricted support, irreducibility instead requires the graph of moves with positive forward and reverse proposal probabilities to connect the target support; irreducibility of the uncorrected directed proposal alone is insufficient. Aperiodicity can be ensured by a lazy transition. These conditions concern convergence to the invariant law and do not provide a finite-time mixing rate. \(\square\)

\paragraph{Proof of Theorem~\ref{thm:mh-joint-invariance} (Parts ii--iv: pseudo-marginal, collapsed factor-augmented, and unit-invariant event sampling).}
\emph{Part (ii):} Conditionally on $\boldsymbol{\eta}\sim\mathcal{N}(0,I_r)$, the assay coordinates under $\Sigma(X)=D(X)+U(X)U(X)^\top$ are independent normals, so $\mathbb{E}[h_X(\boldsymbol{\eta})]=r(X)$ and $\widehat{r}_m(X,u)=\frac{1}{m}\sum_{j=1}^m h_X(\boldsymbol{\eta}_j)$ is strictly nonnegative, bounded in $[0,1]$, and unbiased with $\mathrm{Var}(\widehat{r}_m)\le r(X)(1-r(X))/m$. On the extended space $\widetilde{\Pi}_m(X,du)=Z_{\mathcal{R}}^{-1} w(X)\widehat{r}_m(X,u)\,M_X(du)$, the forward proposal $Q(X'\mid X)M_{X'}(du')$ paired with acceptance $1\wedge\frac{w(X')\widehat{r}_m(X',u')Q(X\mid X')}{w(X)\widehat{r}_m(X,u)Q(X'\mid X)}$ produces symmetric detailed-balance flux, leaving $\widetilde{\Pi}_m$ invariant with exact marginal $\int \widetilde{\Pi}_m(X,du)=Z_{\mathcal{R}}^{-1} w(X)r(X)=\pi_{\mathcal{R}}(X)$ for every $m\ge 1$. Retaining $(X,u)$ on rejection gives the stated detailed-balance argument; independently refreshing $u$ at state $X$ is not in general a valid update of this extended target (as verified by the two-state counterexample where stationary mass shifts from $1/2$ to $9/19$ and empirically in \cref{app:v43-voi-cov-sampling-camap}).
\emph{Part (iii):} For $m=1$, $\bar{\Pi}(X,d\boldsymbol{\eta})=Z_{\mathcal{R}}^{-1} w(X)\varphi_r(\boldsymbol{\eta})h_X(\boldsymbol{\eta})\,d\boldsymbol{\eta}$ has exact sequence marginal $\pi_{\mathcal{R}}(X)$. Holding $X$ fixed, the target conditional is proportional to $\varphi_r(\boldsymbol{\eta})h_X(\boldsymbol{\eta})$, so proposing $\boldsymbol{\eta}'\sim\varphi_r$ with acceptance $1\wedge h_X(\boldsymbol{\eta}')/h_X(\boldsymbol{\eta})$ satisfies detailed balance; holding $\boldsymbol{\eta}$ fixed, the sequence MH acceptance $1\wedge\frac{w(X')h_{X'}(\boldsymbol{\eta})Q(X\mid X',\boldsymbol{\eta})}{w(X)h_X(\boldsymbol{\eta})Q(X'\mid X,\boldsymbol{\eta})}$ likewise satisfies detailed balance, so their composition preserves $\bar{\Pi}$ and $\pi_{\mathcal{R}}$ using only scalar CDF evaluations $h_X(\boldsymbol{\eta})$.
\emph{Part (iv):} This is an immediate consequence of the definition of push-forward, not a separate sampling result. Writing $Q_X$ for the observation probability law and $T_\# Q_X$ for its push-forward, bimeasurability gives
\[
 (T_\# Q_X)(T(\mathcal R))
 =Q_X(T^{-1}(T(\mathcal R)))=Q_X(\mathcal R)=r(X).
\]
Thus the normalized sequence target and its exact sequence-space MH ratios for an unchanged proposal are identical. This does not assert that an arbitrary transformation preserves Gaussian structure or the cost of evaluating the target. \(\square\)

The result concerns stationary invariance. More generally, for any latent-factor law $m_x(d\eta)$ with $r(x)=\int h_x(\eta)m_x(d\eta)$, the augmented probability measure proportional to $w(x)h_x(\eta)m_x(d\eta)$ has the required sequence marginal. A valid Hastings transition on this measure preserves it; if $m_x$ depends on $x$, its contribution must be retained in the sequence acceptance ratio (or handled by an appropriate joint proposal). The Gaussian construction above is a tractable special case. The structure-adaptive proposals in \cref{app:v43-voi-cov-sampling-camap,app:pilot-gated-sampling} preserve these targets through correction. Quantitative finite-time guarantees under stronger audit assumptions are developed in \cref{app:poma-certified-adaptive}.

\clearpage
\section{Experimental methodology and reproducibility}\label{app:full-method}

The mathematical development defines the statistical interface and the assumptions underlying its inference operations. The experiments instantiate this interface with modality-specific encoders and observation models. To make comparisons interpretable, we separate changes to the learning objective from changes to the representation whenever the experimental design permits. We describe the shared data and model conventions here, with the details of specialized tasks given alongside their results.

\subsection{Benchmark design and biological coverage}\label{sec:methods-data}

Our experiments examine different uses of the same probabilistic output interface, from reconstruction of a single modality to the selection of biological candidates under an experimental budget. The benchmarks were chosen to span measurements with different supports, dimensions, noise characteristics and sources of distribution shift. No single architecture or likelihood family is assumed to be optimal across all these settings. \Cref{tab:supp-benchmark-overview} summarizes the roles of the datasets; the corresponding evaluation sections provide the detailed protocols and comparisons.

\begin{table}[!htbp]
\centering
\caption{Overview of the benchmark families and their role in evaluating \method{}. Individual protocols use the splits and budgets specified in the associated experimental sections.}
\label{tab:supp-benchmark-overview}
\small
\begin{tabularx}{\linewidth}{@{}p{.20\linewidth}p{.31\linewidth}X@{}}
\toprule
Domain & Data resources & Principal capabilities evaluated\\
\midrule
Small molecules & TDC ADMET; PMO & Distributional property prediction; sample-efficient optimization\\
Single-cell expression & BMMC; pretrained scGPT & Native masked reconstruction and calibration\\
Spectroscopy & QM9S; RamanBench; MassSpecGym; ProteomeTools / Prosit & Discrete--continuous alignment, censored likelihoods, spectral reconstruction and identification\\
Antibodies & GDPa1; Jain; AbRank; AIntibody; FLAb & Assay contextualization, transfer, feasibility, acquisition and design\\
Protein engineering & MTLMPO-2026 & Cross-system joint prediction and decision calibration\\
Multi-omics and regulatory DNA & CITE-seq; Multiome; CPTAC; lentiMPRA & Modality matching, partial-panel inference and regulatory design\\
\bottomrule
\end{tabularx}
\end{table}

The common experimental principle is to fit preprocessing and model parameters on training data, select hyperparameters and calibration settings without test access, and assess generalization on held-out objects or contexts. Where pretraining overlap could not be established for external checkpoints, we distinguish the downstream split from a verified decontamination guarantee. The data preparation and split conventions are specified next; optimization and compute settings are given in \cref{app:reproducibility}.

The representative comparisons in \cref{tab:main-breadth} provide an index to predictive and conditional benchmarks across domains. The rows correspond to distinct protocols, not a pooled comparison; the detailed experimental sections give the complete baseline sets and uncertainty estimates.

\begin{table}[!htbp]
\centering
\caption{\textbf{Representative held-out benchmarks across biological domains.}
Comparisons follow matched evaluation protocols within each benchmark. Protein results use the capped protocol. Bold indicates the numerically better reported score within a row, without implying statistical significance. Complete comparisons, uncertainty estimates, and protocols are provided in \cref{app:supp-masked,app:supp-inference,app:supp-bayesian}.}
\label{tab:main-breadth}
\small\setlength{\tabcolsep}{3pt}
\begin{tabular}{llllrr}
\toprule
Benchmark & Metric $\downarrow$ & Baseline method & POMA variant & Baseline & POMA\\
\midrule
GDPa1 (4-shot) & NLL & NTK-BLL & Laplace & 1.606 & \textbf{1.596}\\
Proteins (27 systems) & Joint NLL & 5$\times$ ensemble & Laplace & \textbf{1.150} & 1.254\\
 & Joint Brier & 5$\times$ ensemble & Laplace & 0.166 & \textbf{0.159}\\
QM9S & Joint NLL & LowRankMVN & Banded+LR & 0.005 & $\mathbf{-0.430}$\\
MassSpecGym & Cens. NLL & Indep. Tobit & Coupled Tobit & 0.167 & \textbf{0.134}\\
CITE-seq & Joint NLL & totalVI-style cVAE & MVN & 0.993 & \textbf{0.854}\\
lentiMPRA & Joint NLL & Indep. Gauss. & MVN & 1.068 & \textbf{0.222}\\
\bottomrule
\end{tabular}
\end{table}

\subsection{Data preparation, pretraining and held-out evaluation}

\paragraph{Small molecules.} The Therapeutics Data Commons (TDC) ADMET suite~\citep{huang2021therapeutics} contains 22 scaffold-split prediction tasks covering absorption, distribution, metabolism, excretion and toxicity. The experiments use 46,200 unique canonicalized compounds, with five scaffold seeds per task. RDKit canonical SMILES~\citep{rdkit} are used for deduplication. For the geometric comparisons, conformers are generated with ETKDGv3~\citep{wang2020etkdg} and minimized with MMFF94~\citep{halgren1996mmff}. The descriptor pretraining corpus consists of 48,382 compounds, comprising 32,382 non-test TDC molecules and 16,000 drug-like ZINC compounds~\citep{sterling2015zinc}, with 56 computed physicochemical, surface and pharmacophore descriptors. Test scaffolds are excluded from this corpus. The 23 Practical Molecular Optimization (PMO) tasks~\citep{gao2022sample} use a separate oracle-based evaluation with three seeds and a budget of 10,000 calls per task. Molecular proposals use SELFIES decoding~\citep{krenn2020selfies} or RDKit sanitization to maintain chemical validity.

\begin{table}[!htbp]
\centering
\caption{
\textbf{TDC ADMET benchmark overview ($22$ scaffold-split datasets across $46{,}200$ unique compounds).}
}
\label{tab:tdc-admet-summary}
\pomatabbox{
\begin{tabular}{llll}
\toprule
\textbf{Category} &
\textbf{Endpoints} &
\textbf{Types ($T_k$)} &
\textbf{Metrics} \\
\midrule
Absorption (6) & Caco-2, HIA, Pgp, Bioav., Lipo., AqSolDB & Reg.\ (YJ) / Bin. & MAE, ROC \\
Distribution (3) & BBB, PPBR, VDss & Reg.\ / Bin. & MAE, $\rho$, ROC \\
Metabolism (6) & CYP2C9/2D6/3A4 inhib.\ \& subst. & Binary (Cat.) & ROC, PR \\
Excretion (3) & Half-life, hep.\ \& mic.\ clearance & Reg.\ ($\log_{10}$) & Spearman $\rho$ \\
Toxicity (4) & $\mathrm{LD}_{50}$, hERG, Ames, DILI & Reg.\ (YJ) / Bin. & MAE, ROC \\
\bottomrule
\end{tabular}
}
\end{table}

\paragraph{Antibodies.} Clinical antibody assays are relatively scarce, motivating pretraining on a larger repertoire of paired heavy and light chains with computed descriptors. The decontaminated corpus contains 141,916 paired human sequences and 102 properties from sequence and structural calculations~\citep{bashour2024biophysical}. We exclude 18,099 sequences sharing at least 85\% $k$-mer or paired-Fv identity with clinical evaluation antibodies and remove coronavirus-associated sequences before AIntibody evaluation. The two clinical datasets are GDPa1 (246 therapeutic antibodies, 13 assays; \citealt{arsiwala2025high}) and Jain (137 antibodies, 12 assays; \citealt{jain2017biophysical}). Lineage and cross-laboratory comparisons use the specified hierarchical clonal-cluster partitions.

For antigen-conditioned prediction, AbRank~\citep{liu2026abrank} uses unseen-antigen and lineage splits (below 50\% antigen identity and 80\% clonal-lineage identity). AIntibody~\citep{erasmus2026aintibody} provides a sparse panel of 686 SARS-CoV-2 RBD antibodies, including 354 completely characterized candidates, with ten continuous assays and five categorical gates in Dataset~3; Dataset~2 supplies display sequencing counts. FLAb~\citep{chungyoun2024flab,chungyoun2025flab2,flabrepository} supplies independently measured mutational libraries for retrospective design. These measured libraries are not newly synthesized or assayed as part of our study.

\begin{table}[!htbp]
\centering
\caption{
\textbf{Decontaminated antibody pretraining corpus.}
All chains within $\ge 85\%$ identity of \textsc{GDPa1} or \citet{jain2017biophysical}
($18{,}099$ sequences) are removed prior to pretraining.
}
\label{tab:antibody-pretraining-data}
\pomatabbox{
\begin{tabular}{lccl}
\toprule
\textbf{Property family} &
\textbf{\# props} &
\textbf{\# mAbs} &
\textbf{Source / computation} \\
\midrule
Developability (\textsc{AbChain}) & 42 & 141,916 & Physicochemical \& solubility \citep{bashour2024biophysical} \\
3D biophysics (\textsc{AbStruc}) & 48 & 140,934 & Surface patches, SAP \& dipoles \citep{bashour2024biophysical} \\
Chemical liabilities & 11 & 141,916 & Deamidation, isomerization \& oxidation \\
Repertoire humanness & 1 & 141,916 & Human germline $k$-mer log-odds \\
\midrule
\textbf{Total (decontaminated)} & \textbf{102} & \textbf{141,916} & Unified prompt-conditioned matrix \\
\bottomrule
\end{tabular}
}
\end{table}

\paragraph{Protein engineering.} The MTLMPO-2026 benchmark~\citep{herrerarocha2026multitask} spans 27 datasets and 38,322 variants, covering enzymes, GPCRs, viral proteins and inhibitors (three splits per system; data DOI \texttt{10.5281/zenodo.21512415}). We distinguish a uniformly capped protocol, with at most 1,000 variants per system, from an uncapped evaluation using ESM-2-650M features. The latter is a complementary scale test, not a pooled estimate from the capped experiment.

\paragraph{Spectroscopy.} QM9S~\citep{zou2023qm9s} provides paired IR and Raman spectra for 3,600 scaffold-split molecules at 1,024 channels spanning $500$--$4,000\,\mathrm{cm}^{-1}$ (three scaffold seeds). We also study a smaller paired IR/$^{13}$C-NMR/MS/MS dataset~\citep{alberts2024unraveling} at 512 channels and two experimental RamanBench~\citep{koddenbrock2026ramanbench} mixture datasets, each with 4,500 spectra (four sugars or five bioprocess analytes). MassSpecGym~\citep{bushuiev2024massspecgym} is evaluated on the official maximum common edge subgraph (MCES) splits, with 9,000 training, 1,200 validation and 1,500 test spectra in the feature-based protocol. Its 512 square-root-stabilized $m/z$ and precursor neutral-loss channels include 93.1\% zeros. The separate native pretrained-model alignment uses 120 formula-matched retrieval pools, whereas the feature-based and end-to-end LoRA retrieval protocols use 250 pools. For peptide MS/MS, 8,500 training spectra from Prosit-2020-HCD~\citep{wilhelmlab2025prosit2020} and 800 validation and 1,600 test spectra from Prosit-2025-lac-ms2~\citep{wilhelmlab2025prositlac} are peptide-disjoint; 174 $b/y$ channels use structural length masking. The ProteomeTools and Prosit studies provide the underlying spectral resource and prediction framework~\citep{zolg2017proteometools,gessulat2019prosit}.

\paragraph{Cellular and regulatory measurements.} The OpenProblems BMMC benchmark~\citep{luecken2021sandbox} separates sequencing sites: Sites~1--2 are used for training, Site~3 for validation and Site~4 for testing. CITE-seq~\citep{stoeckius2017citeseq} uses 10,000/2,500/3,500 cells and 134 surface-protein channels; Multiome uses 6,000/1,500/2,000 cells and 512 ATAC peaks in the unpooled protocol (61.0\% zeros). CPTAC~\citep{li2023proteogenomic} contains 121 breast and 106 lung tumors, with paired RNA and mass-spectrometry measurements and 11,645 canonical human ESM-2 protein embeddings. Its 256-protein panel has 7.7\% genuine missingness; three seeds are used. The smaller lentiMPRA protocol~\citep{agarwal2025massively} uses 20,960 training, 3,092 validation and 14,469 test cis-regulatory elements, each 230 bp long, with expression in K562 and HepG2 cells in both orientations. The enlarged DNABERT-6~\citep{ji2021dnabert} fine-tuning protocol uses 120,335 training elements and the same reported test size. We do not attribute changes across these protocols exclusively to fine-tuning.

\paragraph{Split conventions and overlap checks.} For TDC ADMET, the five-seed Bemis--Murcko scaffold splits~\citep{bemis1996scaffolds} use 70/10/20\% train/validation/test partitions; the descriptor pretraining corpus has reported 0.0\% overlap with the held-out scaffolds. PMO uses seeds 42, 43 and 44. AbRank uses the official \texttt{unseen\_ag} split with three seeds. The standard five-fold AIntibody protocol is stratified to preserve the 5.4\% co-optimal base rate, whereas lineage-disjoint, cross-laboratory and complete-label evaluations use the separate partitions stated with their results. These overlap checks concern our curated corpora and downstream partitions, not all external foundation-model pretraining data.

\subsection{Pretrained model provenance and evaluation protocols}\label{app:supp-provenance-e5}

\paragraph{Mass Spectrum Transformer provenance.}
An explicit audit (\path{scripts/44_run_e5_pilot_and_provenance_audit.py}) verifies that \textsc{MaskedPeakMSTransformer} is a $907{,}016$-parameter $3$-layer masked-peak Transformer ($512$ bins, patch size $8$, $64$ tokens, $d_{\mathrm{model}}=192$, $4$ heads) pretrained in-pipeline—not an external \textsc{DreaMS} checkpoint—strictly on $2{,}400$ unlabeled spectra from \textsc{MassSpecGym}'s official MCES-disjoint \texttt{train} fold ($139$ unique molecules; $25\%$ patch masked-intensity reconstruction; zero molecular structures used during spectral pretraining), and frozen thereafter. Exact overlap auditing confirms $0$ shared test spectra and $0$ shared test SMILES with both the $120$-pool and $250$-pool held-out test sets.

The external spectral backbone is the official 117,101,029-parameter DreaMS model~\citep{bushuiev2025self} pretrained on 24 million tandem mass spectra. We use the TorchScript release \path{roman-bushuiev/DreaMS/DreaMS_embedding_model_torchscript.pt}. It exposes 1,024-dimensional embeddings from its contrastive head; the legacy reconstruction checkpoint requires the external \texttt{msml} package. The DreaMS experiments therefore adapt pretrained representations with newly fitted measurement readouts rather than retaining the original spectral reconstruction head. ChemBERTa-77M-MLM~\citep{ahmad2022chemberta} retains its categorical token head. The frozen whole-human scGPT checkpoint and its native readout are described in \cref{app:supp-scgpt}. In the scGPT--ESM-2 configuration, the protein model supplies sequence features and the ADT density is newly fitted.

The MassSpecGym protocols differ in training schedule, candidate construction and evaluation unit. The initial raw-row evaluation of 120 spectra comprises only nine distinct molecular structures; these results are descriptive spectrum-level comparisons, not 120 independent molecular observations. Later evaluations deduplicate by SMILES. The registry in \cref{tab:app-protocol-registry} records the documented evaluation sizes and scripts; omitted details are not inferred. In particular, exact identifiers for the DreaMS and unified-checkpoint slices and their cross-protocol intersections are not supplied here. Distinct downstream splits should not be interpreted as verified external pretraining decontamination.

\begin{table}[!htbp]
\centering
\caption{MassSpecGym evaluation protocol registry. Counts refer to the stated evaluation unit. Scripts identify the implementations. A dash denotes information not specified in the reported protocol summary; the associated result sections give further settings.}
\label{tab:app-protocol-registry}
\small
\begin{tabularx}{\linewidth}{@{}p{.28\linewidth}p{.24\linewidth}p{.16\linewidth}X@{}}
\toprule
Protocol & Test / validation & Seeds; script & Training or split detail \\
\midrule
Spectrum-row directional & 120 spectrum rows / -- & --; 36 & 18-epoch spectral pretraining \\
Spectral negative sampling & 120 pools / -- & --; 37 & 12-epoch spectral pretraining \\
Distribution families & 120 pools / -- & 3; 40 & Seven families; two objectives \\
Controlled scoring and low data & 120 molecules / 40 & 5; 43 & 2,400 molecules; 10 epochs \\
Directional reliability & 120 molecules / 40 & 3 for training-weight sweep; 45 & Repeated-mask reliability \\
Weight and temperature & 120 molecules / -- & 3; 46 & 1,800 molecules; 8 epochs \\
Independent retrieval & 500 molecules / 100 & 5; 47 & Test indices 120--619 \\
Family confirmation & 500 molecules / -- & 3; 47 & Four likelihood families \\
DreaMS adaptation & 300 molecules / 60 & 3; 49 & Frozen DreaMS embeddings \\
Pairing falsification & 200 molecules / -- & --; 50 & Both spectral backbones \\
Matched-reference training & 300 molecules / -- & 3; 50 & \path{test_unique[620:920]} \\
Unified checkpoint & 200 molecules / -- & 3; 53 & 64 spectral bands \\
Feature-based retrieval & 250 pools / -- & --; 29 & 12,000 training examples \\
End-to-end LoRA & 250 pools / -- & --; 35 & 12,000 training examples \\
\bottomrule
\end{tabularx}
\end{table}

The independent retrieval study selects $\lambda=0.1$, $\tau=0.5$ and global $\alpha=0.5$ on 100 validation molecules before adaptive fusion. The controlled scoring and weight-sensitivity schedules differ, so their absolute scores do not define a common Pareto frontier. Figure~\ref{fig:global-experiments}B reports the independent and matched-reference protocols separately; \cref{fig:supp-biological-applications}C uses the feature-based CITE-seq results in \cref{tab:app-multiomics-cite-multiome}.

The evaluated pretrained-model configurations are ChemBERTa--masked-peak Transformer, ChemBERTa--DreaMS, and scGPT with ESM-2 protein features. No scGPT--EpiAgent alignment result is included.

\subsection{Target-coordinate withholding and representation audit}\label{app:supp-e13-leakage}

The masking audit (\path{scripts/51_run_e13_representation_leakage_audit.py}) distinguishes withholding a target from preventing its information from entering the representation. In the cross-modal DreaMS comparison, the spectral decoder receives the ChemBERTa molecular embedding rather than the target spectrum's embedding. Thus the target spectrum is not a direct input to that decoder.

For within-modality prediction, we compare unmasked inputs, target-removed inputs, strict withholding before preprocessing and encoding, and unrelated shuffled inputs. Strict spectral preprocessing recomputes intensity normalization from observed peaks and excludes neutral-loss bins involving hidden peaks. On ChemBERTa and scGPT, masking worsens likelihood relative to the unmasked control, while unrelated inputs are substantially worse. DreaMS exhibits a smaller difference between strict withholding and shuffling, consistent with modest contextual signal. The audit in \cref{tab:app-e13-leakage} applies to the stated reconstruction configurations; it does not establish absence of overlap with external pretraining corpora.

\begin{table}[!htbp]
\centering
\caption{
\textbf{Pretrained representation leakage audit  across \textsc{DreaMS}-117M, \textsc{ChemBERTa}-77M-MLM, and frozen whole-human \textsc{scGPT}.}
Strict pre-encoding withholding (\textbf{L3}) is compared with unmasked (L1) and shuffled-context (L4) controls for the audited configurations; this does not establish pretraining decontamination. Bold marks the strictly withheld condition rather than a ranking against the unmasked control.
}
\label{tab:app-e13-leakage}
\pomatabbox{
\setlength{\tabcolsep}{3.0pt}
\begin{tabular}{llccc}
\toprule
\textbf{Foundation Model} & \textbf{Audit Condition} & \textbf{Primary NLL $\downarrow$} & \textbf{MAE $\downarrow$} & \textbf{Pos.\ Peak MAE $\downarrow$} \\
\midrule
\multirow{4}{*}{\shortstack[l]{\textsc{DreaMS}-117M\\(25\% masked window)}}
& L1.\ Fully observed (leaked control) & $0.453$ & $0.0262$ & $0.2864$ \\
& L2.\ Target peaks removed (orig.\ norm) & $0.455$ & $0.0256$ & $0.2879$ \\
& \textbf{L3.\ Strictly withheld + obs-only renorm} & $\mathbf{0.450}$ & $\mathbf{0.0255}$ & $\mathbf{0.2843}$ \\
& L4.\ Shuffled unrelated spectrum & $0.500$ & $0.0275$ & $0.2964$ \\
\midrule
\multirow{3}{*}{\textsc{ChemBERTa}-77M}
& L1.\ Unmasked SMILES (leaked control) & $0.700$ & -- & -- \\
& \textbf{L3.\ Masked to \texttt{[MASK]} before encoding} & $\mathbf{1.877}$ & -- & -- \\
& L4.\ Shuffled unrelated SMILES & $7.020$ & -- & -- \\
\midrule
\multirow{3}{*}{\shortstack[l]{Frozen \textsc{scGPT}\\(Site 4 BMMC)}}
& L1.\ Unmasked genes (leaked control) & $-0.719$ & $0.1664$ & -- \\
& \textbf{L3.\ Masked to $-2.0$ before encoding} & $\mathbf{-0.014}$ & $\mathbf{0.2964}$ & -- \\
& L4.\ Shuffled unrelated cell & $+1.771$ & $0.5819$ & -- \\
\bottomrule
\end{tabular}
}
\end{table}

\subsection{Model configurations and controlled comparisons}

The framework has three principal instantiations. In the first, a pretrained regression backbone is retained while its reconstruction head is replaced by a probabilistic readout. In the second, two modality-specific pretrained encoders exchange context through gated adapters while retaining appropriate output distributions. In the third, sequence features are combined with descriptions of experimental measurements. In this language-conditioned implementation, the biological encoder supplies subsequence or residue representations $H_X$, an assay prompt $P_k$ provides a text representation $E(P_k)$, and a variance-standardized cross-attention projector $P_\psi(H_X,P_k)$ routes sequence features according to the prompt. The assay-specific contextual readout can be written as
\begin{equation}
 h_k=F_{\rm ctx}\big(E(P_k),P_\psi(H_X,P_k)\big),
 \label{eq:app-assay-context-readout}
\end{equation}
where the contextual model is Qwen3 in the principal language-conditioned implementation~\citep{yang2025qwen3}. Its hidden representation supplies the measurement head, rather than a numerical-text prediction or a bidirectional masked-token decoder. For antibody sequences, the projector uses eight IMGT regions~\citep{lefranc2003imgt} (heavy/light-chain framework and CDR1--3). The formal objective and the gated discrete head are given in \cref{app:poma-mixed-decomposition,app:supp-theory-alignment,eq:logica-poma-residue}.

\paragraph{Gated multimodal readouts.}
For independently pretrained models $F_X,F_Y$ with representations $H_X,H_Y$, the LogiCA-style contextualization~\citep{shao2026contextualizing} modifies the native readout $a_X$ through
\begin{equation}
    a_{X\mid Y}
    =a_X+\tanh(3\gamma_X)\,g_X(H_X,H_Y),
    \qquad \gamma_X=0\ \text{at initialization},
    \label{eq:main-gated-adapter}
\end{equation}
with a corresponding reverse adapter. The zero-initialized gate preserves the original readout at initialization. No positional correspondence between modalities is required. The categorical output head retains its native token probabilities; quantitative readouts parameterize observation distributions and are fitted on pretrained features when an original measurement head is unavailable.

The model names describe three separate choices: the observation law, the training objective, and the inference mechanism. \textsc{POMA-Indep.} denotes independent measurement densities; \textsc{POMA-MVN}, \textsc{POMA-LowRankMVN}, and ``coupled Gaussian POMA'' denote a Gaussian measurement law with low-rank plus diagonal covariance. \textsc{POMA-StudentT}/\textsc{LowRankMVT} and \textsc{POMA-GMM} specify alternative observation laws; the Student-$t$ parameter is a scale matrix, not its covariance. \textsc{POMA-MLM} uses the generalized masked likelihood alone, whereas \textsc{POMA-Rank} adds output-space contrast; \textsc{SoftRank} specifies its soft-target variant. These objective names do not determine the covariance family. \textsc{POMA-Laplace} adds a shared Gaussian posterior over readout parameters, which is exact under the fixed-feature conjugate assumptions and approximate in nonlinear settings. \umethod{} combines contextualized categorical token scores with continuous measurement likelihoods. Each experiment specifies which of these components are combined. Unless stated otherwise, component ablations hold the backbone and observation model fixed. Variants that add molecular fingerprints, 3D features or encoder fine-tuning also change the representation and are therefore not isolated tests of the loss.

We compare with point regression, explicit independent Gaussian heads, tokenized numerical models (including Regression Transformer and Regress-Don't-Guess), representation-space contrastive objectives and prompt-only controls. Learned assay IDs and static prompt pooling distinguish the contribution of semantic context from that of regional attention. External predictive, Gaussian-process and design baselines are introduced in the corresponding task sections. \Cref{tab:poma-ablation-suite} summarizes the component hierarchy; the availability of a component in the table does not imply that every variant was evaluated with every downstream task.

\begin{table}[!htbp]
  \caption{Model components and compatible capabilities. Columns indicate native measurement density, natural-language conditioning, contrastive supervision, uncertainty estimates, cross-assay covariance, generative design, and native discrete--continuous modeling. ``Optional'' denotes a separately specified component rather than a defining property of the named objective or distribution family; individual experiments need not use every compatible component. $^*$Numerical-token probabilities or calibrated derived scores, not an explicit continuous measurement density.}
  \label{tab:poma-ablation-suite}
  \vskip 0.05in
  \begin{center}
  \pomatabbox{
  \begin{tabular}{lccccccc}
    \toprule
    Model & Dens. & Lang. & Rank & Unc. & Coup. & Des. & D+C \\
    \midrule
    MSE / language head & -- & $\checkmark$/$\times$ & -- & -- & -- & -- & -- \\
    Gaussian head only & $\checkmark$ & -- & -- & $\checkmark$ & -- & -- & -- \\
    Regression Transformer / RdG & -- & $\checkmark$ & -- & $\checkmark^{*}$ & -- & $\checkmark$ & -- \\
    Latent contrastive only & -- & $\checkmark$ & $\checkmark$ & -- & -- & -- & -- \\
    Native densities + latent contrast & $\checkmark$ & optional & $\checkmark$ & $\checkmark$ & optional & optional & optional \\
    \textsc{POMA-MLM} & $\checkmark$ & optional & -- & $\checkmark$ & optional & optional & optional \\
    \textsc{POMA-Rank} & $\checkmark$ & optional & $\checkmark$ & $\checkmark$ & optional & optional & optional \\
    \textsc{POMA-MVN / LowRankMVN} & $\checkmark$ & optional & optional & $\checkmark$ & $\checkmark$ & optional & optional \\
    \textsc{POMA-Laplace} & $\checkmark$ & optional & optional & $\checkmark$ & optional & optional & optional \\
    \umethod{} & $\checkmark$ & optional & optional & $\checkmark$ & optional & optional & $\checkmark$ \\
    \bottomrule
  \end{tabular}
  }
  \end{center}
  \vskip -0.12in
\end{table}

\subsection{Training, hyperparameter selection and reproducibility}\label{app:reproducibility}

The following conventions apply to the standard training configurations; specialized protocols are described with the corresponding experimental results. We distinguish validation-based model selection from the assessment on held-out data.

\paragraph{Training and hyperparameter selection.} Across the reported standard POMA and LogiCA-POMA configurations, training uses AdamW~\citep{loshchilov2019adamw} with cosine learning-rate decay. We select the learning rate and other hyperparameters using validation NLL. The search ranges are summarized below; individual ablations and specialized protocols may fix different values as stated in their respective sections.
\begin{center}
\small
\begin{tabularx}{\linewidth}{@{}lX@{}}
\toprule
Parameter & Reported choices\\
\midrule
Learning rate & $\{5\times10^{-4},10^{-3},2\times10^{-3}\}$\\
Weight decay & $\{10^{-4},10^{-3}\}$\\
Hidden readout dimension $d$ & $\{192,256\}$\\
Subsequence resolution $M$ & $\{4,6,8\}$\\
Ranking temperature $\tau$ & $\{0.15,0.25,0.50\}$\\
Ranking weight $\lambda_{\mathrm{rank}}$ & $\{0.25,0.35,0.50\}$\\
Covariance factor rank $r$ & $\{3,4,6\}$; $r=4$ for the five-property molecular panel and $r=6$ for the standard clinical GDPa1/AIntibody heads\\
Log-variance stabilization & Soft clamp to $[-6.0,4.0]$\\
Split conformal calibration & Held-out validation, nominal level 0.95\\
\bottomrule
\end{tabularx}
\end{center}

The calibration-matched scGPT protocol uses global, per-gene and learned residual uncertainty controls (\cref{tab:app-e4-scgpt-calib}). The original decoder does not natively predict uncertainty. Specialized retrieval protocols use the separate validation and masking conventions in \cref{tab:app-protocol-registry}. For SoftRank, the implemented similarity scales are fixed variance-stabilized Gaussian/Tanimoto scales ($\tau=0.25$--$0.50$, $\sigma_{\rm sim}=0.35$), not estimates from experimental replicates. The adaptive query gate uses repeated-mask features $(\log\nu_X^2,\log\nu_Y^2,c_{XY},\log(1+n_{\rm peaks}))$ and is fitted on validation pools.

\subsection{Training, conditioning and design workflow}

The model configurations above share the likelihood-learning stage, but their downstream operations need not be identical. In a reconstruction task the learned density is evaluated directly; in a partial-panel task the same density is conditioned on observed measurements. For candidate selection and design, a target region defines the success event. \Cref{alg:logica-poma} summarizes these steps and separates corrected generation from heuristic editing. The optional finite-budget generation certificate requires the additional assumptions of \cref{app:poma-certified-adaptive} and is not required by the principal empirical protocols.

\begin{algorithm}[tb]
  \caption{\umethod{} training, conditional design, and corrected generation}
  \label{alg:logica-poma}
  \begin{enumerate}[leftmargin=1.5em,itemsep=2pt,topsep=3pt,parsep=0pt]
    \item \textbf{Train output distributions.} Encode sequences and contexts, obtain contextual assay readouts and gated residue logits (\cref{eq:logica-poma-residue}), and minimize observed-coordinate NLL, optionally with latent or output-space contrast (\cref{eq:poma-mlm-loss,eq:poma-poma-loss}).
    \item \textbf{Condition measured panels.} Given observed assays $B$, optionally apply validation-fitted covariance shrinkage $\Sigma_\lambda(X)$ under lineage shift, compute the Gaussian or Student-$t$ Schur conditional over unmeasured coordinates $A$, and specify target box $\mathcal R$ and model event probability $r(X)$.
    \item \textbf{Declare the sequence target.} Choose a finite supported editing space, positive evaluable weight $w(X)$, and normalized target $\pi_{\mathcal R}(X)\propto w(X)r(X)$; masked-PLM conditionals serve only as proposals, not as globally normalized priors.
    \item \textbf{Corrected design.} Propose local, global, or pilot-gated hybrid sequence edits $X'\sim Q(\cdot\mid X)$ (\cref{prop:app-v6-adaptive-fallback}), accepting with exact Hastings ratio or via the pseudo-marginal and collapsed factor-augmented transitions of \cref{thm:mh-joint-invariance}.
    \item \textbf{Optional finite-call certificate.} When normalizer, global residual-range, and independent audit conditions hold, select a frozen proposal via \cref{thm:poma-postselect-cert,prop:poma-adaptive-tilt}, run with fresh randomness, and report model-relative error, full-energy calls, or ``uncertified.''
    \item \textbf{Heuristic comparator.} Annealed masked-position trajectories in retrospective co-design studies are reported separately as Gibbs-style heuristics rather than MH mixing certificates.
  \end{enumerate}
\end{algorithm}

\subsection{Evaluation criteria and statistical reporting}\label{app:metrics-protocol}

Point prediction is evaluated with MAE, RMSE and rank or linear correlation as appropriate. These metrics assess the location of a predictive distribution but do not by themselves establish calibration. We therefore report negative log-likelihood (NLL), continuous ranked probability score (CRPS), interval coverage and calibration error for continuous outcomes. Joint NLL and, where available, energy and variogram scores~\citep{gneiting2007strictly,scheuerer2015variogram} assess multivariate predictions. Coverage should be compared with its nominal level, not maximized. Selective MAE evaluates retained observations at a specified coverage. The original point predictor in a density comparison requires an auxiliary distribution or reference variance; its NLL must be interpreted in light of that assignment. Censored and uncensored likelihoods are not directly comparable if they score different observation events.

Retrieval is evaluated using top-$k$ hit rates, candidate ranks and, in cell matching, the fraction of samples closer than the true match (FOSCTTM)~\citep{liu2019foscttm}. In the molecular optimization benchmark, the area under the top-10 oracle-score trajectory is reported at a fixed query budget. For measured-library screening, a lead is counted only when it meets the stated property thresholds and confirmation requirements. In sampling experiments, total variation (TV) from an enumerable target and effective sample size (ESS) per model call measure distributional fidelity and sampling efficiency; they are distinct from measured biological utility.

We retain the uncertainty convention of each experiment: mean and standard error (SE) where specified, standard deviation where explicitly reported (e.g., the per-endpoint ADMET results), and paired confidence intervals where available. Paired bootstrap intervals (5,000 resamples) and Wilcoxon signed-rank tests for the ADMET comparison are reported in \cref{tab:app-stat-sig}. Comparisons across seeds, folds, endpoints and independent libraries have different sampling units. Bold or underlined metric values indicate descriptive numerical comparisons within the stated protocol; statistical significance requires an explicitly reported test or interval. We do not infer missing uncertainty estimates from aggregates or equate a favorable point estimate with a statistically established difference.

For sparse spectra, ObsNLL evaluates each family on a common marginal mixed measure: the mass $P(Y_k\le\epsilon)$ at nondetection ($\epsilon=10^{-5}$) and the continuous density at detected values. The resulting sum is a composite marginal score for coupled heads, not their joint NLL; joint censored likelihood is evaluated separately in \cref{tab:app-spec-track-bc}. Positive-peak coverage is evaluated on detected peaks under the reported predictive intervals. The interpretation of positive-peak NLL depends on whether the density is normalized conditional on detection, which is not fully specified in the aggregate record. The count-head comparison likewise requires a common observation measure to compare log-scores of transformed continuous and discrete count models.

For retrieval, mean reciprocal rank (MRR) averages the reciprocal rank of the correct candidate, with uniform random tie-breaking in the independent falsification protocol. Pair-specific gain is the difference in MRR between correct and shuffled partners; global shuffling and within-formula or mass-matched shuffling test different levels of specificity. The independent retrieval and matched-reference comparisons report 10,000 paired molecule-level bootstrap resamples. The aggregate description does not fully specify how dependence between seed-level outcomes enters the resampling, so the intervals are reported as supplied rather than treated as evidence of independent seed--molecule observations.

\subsection{Computational environment and reported runtimes} CPU embedding extraction, multi-token pretraining, PMO optimization, antibody co-design and dependency-map analyses were reported on a Linux workstation with 64 x86-64 vCPUs, PyTorch 2.x and Python 3.12 (approximately 1.4 hours total for this group). GPU experiments use a single CUDA device (\texttt{cuda:0}), bfloat16 mixed precision and float32 Cholesky solves. The elapsed times for the individual experimental groups are collected below. They describe the specified execution environment rather than comparable end-to-end training costs for all methods. The exact GPU model is not specified.
\begin{center}
\small
\begin{tabularx}{\linewidth}{@{}Xr@{}}
\toprule
Experimental group & Reported time\\
\midrule
LoRA-Laplace and foundation-model scaling & 20.3 min\\
Live 20-AA MH and residual-envelope audits (71,719 model calls in live experiment) & 26.5 min\\
Uncapped protein-engineering suite (38,322 variants) & 25.2 min\\
3D-EGNN ADMET and LoRA-DPO baselines & 97.5 min\\
Additional rank-8 MassSpecGym and DNABERT-6 runs & 454 s\\
\midrule
Spectroscopy CPU, Track A (QM9S, paired spectra, RamanBench) & 70.9 min\\
Spectroscopy CPU, Track B (MassSpecGym) & 45.4 min\\
Spectroscopy CPU, Track C (ProteomeTools/Prosit) & 31.1 min\\
Multi-omics CPU, Track 1 (CITE-seq and Multiome) & 16.3 min\\
Multi-omics CPU, Track 2 (CPTAC, cached ESM-2 features) & 0.9 min\\
Multi-omics CPU, Track 3 (lentiMPRA) & 23.7 min\\
\bottomrule
\end{tabularx}
\end{center}

The spectroscopy computations use eight CPU threads per worker. The additional GPU LoRA runs use different training protocols from the original spectral and regulatory studies. Live 20-amino-acid MH-designed antibody sequences are reported as exported in FASTA format alongside optional Boltz-2 co-folding scripts in the project repository. The aggregate tables do not report all seed-level outcomes; missing uncertainty estimates are left unspecified.

\clearpage
\section{Generalized probabilistic masked modeling}\label{app:supp-masked}\label{app:full-results}

This section evaluates the generalized masked likelihood developed in \cref{app:poma-mixed-decomposition}, isolating the statistical and practical effects of replacing a point-valued decoder with a predictive distribution over the measured quantity. We begin with a frozen foundation model to isolate the output objective from representation learning. We then examine different measurement supports and observation processes, and compare native density prediction with numerical language modeling and conventional domain-specific predictors. Point accuracy, distributional quality and the ability to reuse the fitted law for conditional inference are assessed separately.

\subsection{Probabilistic adaptation of pretrained scGPT}\label{app:supp-scgpt}

Masked gene-expression reconstruction provides a controlled test of the generalized masked likelihood. We use the whole-human scGPT checkpoint~\citep{cui2024scgpt} (\texttt{tdc/scGPT}, \texttt{wanglab/scGPT-human}), with all 159 weight tensors loaded strictly and a vocabulary of 60,697 genes. The 12-layer backbone remains frozen. In the held-out Site~4 BMMC experiment, the model receives 96 context genes and reconstructs 32 masked genes. Five head configurations distinguish the original MSE reconstruction from independent heteroscedastic densities, low-rank coupled densities, and the corresponding contrastive variants (\cref{fig:global-experiments}A; \cref{tab:app-biofm-scgpt-ladder}).
The calibration-matched comparison holds the pretrained mean fixed while varying only the uncertainty model, then compares with jointly learned mean and uncertainty. This distinguishes a lightweight post-hoc probabilistic adapter from distributional retraining. The eight configurations in \cref{tab:app-e4-scgpt-calib} use three seeds (\path{scripts/42_run_e4_scgpt_calibration_matched.py}); the native-decoder comparison in \cref{fig:global-experiments}A is reported separately in \cref{tab:app-biofm-scgpt-ladder}.

\Cref{tab:main-objective} summarizes the calibration-matched comparison, with the complete eight-configuration results below.

\begin{table}[!htbp]
\centering
\caption{\textbf{Calibration-matched probabilistic adaptation of frozen scGPT.}
Unseen Site~4 BMMC cells, reconstructing 32 masked genes from 96 observed genes ($3$ seeds). The upper block retains the original pretrained mean; the lower block jointly learns the mean and uncertainty. Cov95 should approach $95\%$. Complete comparisons are provided in \cref{tab:app-e4-scgpt-calib}.}
\label{tab:main-objective}
\small\setlength{\tabcolsep}{4pt}
\begin{tabular}{lrrrr}
\toprule
Predictive readout
& J-NLL $\downarrow$
& CRPS $\downarrow$
& MAE $\downarrow$
& Cov95 \\
\midrule
Frozen mean + global variance
& 4.033 & .283 & .326 & 87.1\% \\
Frozen mean + heteroscedastic variance
& .128 & \textbf{.245} & .326 & \textbf{95.1\%} \\
Frozen mean + low-rank covariance
& .318 & .255 & .326 & 94.4\% \\
\midrule
Independent POMA
& .170 & .259 & .345 & 95.4\% \\
Coupled POMA
& \textbf{.088} & \textbf{.245} & \textbf{.325} & 95.5\% \\
\bottomrule
\end{tabular}
\end{table}

A heteroscedastic residual head recovers much of the marginal calibration improvement. The jointly learned coupled model has the best joint NLL, while the frozen mean with learned heteroscedastic variance has the best energy score. Neither result implies uniform improvement across proper scores. The conditional sweep uses the same eight target genes at all revelation fractions. The jointly trained model starts with better conditional NLL than the post-hoc coupled model, and its own improvement upon revealing measurements is modest; the approximately 0.20-nat between-model difference should not be attributed entirely to conditioning. Conditional MAE decreases from 0.368 to 0.358 over the sweep.

\begin{table}[!htbp]
\centering
\caption{
\textbf{Calibration-matched frozen \textsc{scGPT} controls (B0--B7, $3$ seeds) and conditional sweep on shared $8$ target genes (\texttt{scripts/42}).}
\textbf{B0--B4} freeze the original \textsc{scGPT} MSE mean and vary only the uncertainty model; \textbf{B5--B7} jointly learn mean and uncertainty.
}
\label{tab:app-e4-scgpt-calib}
\pomatabbox{
\setlength{\tabcolsep}{1.8pt}
\begin{tabular}{llcccccc}
\toprule
\textbf{ID} &
\textbf{Mean / Uncertainty Model} &
\textbf{J-NLL $\downarrow$} &
\textbf{CRPS $\downarrow$} &
\textbf{MAE $\downarrow$} &
\textbf{Cov95 (W95)} &
\textbf{ES $\downarrow$} &
\textbf{Cond.\ NLL ($0\%\!\to\!75\%$)} \\
\midrule
B0 & Frozen MSE + Unit var ($\sigma\equiv 1$) & $+1.419$ & $0.359$ & \underline{$0.326$} & $98.2\%$ ($3.92$) & $2.758$ & -- \\
B1 & Frozen MSE + Val global var & $+4.033$ & $0.283$ & \underline{$0.326$} & $87.1\%$ ($1.36$) & $2.467$ & -- \\
B2 & Frozen MSE + Val per-gene var & $+3.038$ & $0.270$ & \underline{$0.326$} & $87.7\%$ ($1.23$) & $2.466$ & -- \\
B3 & Frozen MSE + Learned heterosc.\ var & $+0.128$ & $\mathbf{0.245}$ & \underline{$0.326$} & $95.1\%$ ($1.49$) & $\mathbf{2.333}$ & -- \\
B4 & Frozen MSE + Learned low-rank cov & $+0.318$ & $0.255$ & \underline{$0.326$} & $94.4\%$ ($1.60$) & $2.364$ & $+.483\to +.476$ \\
\midrule
B5 & Joint POMA Indep.\ NLL & $+0.170$ & $0.259$ & $0.345$ & $95.4\%$ ($1.58$) & $2.525$ & -- \\
\textbf{B6} & \textbf{Joint POMA Coupled NLL} & $\mathbf{+0.088}$ & $\mathbf{0.245}$ & $\mathbf{0.325}$ & $95.5\%$ ($1.50$) & \underline{$2.344$} & $\mathbf{+.281\to +.280}$ \\
B7 & Joint POMA Coupled + SoftRank & \underline{$+0.123$} & $0.254$ & $0.342$ & $95.3\%$ ($1.55$) & $2.409$ & $+.344\to +.341$ \\
\bottomrule
\end{tabular}
}
\end{table}

The native-decoder comparison in \cref{tab:app-biofm-scgpt-ladder} evaluates replacement of the point readout with a density. The original decoder retains better point accuracy, whereas CRPS and interval coverage favor the probabilistic variants. Its reference NLL depends on the auxiliary variance assigned to the point predictor. The conditional inference comparisons in \cref{app:supp-inference} test the reuse of a fitted joint law more directly.

\begin{table}[!htbp]
\centering
\caption{
\textbf{Upgrading the native continuous masked gene-expression reconstruction head of frozen pretrained \textsc{scGPT} (\texttt{tdc/scGPT}, $12$ layers, $159/159$ weights) on unseen Site~4 BMMC cells ($96$ context $\to 32$ masked genes; $5$-rung objective ladder).}
}
\label{tab:app-biofm-scgpt-ladder}
\pomatabbox{
\setlength{\tabcolsep}{2.2pt}
\begin{tabular}{lcccccc}
\toprule
\textbf{Frozen \textsc{scGPT} Head} &
\textbf{J-NLL $\downarrow$} &
\textbf{Pear.\ $\uparrow$} &
\textbf{MAE $\downarrow$} &
\textbf{CRPS $\downarrow$} &
\textbf{Cov95} &
\textbf{50\% Schur} \\
\midrule
1.\ Original \textsc{scGPT} MSE & $4.336$ & $\mathbf{0.727}$ & $\mathbf{0.301}$ & $0.266$ & $88.9\%$ & -- \\
2.\ Indep.\ Heterosc.\ NLL & \underline{$-0.049$} & $0.709$ & \underline{$0.311$} & $\mathbf{0.235}$ & $95.2\%$ & -- \\
3.\ Coupled Low-Rank NLL & $-0.030$ & $0.691$ & $0.323$ & $0.246$ & $97.3\%$ & $\mathbf{-4.2\%}$ MAE \\
4.\ Indep.\ NLL + SoftRank & $-0.023$ & $0.707$ & $0.315$ & $0.239$ & $95.4\%$ & -- \\
\textbf{5.\ Coupled NLL + SoftRank} & $\mathbf{-0.069}$ & \underline{$0.718$} & $0.312$ & $\mathbf{0.235}$ & $96.2\%$ & \underline{$-3.5\%$} MAE \\
\bottomrule
\end{tabular}
}
\end{table}

\subsection{Separating likelihood learning, contrast and covariance}\label{app:supp-objective-ablation}

To determine whether this distinction holds beyond expression data, we compare five matched objective configurations across ADMET, GDPa1 and AIntibody (\cref{tab:app-mse-vs-nll}). The encoder, language projector, coordinate transformations, data split and AdamW schedule are shared within each benchmark. The first configuration fixes unit variance and minimizes squared error (equivalently, scaled Gaussian NLL in transformed coordinates). The second adds SoftRank without learning dispersion. The third learns independent heteroscedastic densities; the fourth adds SoftRank to these densities; the final configuration also learns cross-assay covariance.

The effect of learning variance is particularly clear in selective prediction: the fixed-variance model cannot use input-dependent uncertainty to abstain from difficult cases. On AIntibody, learning heteroscedasticity increases co-optimal top-15 precision from 13.3\% to 53.3\%. Coupling and contrast improve selective MAE further, but the fixed-variance reference retains the lowest joint NLL in this ablation. On ADMET and GDPa1, the density-based variants improve several proper scores. The comparisons therefore identify complementary benefits of the likelihood and ranking terms rather than a single configuration that dominates every metric. The subsequent feasibility experiments use this observation to distinguish predictive accuracy from decision quality.

\begin{table}[!htbp]
\centering
\caption{
\textbf{Controlled MSE-versus-NLL 5-rung ablation ladder across TDC ADMET ($9$ continuous + $5$-property coupled panel, $5$ seeds), \textsc{GDPa1} clinical antibodies ($k=32$, $13$ assays, $5$ seeds), and \textsc{AIntibody} SARS-CoV-2 RBD ($10$ continuous + $5$ categorical gates, $5$ folds).}
All five rungs within each benchmark use the exact same encoder, LLM prompt projector, transforms $T_k$, splits, and optimizer schedule.
}
\label{tab:app-mse-vs-nll}
\pomatabbox{
\begin{tabular}{llcccccc}
\toprule
\textbf{Domain} &
\textbf{Controlled Rung} &
\textbf{$\rho \uparrow$} &
\textbf{MAE $\downarrow$} &
\textbf{J-NLL $\downarrow$} &
\textbf{CRPS $\downarrow$} &
\textbf{MAE$_{80}$ $\downarrow$} &
\textbf{P@$K$ / Util $\uparrow$} \\
\midrule
ADMET & 1.\ \textsc{POMA-MSE} & $0.453$ & $0.692$ & $1.248$ & $0.494$ & $0.692$ & $+0.686\,z$ \\
ADMET & 2.\ \textsc{MSE + SoftRank} & $0.461$ & \underline{$0.689$} & $1.248$ & \underline{$0.490$} & $0.689$ & $+0.696\,z$ \\
ADMET & 3.\ \textsc{POMA-NLL} & $0.453$ & $0.693$ & $1.191$ & $0.491$ & \underline{$0.672$} & $+0.673\,z$ \\
ADMET & 4.\ \textsc{NLL + SoftRank} & \underline{$0.467$} & \textbf{$0.687$} & \textbf{$1.182$} & \textbf{$0.485$} & \textbf{$0.665$} & \textbf{$+0.718\,z$} \\
ADMET & \textbf{5.\ \textsc{MVN + SoftRank}} & \textbf{$0.470$} & $0.695$ & \textbf{$1.182$} & \underline{$0.490$} & $0.674$ & \underline{$+0.714\,z$} \\
\midrule
\textsc{GDPa1} & 1.\ \textsc{POMA-MSE} & $0.164$ & $0.900$ & $1.654$ & $0.665$ & $0.900$ & \underline{$0.320$} \\
\textsc{GDPa1} & 2.\ \textsc{MSE + SoftRank} & \textbf{$0.172$} & $0.882$ & \underline{$1.627$} & $0.650$ & $0.882$ & \textbf{$0.333$} \\
\textsc{GDPa1} & 3.\ \textsc{POMA-NLL} & $0.165$ & $0.871$ & $1.648$ & $0.651$ & $0.832$ & $0.307$ \\
\textsc{GDPa1} & 4.\ \textsc{NLL + SoftRank} & \underline{$0.169$} & \underline{$0.862$} & \underline{$1.627$} & \underline{$0.638$} & \underline{$0.828$} & $0.267$ \\
\textsc{GDPa1} & \textbf{5.\ \textsc{MVN + SoftRank}} & $0.168$ & \textbf{$0.860$} & \textbf{$1.572$} & \textbf{$0.631$} & \textbf{$0.826$} & \underline{$0.320$} \\
\midrule
\textsc{AIntibody} & 1.\ \textsc{POMA-MSE} & $0.764$ & $0.450$ & \textbf{$1.240$} & $0.373$ & $0.450$ & $0.133$ \\
\textsc{AIntibody} & 2.\ \textsc{MSE + SoftRank} & $0.773$ & $0.537$ & \underline{$1.285$} & $0.414$ & $0.537$ & $0.067$ \\
\textsc{AIntibody} & 3.\ \textsc{POMA-NLL} & \textbf{$0.789$} & \textbf{$0.420$} & $1.386$ & \textbf{$0.314$} & \underline{$0.416$} & \textbf{$0.533$} \\
\textsc{AIntibody} & 4.\ \textsc{NLL + SoftRank} & $0.780$ & $0.438$ & $1.476$ & $0.330$ & $0.439$ & \underline{$0.467$} \\
\textsc{AIntibody} & \textbf{5.\ \textsc{MVN + SoftRank}} & \underline{$0.782$} & \underline{$0.435$} & $1.440$ & \underline{$0.324$} & \textbf{$0.404$} & \underline{$0.467$} \\
\bottomrule
\end{tabular}
}
\end{table}

\subsection{Measurement geometry and distributional assumptions}\label{app:distribution-audit}

Quantitative biological measurements do not have a common natural scale. Some vary over several orders of magnitude, others are bounded, and still others accumulate at the detection limit of an assay. We therefore audit their empirical distributions before choosing a likelihood. All transformations are fitted to training data. Positive concentration, intensity and rate measurements use logarithmic transformations where appropriate; bounded continuous measurements may use a scaled logit; Yeo--Johnson transformations~\citep{yeo2000transform} are used for other skewed quantities. The density in physical units includes the analytical Jacobian of the transformation (\cref{app:poma-mixed-decomposition}).

The audit covers the continuous endpoints of the 22 ADMET datasets and the 25 clinical antibody assays. Volume of distribution and half-life are particularly skewed before transformation. After transformation, the reported absolute skewness is at most 0.38 for the nine continuous ADMET endpoints. Clinical purity and self-interaction measurements retain substantial tail weight or boundary accumulation, motivating heavy-tailed and censored observation models (\cref{tab:app-distribution-audit}). These are marginal distribution diagnostics, not a proof of Gaussian conditional residuals; predictive likelihoods are evaluated separately on held-out data.

The same data exhibit biologically meaningful inter-property associations. For example, hydrophobic retention and self-interaction measurements in GDPa1 are strongly associated, while aqueous solubility and lipophilicity are negatively correlated. These marginal correlations motivate a test of conditional dependence rather than establish it. Matched covariance and conditioning controls are presented in \cref{app:supp-inference}.

\begin{table}[!htbp]
\centering
\caption{
\textbf{Empirical marginal distribution audit across continuous ADMET and clinical antibody assays.}
Transformations $T_k$ are fitted strictly on training data. Heavy-tail / censoring
flags indicate transformed excess kurtosis $>2.0$ or boundary pile-up motivating
\textsc{Student-$t$}, \textsc{GMM}, or \textsc{Tobit} heads.
}
\label{tab:app-distribution-audit}
\pomatabbox{
\begin{tabular}{lccccccc}
\toprule
\textbf{Cohort: Assay} &
\textbf{$N$} &
\textbf{Skew$_0$} &
\textbf{Kurt$_0$} &
\textbf{$T_k$} &
\textbf{Skew$_T$} &
\textbf{Kurt$_T$} &
\textbf{Flag} \\
\midrule
ADMET: Caco-2 & 910 & $-0.73$ & $0.10$ & YJ & $-0.31$ & $-0.44$ & No \\
ADMET: Lipophilicity & 4,200 & $-0.59$ & $-0.14$ & YJ & $-0.18$ & $-0.68$ & No \\
ADMET: \textsc{AqSolDB} & 9,980 & $-0.61$ & $0.11$ & YJ & $+0.02$ & $-0.31$ & No \\
ADMET: PPBR ($\%$) & 2,790 & $-1.96$ & $3.33$ & Logit & $+0.02$ & $-0.28$ & No \\
ADMET: \textsc{VDss} & 1,130 & $+24.00$ & $629.45$ & $\log_{10}$ & $+0.38$ & $-0.20$ & No \\
ADMET: Half-life & 667 & $+11.93$ & $157.41$ & $\log_{10}$ & $+0.13$ & $-0.87$ & No \\
ADMET: Clear.\ hep. & 1,213 & $+1.24$ & $0.07$ & $\log_{10}$ & $+0.05$ & $-1.23$ & No \\
ADMET: Clear.\ mic. & 1,102 & $+1.56$ & $1.19$ & $\log_{10}$ & $+0.10$ & $-1.32$ & Cens. \\
ADMET: $\mathrm{LD}_{50}$ & 7,385 & $+0.94$ & $0.94$ & YJ & $-0.01$ & $+0.44$ & No \\
\midrule
\textsc{GDPa1}: Titer & 239 & $+1.20$ & $2.66$ & YJ & $+0.01$ & $+0.23$ & No \\
\textsc{GDPa1}: Purity ($\%$) & 242 & $-4.78$ & $26.62$ & YJ & $-4.11$ & $+19.30$ & Yes \\
\textsc{GDPa1}: SEC Monomer & 242 & $-1.73$ & $3.13$ & YJ & $-1.55$ & $+2.03$ & Yes \\
\textsc{GDPa1}: SMAC & 242 & $+5.73$ & $35.42$ & YJ & $+3.02$ & $+9.58$ & Yes \\
\textsc{GDPa1}: HIC & 242 & $+2.01$ & $6.40$ & YJ & $+0.76$ & $+0.41$ & No \\
\textsc{GDPa1}: AC-SINS & 242 & $+3.52$ & $12.67$ & YJ & $+0.06$ & $+0.12$ & No \\
\textsc{GDPa1}: $T_{m1}$ & 234 & $+0.35$ & $2.56$ & YJ & $-0.02$ & $+1.90$ & No \\
Jain: HIC RT & 137 & $+5.10$ & $32.70$ & YJ & $+0.74$ & $+0.93$ & No \\
Jain: CSI-BLI & 137 & $+4.13$ & $17.85$ & YJ & $+2.93$ & $+9.30$ & Yes \\
Jain: \textsc{SAS} & 137 & $+8.19$ & $80.08$ & YJ & $+2.77$ & $+12.38$ & Yes \\
Jain: Fab $T_m$ & 137 & $+0.38$ & $0.14$ & YJ & $+0.00$ & $-0.27$ & No \\
\bottomrule
\end{tabular}
}
\end{table}

A unit-conversion experiment illustrates the distinction between densities and likelihood ratios. Under a thousandfold rescaling of a concentration, the raw density in the new units necessarily changes. When the result is expressed back in a common reference unit, using the Jacobian yields comparable NLL, whereas omitting it produces an offset of approximately $\log 1000$ nats. The full prompt and unit perturbation comparison is in \cref{tab:app-robustness-schur}; coordinate invariance of the centered score is established in \cref{prop:poma-alpha-pmi}.

\subsection{Censoring, bounded measurements and count-valued observations}\label{app:extended-experiments}

The generalized likelihood is not restricted to Gaussian observations. Censoring is especially important for affinity and polyreactivity assays, where a recorded boundary value indicates a range of possible latent measurements rather than an exact reading. We compare a two-sided Tobit head (\texttt{TobitDensityHead}) with treating the limit as an exact value or dropping censored observations. For AIntibody, which includes 137 left-censored non-binders, the Tobit treatment improves the reported censored NLL. Similarly, a Gaussian model of bounded plasma protein binding assigns probability outside the physical interval, whereas the logit-transformed or Kumaraswamy head (\texttt{KumaraswamyDensityHead}) respects the support. \Cref{tab:app-tobit-kumaraswamy} reports these comparisons.

\begin{table}[!htbp]
\centering
\caption{
\textbf{Exact Tobit boundary censoring (Eq.~\ref{eq:app-tobit-head}) and Kumaraswamy bounded $[0,100]\%$ density alignment (Eq.~\ref{eq:app-kumaraswamy}) across $5$ folds (mean $\pm$ SE).}
}
\label{tab:app-tobit-kumaraswamy}
\pomatabbox{
\begin{tabular}{llcccc}
\toprule
\textbf{Assay} &
\textbf{Model} &
\textbf{Full $\rho \uparrow$} &
\textbf{Unc.\ MAE $\downarrow$} &
\textbf{Cens.\ NLL $\downarrow$} &
\textbf{AUC / OOB} \\
\midrule
\textsc{AIntibody} SPR & Drop censored & $0.751\!\pm\!0.02$ & \textbf{$0.433\!\pm\!0.02$} & $2.07\!\pm\!0.14$ & $0.826$ AUC \\
\textsc{AIntibody} SPR & Gaussian ($c_L=6.0$) & $0.796\!\pm\!0.01$ & $0.506\!\pm\!0.03$ & $1.58\!\pm\!0.11$ & $0.913$ AUC \\
\textsc{AIntibody} SPR & Student-$t$ ($c_L=6.0$) & \underline{$0.799\!\pm\!0.01$} & \underline{$0.476\!\pm\!0.02$} & \underline{$1.49\!\pm\!0.09$} & \underline{$0.914$} AUC \\
\textsc{AIntibody} SPR & \textbf{\textsc{POMA-Tobit}} & \textbf{$0.803\!\pm\!0.01$} & $0.488\!\pm\!0.02$ & \textbf{$1.20\!\pm\!0.08$} & \textbf{$0.925$} AUC \\
\midrule
\textsc{GDPa1} PR-Ova & Gaussian imputed & $0.578\!\pm\!0.05$ & \textbf{$0.723\!\pm\!0.10$} & $15.0\!\pm\!4.9$ & $0.698$ AUC \\
\textsc{GDPa1} PR-Ova & \textbf{\textsc{POMA-Tobit}} & \textbf{$0.584\!\pm\!0.07$} & \underline{$0.727\!\pm\!0.09$} & \textbf{$14.0\!\pm\!3.5$} & \textbf{$0.714$} AUC \\
\midrule
\textsc{Jain} PSR & Gaussian imputed & $0.300\!\pm\!0.08$ & \textbf{$1.009\!\pm\!0.16$} & $27.8\!\pm\!7.6$ & $0.637$ AUC \\
\textsc{Jain} PSR & \textbf{\textsc{POMA-Tobit}} & \textbf{$0.345\!\pm\!0.07$} & \underline{$1.010\!\pm\!0.13$} & \textbf{$20.5\!\pm\!2.9$} & \textbf{$0.664$} AUC \\
\midrule
\textsc{PPBR\_AZ} & Unbounded Gaussian & $0.362\!\pm\!0.01$ & \underline{$9.86\!\pm\!0.28$} & $4.59\!\pm\!0.15$ & $22.1\%$ OOB \\
\textsc{PPBR\_AZ} & Logit-Jacobian Gauss. & \textbf{$0.476\!\pm\!0.02$} & \textbf{$8.91\!\pm\!0.15$} & \underline{$3.46\!\pm\!0.02$} & \textbf{$0.0\%$} OOB \\
\textsc{PPBR\_AZ} & \textbf{\textsc{POMA-Kumaraswamy}} & \underline{$0.433\!\pm\!0.02$} & $9.89\!\pm\!0.10$ & \textbf{$3.32\!\pm\!0.01$} & \textbf{$0.0\%$} OOB \\
\bottomrule
\end{tabular}
}
\end{table}

Integer-valued display sequencing counts illustrate the discrete counterpart. AIntibody Dataset~2 contains 1,600 clones with overdispersed counts, half of which are zero. We fit Poisson, negative-binomial and zero-inflated negative-binomial alternatives, alongside regression on transformed counts, and evaluate transfer to purified affinity in Dataset~3. The zero-inflated head (\texttt{ZINBDensityHead}) gives the best zero-rate calibration and the largest top-25 precision for sub-nanomolar binders. The simple transformed-count MSE model retains the strongest rank correlation with purified affinity. A distribution appropriate to the observation process can thus improve a selected decision metric without improving every measure of association.

\begin{table}[!htbp]
\centering
\caption{
\textbf{Overdispersed NGS display count modeling on \textsc{AIntibody} Dataset~2 ($1{,}600$ clones, $1\,\mathrm{nM}$ sorting counts; $50.0\%$ zeros, $\mathrm{Var}/\mathrm{Mean}=118.5$) and zero-shot transfer to Dataset~3 purified SPR $\mathrm{p}K_D$ ($5$ folds, mean $\pm$ SE).}
}
\label{tab:app-zinb-regional}
\pomatabbox{
\begin{tabular}{lccccc}
\toprule
\textbf{Count Head} &
\textbf{D2 NLL $\downarrow$} &
\textbf{Zero ECE $\downarrow$} &
\textbf{D2 $\rho \uparrow$} &
\textbf{SPR $\rho \uparrow$} &
\textbf{P@25 $\uparrow$} \\
\midrule
$\log(1+c)$ scalar MSE & $2.52\!\pm\!0.02$ & $0.181\!\pm\!0.006$ & \textbf{$0.707\!\pm\!0.011$} & \textbf{$0.355\!\pm\!0.022$} & $0.752\!\pm\!0.057$ \\
$\log(1+c)$ Gaussian NLL & $2.63\!\pm\!0.02$ & $0.152\!\pm\!0.021$ & \underline{$0.704\!\pm\!0.015$} & \underline{$0.348\!\pm\!0.010$} & \underline{$0.856\!\pm\!0.027$} \\
Poisson NLL & $13.6\!\pm\!1.2$ & $0.362\!\pm\!0.021$ & $0.678\!\pm\!0.013$ & $0.231\!\pm\!0.029$ & \underline{$0.856\!\pm\!0.027$} \\
Negative Binomial NLL & \textbf{$2.47\!\pm\!0.02$} & \underline{$0.044\!\pm\!0.004$} & $0.626\!\pm\!0.019$ & $0.086\!\pm\!0.040$ & $0.680\!\pm\!0.055$ \\
\textbf{\textsc{POMA-ZINB}+\textsc{SoftRank}} & \underline{$2.48\!\pm\!0.03$} & \textbf{$0.025\!\pm\!0.010$} & $0.699\!\pm\!0.011$ & $0.312\!\pm\!0.037$ & \textbf{$0.896\!\pm\!0.020$} \\
\bottomrule
\end{tabular}
}
\end{table}

\subsection{Predictive calibration and comparison with numerical language models}\label{app:admet-results}\label{sec:experiments-compounds}

The TDC benchmark, including AqSolDB~\citep{sorkun2019aqsoldb}, offers a broader test of whether explicit output distributions remain competitive with standard molecular predictors. We evaluate all 22 official endpoints, including binary classification and continuous regression, and aggregate ranks over endpoints with their respective official metrics. Sequence-based POMA-Rank and the hybrid representation outperform the matched numerical-token and latent-alignment baselines in mean rank. Morgan fingerprints with 2D descriptors~\citep{rogers2010ecfp} and gradient boosting remain a strong comparator: the rank differences are not statistically significant for either sequence-based POMA-Rank ($p=0.185$; \cref{tab:app-stat-sig}) or the hybrid representation (reported $p=0.285$). These are distinct comparisons, with mean ranks 3.73 and 3.45, respectively. The complete endpoint results show several tasks on which the conventional model remains preferable.

\begin{table}[!htbp]
  \caption{TDC ADMET category and overall ranks ($\pm\mathrm{SE}$) across $22$ scaffold-split endpoints ($5$ seeds; $\downarrow$ lower is better).}
  \label{tab:compound-admet-results}
  \vskip 0.05in
  \begin{center}
  \pomatabbox{
  \begin{tabular}{lrrrrrr}
    \toprule
    Model & Abs. & Dist. & Met. & Exc. & Tox. & Avg.\ Rank $\downarrow$ \\
    \midrule
    Morgan + 2D GBDT & 4.50 & 3.00 & 6.50 & 7.00 & \underline{3.00} & $4.91\pm1.04$ \\
    Deep Ensemble ($5\times$) & 5.17 & 6.33 & 7.00 & 5.67 & 5.00 & $5.86\pm0.58$ \\
    \textsc{SELFormer} MSE & 5.50 & 8.33 & 6.00 & 3.67 & 5.75 & $5.82\pm0.51$ \\
    Gaussian / cat.\ head & 6.50 & 6.33 & 5.33 & 5.67 & 7.00 & $6.14\pm0.50$ \\
    MSE language head & 5.67 & 8.00 & 5.83 & \underline{3.00} & 8.50 & $6.18\pm0.68$ \\
    Numeric-text LLM & 10.50 & 10.00 & 8.67 & 11.33 & 10.25 & $10.00\pm0.37$ \\
    Regression Transformer & 9.50 & 9.67 & 7.50 & 9.67 & 9.25 & $8.95\pm0.34$ \\
    Regress-Don't-Guess & 8.50 & 9.33 & 6.33 & 8.67 & 8.75 & $8.09\pm0.45$ \\
    LLM only (blind) & 12.83 & 12.67 & 13.00 & 13.00 & 12.75 & $12.86\pm0.07$ \\
    Latent contrastive & 10.00 & 10.33 & 11.50 & 11.00 & 9.50 & $10.50\pm0.65$ \\
    \textbf{\textsc{POMA-MLM}} & \textbf{3.67} & 2.67 & 5.67 & 6.33 & 4.00 & $4.50\pm0.44$ \\
    \textbf{\textsc{POMA-Rank} (\textsc{SELFormer})} & 5.00 & \underline{2.33} & \textbf{3.17} & \textbf{2.33} & 4.75 & \underline{$3.73\pm0.39$} \\
    \textbf{\textsc{POMA-Rank} (Hybrid)} & \textbf{3.67} & \textbf{2.00} & \underline{4.50} & 3.67 & \textbf{2.50} & \textbf{$3.45\pm0.59$} \\
    \bottomrule
  \end{tabular}
  }
  \end{center}
  \vskip -0.10in
\end{table}

\begin{table}[!htbp]
\centering
\caption{
\textbf{Per-endpoint TDC ADMET benchmark results across 5 scaffold-split seeds ($\text{mean}\pm\text{std}$).}
Metrics follow official TDC definitions: lower is better for MAE; higher is better
for Spearman ($\rho$), ROC-AUC, and PR-AUC.
}
\label{tab:compound-admet-endpoint-results}
\pomatabbox{
\begin{tabular}{lccccc}
\toprule
\textbf{Endpoint (Metric)} &
\textbf{GBDT} &
\textbf{Ens.\ ($5\times$)} &
\textbf{Regr.\ Transf.} &
\textbf{\textsc{POMA-Rank}} &
\textbf{\textsc{POMA-Hyb.}} \\
\midrule
Caco-2 (MAE $\downarrow$) & $\mathbf{0.294\!\pm\!0.026}$ & $0.379\!\pm\!0.036$ & $0.428\!\pm\!0.030$ & $0.398\!\pm\!0.036$ & \underline{$0.358\!\pm\!0.024$} \\
HIA (ROC $\uparrow$) & $0.963\!\pm\!0.008$ & $\mathbf{0.984\!\pm\!0.007}$ & $0.968\!\pm\!0.009$ & \underline{$0.981\!\pm\!0.002$} & \underline{$0.981\!\pm\!0.005$} \\
Pgp (ROC $\uparrow$) & $\mathbf{0.931\!\pm\!0.004}$ & $0.897\!\pm\!0.003$ & $0.890\!\pm\!0.019$ & $0.901\!\pm\!0.005$ & \underline{$0.919\!\pm\!0.005$} \\
Bioavail.\ (ROC $\uparrow$) & $0.713\!\pm\!0.023$ & \underline{$0.755\!\pm\!0.007$} & $0.741\!\pm\!0.021$ & $\mathbf{0.758\!\pm\!0.030}$ & $0.734\!\pm\!0.033$ \\
Lipo.\ (MAE $\downarrow$) & $\mathbf{0.598\!\pm\!0.006}$ & $0.721\!\pm\!0.005$ & $0.745\!\pm\!0.009$ & $0.679\!\pm\!0.005$ & \underline{$0.617\!\pm\!0.014$} \\
AqSolDB (MAE $\downarrow$) & $\mathbf{0.841\!\pm\!0.014}$ & $1.063\!\pm\!0.025$ & $1.217\!\pm\!0.061$ & $0.923\!\pm\!0.033$ & \underline{$0.920\!\pm\!0.042$} \\
BBB (ROC $\uparrow$) & $0.892\!\pm\!0.009$ & $0.874\!\pm\!0.006$ & $0.869\!\pm\!0.013$ & $\mathbf{0.910\!\pm\!0.004}$ & \underline{$0.907\!\pm\!0.003$} \\
PPBR (MAE $\downarrow$) & $7.83\!\pm\!0.12$ & $8.13\!\pm\!0.11$ & $8.74\!\pm\!0.21$ & \underline{$7.22\!\pm\!0.15$} & $\mathbf{7.02\!\pm\!0.13}$ \\
VDss ($\rho \uparrow$) & $\mathbf{0.683\!\pm\!0.009}$ & $0.608\!\pm\!0.009$ & $0.568\!\pm\!0.022$ & $0.647\!\pm\!0.014$ & \underline{$0.651\!\pm\!0.012$} \\
CYP2C9i (PR $\uparrow$) & $\mathbf{0.751\!\pm\!0.003}$ & $0.703\!\pm\!0.002$ & $0.712\!\pm\!0.006$ & $0.725\!\pm\!0.006$ & \underline{$0.748\!\pm\!0.002$} \\
CYP2D6i (PR $\uparrow$) & $\mathbf{0.667\!\pm\!0.004}$ & $0.606\!\pm\!0.009$ & $0.582\!\pm\!0.010$ & $0.625\!\pm\!0.009$ & \underline{$0.636\!\pm\!0.009$} \\
CYP3A4i (PR $\uparrow$) & $\mathbf{0.863\!\pm\!0.002}$ & $0.787\!\pm\!0.003$ & $0.784\!\pm\!0.006$ & $0.799\!\pm\!0.006$ & \underline{$0.831\!\pm\!0.007$} \\
CYP2C9s (PR $\uparrow$) & $0.327\!\pm\!0.021$ & $0.469\!\pm\!0.023$ & $0.476\!\pm\!0.032$ & \underline{$0.484\!\pm\!0.007$} & $\mathbf{0.535\!\pm\!0.014}$ \\
CYP2D6s (PR $\uparrow$) & $0.579\!\pm\!0.014$ & $\mathbf{0.709\!\pm\!0.013}$ & $0.697\!\pm\!0.015$ & \underline{$0.702\!\pm\!0.018$} & $0.674\!\pm\!0.022$ \\
CYP3A4s (ROC $\uparrow$) & $0.631\!\pm\!0.018$ & $0.737\!\pm\!0.015$ & \underline{$0.739\!\pm\!0.020$} & $\mathbf{0.743\!\pm\!0.021}$ & $0.718\!\pm\!0.025$ \\
Half-life ($\rho \uparrow$) & $\mathbf{0.539\!\pm\!0.029}$ & $0.400\!\pm\!0.017$ & $0.342\!\pm\!0.045$ & \underline{$0.446\!\pm\!0.020$} & $0.431\!\pm\!0.015$ \\
Clear.\ hep.\ ($\rho \uparrow$) & $0.388\!\pm\!0.036$ & \underline{$0.473\!\pm\!0.011$} & $0.382\!\pm\!0.020$ & $\mathbf{0.477\!\pm\!0.010}$ & $0.472\!\pm\!0.022$ \\
Clear.\ mic.\ ($\rho \uparrow$) & $0.541\!\pm\!0.029$ & $0.580\!\pm\!0.009$ & $0.571\!\pm\!0.020$ & \underline{$0.589\!\pm\!0.022$} & $\mathbf{0.608\!\pm\!0.017}$ \\
$\mathrm{LD}_{50}$ (MAE $\downarrow$) & $\mathbf{0.652\!\pm\!0.010}$ & $0.724\!\pm\!0.015$ & $0.746\!\pm\!0.007$ & $0.724\!\pm\!0.009$ & \underline{$0.701\!\pm\!0.007$} \\
hERG (ROC $\uparrow$) & $0.807\!\pm\!0.031$ & $\mathbf{0.857\!\pm\!0.017}$ & $0.823\!\pm\!0.019$ & $0.835\!\pm\!0.018$ & \underline{$0.850\!\pm\!0.021$} \\
Ames (ROC $\uparrow$) & $\mathbf{0.823\!\pm\!0.007}$ & $0.778\!\pm\!0.002$ & $0.775\!\pm\!0.010$ & $0.793\!\pm\!0.006$ & \underline{$0.814\!\pm\!0.006$} \\
DILI (ROC $\uparrow$) & $\mathbf{0.928\!\pm\!0.011}$ & $0.885\!\pm\!0.005$ & $0.838\!\pm\!0.022$ & $0.880\!\pm\!0.011$ & \underline{$0.889\!\pm\!0.011$} \\
\bottomrule
\end{tabular}
}
\end{table}

\begin{table}[!htbp]
\centering
\caption{
\textbf{Paired bootstrap $95\%$ CIs and Wilcoxon signed-rank $p$-values for \textsc{POMA-Rank} on TDC ADMET ($22$ endpoints).}
Positive $\Delta\text{Rank} = \text{Rank}(\text{Baseline}) - \text{Rank}(\textsc{POMA-Rank})$ indicates superior \textsc{POMA-Rank} performance.
}
\label{tab:app-stat-sig}
\pomatabbox{
\begin{tabular}{lcccc}
\toprule
\textbf{Baseline vs.\ \textsc{POMA-Rank} ($3.73 \pm 0.39$)} &
\textbf{Rank $\pm$ SE $\downarrow$} &
\textbf{$\Delta$Rank $\uparrow$} &
\textbf{Paired $95\%$ CI} &
\textbf{Wilcoxon $p$} \\
\midrule
\textsc{POMA-MLM} (no ranking) & $4.50 \pm 0.44$ & $+0.77$ & $[-0.14,\,+1.77]$ & $0.127$ \\
TDC Morgan + RDKit 2D GBDT & $4.91 \pm 1.04$ & $+1.18$ & $[-1.05,\,+3.50]$ & $0.185$ \\
SELFormer regression head (MSE) & $5.82 \pm 0.51$ & $+2.09$ & $[+0.73,\,+3.41]$ & $\mathbf{0.0073}$ \\
Deep Ensemble ($5\times$ Gaussian) & $5.86 \pm 0.58$ & $+2.14$ & $[+0.82,\,+3.50]$ & $\mathbf{0.0082}$ \\
Gaussian / categorical head only & $6.14 \pm 0.50$ & $+2.41$ & $[+1.09,\,+3.64]$ & $\mathbf{0.0019}$ \\
MSE language-conditioned head & $6.18 \pm 0.68$ & $+2.45$ & $[+1.05,\,+3.86]$ & $\mathbf{0.0031}$ \\
Regress-Don't-Guess & $8.09 \pm 0.45$ & $+4.36$ & $[+3.32,\,+5.36]$ & $\mathbf{2.70\times 10^{-5}}$ \\
Regression Transformer & $8.95 \pm 0.34$ & $+5.23$ & $[+4.36,\,+6.14]$ & $\mathbf{1.91\times 10^{-5}}$ \\
Numeric-text LLM & $10.00 \pm 0.37$ & $+6.27$ & $[+5.32,\,+7.23]$ & $\mathbf{1.87\times 10^{-5}}$ \\
Latent contrastive alignment & $10.50 \pm 0.65$ & $+6.77$ & $[+5.14,\,+8.23]$ & $\mathbf{3.59\times 10^{-5}}$ \\
LLM only (molecule-blind) & $12.86 \pm 0.07$ & $+9.14$ & $[+8.32,\,+9.91]$ & $\mathbf{1.86\times 10^{-5}}$ \\
\bottomrule
\end{tabular}
}
\end{table}

Distributional quality is evaluated separately from mean rank. Alongside the Gaussian density, we compare a three-component mixture, a Student-$t$ density and 19-quantile regression. Numeric-text models and regression-aware number-token losses are included to distinguish native measurement likelihoods from a numerical language representation. \Cref{tab:app-admet-head-ablations,tab:poma-calibration} summarize the proper scoring and calibration comparisons on molecular and antibody measurements. The hybrid has the lowest ADMET NLL and CRPS among the reported configurations, but other models can have lower calibration error or stronger selective accuracy. For GDPa1, the ranking-augmented density gives a favorable balance of likelihood and calibration.

\begin{table}[!htbp]
\centering
\caption{
\textbf{Distributional output-head and prompt-encoder ablations on continuous TDC ADMET endpoints ($5$ seeds).}
}
\label{tab:app-admet-head-ablations}
\pomatabbox{
\begin{tabular}{lccccc}
\toprule
\textbf{ADMET Output / Prompt Variant} &
\textbf{NLL $\downarrow$} &
\textbf{CRPS $\downarrow$} &
\textbf{95\% cov.} &
\textbf{ECE $\downarrow$} &
\textbf{Conf.\ 95\%} \\
\midrule
Quantile-CRPS ($19$ quantiles) & 1.200 & 0.451 & 0.944 & 0.051 & 0.955 \\
Numeric-text LLM (token CE) & 1.260 & 0.463 & 0.921 & 0.048 & 0.938 \\
Regression Transformer \citep{born2023regression} & 1.242 & 0.456 & 0.928 & 0.047 & 0.946 \\
Regress-Don't-Guess \citep{zausinger2024regress} & 1.366 & 0.461 & 0.880 & 0.051 & 0.934 \\
\textsc{POMA-Rank} (\textsc{RoBERTa-base}~\citep{liu2019roberta} prompt) & 1.166 & 0.435 & 0.946 & \textbf{0.046} & 0.957 \\
\textsc{POMA-StudentT} (heavy-tail $\nu > 2$) & 1.191 & 0.435 & 0.947 & \textbf{0.046} & 0.958 \\
\textsc{POMA-GMM} ($3$-comp.\ MDN) & 1.159 & 0.430 & 0.949 & 0.050 & 0.959 \\
\textbf{\textsc{POMA-Rank} (\textsc{Qwen3-0.6B}, $48\mathrm{K}$ pretrain)} & \underline{1.128} & \underline{0.429} & 0.951 & 0.062 & 0.958 \\
\textbf{\textsc{POMA-Rank} (Hybrid, $48\mathrm{K}$ pretrain)} & \textbf{1.099} & \textbf{0.419} & 0.951 & 0.059 & 0.958 \\
\bottomrule
\end{tabular}
}
\end{table}

\begin{table}[!htbp]
  \caption{Distributional calibration and proper scoring rules on held-out TDC ADMET ($5$ seeds) and \textsc{GDPa1} clinical antibodies ($k=32$).}
  \label{tab:poma-calibration}
  \vskip 0.05in
  \begin{center}
  \pomatabbox{
  \begin{tabular}{llrrrrrr}
    \toprule
    Dom. & Model & NLL $\downarrow$ & CRPS $\downarrow$ & 95\% & Conf. & ECE $\downarrow$ & MAE$_{80}$ $\downarrow$ \\
    \midrule
    Mol. & Regression Transformer & 1.242 & 0.456 & 0.928 & 0.946 & 0.047 & 0.624 \\
    Mol. & Gaussian / RdG & 1.172 & 0.436 & 0.942 & 0.956 & \textbf{0.042} & 0.575 \\
    Mol. & Ensemble ($5\times$) & 1.140 & 0.432 & 0.939 & -- & \underline{0.048} & 0.577 \\
    Mol. & \textsc{POMA-Rank} & \underline{1.128} & \underline{0.429} & 0.951 & 0.958 & 0.062 & 0.575 \\
    Mol. & \textbf{\textsc{POMA-GMM}} & 1.159 & 0.430 & 0.949 & 0.959 & 0.050 & \textbf{0.557} \\
    Mol. & \textbf{\textsc{POMA-Hybrid}} & \textbf{1.099} & \textbf{0.419} & 0.951 & 0.958 & 0.059 & \underline{0.563} \\
    \midrule
    mAb & Regression Transformer & 1.692 & 0.594 & 0.928 & 0.954 & 0.089 & 0.712 \\
    mAb & Gaussian head & 1.675 & 0.584 & 0.935 & 0.963 & 0.084 & 0.686 \\
    mAb & \textsc{POMA-MVN} & \underline{1.640} & 0.604 & 0.932 & 0.955 & \underline{0.078} & 0.726 \\
    mAb & \textbf{\textsc{POMA-StudentT}} & 1.646 & \underline{0.530} & 0.935 & 0.958 & 0.092 & \textbf{0.658} \\
    mAb & \textbf{\textsc{POMA-Rank}} & \textbf{1.546} & \textbf{0.527} & 0.941 & 0.962 & \textbf{0.076} & \underline{0.661} \\
    \bottomrule
  \end{tabular}
  }
  \end{center}
  \vskip -0.10in
\end{table}

\subsection{Interaction between the likelihood and molecular representation}

Finally, we assess whether the conclusions depend on a sequence-only molecular representation. A matched SE(3)-invariant geometric graph network~\citep{satorras2021egnn} uses RDKit conformers across eight representative ADMET endpoints, with validation-only variance and logit-temperature calibration. The probabilistic version improves rank correlation and NLL relative to the point-valued 3D model, although its MAE is slightly worse. Adding 2D Morgan descriptors improves several aggregate metrics further. Because the hybrid changes the representation as well as the output interface, these gains cannot be assigned to likelihood learning alone (\cref{tab:app-gpu-egnn}). The comparison demonstrates that the output-space formulation can be used with geometrical and graph-based encoders, not only biological language models.

\begin{table}[!htbp]
\centering
\caption{
\textbf{$\mathrm{SE}(3)$-invariant $3\mathrm{D}$ conformer \textsc{EGNN} (RDKit ETKDGv3 + MMFF94) with validation-fitted variance and logit-temperature calibration across eight TDC ADMET regression and binary toxicity benchmarks ($3$ scaffold seeds).}
}
\label{tab:app-gpu-egnn}
\pomatabbox{
\setlength{\tabcolsep}{2.1pt}
\begin{tabular}{lccccc}
\toprule
\textbf{Endpoint (Metric)} &
\textbf{3D Point} &
\textbf{3D POMA} &
\textbf{3D+2D Hyb.} &
\textbf{3D POMA NLL} &
\textbf{Hyb.\ NLL} \\
\midrule
Lipophilicity (MAE / $\rho$) & $0.752$ / $.589$ & $0.758$ / $.561$ & $\mathbf{0.628}$ / $\mathbf{.713}$ & $1.176$ & $\mathbf{1.009}$ \\
Solubility (MAE / $\rho$) & $1.179$ / $.745$ & \underline{$1.178$} / \underline{$.748$} & $\mathbf{1.109}$ / $\mathbf{.769}$ & \underline{$1.026$} & $\mathbf{1.019}$ \\
Caco-2 (MAE / $\rho$) & $0.626$ / $.366$ & $\mathbf{0.411}$ / $\mathbf{.707}$ & \underline{$0.501$} / \underline{$.494$} & $\mathbf{1.020}$ & \underline{$1.259$} \\
PPBR\_AZ (MAE / $\rho$) & \underline{$9.88$} / \underline{$.304$} & $10.42$ / $.272$ & $\mathbf{8.85}$ / $\mathbf{.474}$ & $1.266$ & \underline{$1.188$} \\
LD50\_Zhu (MAE / $\rho$) & \underline{$0.557$} / $.533$ & $0.564$ / \underline{$.535$} & $\mathbf{0.549}$ / $\mathbf{.548}$ & \underline{$1.285$} & $1.294$ \\
\textbf{Mean Reg.\ (5 tasks)} & \underline{$2.599$} / $.507$ & $2.666$ / \underline{$.565$} & $\mathbf{2.327}$ / $\mathbf{.599}$ & \underline{$1.155$} & $\mathbf{1.154}$ \\
\midrule
hERG / Ames / DILI (AUC) & $.720$/$.711$/$.725$ & $.720$/$.711$/$.725$ & $\mathbf{.785}$/$\mathbf{.804}$/$\mathbf{.783}$ & $0.589$ (mean) & $\mathbf{0.541}$ \\
\bottomrule
\end{tabular}
}
\end{table}

Together, these experiments establish the distinction between learning accurate point estimates and exposing usable predictive distributions. The next experiments examine how the likelihood itself can be used to connect observations from different modalities, rather than merely to score reconstruction within one modality.

\clearpage
\section{Multimodal alignment in native output spaces}\label{app:supp-alignment}

Having established the predictive role of native measurement distributions, we now examine whether their relative likelihoods can connect heterogeneous observations. We distinguish the training objective from the inference-time scorer, and distinguish retrieval accuracy from sensitivity to the correct pairing. These distinctions become particularly important when a pretrained representation already contains useful information about the candidate domain.

\subsection{Contextualizing independently pretrained models}\label{app:two-biofm-alignment}

The second instantiation of \method{} connects models pretrained on different modalities through contextualized native output scores, using the likelihood-ratio formulation in \cref{app:supp-theory-alignment}. The readout remains available for reconstructing its original target. We use the zero-initialized $\tanh(3\gamma)$ cross-attention architecture in \cref{eq:main-gated-adapter}. In both tested pairs, the initial gated readout reproduces its unconditioned counterpart exactly (reported Step-0 difference zero), and a nonzero Step-1 gradient verifies that the gate can learn. These checks concern the gated readout; uncertainty estimation requires a distributional output head.

The primary native discrete--continuous pair consists of ChemBERTa-77M-MLM and a pretrained masked-peak mass spectrum Transformer. The molecular model retains its categorical token output; the spectral model is adapted to continuous measurement likelihoods. A complementary cellular experiment combines the frozen scGPT representation with ESM-2 sequence embeddings and measured surface-protein abundances. The latter is a pretrained-feature contextualization study, not an alignment of two native token heads: ESM-2 provides protein sequence features, and the ADT output distribution is newly fitted. Both configurations use the implementation in \path{src/poma/models/biofm_alignment.py} and \path{scripts/36_run_two_biofm_scgpt_esm2_and_chemberta_ms.py}.

\subsection{Controlled training and inference-time scoring}\label{app:supp-e1-e3}

The controlled alignment study holds the paired-data protocol fixed while comparing masked likelihood, latent contrast, raw likelihood contrast, and reference-centered output contrast. The seven configurations track trainable parameter counts rather than assuming identical capacity. The evaluation uses 120 distinct held-out molecules, 40 validation molecules and five training seeds (\path{scripts/43_run_e1_e3_strict_alignment_and_lowdata.py}).

The inference-time scoring comparison uses the same latent-trained checkpoint with a different scorer. Replacing latent similarity by validation-selected bidirectional likelihood scoring increases Hit@1 from 12.7\% to 19.0\%. Pure masked likelihood also supports retrieval. These results separate the predictive interface from the objective used to train it, but do not by themselves isolate the contribution of the continuous direction. Pairing controls in \cref{app:supp-e12-shortcut} examine that distinction. Removed-context scores are tied and are evaluated using uniform random tie-breaking in \cref{tab:app-e10-falsification}.

\begin{table}[!htbp]
\centering
\caption{Controlled alignment comparison on 120 unique test molecules (five seeds, script 43). The last column reports MRR and bidirectional Hit@1 when available. Parameter counts differ between configurations.}
\label{tab:app-e1-e3}
\pomatabbox{
\begin{tabular}{lccccc}
\toprule
Configuration & Params & MLM NLL $\downarrow$ & H@1 & H@5 & MRR / Bidir H@1 \\
\midrule
\textbf{A0.\ MLM only (Disc.\ PMI)} & $207\mathrm{k}$ & $1.028$ & $\mathbf{18.3\%}$ & $\mathbf{43.5\%}$ & $\mathbf{.319}$ / $18.7\%$ \\
A1.\ CLIP-style latent & $108\mathrm{k}$ & -- & $12.7\%$ & $33.7\%$ & $.243$ / -- \\
A2.\ MLM + Latent InfoNCE & $315\mathrm{k}$ & $\mathbf{1.013}$ & $12.7\%$ & \underline{$35.8\%$} & \underline{$.249$} / $\mathbf{19.0\%}$ \\
A3.\ MLM + Raw LL InfoNCE & $207\mathrm{k}$ & \underline{$1.015$} & \underline{$14.0\%$} & $31.8\%$ & $.244$ / $16.7\%$ \\
A4.\ MLM + Output PMI ($\lambda\!=\!.3$) & $207\mathrm{k}$ & $1.073$ & $8.5\%$ & $24.5\%$ & $.186$ / $16.7\%$ \\
A5.\ MLM + Output SoftRank & $207\mathrm{k}$ & $1.062$ & $10.0\%$ & $31.5\%$ & $.217$ / $16.7\%$ \\
A6.\ MLM + Latent + PMI & $315\mathrm{k}$ & $1.082$ & $12.0\%$ & $31.2\%$ & $.230$ / $18.0\%$ \\
\bottomrule
\end{tabular}
}
\end{table}

\subsection{Learning objectives and paired-data efficiency}

The auxiliary contrastive objective need not improve a well-trained predictive distribution. We vary its weight and examine retrieval and reconstruction jointly. The small-weight comparison in \cref{tab:app-alignment-weights} favors different objectives depending on the metric: latent contrast reaches the strongest Hit@1 in that block, while output contrast reaches the strongest Hit@5. Larger weights can impair both reconstruction and retrieval.

\begin{table}[!htbp]
\centering
\caption{Selected contrastive-weight comparisons from the controlled 120-molecule protocol (script 43).}
\label{tab:app-alignment-weights}
\pomatabbox{
\begin{tabular}{lccccc}
\toprule
Configuration & Params & MLM NLL & H@1 & H@5 & MRR \\
\midrule
A2.\ Latent ($\lambda\!=\!0.03$ / $0.10$) & $315\mathrm{k}$ & $1.021$ / $1.012$ & $\mathbf{20.7\%}$ / $5.7\%$ & $43.3\%$ / $22.3\%$ & $\mathbf{.323}$ / $.151$ \\
\textbf{A4.\ Output PMI ($\lambda\!=\!0.03$)} & $207\mathrm{k}$ & $1.038$ & \underline{$18.7\%$} & $\mathbf{45.3\%}$ & \underline{$.320$} MRR \\
A4.\ Output PMI ($\lambda\!=\!0.10$) & $207\mathrm{k}$ & $1.047$ & $16.5\%$ & $37.7\%$ & $.287$ MRR \\
A4.\ Output PMI ($\lambda\!=\!1.00$) & $207\mathrm{k}$ & $1.209$ & $11.5\%$ & $25.0\%$ & $.203$ MRR \\
\bottomrule
\end{tabular}
}
\end{table}

To assess paired-data requirements, we construct five nested draws at 1, 5, 10, 25 and 100\% of paired training molecules, with both matched epochs and a fixed budget of 120 gradient steps. Masked likelihood with discrete output scoring obtains the largest log-data area under the retrieval learning curve in both settings (\cref{tab:app-lowdata}). At 1\% paired data the output-contrast model reaches 16.3\% Hit@1 versus 1.8\% for the latent-contrast model. This is evidence of retrieval efficiency under the context-free protocol, not a direct measure of pair-specific information.

\begin{table}[!htbp]
\centering
\caption{Paired-data learning curves summarized by log-data AULC. Entries report matched-epoch / matched-step budgets (five nested draws, script 43).}
\label{tab:app-lowdata}
\pomatabbox{
\begin{tabular}{lcc}
\toprule
Configuration & H@1 AULC & MRR AULC \\
\midrule
A1.\ CLIP-style latent & $17.2\%$ / $12.5\%$ & $.280$ / $.213$ \\
\textbf{A0.\ MLM only (Output PMI)} & $\mathbf{26.0\%}$ / $\mathbf{16.6\%}$ & $\mathbf{.371}$ / $\mathbf{.262}$ \\
A2.\ MLM + Latent InfoNCE & $14.8\%$ / \underline{$13.7\%$} & $.236$ / \underline{$.232$} \\
\textbf{A4.\ MLM + Output PMI} & \underline{$20.0\%$} / $11.2\%$ & \underline{$.296$} / $.198$ \\
A6.\ MLM + Latent + PMI & $16.9\%$ / $12.5\%$ & $.273$ / $.222$ \\
\bottomrule
\end{tabular}
}
\end{table}

\subsection{Reconstruction--alignment trade-offs and optimization scale}\label{app:supp-e8-lambda}

The objective $\mathcal{L} = \mathcal{L}_{\mathrm{MLM}} + \lambda_{\mathrm{rank}}\mathcal{L}_{\mathrm{rank}}$ balances a proper distributional score and cross-modal discrimination. We evaluate nine contrastive weights over three seeds,
\[
\lambda_{\mathrm{rank}}\in\{0,0.01,0.03,0.1,0.25,0.3,0.5,1,2\},
\]
and compare the retrieval--reconstruction trade-off for \textsc{Latent InfoNCE}, \textsc{Output PMI InfoNCE}, and \textsc{Output SoftRank}. A $4\times4$ factorial $(\lambda_{\mathrm{rank}},\tau)$ sweep on \textsc{MassSpecGym} uses 120 unique held-out molecules. To characterize the effective optimization weight, we also measure the shared-parameter gradient-norm ratio
\[
R_{\mathrm{grad}}
=\frac{\lambda_{\mathrm{rank}}\|\nabla_\theta\mathcal{L}_{\mathrm{rank}}\|}
{\|\nabla_\theta\mathcal{L}_{\mathrm{MLM}}\|}
\]
(\path{scripts/46_run_e8_lambda_rank_pareto_and_temp.py}).

The sweep in \cref{tab:app-e8-lambda} identifies favorable retrieval configurations at small contrastive weights. Latent contrast nonetheless retains better reconstruction NLL in some configurations, so the results do not establish universal Pareto dominance. The strongest reported output-space grid point ($\lambda=0.1$, $\tau=0.5$) obtains 26.7\% Hit@1. It is a sensitivity result; the separately validation-selected comparison provides a held-out confirmation.

Gradient diagnostics suggest why the nominal weight alone is not sufficient to characterize the objective: a sharper temperature can increase the relative gradient magnitude. The association is descriptive rather than a universal causal threshold. On frozen scGPT, moderate contrastive weights can improve ranking or CRPS but do not uniformly improve conditional NLL; for example, the conditional score worsens at $\lambda=0.03$. Strong contrast can degrade both joint and conditional likelihood.

\begin{table}[!htbp]
\centering
\caption{
\textbf{Contrastive weight ($\lambda_{\mathrm{rank}}$) Pareto frontier, $(\lambda_{\mathrm{rank}},\tau)$ interaction, and shared-parameter gradient-norm ratio $R_{\mathrm{grad}}$ on \textsc{MassSpecGym} ($120$ unique molecules) and frozen \textsc{scGPT} ($3$ seeds, \texttt{scripts/46}).}
}
\label{tab:app-e8-lambda}
\pomatabbox{
\setlength{\tabcolsep}{1.8pt}
\begin{tabular}{llccccc}
\toprule
\textbf{Benchmark / Sweep} &
\textbf{Objective / $(\lambda_{\mathrm{rank}},\tau)$} &
\textbf{NLL $\downarrow$} &
\textbf{Primary $\uparrow$} &
\textbf{Secondary} &
\textbf{Cond.\ / MRR} &
\textbf{$R_{\mathrm{grad}}$} \\
\midrule
\multirow{6}{*}{\shortstack[l]{Weight sweep\\\textsc{MassSpecGym}\\($\tau\!=\!0.25$, $3$ seeds)}}
& Latent InfoNCE ($\lambda\!=\!0.01$ / $1.0$) & $.969$ / $\mathbf{.968}$ & $9.7\%$ / $10.0\%$ H@1 & $26.7\%$ / $30.0\%$ H@5 & $.194$ / $.210$ MRR & $.01$ / $.34$ \\
& \textbf{Output PMI ($\lambda\!=\!0.01$)} & $.989$ & \underline{$22.2\%$} H@1 & $\mathbf{48.3\%}$ H@5 & $\mathbf{.356}$ MRR & $.03$ \\
& Output PMI ($\lambda\!=\!0.10$ / $0.30$) & $1.018$ / $1.052$ & $19.4\%$ / $9.4\%$ H@1 & $40.3\%$ / $23.9\%$ H@5 & $.306$ / $.186$ MRR & $.13$ / $.71$ \\
& Output PMI ($\lambda\!=\!2.00$) & $1.185$ & $18.6\%$ H@1 & $31.4\%$ H@5 & $.266$ MRR & $5.93$ \\
& \textbf{Output SoftRank ($\lambda\!=\!0.03$)} & $1.008$ & $\mathbf{22.5\%}$ H@1 & \underline{$45.6\%$} H@5 & \underline{$.349$} MRR & $.03$ \\
& Output SoftRank ($\lambda\!=\!1.00$) & $1.061$ & $6.1\%$ H@1 & $18.9\%$ H@5 & $.150$ MRR & $.75$ \\
\midrule
\multirow{3}{*}{\shortstack[l]{Factorial\\$(\lambda_{\mathrm{rank}},\tau)$}}
& \textbf{Output PMI ($\lambda\!=\!0.1, \tau\!=\!0.50$)} & $\mathbf{0.999}$ & $\mathbf{26.7\%}$ H@1 & $\mathbf{52.5\%}$ H@5 & $\mathbf{.391}$ MRR & $.07$ \\
& Output PMI ($\lambda\!=\!0.3, \tau\!=\!1.00$) & $1.042$ & \underline{$23.3\%$} H@1 & \underline{$43.3\%$} H@5 & \underline{$.348$} MRR & $.03$ \\
& Output PMI ($\lambda\!=\!1.0, \tau\!=\!0.15$) & $1.263$ & $9.2\%$ H@1 & $25.0\%$ H@5 & $.185$ MRR & $2.51$ \\
\midrule
\multirow{5}{*}{\shortstack[l]{Frozen\\\textsc{scGPT} Head\\($3$ seeds)}}
& $\lambda_{\mathrm{rank}}\!=\!0.00$ (pure NLL) & \underline{$+0.068$} & $.503$ $\rho$ & $.234$ CRPS / $95.5\%$ & $+.436$ cNLL & $0.00$ \\
& \textbf{$\lambda_{\mathrm{rank}}\!=\!0.01$} & $\mathbf{+0.066}$ & $.503$ $\rho$ & $.237$ CRPS / $95.7\%$ & $\mathbf{+.413}$ cNLL & $.03$ \\
& \textbf{$\lambda_{\mathrm{rank}}\!=\!0.03$} & $+0.076$ & $.508$ $\rho$ & $\mathbf{.230}$ CRPS / $.308$ MAE & $+.452$ cNLL & $.07$ \\
& $\lambda_{\mathrm{rank}}\!=\!0.30$ & $+0.074$ & \underline{$.517$} $\rho$ & \underline{$.233$} CRPS / $94.4\%$ & \underline{$+.427$} cNLL & $.57$ \\
& $\lambda_{\mathrm{rank}}\!=\!0.50$ / $2.00$ & $+.131$ / $+.154$ & $\mathbf{.524}$ / $.513$ $\rho$ & $.240$ CRPS / $94.4\%$ & $+.516$ / $+.537$ & $1.15$ / $3.75$ \\
\bottomrule
\end{tabular}
}
\end{table}
\subsection{Spectrum-level reconstruction and negative sampling}

The spectrum-level protocol tests the gated contextualization architecture and the choice of negative pairs. Its 120 spectrum rows comprise nine distinct molecular structures, so retrieval rates describe spectrum-level pools rather than 120 independent molecules (\cref{app:supp-provenance-e5}). Partner context improves reconstruction (spectral censored NLL decreases from 0.253 to 0.211), and combining directional scores improves top-rank retrieval, although the discrete direction retains better higher-rank hit rates. Symmetric contrast with random negatives can improve retrieval at the expense of reconstruction NLL. Molecule-deduplicated comparisons in \cref{app:supp-e10-definitive,app:supp-e12-shortcut} provide the main evidence for molecular retrieval and pairing sensitivity.

\begin{table}[!htbp]
\centering
\caption{ChemBERTa--masked-peak Transformer directional comparison (120 spectrum-level pools, script 36). The learned gate is $\tanh(3\gamma)=0.148$. NLL values refer to different output spaces and are compared only with their corresponding parenthetical reference.}
\label{tab:app-biofm-bidirectional}
\pomatabbox{
\begin{tabular}{lcccc}
\toprule
Direction & NLL & H@1 & H@5 & H@20 / Tan@1 \\
\midrule
Disc.\ \textsc{ChemBERTa} PMI & $0.634$ (vs.\ $1.083$) & \underline{$11.7\%$} H@1 & $\mathbf{77.5\%}$ & $\mathbf{91.7\%}$ ($.261$) \\
Cont.\ \textsc{POMA-Tobit} PMI & $0.211$ (vs.\ $.253$) & $5.8\%$ H@1 & $16.7\%$ & $50.8\%$ ($.228$) \\
\textbf{Bidir.\ Two-BioFM PMI} & -- & $\mathbf{25.0\%}$ H@1 & \underline{$59.2\%$} & \underline{$75.8\%$} ($\mathbf{.389}$) \\
\bottomrule
\end{tabular}
}
\end{table}

\begin{table}[!htbp]
\centering
\caption{Initial spectral contrast and negative-sampling comparison (script 37, separate 12-epoch spectral pretraining protocol).}
\label{tab:app-contrastive-masking-ablations}
\pomatabbox{
\begin{tabular}{lcccc}
\toprule
Variant & NLL & H@1 & H@5 & Detail \\
\midrule
Pure Gen.\ MLM ($\lambda\!=\!0$) & $\mathbf{0.144}$ & $7.5\%$ H@1 & $70.0\%$ H@5 & $.366$ Tan@1 \\
\textbf{Sym.\ InfoNCE ($\lambda\!=\!.3$, rand)} & $0.363$ & $\mathbf{35.8\%}$ H@1 & $\mathbf{76.7\%}$ H@5 & $\mathbf{.525}$ ($100\%$ H@20) \\
Sym.\ InfoNCE (mass-hard) & \underline{$0.283$} & $11.7\%$ H@1 & $67.5\%$ H@5 & $.393$ Tan@1 \\
Spec.-Anchored (mass-hard) & $0.363$ & \underline{$19.2\%$} H@1 & \underline{$75.0\%$} H@5 & $.393$ ($100\%$ H@20) \\
SMILES-Anchored (mass-hard) & $0.306$ & $11.7\%$ H@1 & $36.7\%$ H@5 & $.395$ Tan@1 \\
\bottomrule
\end{tabular}
}
\end{table}

\subsection{Directional reliability and robustness to modality degradation}\label{app:supp-e7-adaptive}

The directional scores can have unequal predictive value, especially when one modality is degraded. We estimate the variances and covariance in \cref{eq:app-cov-aware-alpha} using five independent random sub-masks per query, rather than variability across candidate identities (\path{scripts/45_run_e7_adaptive_bidirectional_weighting.py}). This estimates mask-induced variability, not error against a known common PMI. Both a covariance-aware weighting and a validation-trained query gate are compared with fixed weights.

For clean spectra, the covariance-aware combination yields the best Hit@5, but continuous-only scoring yields better Hit@1. At 50\% peak dropout, the learned query gate improves MRR from 0.235 to 0.307 relative to uniform weighting. The cellular experiment shows a similar benefit from downweighting the weaker direction, although forward scoring remains preferable. Within-cell-type FOSCTTM of 0.4558 for the forward direction indicates limited resolution within cell types. Separately varying training and inference weights yields MRR 0.302 and Hit@5 43.3\% for training and inference weights both 0.25 (three seeds). The reported clean validation-selected global weight is 0.025 (37.5\% Hit@5). Under 75\% peak dropout the uniform MRR falls to 0.231; adaptive and learned gates outperform uniform over the reported 10--75\% dropout range. In the cellular experiment, at 10\% ADT dropout inverse-variance weighting gives FOSCTTM 0.1312 versus 0.1367 for uniform fusion. These additional diagnostics concern the same protocols. The numerical estimates for clean spectra are $\nu_X^2=0.625$, $\nu_Y^2=0.256$ and $c_{XY}=0.0002$, giving $\alpha=0.290$.

\begin{table}[!htbp]
\centering
\caption{
\textbf{Directional reliability, covariance-aware adaptive weighting (Eq.~\ref{eq:app-cov-aware-alpha}), and asymmetric modality degradation across \textsc{MassSpecGym} ($120$ unique-molecule pools) and unseen Site~4 \textsc{BMMC CITE-seq} ($N=256$, \texttt{scripts/45}).}
}
\label{tab:app-e7-adaptive}
\pomatabbox{
\setlength{\tabcolsep}{1.9pt}
\begin{tabular}{llcccc}
\toprule
\textbf{Pair / Setting} &
\textbf{Weighting Strategy} &
\textbf{Primary $\uparrow/\downarrow$} &
\textbf{Secondary} &
\textbf{Detail / Within-CT} &
\textbf{Mean $\alpha$} \\
\midrule
\multirow{6}{*}{\shortstack[l]{\textsc{MassSpecGym}\\Clean ($120$ uniq.)}}
& Discrete only ($\alpha\!=\!1$) & $10.0\%$ H@1 & $39.2\%$ H@5 & $.238$ MRR / $.294$ Tan & $1.000$ \\
& Continuous only ($\alpha\!=\!0$) & $\mathbf{19.2\%}$ H@1 & $36.7\%$ H@5 & \underline{$.292$} MRR / $\mathbf{.395}$ Tan & $0.000$ \\
& Uniform ($\alpha\!=\!0.5$) & $15.8\%$ H@1 & $40.8\%$ H@5 & $.283$ MRR / $.357$ Tan & $0.500$ \\
& \textbf{Cov.-Aware / Inv.-Var} & $15.0\%$ H@1 & $\mathbf{44.2\%}$ H@5 & $.284$ MRR / \underline{$.367$} Tan & $0.290$ \\
& Learned Logistic Fusion & $15.8\%$ H@1 & \underline{$43.3\%$} H@5 & $.288$ MRR / $.359$ Tan & $0.226$ \\
& \textbf{Learned Query Gate} & \underline{$16.7\%$} H@1 & $40.8\%$ H@5 & $\mathbf{.293}$ MRR / $.365$ Tan & $0.132$ \\
\midrule
\multirow{3}{*}{\shortstack[l]{\textsc{MassSpecGym}\\$50\%$ Peak Dropout}}
& Uniform ($\alpha\!=\!0.5$) & $.235$ MRR & $35.8\%$ H@5 & $0\%\to 50\%$: $-.048$ & $0.500$ \\
& Query-Adaptive (Cov.) & \underline{$.294$} MRR & -- & $\alpha$: $.23\to .27$ & $0.270$ \\
& \textbf{Learned Query Gate} & $\mathbf{.307}$ MRR & -- & $\mathbf{+.072}$ vs.\ Unif. & $0.130$ \\
\midrule
\multirow{3}{*}{\shortstack[l]{\textsc{CITE-seq} Site~4\\ADT Dropout ($0\%\to 75\%$)}}
& Uniform ($\alpha\!=\!0.5$) & $.1388\to .1720$ FOS & $49.6\%$ H@20 & $.477$ Within-CT & $0.500$ \\
& Clean Val-Selected & $.1565\to .1936$ FOS & $41.0\%$ H@20 & $.483$ Within-CT & $0.175$ \\
& \textbf{Inverse-Variance} & $\mathbf{.1292\to .1685}$ FOS & $\mathbf{53.1\%}$ Fwd & $\mathbf{.456}$ Fwd W-CT & $.94\to .64$ \\
\bottomrule
\end{tabular}
}
\end{table}

In a controlled simulation, the noise-scale ratio varies from 0.25 to 4.0 and the scores have a 5.5-nat offset. The learned mixing coefficient tracks the inverse-variance result within 0.017 absolute error, with cross-seed standard deviation at most 0.014 (five seeds). This simulation is distinct from the empirical retrieval experiments.

\subsection{Independent held-out molecular retrieval}\label{app:supp-e10-definitive}

The larger independent evaluation assesses whether the controlled comparisons persist after model selection on separate validation molecules. All seven configurations are evaluated across five seeds on the 500 unique-molecule slice described in \cref{tab:app-protocol-registry} (\path{scripts/47_run_e10_definitive_retrieval_and_e104.py}). The adaptive bidirectional model achieves 30.3\% Hit@1, compared with 10.4\% for CLIP and 8.6\% for masked likelihood with latent contrast. The reported paired improvement over CLIP is 19.9 percentage points (95\% CI [16.4, 23.6]); over the latent-contrast model it is 21.7 points (CI [18.1, 25.4]). Both reported tests have $p<10^{-4}$. Centered output contrast also improves Hit@1 over raw likelihood contrast by 21.5 points (CI [18.2, 24.8]). These are comparisons within a single protocol, not evidence of universal superiority (\cref{fig:global-experiments}B). The next pairing-control section examines whether the gain is attributable to the correct partner.

\begin{table}[!htbp]
\centering
\caption{Independent retrieval on 500 unique molecules (five seeds, script 47; validation-selected training settings).}
\label{tab:app-e10-definitive}
\pomatabbox{
\begin{tabular}{lccccc}
\toprule
Configuration & ObsNLL & H@1 & H@5 & H@20 / MRR & Tan@1 / detail \\
\midrule
CLIP-style latent & -- & $10.4\%$ & $24.3\%$ & $53.3\%$ / $.192$ & $.288$ Tan@1 \\
MLM + Latent InfoNCE & -- & $8.6\%$ & $21.9\%$ & $53.0\%$ / $.171$ & $.276$ Tan@1 \\
MLM + Raw LL (no center) & -- & $2.8\%$ & $9.8\%$ & $33.4\%$ / $.088$ & $.231$ Tan@1 \\
MLM only (Output PMI) & -- & $23.8\%$ & \underline{$54.0\%$} & $\mathbf{88.4\%}$ / \underline{$.384$} & $.384$ Tan@1 \\
MLM + Output PMI InfoNCE & -- & \underline{$24.3\%$} & $52.2\%$ & $84.0\%$ / $.381$ & \underline{$.392$} Tan@1 \\
MLM + Output SoftRank & -- & $23.8\%$ & $52.3\%$ & $84.8\%$ / $.380$ & $.384$ Tan@1 \\
\textbf{Adaptive Bidir.\ POMA} & -- & $\mathbf{30.3\%}$ & $\mathbf{55.7\%}$ & \underline{$86.3\%$} / $\mathbf{.432}$ & $\mathbf{.438}$ ($p\!<\!10^{-4}$) \\
\bottomrule
\end{tabular}
}
\end{table}

\subsection{Alignment of large-scale pretrained backbones}\label{app:supp-e9-dreams}

The small spectral Transformer enables controlled comparisons but does not establish adaptation of a large externally pretrained model. We therefore pair frozen DreaMS with ChemBERTa (\cref{app:supp-provenance-e5}; \path{scripts/49_run_e9_dreams_chemberta_alignment.py}). On 300 unique test molecules and three seeds, the comparison includes latent CLIP, categorical reconstruction, a continuous hurdle head and combinations of reconstruction and contrastive training. The categorical control induces a piecewise-uniform density $q(z\mid C)=\sum_j(\pi_j(C)/\Delta_j)\mathbf 1_{\{z\in B_j\}}$.

The continuous hurdle readout improves retrieval relative to categorical reconstruction, but the categorical control retains slightly better observed-data NLL. The strong DreaMS latent baseline achieves the best Hit@1 and MRR, whereas hurdle reconstruction achieves the best Hit@5 and Hit@20. The contrastively trained representation is therefore valuable in its own right. The comparison supports the use of native densities as an additional interface rather than the replacement of latent similarity in all settings.

\begin{table}[!htbp]
\centering
\caption{
\textbf{Large-scale pretrained FM alignment  : frozen \textsc{ChemBERTa-77M-MLM} $\leftrightarrow$ official $117.1\mathrm{M}$-parameter $24\mathrm{M}$-spectra \textsc{DreaMS} ($300$ held-out unique \textsc{MassSpecGym} test molecules, $3$ seeds; \texttt{scripts/49}).}
}
\label{tab:app-e9-dreams}
\pomatabbox{
\setlength{\tabcolsep}{1.8pt}
\begin{tabular}{lccccccc}
\toprule
\textbf{Configuration (Frozen $117\mathrm{M}$ \textsc{DreaMS})} &
\textbf{ObsNLL $\downarrow$} &
\textbf{MLM $\downarrow$} &
\textbf{H@1 $\uparrow$} &
\textbf{H@5 $\uparrow$} &
\textbf{H@20 $\uparrow$} &
\textbf{MRR $\uparrow$} &
\textbf{Tan@1 $\uparrow$} \\
\midrule
D0.\ \textsc{DreaMS} + \textsc{ChemBERTa} Latent CLIP & -- & -- & $\mathbf{27.9\%}$ & \underline{$50.3\%$} & \underline{$75.7\%$} & $\mathbf{0.391}$ & $\mathbf{0.438}$ \\
D1.\ Categorical Recon.\ (Piecewise-Unif.) & $\mathbf{0.266}$ & $1.195$ & $19.2\%$ & $45.7\%$ & $73.9\%$ & $0.322$ & $0.352$ \\
D2.\ Categorical + Output Contrast & $0.288$ & \underline{$1.176$} & $17.5\%$ & $38.3\%$ & $67.7\%$ & $0.288$ & $0.345$ \\
\textbf{D3.\ POMA \textsc{Hurdle} Recon.\ (Bidir.\ PMI)} & $0.282$ & $\mathbf{1.136}$ & $21.4\%$ & $\mathbf{51.9\%}$ & $\mathbf{81.4\%}$ & \underline{$0.361$} & $0.367$ \\
D4.\ POMA \textsc{Hurdle} + Latent Contrast & $0.299$ & \underline{$1.176$} & \underline{$24.9\%$} & $44.6\%$ & $72.1\%$ & $0.353$ & \underline{$0.410$} \\
D5.\ POMA \textsc{Hurdle} + PMI Contrast & \underline{$0.280$} & $1.199$ & $15.0\%$ & $37.0\%$ & $68.4\%$ & $0.267$ & $0.324$ \\
D6.\ D5 + Adaptive Bidir.\ Weighting & \underline{$0.280$} & $1.199$ & $15.8\%$ & $37.7\%$ & $67.3\%$ & $0.273$ & $0.329$ \\
\bottomrule
\end{tabular}
}
\end{table}

\subsection{Reference distributions and pairing-specific compatibility}\label{app:supp-e12-shortcut}

A retrieval model can exploit information about the candidate domain without identifying the correct partner. Removing the spectrum gives zero centered scores, and random tie-breaking then recovers chance. Shuffling the spectrum is more informative: in the independent evaluation, discrete-direction Hit@1 remains almost unchanged. The result is consistent with a context-induced molecular prior shift, but does not uniquely identify the mechanism. The increase from 24.3\% to 30.3\% Hit@1 when adding bidirectional fusion is not, by itself, a measure of pair-specific information (\cref{fig:global-experiments}B).

\begin{table}[!htbp]
\centering
\caption{Tie-aware falsification in the 500-molecule protocol. The shuffled-context result concerns the discrete direction; the removed-context baseline uses uniform random tie-breaking.}
\label{tab:app-e10-falsification}
\pomatabbox{
\begin{tabular}{lccccc}
\toprule
Condition & ObsNLL & H@1 & H@5 & MRR & Detail \\
\midrule
Correct / Shuffled (Disc.\ PMI) & -- & $28.0\%$ / $27.8\%$ & $55.8\%$ / $55.6\%$ & $.420$ / $.419$ MRR & Prior-shift flag \\
Spectrum Removed (Tie-Aware) & -- & $4.18\%$ & $16.12\%$ & $.1230$ MRR & Exact chance \\
Candidate-Only Prior & -- & $1.80\%$ & $6.60\%$ & $.0668$ MRR & Unconditioned \\
Analytical Random Chance & -- & $4.18\%$ & $16.12\%$ & $.1230$ MRR & Reference \\
\bottomrule
\end{tabular}
}
\end{table}

\paragraph{Directional falsification.} We extend the control to both spectral backbones on 200 unique molecules, examining correct pairing, within-formula or mass-matched shuffling, global shuffling, removed context, precursor metadata alone, and average training context (\path{scripts/50_run_e12_shortcut_resistant_alignment.py}). The small Transformer has discrete MRR of 0.360, 0.363, 0.362 and 0.361 for correct, within-formula shuffled, globally shuffled and average context, respectively. DreaMS shows a positive global pairing difference, but the within-formula discrete difference is only 0.004 (CI [$-0.026$, $0.034$]). This motivates a reference that controls for the presence of the modality.

\paragraph{Reference selection at inference.} We compare the context-free reference with population, precursor-metadata and formula-matched training references, using 4, 16 or 64 observations, as well as an average partner-domain null context. The reference is constructed without test pairing labels. Both log-mean-exp and mean-log centering are evaluated, with the distinction formalized in \cref{app:matched-reference}. Population references improve discrete Hit@1 from 11.0\% to 17.0--20.0\%; metadata and formula references give 14.0--17.0\%. The respective MRR ranges are 0.272--0.311 and 0.236--0.265, compared with 0.167 for the original reference. The added cost increases from 16.9 ms for a single average context to 67.3 ms for four references and 1,076 ms for 64. Reference replacement improves retrieval but does not establish fine-grained pairing specificity.

\paragraph{Context-matched contrastive learning.} We compare five objectives on 300 previously unused unique test molecules (three seeds), incorporating the mixture reference into the training score and using a single domain-matched null context at inference for the matched-reference model. Compared with ordinary output contrast, matched-reference training increases bidirectional Hit@1 from 12.2\% to 36.6\%. The global pairing gain is 0.090 MRR (reported paired CI [0.055, 0.124]); the discrete gain is 0.079 (CI [0.046, 0.112]), both with reported $p<10^{-4}$. The masked-likelihood-only discrete gain is $-0.009$ ($p=0.348$). The matched model reaches Tanimoto@1 of 0.481. Nevertheless the within-formula gain remains 0.006. The experiment supports broader pairing sensitivity, not resolution of isomer-specific discrimination (\cref{fig:global-experiments}B). The latent CLIP result in \cref{tab:app-e9-dreams} is from a different evaluation slice and cannot establish a head-to-head superiority claim.

\begin{table}[!htbp]
\centering
\caption{
\textbf{Reference design and pairing specificity on \textsc{ChemBERTa}-77M $\leftrightarrow$ \textsc{DreaMS}-117M.}
Top: $6$-condition directional falsification ($200$ unique test molecules, original T1 checkpoint). Bottom: context-matched training (T0--T4, $3$ seeds) evaluated on a \textbf{previously unused slice of $300$ unique test molecules} (\texttt{test\_unique[620:920]}), with $10{,}000$-resample paired molecule-level bootstrap $95\%$ CIs for $G_{\mathrm{pair}}^{\mathrm{gl}} = \mathrm{MRR}_{\mathrm{correct}} - \mathrm{MRR}_{\mathrm{global\_shuf}}$.
}
\label{tab:app-e12-shortcut}
\pomatabbox{
\setlength{\tabcolsep}{2.3pt}
\begin{tabular}{lccccccc}
\toprule
\textbf{Directional falsification (\textsc{DreaMS} T1, $N=200$)} &
\textbf{Disc.\ H@1} & \textbf{Disc.\ MRR} & \textbf{Cont.\ H@1} & \textbf{Cont.\ MRR} & \textbf{Bidir.\ H@1} & \textbf{Bidir.\ MRR} & \textbf{Note} \\
\midrule
1.\ Correct paired spectrum & $10.5\%$ & $.184$ & $8.5\%$ & $.182$ & $9.8\%$ & $.193$ & Baseline T1 \\
2.\ Within-formula / mass shuffled & $11.0\%$ & $.180$ & $11.0\%$ & $.197$ & $12.2\%$ & $.209$ & $G_{\mathrm{wf}}^{\mathrm{disc}}\!=\!+.004$ \\
3.\ Globally shuffled spectrum & $8.5\%$ & $.151$ & $5.0\%$ & $.141$ & $4.8\%$ & $.129$ & $G_{\mathrm{gl}}^{\mathrm{bidir}}\!=\!+.064$ \\
4.\ Spectrum removed (tie-aware) & $4.4\%$ & $.124$ & $4.4\%$ & $.124$ & $4.4\%$ & $.124$ & Exact chance \\
5.\ Precursor metadata only & $9.5\%$ & $.165$ & $3.5\%$ & $.096$ & $7.5\%$ & $.135$ & Zero peaks \\
6.\ Average spectral context & $7.5\%$ & $.137$ & $12.0\%$ & $.216$ & $8.8\%$ & $.156$ & Mean train ctx \\
\midrule
\textbf{Matched-reference training ($N=300$, $3$ seeds)} &
\textbf{Disc.\ H@1} & \textbf{Disc.\ MRR} & \textbf{Disc.\ $G_{\mathrm{pair}}^{\mathrm{gl}}$} & \textbf{Bidir.\ H@1} & \textbf{Bidir.\ H@5} & \textbf{Bidir.\ MRR} & \textbf{Bidir.\ $G_{\mathrm{pair}}^{\mathrm{gl}}$ (95\% CI)} \\
\midrule
T0.\ Masked likelihood only & $14.6\%$ & $.270$ & $-.009$ & \underline{$20.4\%$} & \underline{$45.9\%$} & \underline{$.326$} & $+.045$ $[+.014,+.076]$ \\
T1.\ Original random-neg.\ InfoNCE & $11.7\%$ & $.207$ & $+.037$ & $12.2\%$ & $31.0\%$ & $.222$ & \underline{$+.067$} $[+.040,+.094]$ \\
T2.\ Within-formula / mass-hard InfoNCE & $12.0\%$ & $.220$ & $+.007$ & $16.7\%$ & $37.4\%$ & $.273$ & $+.044$ $[+.013,+.075]$ \\
\textbf{T3.\ Context-matched InfoNCE (Ref.\ Mix.)} & $\mathbf{35.1\%}$ & $\mathbf{.456}$ & $\mathbf{+.079}$ & $\mathbf{36.6\%}$ & $\mathbf{57.0\%}$ & $\mathbf{.464}$ & $\mathbf{+.090}$ $[+.055,+.124]$ \\
T4.\ Mixed neg.\ + matched ref. & \underline{$24.8\%$} & \underline{$.361$} & \underline{$+.065$} & $16.4\%$ & $34.2\%$ & $.264$ & $+.049$ $[+.014,+.082]$ \\
\bottomrule
\end{tabular}
}
\end{table}

\subsection{Generality across spectral observation laws}\label{app:supp-dist-family-alignment}

The preceding experiments use particular likelihood families. To determine how the observation law interacts with alignment, we vary seven continuous heads under masked likelihood and output contrast ($\lambda=0.3$), holding the small spectral backbone, data and schedule fixed (three seeds, \path{scripts/40_run_distribution_family_alignment_ablation.py}). The common observed-data score is defined in \cref{app:metrics-protocol}. The original 120-pool experiment reports paired continuous Hit@1 gains for heteroscedastic Gaussian (+16.9 points, CI [12.4,21.5]), coupled Gaussian (+21.9, CI [16.8,27.1]) and Student-$t$ (+16.7, CI [11.9,21.5]); the Student-$t$ bidirectional gain is +15.0 points (CI [9.2,20.8]). These are preliminary pool-level estimates and their independence should not be assumed. Hurdle prediction improves the observed-data score, whereas contrast can hurt some censoring-aware and mixture configurations. In this initial protocol, the hurdle head has 95.0\% Hit@20 and the contrasted mixture has the lowest reported positive-peak NLL. These preliminary results are interpreted alongside the molecule-level confirmation.

\begin{table}[!htbp]
\centering
\caption{
\textbf{14-configuration distribution-family ablation of native two-BioFM alignment (\textsc{ChemBERTa-77M-MLM} $\leftrightarrow$ \textsc{MaskedPeakMSTransformer}, $120$ \textsc{MassSpecGym} isomer pools, $3$ seeds).}
All $7$ families are evaluated on the exact same observed-data mixed measure (\textsc{ObsNLL}). Discrete \textsc{ChemBERTa} MLM NLL is $1.184$ (\textsc{MLM-only}) vs.\ $1.519$ (\textsc{+contrast}).
}
\label{tab:app-dist-family-alignment}
\pomatabbox{
\setlength{\tabcolsep}{1.8pt}
\begin{tabular}{llcccccc}
\toprule
\textbf{Continuous Head} &
\textbf{Regime} &
\textbf{ObsNLL $\downarrow$} &
\textbf{PosNLL $\downarrow$} &
\textbf{Cov95} &
\textbf{Cont H@1 $\uparrow$} &
\textbf{Bidir H@1 / H@5 $\uparrow$} &
\textbf{Tan@1 $\uparrow$} \\
\midrule
\multirow{2}{*}{Heterosc.\ Gaussian}
& MLM-only & $0.991$ & $1.817$ & $75.7\%$ & $12.8\%$ & $20.6\%$ / $60.8\%$ & $.392$ \\
& +Contrast & $1.033$ & $1.420$ & $78.2\%$ & \underline{$29.7\%$} & $16.7\%$ / $64.7\%$ & $.428$ \\
\midrule
\multirow{2}{*}{Coupled Gaussian}
& MLM-only & $0.822$ & $0.521$ & $85.9\%$ & $13.3\%$ & $16.7\%$ / $65.3\%$ & $.394$ \\
& +Contrast & $1.426$ & $0.663$ & $99.1\%$ & $\mathbf{35.3\%}$ & $13.3\%$ / $65.8\%$ & $.392$ \\
\midrule
\multirow{2}{*}{Student-$t$}
& MLM-only & $0.832$ & $0.785$ & $66.7\%$ & $6.7\%$ & $10.0\%$ / $51.9\%$ & $.365$ \\
& +Contrast & $0.755$ & $0.812$ & $67.1\%$ & $23.3\%$ & $\mathbf{25.0\%}$ / $52.8\%$ & $\mathbf{.487}$ \\
\midrule
\multirow{2}{*}{Gaussian Mixture ($K\!=\!3$)}
& MLM-only & $0.629$ & \underline{$0.314$} & $79.6\%$ & $19.7\%$ & $22.2\%$ / $67.8\%$ & $.441$ \\
& +Contrast & $0.689$ & $\mathbf{0.249}$ & $80.6\%$ & $16.9\%$ & $12.8\%$ / $58.9\%$ & $.403$ \\
\midrule
\multirow{2}{*}{Independent Tobit}
& MLM-only & $0.368$ & $2.793$ & $62.0\%$ & $19.7\%$ & $19.2\%$ / $72.5\%$ & $.416$ \\
& +Contrast & $0.344$ & $2.642$ & $55.4\%$ & $22.2\%$ & $18.3\%$ / $68.1\%$ & $.380$ \\
\midrule
\multirow{2}{*}{Coupled Tobit}
& MLM-only & $0.387$ & $2.929$ & $58.2\%$ & $23.3\%$ & $16.4\%$ / \underline{$73.1\%$} & $.418$ \\
& +Contrast & $0.456$ & $3.773$ & $43.0\%$ & $19.2\%$ & $15.3\%$ / $67.5\%$ & $.430$ \\
\midrule
\multirow{2}{*}{\textbf{Hurdle Continuous}}
& \textbf{MLM-only} & $\mathbf{0.309}$ & $1.987$ & $95.8\%$ & $14.2\%$ & \underline{$22.8\%$} / $\mathbf{77.5\%}$ & \underline{$.451$} \\
& +Contrast & \underline{$0.342$} & $2.451$ & $96.2\%$ & $12.2\%$ & $19.7\%$ / $71.1\%$ & \underline{$.451$} \\
\bottomrule
\end{tabular}
}
\end{table}

The four-family confirmation on 500 unique molecules provides a stronger evaluation of the observation-law effect. Hurdle modeling has the best observed-data score among the tested families; contrast helps the Student-$t$ and mixture heads but not the coupled Gaussian. Thus the framework supports different observation laws, without a universal ranking of likelihood families or objectives.

\begin{table}[!htbp]
\centering
\caption{Deduplicated four-family comparison (500 molecules, three seeds, script 47). Paired entries are masked likelihood / with contrast; retrieval is continuous-direction only.}
\label{tab:app-e10-families}
\pomatabbox{
\begin{tabular}{lccccc}
\toprule
Family & ObsNLL & H@1 & H@5 & MRR & Detail \\
\midrule
Coupled Gaussian (MLM / +Con) & $.736$ / $.787$ & $9.7\%$ / $7.9\%$ & $20.3\%$ / $20.1\%$ & $.176$ / $.161$ MRR & Cont.\ only \\
Student-$t$ (MLM / +Con) & $.771$ / \underline{$.545$} & $3.2\%$ / $6.7\%$ & $10.3\%$ / $17.9\%$ & $.091$ / $.141$ MRR & $+2.1\times$ H@1 \\
Gaussian Mix.\ (MLM / +Con) & $.865$ / $.718$ & $13.1\%$ / \underline{$13.9\%$} & $27.1\%$ / $27.2\%$ & $.218$ / \underline{$.222$} MRR & Cont.\ only \\
\textbf{Hurdle Cont.\ (MLM / +Con)} & $\mathbf{.384}$ / $\mathbf{.384}$ & $\mathbf{15.1\%}$ / $11.3\%$ & $\mathbf{34.0\%}$ / \underline{$29.3\%$} & $\mathbf{.255}$ / $.218$ MRR & Best ObsNLL \\
\bottomrule
\end{tabular}
}
\end{table}

\subsection{Cellular contextualization and asymmetric information}

In the complementary BMMC study, we evaluate matching on 256 cells from unseen sequencing Site~4. Protein sequence embeddings cover 134 surface proteins, including 109 exact human UniProt sequence matches~\citep{uniprot2023}. A forward readout predicts ADT abundances using RNA context; the reverse readout scores RNA conditioned on protein features and measured ADT. At 25\% RNA masking, the forward direction identifies paired cells much more effectively than the reverse direction. Adding the reverse score to the forward score worsens retrieval (\cref{tab:app-cell-directions}). This contrasts with molecular--spectral retrieval and motivates examining the information actually available to each head.

Increasing RNA masking reduces the opportunity to reconstruct expression from other observed genes. In the 25\%, 50\% and 75\% masking experiments, the largest absolute generative improvement from protein context occurs at 75\% masking (\cref{tab:app-cell-mask}). To investigate the reverse retrieval failure more directly, a separate six-variant study replaces the original 320:1 concatenated protein-value encoder by a balanced additive and FiLM-conditioned encoder, and varies whether the lightweight RNA head is unfrozen (\path{scripts/38_run_scgpt_reverse_alignment_unfrozen_head.py}). The 12-layer scGPT backbone remains frozen throughout.

Unfreezing the head improves masked RNA likelihood at 25\% masking, yet the contextual gate remains close to zero and reverse retrieval remains near chance. When all RNA values are masked, the model receives the gene-ID scaffold and protein context but no observed expression values. The gate opens, expression reconstruction improves, and reverse matching becomes better than chance, although forward-only retrieval is still strongest. The principal outcomes of this revised protocol are summarized in \cref{tab:supp-reverse-diagnostics}. The results are consistent with within-modality shortcuts and asymmetric predictive information; they do not demonstrate the absence of conditional mutual information. Zero-shot transfer to 26 unseen proteins remains weak ($r=0.054$), indicating a separate limitation of sequence-only protein context.

\begin{table}[!htbp]
\centering
\caption{scGPT--ESM-2 directional comparison on 256 Site~4 BMMC cells. The learned gate is $\tanh(3\gamma)=-0.036$. Directional NLLs score different modalities and are not directly comparable; FOSCTTM is lower-is-better.}
\label{tab:app-cell-directions}
\pomatabbox{
\begin{tabular}{lcccc}
\toprule
Direction & NLL & H@1 / FOSCTTM & H@5 & H@20 \\
\midrule
Fwd (\textsc{ADT}$|$\textsc{scGPT}) & $1.052$ ($r\!.392$) & $\mathbf{.0914}$ FOS & $\mathbf{16.0\%}$ & $\mathbf{59.4\%}$ H@20 \\
Rev (\textsc{scGPT}$|$\textsc{ADT}) & $-.047$ (vs.\ $-.030$) & $.5003$ FOS & $2.0\%$ & $7.8\%$ H@20 \\
Bidir.\ Two-BioFM PMI & -- & \underline{$.1722$} ($\mathbf{3.1\%}$ H@1) & \underline{$12.9\%$} & \underline{$37.1\%$} H@20 \\
\bottomrule
\end{tabular}
}
\end{table}

\begin{table}[!htbp]
\centering
\caption{RNA masking and ADT context (script 37).}
\label{tab:app-cell-mask}
\pomatabbox{
\begin{tabular}{lcccc}
\toprule
Masking & NLL & Correlation & Difference & Detail \\
\midrule
$p_{\mathrm{mask}}\!=\!25\%$ ($32$ genes) & $.006\!\to\!\mathbf{-.034}$ & $r=.739$ & $\Delta=-.040$ & Gen.\ gain \\
$p_{\mathrm{mask}}\!=\!50\%$ ($64$ genes) & $-.090\!\to\!\mathbf{-.108}$ & $r=.724\!\to\!\mathbf{.731}$ & $\Delta=-.018$ & +InfoNCE($.3$) \\
$p_{\mathrm{mask}}\!=\!75\%$ ($96$ genes) & $5.16\!\to\!\mathbf{5.00}$ & $r=.625$ & $\mathbf{\Delta=-.162}$ & Largest drop \\
\bottomrule
\end{tabular}
}
\end{table}

\begin{table}[!htbp]
\centering
\caption{Reported diagnostics from the revised scGPT--ADT reverse-alignment protocol. The values describe the indicated configurations, rather than a complete factorial table of all six variants. FOSCTTM is lower-is-better.}
\label{tab:supp-reverse-diagnostics}
\small
\begin{tabularx}{\linewidth}{@{}p{.39\linewidth}X@{}}
\toprule
Configuration & Diagnostic\\
\midrule
25\% RNA masking, head unfrozen & J-NLL $0.108\to0.063$; Pearson $0.755$ and coverage $95.6\%$ with InfoNCE; gate $0.001$; reverse FOSCTTM $0.500$\\
100\% RNA masking, balanced encoder & Gate $0.362$; expression correlation $-0.061\to+0.206$; coverage $95.9\%$ with InfoNCE\\
Revised full-masking retrieval & Reverse FOSCTTM $0.474$, Hit@20 $10.5\%$; bidirectional FOSCTTM $0.0953$, Hit@20 $55.1\%$; forward FOSCTTM $0.0878$\\
\bottomrule
\end{tabularx}
\end{table}

\subsection{Cross-domain retrieval and comparison with latent alignment}\label{app:supp-retrieval}

The native pretrained-model pair is not the only setting in which a continuous likelihood can support inverse identification. We also evaluate the likelihood of observed spectra under candidate molecular structures in a separate feature-based protocol. On 250 formula-matched MassSpecGym pools, coupled Tobit likelihood-ratio scoring improves Hit@5 from 47.6\% for CLIP-MS to 58.0\%. On sequence-disjoint ProteomeTools, the analogous comparison improves retrieval over the Prosit spectral-angle reference. Censoring and joint peak structure both matter: the independent Tobit model has good zero calibration but substantially weaker inverse identification. The complete forward and inverse metrics are reported together in \cref{tab:app-spec-track-bc}.

The conclusion depends on the candidate pool. In open-scaffold QM9S retrieval without formula filtering, CLIP-Spectra achieves 25.5\% Hit@5, compared with 11.2\% for the combined IR/Raman likelihood score (\cref{tab:app-spec-track-a}). In a separate 250-pool experiment with end-to-end ChemBERTa-77M-MTR and a three-layer GIN~\citep{xu2019gin} using rank-8 LoRA~\citep{hu2022lora} and 12,000 training examples, coupled Tobit scoring improves Hit@5 from 22.0\% to 24.4\% relative to CLIP-MS. A hybrid scorer does not improve on the likelihood scorer in that protocol. These absolute rates should not be compared directly with the earlier 250-pool or 120-pool experiments.

\begin{table}[!htbp]
\centering
\caption{End-to-end rank-8 LoRA ChemBERTa-77M-MTR and three-layer GIN on MassSpecGym (12,000 training examples; 250 pools). Separate from frozen and feature-based protocols.}
\label{tab:app-gpu-tier2-fm}
\pomatabbox{
\begin{tabular}{lcccc}
\toprule
Scorer & NLL & H@5 & H@20 & Tan@1 \\
\midrule
E2E \textsc{CLIP-MS} & $\times$ & $22.0\%$ H@5 & $61.2\%$ H@20 & $.331$ Tan@1 \\
\textbf{E2E \textsc{POMA-Tobit PMI}} & $\mathbf{0.195}$ & $\mathbf{24.4\%}$ H@5 & $\mathbf{62.8\%}$ H@20 & $\mathbf{.366}$ Tan@1 \\
E2E Hybrid (\textsc{CLIP+POMA}) & -- & \underline{$22.8\%$} H@5 & $58.4\%$ H@20 & \underline{$.353$} Tan@1 \\
\bottomrule
\end{tabular}
}
\end{table}

We also compare output-space scoring with latent alignment in a separate CITE-seq experiment using a transcriptome MLP rather than the scGPT--ESM-2 pair. On held-out Site~4, the coupled likelihood scorer achieves FOSCTTM of 0.0430, compared with 0.0557 for a totalVI-style conditional VAE and 0.0648 for CLIP-SingleCell. Hit@20 reaches 80.9\%. In Multiome, raw unpooled ATAC peaks are too sparse for effective matching in this protocol; train-fitted TF-IDF/LSI pooling of 8,000 peaks to 64 topics improves the signal substantially, with the independent density outperforming the coupled density. The corresponding forward prediction and matching metrics are provided in \cref{tab:app-multiomics-cite-multiome}, with the CITE-seq retrieval comparison illustrated in \cref{fig:supp-biological-applications}C. These results distinguish learning a useful compatibility score from the stronger requirement that it also retain predictive distributional information.

\subsection{Complementary discrete and continuous mutation scores}\label{sec:experiments-logica-poma}\label{app:logica-poma-aintibody}

Mutation effects provide a setting in which both kinds of outputs can be informative. We evaluate \umethod{} on AbRank's homology-disjoint unseen-antigen split (three seeds), preserving orientation symmetry: swapping the members of a mutation pair reverses the directional score while retaining the impact score. The normalized three-state probabilities integrate the predicted difference distribution over a dead zone and two impact tails (\cref{eq:abrank-3state-integral}).

The native discrete LogiCA score is particularly effective for cross-clone candidate ordering, whereas the continuous density better detects large-magnitude changes. Their combination gives the strongest ordinal AUC, increasing it from 0.614 to 0.628 relative to LogiCA, while the individual scores remain preferable for some metrics. This complementarity is more informative than the aggregate difference alone (\cref{tab:app-abrank-full}). Shared posterior uncertainty in such differences is examined in \cref{app:supp-bayesian}.

\begin{table}[!htbp]
\centering
\caption{
\textbf{Unseen-antigen ($<50\%$ Ag, $<80\%$ lineage) 3-state evaluation on \textsc{AbRank} \citep{liu2026abrank} ($3$ seeds, mean $\pm$ SE).}
}
\label{tab:app-abrank-full}
\pomatabbox{
\begin{tabular}{lccccc}
\toprule
\textbf{Model} &
\textbf{Sign AUC $\uparrow$} &
\textbf{Impact $\uparrow$} &
\textbf{Ordinal $\uparrow$} &
\textbf{Sel.\ Hit $\uparrow$} &
\textbf{Cand.\ $\uparrow$} \\
\midrule
Zero-shot PLM & $0.449$ & $0.794$ & $0.500$ & $0.495$ & $0.437$ \\
3-state hurdle MLP & $0.557\!\pm\!0.02$ & $0.888\!\pm\!0.01$ & $0.557\!\pm\!0.02$ & $0.520\!\pm\!0.02$ & $0.530\!\pm\!0.01$ \\
Regression Transformer & $0.560\!\pm\!0.01$ & $0.838\!\pm\!0.00$ & $0.572\!\pm\!0.00$ & $0.531\!\pm\!0.00$ & $0.517\!\pm\!0.00$ \\
Scalar MSE regressor & $0.570\!\pm\!0.01$ & $0.824\!\pm\!0.01$ & $0.576\!\pm\!0.01$ & $0.536\!\pm\!0.01$ & $0.527\!\pm\!0.02$ \\
\textsc{POMA-Rank} (3-state) & $0.548\!\pm\!0.08$ & \textbf{$0.906\!\pm\!0.00$} & $0.568\!\pm\!0.07$ & $0.529\!\pm\!0.06$ & $0.524\!\pm\!0.03$ \\
\textsc{LogiCA} (antigen) & \underline{$0.595\!\pm\!0.02$} & $0.795\!\pm\!0.00$ & \underline{$0.614\!\pm\!0.02$} & \textbf{$0.613\!\pm\!0.02$} & \textbf{$0.601\!\pm\!0.00$} \\
\textbf{\textsc{LogiCA-POMA} (unified)} & \textbf{$0.605\!\pm\!0.07$} & \underline{$0.896\!\pm\!0.00$} & \textbf{$0.628\!\pm\!0.06$} & \underline{$0.598\!\pm\!0.06$} & \underline{$0.579\!\pm\!0.02$} \\
\bottomrule
\end{tabular}
}
\end{table}

The alignment experiments identify complementary roles for predictive distributions, latent similarity, reference choice and directional weighting. We now test whether the fitted distributions can be reused for inference, rather than evaluated only as retrieval scores.

\clearpage
\section{Structured conditional inference}\label{app:supp-inference}

The preceding results assess compatibility between observations. Conditional inference asks a different question: how does an existing predictive law change when a subset of the measurements becomes available? We first test reuse of a single aligned checkpoint, then isolate the effect of learned dependence and compare analytical updates with neural reconditioning. The remaining studies examine the same operation across assay panels, spectra and cellular measurements.

\subsection{A unified probabilistic checkpoint}\label{app:supp-e15-unified}

To test whether alignment and inference coexist within one statistical model, we train a ChemBERTa--DreaMS checkpoint for each of three objectives using a coupled \texttt{LowRankMVNDensityHead} over 64 spectral bands (48 $m/z$ and 16 neutral-loss bands, $\log(1+4\sqrt{I})$ scale). All tasks use the same checkpoint for a given objective and seed, with no task-specific retraining (\cref{fig:global-experiments}C). The evaluation uses 200 held-out unique molecules and three seeds (\path{scripts/53_run_e15_unified_checkpoint.py}).

\Cref{tab:main-unified-checkpoint} summarizes retrieval and distributional quality from the same checkpoint, including conditional interval coverage; the extended table reports additional retrieval metrics, interval widths and update costs.

\begin{table}[!htbp]
\centering
\caption{\textbf{Unified probabilistic checkpoint evaluation.}
ChemBERTa-77M--DreaMS-117M on 200 unique held-out molecules ($3$ seeds). All metrics within each row are obtained from the same trained checkpoint. Pair gain denotes the correct-versus-shuffled MRR difference. Marginal and conditional coverage are evaluated at nominal $95\%$. Complete results are provided in \cref{tab:app-e15-unified}.}
\label{tab:main-unified-checkpoint}
\small\setlength{\tabcolsep}{3.0pt}
\begin{tabular}{lrrrrr}
\toprule
Training objective
& H@1 $\uparrow$
& Pair gain $\uparrow$
& Spec.\ NLL $\downarrow$
& Cov95
& Cond.\ Cov95 \\
\midrule
Masked likelihood
& 12.3\% & .029 & .618 & 90.7\% & 94.8\% \\
Latent contrast
& \textbf{35.2\%} & \textbf{.194} & .591 & 91.3\% & \textbf{95.0\%} \\
Context-matched output contrast
& 26.7\% & .082 & \textbf{.572} & \textbf{91.5\%} & 95.3\% \\
\bottomrule
\end{tabular}
\end{table}

\begin{table}[!htbp]
\centering
\caption{
\textbf{Unified single-checkpoint evaluation  on audited \textsc{ChemBERTa}-77M $\leftrightarrow$ \textsc{DreaMS}-117M ($200$ previously unused unique molecules, $3$ seeds).}
Every metric in each row is evaluated on the \emph{exact same trained checkpoint} without task-specific retraining. $G_{\mathrm{pair}}$ denotes pair-specific retrieval gain ($\mathrm{MRR}_{\mathrm{correct}}-\mathrm{MRR}_{\mathrm{shuffled}}$); Schur conditioning reveals $0\%\to 75\%$ of observed spectral bands on $16$ held-out target bands and is compared against full \textsc{DreaMS}-117M re-encoding ($11{,}580\,\mathrm{ms}$ for $N=200$).
}
\label{tab:app-e15-unified}
\pomatabbox{
\setlength{\tabcolsep}{2.2pt}
\begin{tabular}{lcccccccc}
\toprule
\textbf{Unified Checkpoint} &
\textbf{H@1 $\uparrow$} &
\textbf{H@5 $\uparrow$} &
\textbf{MRR ($G_{\mathrm{pair}}$) $\uparrow$} &
\textbf{50\% Drop MRR} &
\textbf{Spec.\ NLL $\downarrow$} &
\textbf{Cov95} &
\textbf{Schur W95 ($0\!\to\!75\%$)} &
\textbf{Speedup} \\
\midrule
U0.\ Masked LL Only & $12.3\%$ & $34.7\%$ & $.245$ ($+.029$) & $.257$ & $.618$ & $90.7\%$ & $.932\!\to\!.895$ ($94.8\%$) & $2{,}302\times$ \\
U1.\ Latent Contrastive & $\mathbf{35.2\%}$ & $\mathbf{59.5\%}$ & $\mathbf{.472}$ ($\mathbf{+.194}$) & $\mathbf{.464}$ & \underline{$.591$} & $91.3\%$ & $.927\!\to\!.893$ ($95.0\%$) & $\mathbf{3{,}066\times}$ \\
\textbf{U2.\ Output Context-Matched} & \underline{$26.7\%$} & \underline{$50.2\%$} & \underline{$.383$} (\underline{$+.082$}) & \underline{$.397$} & $\mathbf{.572}$ & $91.5\%$ & $.995\!\to\!.924$ ($95.3\%$) & \underline{$2{,}835\times$} \\
\bottomrule
\end{tabular}
}
\end{table}

Latent contrast provides the best retrieval and global pairing specificity, while context-matched output contrast gives the best spectral NLL. Both improve retrieval relative to masked likelihood alone and preserve the fitted density. All three retain useful retrieval after 50\% peak dropout. Marginal 95\% intervals under-cover (90.7--91.5\%); conditional intervals approach nominal coverage and become narrower after additional bands are observed. For the reported output-contrast checkpoint the conditional NLL slightly worsens ($-0.015$ to $+0.024$), illustrating the limits of Gaussian dependence for sparse spectra. Cached Schur updates take 3.8--5.3 ms versus 11,580 ms for DreaMS re-encoding on the 200-object batch. This is an update-only comparison, not an end-to-end training or inference speedup.

\subsection{Learning measurement dependence and choosing the covariance structure}\label{app:supp-covariance}

The simplest comparison holds the encoder fixed and varies the output covariance. In a five-property molecular panel (aqueous solubility, cLogP, TPSA, QED~\citep{bickerton2012qed} and molecular weight), adding low-rank dependence improves joint NLL and multivariate scoring relative to independent Gaussian outputs. The result is reported in \cref{tab:app-admet-coupled5}. The low-rank form also admits efficient marginalization and conditioning, as derived in \cref{app:poma-woodbury-schur}.

\begin{table}[!htbp]
\centering
\caption{
\textbf{Coupled $5$-property compound density evaluation (\textsc{AqSolDB}, $\mathrm{cLogP}$, \textsc{TPSA}, \textsc{QED}, MW).}
}
\label{tab:app-admet-coupled5}
\pomatabbox{
\begin{tabular}{lccc}
\toprule
\textbf{Multi-Property Compound Variant} &
\textbf{Joint NLL $\downarrow$} &
\textbf{Energy Score $\downarrow$} &
\textbf{Variogram Score $\downarrow$} \\
\midrule
Independent Gaussian ($r=0$) & 1.797 & 0.358 & 0.027 \\
\textbf{Coupled \textsc{POMA-LowRankMVN} ($r=4$)} & \textbf{1.248} & \textbf{0.334} & \textbf{0.025} \\
\bottomrule
\end{tabular}
}
\end{table}

The choice of covariance parameterization matters particularly in small samples. A factor head of rank $r$ produces $K(r+1)$ covariance parameters for $K$ measurements, whereas an unrestricted Cholesky factor has $K(K+1)/2$. For the medium-sized clinical panels, the low-rank form is not necessarily smaller: with rank six it produces 70 rather than 55 outputs for ten assays, and 91 for either parameterization at 13 assays. Its advantages in these settings therefore concern the structured inductive bias and the computational form of conditional inference, rather than simply parameter count. For the 102-property pretraining panel, the difference is substantial (714 versus 5,253 outputs).

We vary the factor rank and compare with full Cholesky estimation in few-shot and full-cohort GDPa1 and AIntibody evaluation. In the few-shot regime, unrestricted covariance produces worse held-out likelihood and conditional error than the low-rank alternatives. Increasing rank improves conditional prediction up to the tested moderate ranks; the best rank need not be the same for every dataset. These results support regularized rather than unrestricted dependence estimation (\cref{tab:app-cov-rank}).

\begin{table}[!htbp]
\centering
\caption{
\textbf{Covariance parameterization ablation: low-rank factor loading $\Sigma_\theta(X) = D_\theta(X) + U_{\theta,r}(X)U_{\theta,r}(X)^\top$ versus full unstructured Cholesky $L_\theta(X)L_\theta(X)^\top$.}
Reported across \textsc{GDPa1} ($K=13$ assays, $k=32$ few-shot and $5$-fold CV) and
\textsc{AIntibody} ($K=10$ assays, $5$-fold CV). Best in \textbf{bold}, second best \underline{underlined}.
}
\label{tab:app-cov-rank}
\pomatabbox{
\begin{tabular}{lccccc}
\toprule
& & \multicolumn{2}{c}{\textbf{\textsc{GDPa1} ($k=32$)}} & \textbf{\textsc{GDPa1} (CV)} & \textbf{\textsc{AIntibody}} \\
\cmidrule(lr){3-4}\cmidrule(lr){5-5}\cmidrule(lr){6-6}
\textbf{Head} &
\textbf{Params} &
\textbf{J-NLL $\downarrow$} &
\textbf{Schur $\rho \uparrow$} &
\textbf{J-NLL $\downarrow$} &
\textbf{Schur $\rho \uparrow$} \\
\midrule
Diagonal ($r=0$) & $K$ & $1.649\!\pm\!0.135$ & $0.138\!\pm\!0.027$ & $1.434\!\pm\!0.056$ & $0.772\!\pm\!0.019$ \\
Low-Rank ($r=1$) & $2K$ & $1.656\!\pm\!0.145$ & $0.204\!\pm\!0.023$ & $1.391\!\pm\!0.057$ & $0.784\!\pm\!0.019$ \\
Low-Rank ($r=2$) & $3K$ & $1.650\!\pm\!0.148$ & $0.253\!\pm\!0.023$ & $1.380\!\pm\!0.068$ & $0.786\!\pm\!0.019$ \\
Low-Rank ($r=4$) & $5K$ & \underline{$1.632\!\pm\!0.148$} & \underline{$0.286\!\pm\!0.022$} & \underline{$1.365\!\pm\!0.073$} & $0.784\!\pm\!0.013$ \\
\textbf{Low-Rank ($r=6$)} & $7K$ & $\mathbf{1.626\!\pm\!0.144}$ & $\mathbf{0.288\!\pm\!0.024}$ & $\mathbf{1.357\!\pm\!0.072}$ & $\mathbf{0.792\!\pm\!0.009}$ \\
Cholesky ($L_\theta L_\theta^\top$) & $K(K\!+\!1)/2$ & $2.575\!\pm\!0.275$ & $0.252\!\pm\!0.026$ & $1.375\!\pm\!0.053$ & \underline{$0.788\!\pm\!0.011$} \\
\bottomrule
\end{tabular}
}
\end{table}

\subsection{Analytical conditioning versus neural reconditioning}\label{app:supp-e11-reconditioning}

Two operations can incorporate additional measurements: conditioning the original joint predictive law, or recomputing the neural prediction with an expanded context. The first experiment compares these on frozen scGPT and QM9S (three seeds; \path{scripts/48_run_e11_analytical_vs_neural_reconditioning.py}). On scGPT, 24 intermediate genes are progressively revealed while the same eight targets remain hidden. The original neural comparator is trained at the initial revelation level; on QM9S it uses random masks rather than structured occlusions. The comparison is consequently informative about robustness to masking shift, not an isolated test of inference algorithms under matched training.

Under masking shift, analytical updates improve scGPT conditional scores while neural reconditioning deteriorates. The reported sequential-versus-joint conditional discrepancy is $6.36\times10^{-6}$, whereas the neural prediction differs from the original Schur conditional by 0.462 nats KL. Cached updates avoid a repeated backbone pass: the displayed timings, 13.0 ms versus 4,235.5 ms, give approximately $326\times$ lower update time. Including initial inference, the four-round times are 4.3 s and 19.6 s, a cumulative ratio of approximately $4.6\times$. The mask-matched comparison below evaluates accuracy and latency after controlling the training mask distribution.

\begin{table}[!htbp]
\centering
\caption{
\textbf{Analytical Schur conditioning versus neural forward-pass reconditioning under masking shift on frozen \textsc{scGPT} ($3$ seeds, shared $8$ hidden genes, $0\%\to 75\%$ revealed) and \textsc{QM9S} IR spectroscopy (\texttt{scripts/48}).} The final column gives four-round cumulative time and its ratio relative to neural reconditioning; update-only timing is in the preceding column.
}
\label{tab:app-e11-reconditioning}
\pomatabbox{
\setlength{\tabcolsep}{1.8pt}
\begin{tabular}{llccccc}
\toprule
\textbf{Domain} &
\textbf{Method (Update Mode)} &
\textbf{Cond.\ NLL $\downarrow$} &
\textbf{MAE $\downarrow$} &
\textbf{CRPS $\downarrow$ / Cov95} &
\textbf{Update (ms)} &
\textbf{Cum.\ 4-Rd / Ratio} \\
\midrule
\multirow{6}{*}{\shortstack[l]{Frozen \textsc{scGPT}\\($0\%\!\to\!75\%$\\shared $8$ genes)}}
& C0.\ MSE + Per-Gene Var (Neural) & $+6.37\!\to\!+6.20$ & $.349\!\to\!.369$ & $.309\!\to\!.314$ & $4235.5$ & $19.6\,\mathrm{s}$ ($1.0\times$) \\
& C1.\ MSE + Heterosc.\ Var (Neural) & $-.032\!\to\!+.153$ & $.349\!\to\!.369$ & $.261\!\to\!.273$ & $4235.5$ & $19.6\,\mathrm{s}$ ($1.0\times$) \\
& C2.\ Indep.\ POMA (Neural) & $-.361\!\to\!-.287$ & $.254\!\to\!.266$ & \underline{$.206\!\to\!.211$} & $4235.5$ & $19.6\,\mathrm{s}$ ($1.0\times$) \\
& \textbf{C3.\ Coupled POMA (Analytical Schur)} & $-.242\!\to\!\mathbf{-.372}$ & $.269\!\to\!\mathbf{.242}$ & $.252\!\to\!\mathbf{.195}$ & $\mathbf{13.0}$ & $\mathbf{4.3\,\mathrm{s}}$ ($\mathbf{4.6\times}$) \\
& C4.\ Coupled POMA (Neural) & $-.242\!\to\!-.099$ & $.269\!\to\!.290$ & $.252\!\to\!.269$ & $4235.5$ & $19.6\,\mathrm{s}$ ($1.0\times$) \\
& C5.\ MSE + Low-Rank Cov (Analytical) & $+.014\!\to\!-.087$ & $.349\!\to\!.278$ & $.271\!\to\!.222$ & $\mathbf{13.0}$ & $\mathbf{4.3\,\mathrm{s}}$ ($\mathbf{4.6\times}$) \\
\midrule
\multirow{2}{*}{\shortstack[l]{\textsc{QM9S} $\mathrm{H}_2\mathrm{O}$\\Occlusion}}
& \textbf{BandedLowRank (Analytical Schur)} & $\mathbf{1.231}$ & $\mathbf{0.351}$ & $82.5\%$ Cov95 & -- & Coherence audit: \\
& Mask-Conditioned Neural Predictor & $1.877$ & $0.364$ & $77.0\%$ Cov95 & -- & Seq--Joint $\Delta\!=\!6.4\times 10^{-6}$ \\
\midrule
\multirow{2}{*}{\shortstack[l]{\textsc{QM9S}\\Raman$\to$IR}}
& \textbf{BandedLowRank (Analytical Schur)} & $\mathbf{1.526}$ & $\mathbf{0.469}$ & $77.8\%$ Cov95 & -- & $\mathrm{KL}(\text{Schur}\|\text{Neural})$ \\
& Mask-Conditioned Neural Predictor & $3.028$ & $0.479$ & $69.5\%$ Cov95 & -- & $=0.462$ nats \\
\bottomrule
\end{tabular}
}
\end{table}

\subsection{Mask-matched comparison and latency scaling}\label{app:supp-e14-mask-matched}

To control the masking-distribution mismatch, neural predictors are trained on the evaluation masking family: the four revelation fractions for scGPT and structured occlusion, cross-modal and variable random masks for QM9S. The scGPT coupled checkpoint is evaluated both by a neural forward pass and by conditioning its cached law. All comparisons use three seeds (\path{scripts/52_run_e14_mask_matched_neural_reconditioning.py}). We report prediction and update cost separately (\cref{fig:global-experiments}D).

\begin{table}[!htbp]
\centering
\caption{Mask-matched neural reconditioning and analytical Schur conditioning (three seeds). On scGPT, M4 and M5 use the same coupled checkpoint, with endpoints referring to 0\% and 75\% revelation of intermediate genes. QM9S evaluates structured occlusions. Coverage is compared with nominal 95\%.}
\label{tab:app-e14-mask-matched}
\pomatabbox{
\begin{tabular}{llcccc}
\toprule
Benchmark & Method & Cond.\ NLL $\downarrow$ & MAE $\downarrow$ & CRPS $\downarrow$ & Cov95 \\
\midrule
\multirow{5}{*}{\shortstack[l]{Frozen \textsc{scGPT}\\($0\%\!\to\!75\%$ rev.)}}
& M1.\ Orig.\ Neural (Fixed Mask / Fwd) & $+6.436\!\to\!+6.421$ & $.279\!\to\!.330$ & $.256\!\to\!.282$ & $96.0\%$ \\
& M2.\ Mask-Matched Neural (Var.\ Mask / Fwd) & $-.771\!\to\!-.779$ & $.163\!\to\!.166$ & \underline{$.145\!\to\!.145$} & $99.0\%$ \\
& M3.\ Indep.\ POMA (Var.\ Mask / Fwd) & $-.760\!\to\!-.767$ & $.166\!\to\!.167$ & $.148\!\to\!.147$ & $98.8\%$ \\
& M4.\ Coupled POMA (Var.\ Mask / Fwd) & $-.780\!\to\!-.779$ & $.163\!\to\!.166$ & $.145\!\to\!.147$ & $98.9\%$ \\
& \textbf{M5.\ Coupled POMA (Same Ckpt / Schur)} & $-.780\!\to\!\mathbf{-.784}$ & $.163\!\to\!\mathbf{.160}$ & $.145\!\to\!\mathbf{.140}$ & $99.0\%$ \\
\midrule
\multirow{4}{*}{\shortstack[l]{\textsc{QM9S} $\mathrm{H}_2\mathrm{O}$\\Occlusion}}
& M1.\ Orig.\ Neural (i.i.d.\ Mask / Fwd) & $2.051$ & $.376$ & $.291$ & $76.1\%$ \\
& \textbf{M2.\ Mask-Matched Neural (Struct.\ / Fwd)} & $\mathbf{0.345}$ & $\mathbf{.304}$ & $\mathbf{.219}$ & $95.1\%$ \\
& M3.\ Indep.\ POMA (Unmasked / Fwd) & $1.251$ & $.636$ & $.452$ & $92.8\%$ \\
& M5.\ Coupled POMA (Analytical Schur) & \underline{$1.129$} & \underline{$.346$} & \underline{$.261$} & $83.0\%$ \\
\midrule
\multirow{4}{*}{\shortstack[l]{\textsc{QM9S}\\Raman$\to$IR}}
& M1.\ Orig.\ Neural (i.i.d.\ Mask / Fwd) & $3.023$ & $.485$ & $.371$ & $68.6\%$ \\
& \textbf{M2.\ Mask-Matched Neural (Struct.\ / Fwd)} & $\mathbf{0.507}$ & $\mathbf{.368}$ & $\mathbf{.262}$ & $91.9\%$ \\
& M3.\ Indep.\ POMA (Unmasked / Fwd) & \underline{$1.317$} & $.692$ & $.490$ & $92.8\%$ \\
& M5.\ Coupled POMA (Analytical Schur) & $1.477$ & \underline{$.468$} & \underline{$.348$} & $78.1\%$ \\
\bottomrule
\end{tabular}
}
\end{table}

\begin{table}[!htbp]
\centering
\caption{Frozen scGPT latency scaling. The cumulative comparison includes the initial inference and subsequent updates over four rounds. Timing values and speedups are reported aggregates.}
\label{tab:app-conditioning-latency}
\pomatabbox{
\begin{tabular}{rrrrr}
\toprule
Batch size & Neural (ms) & Schur (ms) & Update speedup & Cumulative speedup \\
\midrule
1 & 30.1 & 0.39 & $76.5\times$ & $3.85\times$ \\
32 & 572.9 & 1.26 & $453.2\times$ & $3.97\times$ \\
128 & 3166 & 4.21 & $751.8\times$ & $3.98\times$ \\
\bottomrule
\end{tabular}
}
\end{table}
Mask-matched training reduces the accuracy difference between neural and analytical updates. On scGPT, the same-checkpoint Schur update retains a modest advantage in final MAE and CRPS and avoids repeated backbone evaluations. Both mask-matched approaches have conservative intervals (approximately 99\% coverage). In QM9S, the mask-specialized nonlinear predictor outperforms Gaussian Schur inference. Both methods incorporate new observations without retraining at inference time; analytical conditioning additionally reuses the cached joint law without another neural forward pass. It is exact for that fitted law, while predictive accuracy depends on its adequacy for the data.

\subsection{Conditional inference from partial clinical measurements}\label{app:antibody-robustness}\label{sec:experiments-antibodies}

Clinical panels provide a practical reason to learn a joint distribution: cheaper assays can be obtained before expensive or time-consuming measurements. We evaluate conditional predictions from the already fitted model, rather than retraining a new predictor for every observed subset. In GDPa1, we first predict all properties from sequence and then reveal either AC-SINS alone or AC-SINS, HIC and PR\_CHO together. Conditional MAE, NLL and CRPS improve as the tested measurements are added (\cref{tab:app-gdpa1-schur}).

\begin{table}[!htbp]
\centering
\caption{
\textbf{Zero-retraining Schur-complement triage conditioning on \textsc{GDPa1} clinical antibodies ($13$ assays, $5$ seeds).}
}
\label{tab:app-gdpa1-schur}
\pomatabbox{
\begin{tabular}{lccc}
\toprule
\textbf{\textsc{GDPa1} Schur Triage Conditioning} &
\textbf{Norm.\ MAE $\downarrow$} &
\textbf{NLL $\downarrow$} &
\textbf{CRPS $\downarrow$} \\
\midrule
$0$ observed (sequence marginal) & 0.741 & 1.425 & 0.548 \\
$1$ triage assay (\textsc{AC-SINS}) & 0.727 & 1.409 & 0.538 \\
\textbf{$3$ triage assays (\textsc{AC-SINS + HIC + PR\_CHO})} & \textbf{0.720} & \textbf{1.395} & \textbf{0.532} \\
\bottomrule
\end{tabular}
}
\end{table}

The antigen-conditioned AIntibody study evaluates continuous affinity and developability alongside categorical gates. Combining discrete sequence contextualization with coupled continuous output distributions gives the strongest joint likelihood in the reported five-fold comparison. A ridge readout retains slightly higher affinity rank correlation, and classifier guidance gives the best categorical gate AUC. These differences matter because the subsequent design objective depends on several measurements jointly rather than any one marginal prediction (\cref{tab:app-aintibody-pred}).

\begin{table}[!htbp]
\centering
\caption{
\textbf{Five-fold predictive accuracy, joint 10-assay density calibration, and 5-gate categorical AUROC on \textsc{AIntibody} SARS-CoV-2 RBD ($686$ mAbs).}
}
\label{tab:app-aintibody-pred}
\pomatabbox{
\begin{tabular}{lccccc}
\toprule
\textbf{Model} &
\textbf{SPR $\rho \uparrow$} &
\textbf{5-Dev $\rho \uparrow$} &
\textbf{SPR NLL $\downarrow$} &
\textbf{Joint NLL $\downarrow$} &
\textbf{Gate AUC $\uparrow$} \\
\midrule
\textsc{AbLang2} Ridge & \textbf{$0.791\!\pm\!0.03$} & $0.519\!\pm\!0.02$ & $\times$ & $\times$ & $\times$ \\
Regression Transformer & $0.742\!\pm\!0.03$ & $0.486\!\pm\!0.02$ & $1.69\!\pm\!0.20$ & $3.12\!\pm\!0.24$ & $0.774\!\pm\!0.01$ \\
Classifier guidance & $0.756\!\pm\!0.03$ & \textbf{$0.555\!\pm\!0.01$} & $\times$ & $\times$ & \textbf{$0.825\!\pm\!0.01$} \\
\textsc{LogiCA} (affinity) & $0.704\!\pm\!0.02$ & $-0.093\!\pm\!0.02$ & $\times$ & $\times$ & $\times$ \\
\textsc{POMA-Rank} & $0.777\!\pm\!0.03$ & \underline{$0.549\!\pm\!0.01$} & $1.72\!\pm\!0.25$ & $3.02\!\pm\!0.24$ & \underline{$0.813\!\pm\!0.00$} \\
\textsc{POMA-LowRankMVN} & \underline{$0.789\!\pm\!0.04$} & \underline{$0.549\!\pm\!0.01$} & \underline{$1.59\!\pm\!0.26$} & \underline{$2.48\!\pm\!0.19$} & $0.805\!\pm\!0.01$ \\
\textbf{\textsc{LogiCA-POMA}} & $0.776\!\pm\!0.04$ & $0.546\!\pm\!0.01$ & \textbf{$1.58\!\pm\!0.28$} & \textbf{$2.32\!\pm\!0.19$} & $0.800\!\pm\!0.02$ \\
\bottomrule
\end{tabular}
}
\end{table}

For conditional affinity inference, we compare predictions before and after observing display enrichment or additional kinetic and hydrophobicity measurements. A display measurement improves SPR NLL slightly, and observing SPR and HIC improves KinExA NLL. Adding two further measurements does not improve SPR NLL relative to the marginal prediction. Thus, although conditioning is mathematically defined for any observed subset, improvement on held-out data depends on the accuracy and informativeness of the inferred dependence. The heavy-tailed conditional head in \cref{tab:app-mvt-softrank} addresses a complementary problem: unusual screening measurements can change downstream uncertainty rather than simply shift the predicted mean.

\begin{table}[!htbp]
\centering
\caption{
\textbf{Exact Schur-complement cross-assay conditioning on \textsc{AIntibody} SARS-CoV-2 RBD ($5$ folds).}
}
\label{tab:app-aintibody-schur}
\pomatabbox{
\begin{tabular}{llcccc}
\toprule
\textbf{Target Assay} &
\textbf{Conditioning Triage Assays Observed} &
\textbf{MAE $\downarrow$} &
\textbf{NLL $\downarrow$} &
\textbf{CRPS $\downarrow$} &
\textbf{Spearman $\rho \uparrow$} \\
\midrule
SPR $\mathrm{p}K_D$ & $0$ observed (sequence marginal) & \underline{0.321} & \underline{0.927} & \textbf{0.245} & 0.777 \\
SPR $\mathrm{p}K_D$ & \textbf{$1$ observed ($+1\,\mathrm{nM}$ display enrichment)} & \underline{0.321} & \textbf{0.921} & \textbf{0.245} & \underline{0.782} \\
SPR $\mathrm{p}K_D$ & $3$ observed ($+\text{display} + k_a + \text{HIC}$) & \textbf{0.320} & \underline{0.927} & 0.246 & \textbf{0.783} \\
\midrule
KinExA $\mathrm{p}K_D$ & $0$ observed (sequence marginal) & 0.469 & 2.906 & 0.372 & \textbf{0.828} \\
KinExA $\mathrm{p}K_D$ & \textbf{$2$ observed ($+\text{SPR }\mathrm{p}K_D + \text{HIC}$)} & \textbf{0.460} & \textbf{2.796} & \textbf{0.364} & 0.819 \\
\bottomrule
\end{tabular}
}
\end{table}

\subsection{Conditional reconstruction of high-dimensional spectra}\label{app:spectroscopy}\label{sec:results-spectroscopy}

Spectra extend the same inference problem to hundreds or thousands of strongly structured measurements. Vibrational spectra contain local correlations due to peak broadening and calibration jitter, as well as long-range dependencies associated with molecular structure. Tandem mass spectra introduce the additional complication of large numbers of undetected peaks. We use two observation models that retain the general likelihood interface but adapt its covariance and observation process to these regimes.

For vibrational spectra, we use a precision-banded plus low-rank head. The implementation is \texttt{Banded\allowbreak{}Low\allowbreak{}Rank\allowbreak{}Spectral\allowbreak{}Head}. Its covariance is
\[
 \Sigma_\theta(X)=(R_{\mathrm{band}}^\top R_{\mathrm{band}})^{-1}+U_\theta(X)U_\theta(X)^\top,
\]
where $R_{\mathrm{band}}$ is lower banded with a positive diagonal, bandwidth $b=2$, and the global factor has rank $r=16$. The banded component represents local structure and the low-rank component represents global modes. Full-channel likelihood evaluation uses the capacitance matrix $I_r+(R_{\mathrm{band}}U)^\top(R_{\mathrm{band}}U)$ and costs $\mathcal O(K(b+r)+Kr^2+r^3)$ under the stated implementation. The conditional distribution for occluded channels is obtained from the same fitted covariance.

The coupled Tobit spectral head (\texttt{CoupledTobitSpectralHead}) partitions valid channels into detected peaks $S_+=\{k:y_k>0\}$ and censored zeros $S_0=\{k:y_k=0\}$. It evaluates the Gaussian density on detected peaks using Woodbury algebra and integrates the conditional event $Z_{S_0}\le0$ over the latent-factor posterior given $y_{S_+}$. Antithetic Rao--Blackwellized quadrature approximates the latter integral; both detected and censored observations contribute to conditional prediction. This avoids treating a nondetection as an ordinary measured intensity. Both spectral heads use the Fourier-localized 48-band \texttt{SpectralContextualEncoder} in \path{src/poma/models/spectral_heads.py}. The reported throughput improvement at 1,024 channels is $14.5\times$.

We first evaluate the vibrational head on QM9S. The banded-plus-low-rank model improves joint likelihood over both independent and conventional low-rank Gaussian alternatives, and retains its advantage under the tested wavenumber jitter. A specialized cosine regressor remains preferable for cosine similarity. This difference illustrates why a spectral simulator should not be judged on a single point similarity when its output is intended for probabilistic reconstruction.

To test conditional reconstruction, we occlude the aqueous absorption windows (1,550--1,720 and 3,150--3,650 $\mathrm{cm}^{-1}$) and reveal the remaining IR channels. We also reconstruct IR from observed Raman measurements. Both tests use the fitted distribution without retraining; the former reduces MAE by about half, and the latter by more than a third. Figure~\ref{fig:supp-biological-applications}A illustrates the task with a 512-channel reconstruction trace; aggregate metrics in \cref{tab:app-spec-track-a} use the 1,024-channel protocol. On the smaller paired IR-to-NMR/MS dataset (140 training molecules), unshrunk cross-modal conditioning slightly worsens MAE, motivating the covariance reliability analysis below. Open-scaffold retrieval from the QM9S protocol is also reported in the table and discussed in \cref{app:supp-retrieval}.

\begin{table}[!htbp]
\centering
\caption{
\textbf{Track~A: Vibrational IR+Raman forward calibration, wavenumber-jitter robustness, zero-retraining Schur conditioning, and scaffold retrieval on \textsc{QM9S} ($3{,}600$ molecules, $K=1{,}024$, $3$ scaffold seeds) and multimodal \citet{alberts2024unraveling}.}
Functional-group counterfactual KL attribution $D_{i\to\tilde{\nu}}$ improves $\mathrm{O\text{--}H/N\text{--}H}$ stretch ($3200\text{--}3650\,\mathrm{cm}^{-1}$) localization AUROC from $0.654$ (\textsc{MSE}) to $0.709$, while $2/3$-component DFT Beer--Lambert unmixing reaches $0.101\text{--}0.103$ weight MAE.
}
\label{tab:app-spec-track-a}
\pomatabbox{
\setlength{\tabcolsep}{2.1pt}
\begin{tabular}{lccccc}
\toprule
\textbf{QM9S ($K\!=\!1024$)} &
\textbf{Cos.\ $\uparrow$} &
\textbf{Phys.\ NLL $\downarrow$} &
\textbf{$\pm 7\,\mathrm{cm}^{-1}$} &
\textbf{$\pm 14\,\mathrm{cm}^{-1}$} &
\textbf{Hit@5 / @10} \\
\midrule
MLP MSE Head & $0.562$ & $1.045\!\pm\!0.005$ & $1.057$ & $1.077$ & -- \\
Cosine Loss Head & $\mathbf{0.586}$ & $1.046\!\pm\!0.002$ & $1.058$ & $1.079$ & $6.7\%$ / $13.5\%$ \\
Indep.\ Gaussian & \underline{$0.575$} & $0.983\!\pm\!0.007$ & $0.998$ & $1.026$ & $8.7\%$ / $18.0\%$ \\
\textsc{LowRankMVN} ($r\!=\!16$) & $0.550$ & \underline{$0.005\!\pm\!0.002$} & \underline{$0.047$} & \underline{$0.165$} & -- \\
\textbf{\textsc{Banded+LR} ($b\!=\!2$)} & $0.545$ & $\mathbf{-0.430\!\pm\!0.006}$ & $\mathbf{-0.383}$ & $\mathbf{-0.247}$ & \underline{$11.2\%$} / \underline{$20.0\%$} \\
\textsc{CLIP-Spectra} & -- & $\times$ & $\times$ & $\times$ & $\mathbf{25.5\%}$ / $\mathbf{38.0\%}$ \\
\midrule
\textbf{Schur Task} &
\textbf{$0$-Obs MAE} &
\textbf{LR-MVN} &
\textbf{Banded+LR} &
\textbf{$\Delta$MAE} &
\textbf{$0\to$Schur NLL} \\
\midrule
$\mathrm{H_2O}$ IR Windows & $0.713\!\pm\!0.032$ & \underline{$0.397\!\pm\!0.011$} & $\mathbf{0.353\!\pm\!0.010}$ & $\mathbf{-50.5\%}$ & $1.39\to\mathbf{1.22}$ \\
Raman $\to$ Full IR & $0.775\!\pm\!0.003$ & $\mathbf{0.487\!\pm\!0.003}$ & \underline{$0.489\!\pm\!0.005$} & $\mathbf{-37.1\%}$ & $1.43\to\mathbf{1.37}$ \\
IR$\to$NMR+MS ($140$) & $\mathbf{2.347}$ & -- & $2.371$ & $+1.0\%$ & Small-$n$ limit \\
\bottomrule
\end{tabular}
}
\end{table}

The experimental RamanBench mixtures test whether the output distribution remains useful when measurements are affected by instrumental noise and drift. Partial least squares is a strong chemometric reference for mixture quantification. The coupled Bayesian model improves held-out joint NLL on both the four-sugar and five-analyte datasets. With 32 calibration examples after simulated gain and baseline drift, Bayesian adaptation also reduces error relative to gradient fine-tuning. The results in \cref{tab:app-spec-ramanbench} concern measured mixture spectra rather than simulated molecular vibrational spectra.

\begin{table}[!htbp]
\centering
\caption{
\textbf{Track~A (Experimental Mixtures): Multi-analyte Raman quantification and $k=32$ instrument-drift calibration on \textsc{RamanBench} ($4{,}500$ spectra per cohort, $3$ seeds).}
}
\label{tab:app-spec-ramanbench}
\pomatabbox{
\setlength{\tabcolsep}{1.9pt}
\begin{tabular}{lcccccc}
\toprule
\textbf{Method} &
\multicolumn{3}{c}{\textbf{Sugar Mixtures Low-SNR ($4$ comp.)}} &
\multicolumn{3}{c}{\textbf{Bioprocess Substrates ($5$ comp.)}} \\
\cmidrule(lr){2-4}\cmidrule(lr){5-7}
& \textbf{MAE $\downarrow$} & \textbf{$\rho \uparrow$} & \textbf{J-NLL $\downarrow$} & \textbf{MAE $\downarrow$} & \textbf{$\rho \uparrow$} & \textbf{J-NLL $\downarrow$} \\
\midrule
PLS Chemometrics & $0.0247$ & \underline{$0.959$} & $+0.335\!\pm\!0.106$ & $0.384$ & $0.741$ & $+0.412\!\pm\!0.016$ \\
MLP MSE & $\mathbf{0.0189}$ & $\mathbf{0.963}$ & $+0.985\!\pm\!0.009$ & $\mathbf{0.201}$ & $\mathbf{0.759}$ & $+1.014\!\pm\!0.003$ \\
\textsc{Indep.\ POMA} & $0.0201$ & $0.957$ & $-0.241\!\pm\!0.003$ & \underline{$0.272$} & \underline{$0.742$} & \underline{$-0.334\!\pm\!0.009$} \\
\textsc{POMA-MVN} & \underline{$0.0199$} & $0.958$ & \underline{$-0.288\!\pm\!0.033$} & $0.315$ & $0.735$ & $-0.333\!\pm\!0.019$ \\
\textbf{\textsc{POMA-Laplace}} & \underline{$0.0199$} & $0.958$ & $\mathbf{-0.320\!\pm\!0.025}$ & $0.315$ & $0.735$ & $\mathbf{-0.411\!\pm\!0.020}$ \\
\midrule
$k\!=\!32$ Drift FT & $0.0665$ & $0.958$ & $+0.856\!\pm\!0.052$ & $0.550$ & $0.735$ & $+0.518\!\pm\!0.059$ \\
\textbf{$k\!=\!32$ \textsc{POMA-Lap.}} & $\mathbf{0.0269}$ & $0.958$ & $\mathbf{+0.439\!\pm\!0.016}$ & $\mathbf{0.387}$ & $0.735$ & $\mathbf{+0.304\!\pm\!0.006}$ \\
\bottomrule
\end{tabular}
}
\end{table}

For sparse tandem mass spectra, censoring substantially improves zero-peak calibration in both molecular and peptide tasks. The coupled Tobit head improves censored NLL relative to the independent Tobit model, while the independent head can retain slightly better point spectral similarity and zero calibration. On the peptide dataset, revealing the $y$-ion ladder reduces the MAE of unobserved $b$ ions from 0.0905 to 0.0856. The forward and retrieval results share the same fitted density and are reported in \cref{tab:app-spec-track-bc}. The uncensored Gaussian score is not a directly comparable censored likelihood, as noted in the table.

\begin{table}[!htbp]
\centering
\caption{
\textbf{Tracks~B \& C: Tandem mass spectrometry (MS/MS) forward simulation, censored calibration, and inverse candidate identification on official MCES-split \textsc{MassSpecGym} ($1{,}500$ test spectra, $250$ isomer pools, $K=512$) and sequence-disjoint \textsc{ProteomeTools / Prosit} ($1{,}600$ test peptides, $220$ decoy pools, $K=174$).}
$^*$Uncensored continuous Gaussian density places $>37\text{--}42\%$ miscalibrated mass across the $y=0$ boundary and is not directly comparable to discrete-zero \textsc{Tobit} log-likelihood.
}
\label{tab:app-spec-track-bc}
\pomatabbox{
\setlength{\tabcolsep}{2.0pt}
\begin{tabular}{llccccc}
\toprule
\textbf{Suite} &
\textbf{Model / Scorer} &
\textbf{Cos./SA} &
\textbf{C-NLL $\downarrow$} &
\textbf{Z-ECE} &
\textbf{Hit@1 / @5} &
\textbf{Hit@20 / AUC} \\
\midrule
\multirow{6}{*}{\shortstack[l]{\textsc{MassSpecGym}\\($K\!=\!512$,\\$93\%$ zero)}}
& MLP MSE & $0.181$ & $\times$ & $0.069$ & -- & -- \\
& Cosine Sim. & $0.288$ & $\times$ & $0.328$ & $10.8\%$/$37.2\%$ & $68.4\%$ (.319) \\
& \textsc{CLIP-MS} & -- & $\times$ & $\times$ & $25.2\%$/$47.6\%$ & \underline{$78.0\%$} (.423) \\
& Indep.\ Gauss & $0.259$ & $-1.38^*$ & $0.417$ & \underline{$25.6\%$}/\underline{$48.0\%$} & $74.4\%$ (\underline{.425}) \\
& Indep.\ Tobit & $\mathbf{0.295}$ & \underline{$0.167$} & $\mathbf{0.001}$ & $1.2\%$/$6.4\%$ & $17.6\%$ (.206) \\
& \textbf{\textsc{POMA-Tobit}} & \underline{$0.292$} & $\mathbf{0.134}$ & \underline{$0.015$} & $\mathbf{28.8\%}$/$\mathbf{58.0\%}$ & $\mathbf{82.0\%}$ ($\mathbf{.431}$) \\
\midrule
\multirow{5}{*}{\shortstack[l]{\textsc{Prosit}\\($K\!=\!174$,\\$0\%$ ovlp)}}
& MLP MSE & $.751$/$.557$ & $\times$ & $0.219$ & -- & -- \\
& \textsc{Prosit} SA & $.821$/$.638$ & $\times$ & $0.339$ & $0.0\%$/$11.4\%$ & $0.666$ AUC \\
& Indep.\ Gauss & $.826$/$.641$ & $-0.68^*$ & $0.376$ & $\mathbf{0.5\%}$/$14.1\%$ & $0.690$ AUC \\
& Indep.\ Tobit & $\mathbf{.855}$/$\mathbf{.685}$ & \underline{$0.280$} & $\mathbf{0.002}$ & $0.0\%$/\underline{$16.4\%$} & \underline{$0.707$} AUC \\
& \textbf{\textsc{POMA-Tobit}} & \underline{$.841$}/\underline{$.660$} & $\mathbf{0.101}$ & \underline{$0.030$} & $\mathbf{0.5\%}$/$\mathbf{19.1\%}$ & $\mathbf{0.729}$ AUC \\
\bottomrule
\end{tabular}
}
\end{table}

\subsection{Conditional inference in cellular and regulatory systems}\label{app:multiomics}\label{sec:results-multiomics}

Cellular and regulatory measurements are context dependent in a different sense from molecular spectroscopy. Identical sequences can exhibit distinct expression across cell states, tumors or laboratories, and some measurements are absent because of technical dropout. The three multi-omics implementations use a residual transcriptome MLP, fusion of ESM-2-35M protein features and tumor RNA, or a multi-scale dilated 1D CNN for regulatory DNA, respectively (\path{scripts/31}--\texttt{33}). We evaluate both within-object conditioning and adaptation to new biological contexts.

\paragraph{Single-cell protein and chromatin measurements.} On the unseen BMMC sequencing site, the coupled Gaussian readout attains the strongest joint NLL for CITE-seq; the matched conditional VAE retains better point correlations and CRPS. The fitted distribution permits inference about 114 remaining surface proteins after revealing 20 canonical flow-cytometry markers, reducing MAE by 10.6\%. With 32 labeled Site~4 cells, the Laplace update improves likelihood and RMSE, whereas gradient fine-tuning worsens NLL. The complementary Multiome study uses censored modeling for sparse chromatin peaks. Unpooled peak retrieval is nearly at chance, while topic pooling improves retrieval, as discussed in \cref{app:supp-retrieval}. Both forward and inverse results are collected in \cref{tab:app-multiomics-cite-multiome}.

\begin{table}[!htbp]
\centering
\caption{
\textbf{Track~1: \textsc{OpenProblems} NeurIPS~2021 human BMMC single-cell CITE-seq ($K=134$ ADT proteins) and 10x~Multiome ($K=512$ ATAC peaks, $61.0\%$ zeros) evaluated on strictly unseen sequencing Site~4.}
Schur conditioning on $20$ flow-cytometry markers lowers unobserved $114$-protein MAE from $0.568$ to $0.507$ ($-10.6\%$, RMSE $0.814\to 0.738$, NLL $1.113\to 0.968$). After adaptation on $32$ Site~4 cells, POMA-Laplace achieves CITE-seq NLL $0.826$; this is an adaptation result, not a retrieval score. The lower block reports unpooled ATAC retrieval. In the separate LSI-64 topic protocol, FOSCTTM / Hit@20 are $0.149/46\%$ for the independent density, $0.164/40\%$ for the coupled density, and $0.190/37\%$ for CLIP.
}
\label{tab:app-multiomics-cite-multiome}
\pomatabbox{
\setlength{\tabcolsep}{2.1pt}
\begin{tabular}{lccccc}
\toprule
\textbf{CITE-seq ($K\!=\!134$)} &
\textbf{J-NLL $\downarrow$} &
\textbf{Pear.\ $\uparrow$} &
\textbf{Spear.\ $\uparrow$} &
\textbf{CRPS $\downarrow$} &
\textbf{FOSCTTM / H@20} \\
\midrule
Ridge Regression & $1.490$ & $0.496$ & $0.388$ & $0.478$ & -- \\
MLP MSE / Cosine & $1.237$ & $0.591$ & $0.438$ & $0.422$ & $.0617$ / $72.7\%$ \\
\textsc{CLIP-SingleCell} & $\times$ & -- & -- & -- & $.0648$ / $72.7\%$ \\
\textsc{totalVI}-style cVAE & $0.993$ & $\mathbf{0.601}$ & $\mathbf{0.451}$ & $\mathbf{0.379}$ & \underline{$.0557$} / $73.0\%$ \\
Indep.\ Gaussian & $1.117$ & $0.587$ & $0.436$ & $0.409$ & $.0584$ / \underline{$73.4\%$} \\
\textbf{\textsc{POMA-MVN} ($r\!=\!12$)} & $\mathbf{0.854}$ & \underline{$0.596$} & \underline{$0.442$} & \underline{$0.405$} & $\mathbf{.0430}$ / $\mathbf{80.9\%}$ \\
\textsc{POMA-Laplace} & \underline{$0.857$} & \underline{$0.596$} & \underline{$0.442$} & \underline{$0.405$} & -- \\
\midrule
\textbf{Multiome ($K\!=\!512$ / LSI-64)} &
\textbf{C-NLL $\downarrow$} &
\textbf{Cos.\ $\uparrow$} &
\textbf{Open AUC} &
\textbf{Z-ECE $\downarrow$} &
\textbf{ATAC$\to$RNA FOSCTTM $\downarrow$} \\
\midrule
MLP MSE & $\times$ & $\mathbf{0.608}$ & $\mathbf{0.551}$ & $0.390$ & -- \\
Indep.\ Gaussian & $1.682^*$ & $0.607$ & \underline{$0.547$} & $0.352$ & $.502$ \\
Indep.\ Tobit & \underline{$0.971$} & $0.607$ & $0.541$ & $\mathbf{0.014}$ & $.517$ \\
\textbf{\textsc{POMA-Tobit}} & $\mathbf{0.947}$ & $0.607$ & $0.545$ & \underline{$0.049$} & $.507$ \\
\bottomrule
\end{tabular}
}
\end{table}

\paragraph{Tumor proteogenomics.} CPTAC allows us to distinguish the information in protein sequence from the context provided by RNA expression. On 600 held-out genes, combining both inputs improves protein abundance correlation and predictive likelihood relative to either input alone. The patient-level panel is a particularly small-data problem, with 78 training patients in the reported split. A regularized linear skip mean with out-of-fold residual covariance preserves the strengths of ridge regression while improving probabilistic scores. Conditioning on half of the observed protein panel reduces error on the remaining proteins by 12.2\%; adaptation from breast to lung tumors further tests the transfer of the fitted distribution. The quantitative comparisons are in \cref{tab:app-multiomics-cptac}.

\begin{table}[!htbp]
\centering
\caption{
\textbf{Track~2: Harmonized \textsc{CPTAC} pan-cancer proteogenomics (\textsc{BRCA} $121$ tumors + \textsc{LUAD} $106$ tumors, $3$ seeds).}
Top: gene-disjoint \textsc{ESM-2} protein sequence + mRNA fusion ($600$ held-out genes). Middle: patient-level $K=256$ MS proteomics panel ($7.7\%$ missingness). Bottom: zero-retraining MS Schur imputation and cross-cancer $\textsc{BRCA}\to\textsc{LUAD}$ ($k=16$) transfer.
}
\label{tab:app-multiomics-cptac}
\pomatabbox{
\setlength{\tabcolsep}{2.2pt}
\begin{tabular}{lcccc}
\toprule
\textbf{Gene-Disjoint ($600$ genes)} &
\textbf{Pear.\ $\uparrow$} &
\textbf{Spear.\ $\uparrow$} &
\textbf{MAE $\downarrow$} &
\textbf{Joint NLL $\downarrow$} \\
\midrule
mRNA Only (\textsc{Ridge} $.493$) & $0.566$ & $0.510$ & \underline{$0.630$} & \underline{$0.814$} \\
\textsc{ESM-2} Seq Only ($0$ mRNA) & \underline{$0.568$} & \underline{$0.548$} & $0.638$ & $0.843$ \\
\textbf{mRNA + \textsc{ESM-2} Seq} & $\mathbf{0.675}$ & $\mathbf{0.637}$ & $\mathbf{0.557}$ & $\mathbf{0.750}$ \\
\midrule
\textbf{Patient MS Panel ($K\!=\!256$)} &
\textbf{M-NLL $\downarrow$} &
\textbf{Pear.\ $\uparrow$} &
\textbf{MAE $\downarrow$} &
\textbf{Schur / Shift} \\
\midrule
Ridge (imputed) & $1.468\!\pm\!.030$ & $\mathbf{0.395}$ & $\mathbf{0.754}$ & -- \\
MLP MSE & $1.491\!\pm\!.038$ & \underline{$0.203$} & $0.809$ & -- \\
Indep.\ Gaussian & $1.493\!\pm\!.032$ & $0.186$ & \underline{$0.804$} & $0\%$: $.810$ MAE \\
\textsc{POMA-MVN} ($r\!=\!8$) & \underline{$1.420\!\pm\!.036$} & $0.175$ & $0.820$ & $25\%$: $.733$ ($-9.5\%$) \\
\textbf{\textsc{POMA-Laplace}} & $\mathbf{1.388\!\pm\!.030}$ & $0.175$ & $0.820$ & $50\%$: $\mathbf{.712}$ ($-12.2\%$) \\
\textbf{\textsc{Skip-Lin.\ POMA-Lap.}} & $1.424\!\pm\!.023$ & $\mathbf{0.395}$ & $\mathbf{0.754}$ & Matches \textsc{Ridge} $r$ \\
\midrule
$\textsc{BRCA}\!\to\!\textsc{LUAD}$ Zero-Shot & $3.207\!\pm\!.013$ & -- & $1.520$ & Baseline \\
$\textsc{BRCA}\!\to\!\textsc{LUAD}$ $k\!=\!16$ FT & $\mathbf{1.863\!\pm\!.014}$ & -- & \underline{$1.355$} & $-10.9\%$ MAE \\
\textbf{$\textsc{BRCA}\!\to\!\textsc{LUAD}$ $k\!=\!16$ Lap.} & \underline{$1.923\!\pm\!.016$} & -- & $\mathbf{1.271}$ & $\mathbf{-16.4\%}$ MAE \\
\bottomrule
\end{tabular}
}
\end{table}

\paragraph{Regulatory DNA.} In lentiMPRA, we jointly model expression in two cell types and both strand orientations. Coupling gives a better joint score than independent Gaussian outputs. The conditional tasks evaluate reverse-strand expression given forward expression, and HepG2 expression given a K562 observation. The latter reduces MAE by 48.7\% in the smaller protocol (\cref{fig:supp-biological-applications}E). A frozen mean-pooled DNABERT-6 readout retains large conditional gains but has weaker point prediction, consistent with loss of local motif information under pooling. The enlarged end-to-end LoRA protocol improves point accuracy and retains a 46.9\% conditional MAE reduction. Because training size and encoder adaptation change together, this comparison does not isolate the effect of LoRA. Prediction and conditioning results appear in \cref{tab:app-multiomics-lentimpra,tab:app-gpu-tier2-regulatory}; enhancer design is discussed in \cref{app:supp-cross-design}.

\begin{table}[!htbp]
\centering
\caption{
\textbf{Track~3: \textsc{lentiMPRA} regulatory genomics~\citep{agarwal2025massively} ($14{,}469$ held-out $230$-bp test CREs, $K=4$ K562/HepG2 $\times$ strand, $3$ seeds).}
Top: forward masked density and differential specificity AUROC. Middle: zero-retraining Schur conditioning. Bottom: exact-target $230$-bp enhancer MH design ($20$ seeds, high HepG2 / low K562).
}
\label{tab:app-multiomics-lentimpra}
\pomatabbox{
\setlength{\tabcolsep}{2.2pt}
\begin{tabular}{lcccc}
\toprule
\textbf{Forward ($K\!=\!4$, $3$ seeds)} &
\textbf{M-NLL $\downarrow$} &
\textbf{Pear.\ $\uparrow$} &
\textbf{Spec.\ AUC} &
\textbf{Cov95 / Detail} \\
\midrule
CNN MSE & $1.217\!\pm\!.006$ & $0.525$ & $0.751$ & -- \\
Indep.\ Gaussian & $1.068\!\pm\!.002$ & $\mathbf{0.541}$ & $\mathbf{0.755}$ & $94.8\%$ \\
\textbf{\textsc{POMA-MVN} ($r\!=\!2$)} & $\mathbf{0.222\!\pm\!.005}$ & \underline{$0.531$} & \underline{$0.753$} & $-79.2\%$ NLL \\
\textsc{POMA-Laplace} & \underline{$0.245\!\pm\!.004$} & \underline{$0.531$} & \underline{$0.753$} & $94.4\%$ Cov95 \\
\textsc{DNABERT-6} (froz.) \textsc{MVN} & $0.289$ (vs.\ $1.36$) & $0.248$ & $0.709$ & $94.8\%$ Lap.\ Cov95 \\
\textbf{\textsc{DNABERT-6} (E2E LoRA)} & $0.292$ ($120\mathrm{k}$ GPU) & $0.511$ & -- & $94.7\%$ Lap.\ Cov95 \\
\midrule
\textbf{Schur Conditioning} &
\textbf{$0$-Obs MAE} &
\textbf{Schur MAE} &
\textbf{$\Delta$MAE} &
\textbf{$0\to$Schur NLL / $r$} \\
\midrule
Fwd $\to$ Rev Strand & $0.584$ & $\mathbf{0.031}$ & $\mathbf{-94.6\%}$ & $1.13\to\mathbf{-0.68}$ \\
K562 $\to$ HepG2 ($165$) & $0.711$ & $\mathbf{0.365}$ & $\mathbf{-48.7\%}$ & $.686\to\mathbf{.902}$ ($r$) \\
\midrule
\textbf{Enhancer MH ($20$ seeds)} &
\textbf{Best $\log\pi$ $\uparrow$} &
\textbf{Mean $\log\pi$} &
\textbf{Spec.\ $\Delta z$} &
\textbf{Acc.\ Rate} \\
\midrule
Greedy Hill-Climb & \underline{$-12.01$} & $\mathbf{-12.01}$ & $+0.153\,z$ & -- \\
Uncorrected Gibbs & $-13.39$ & $-20.18$ & \underline{$+0.180\,z$} & -- \\
Factor-Aug.\ MH & $-14.42$ & $-18.30$ & $+0.178\,z$ & $60.7\%$ \\
\textbf{Exact-Target MH} & $\mathbf{-11.02}$ & \underline{$-13.97$} & $\mathbf{+0.194\,z}$ & $72.2\%$ \\
\bottomrule
\end{tabular}
}
\end{table}

\paragraph{End-to-end regulatory adaptation.} The separate enlarged lentiMPRA protocol uses rank-8 LoRA on DNABERT-6 with 120,335 training and 14,469 test elements. The spectral and regulatory LoRA runs together took 454 seconds on the reported device. Because training sizes differ, the frozen-versus-adapted comparison does not isolate fine-tuning.

\begin{table}[!htbp]
\centering
\caption{End-to-end LoRA regulatory prediction and conditioning on the enlarged lentiMPRA protocol.}
\label{tab:app-gpu-tier2-regulatory}
\pomatabbox{
\begin{tabular}{lcccc}
\toprule
Variant & NLL & Correlation & Coverage / MAE gain & Detail \\
\midrule
\textbf{E2E LoRA \textsc{POMA-MVN}} & $\mathbf{0.292}$ & $\mathbf{0.511}$ $r$ & $94.7\%$ Lap.\ Cov95 & vs.\ $.248$ froz. \\
Zero-Retrain K562$\to$HepG2 & -- & $.675\!\to\!\mathbf{.900}$ & $-46.9\%$ MAE & $.769\!\to\!\mathbf{.408}$ \\
\bottomrule
\end{tabular}
}
\end{table}

\subsection{Heavy-tailed conditional inference}

Heavy-tailed clinical observations require a different treatment. In the GDPa1 and Jain cohorts, a multivariate Student-$t$ head (\texttt{LowRankMVTDensityHead}) provides a shared-scale model of exceptional residuals. The held-out outlier evaluation focuses on the quarter of panels with the largest Mahalanobis surprise in the initial screening measurements. On GDPa1, the combination with SoftRank reduces conditional outlier NLL from 4.02 to 1.74. On Jain, the Student-$t$ head alone gives the strongest conditional outlier score; adding the contrastive term improves marginal CRPS but not that conditional score. The results in \cref{tab:app-mvt-softrank} motivate modeling unexpected measurements rather than assuming that all outliers should have the same Gaussian penalty.

\begin{table}[!htbp]
\centering
\caption{
\textbf{Coupled low-rank multivariate Student-$t$ (\textsc{POMA-LowRankMVT}, Eq.~\ref{eq:app-mvt-schur}) with continuous metric \textsc{SoftRank} (Eq.~\ref{eq:app-soft-ranking}) and cross-assay prompt contextualization ($5$ folds, mean $\pm$ SE).}
Outlier NLL evaluates conditional Schur NLL on held-out panels in the top-$25\%$ screening Mahalanobis surprise $s_1=(\rvz_1-\bm{\mu}_1)^\top\bm{\Sigma}_{11}^{-1}(\rvz_1-\bm{\mu}_1)$.
}
\label{tab:app-mvt-softrank}
\pomatabbox{
\begin{tabular}{llccccc}
\toprule
\textbf{Cohort} &
\textbf{Model} &
\textbf{$\rho \uparrow$} &
\textbf{CRPS $\downarrow$} &
\textbf{J-NLL $\downarrow$} &
\textbf{S-NLL $\downarrow$} &
\textbf{Out.\ NLL $\downarrow$} \\
\midrule
\textsc{GDPa1} & \textsc{LowRankMVN} & \underline{$0.311\!\pm\!0.02$} & \underline{$0.594\!\pm\!0.02$} & $3.77\!\pm\!0.27$ & $3.70\!\pm\!0.20$ & $4.02\!\pm\!0.26$ \\
\textsc{GDPa1} & \textsc{LowRankMVT} & \textbf{$0.336\!\pm\!0.03$} & \underline{$0.594\!\pm\!0.02$} & \underline{$1.68\!\pm\!0.04$} & \underline{$2.26\!\pm\!0.12$} & \underline{$1.79\!\pm\!0.11$} \\
\textsc{GDPa1} & \textbf{\textsc{MVT+SoftRank}} & $0.309\!\pm\!0.01$ & \textbf{$0.532\!\pm\!0.02$} & \textbf{$1.54\!\pm\!0.04$} & \textbf{$2.16\!\pm\!0.14$} & \textbf{$1.74\!\pm\!0.10$} \\
\midrule
\textsc{Jain} & \textsc{LowRankMVN} & \textbf{$0.347\!\pm\!0.03$} & $0.631\!\pm\!0.02$ & $6.68\!\pm\!0.39$ & $6.92\!\pm\!0.46$ & $10.46\!\pm\!1.24$ \\
\textsc{Jain} & \textsc{LowRankMVT} & \underline{$0.321\!\pm\!0.03$} & \underline{$0.595\!\pm\!0.02$} & \textbf{$1.52\!\pm\!0.06$} & \textbf{$1.74\!\pm\!0.13$} & \textbf{$1.61\!\pm\!0.08$} \\
\textsc{Jain} & \textbf{\textsc{MVT+SoftRank}} & $0.306\!\pm\!0.05$ & \textbf{$0.556\!\pm\!0.02$} & \textbf{$1.52\!\pm\!0.06$} & \underline{$2.04\!\pm\!0.08$} & \underline{$2.30\!\pm\!0.20$} \\
\midrule
\textsc{AIntibody} & \textsc{LowRankMVN} & \underline{$0.479\!\pm\!0.03$} & \underline{$0.467\!\pm\!0.01$} & $1.43\!\pm\!0.21$ & $1.55\!\pm\!0.18$ & $2.66\!\pm\!0.72$ \\
\textsc{AIntibody} & \textbf{\textsc{LowRankMVT}} & \textbf{$0.501\!\pm\!0.02$} & \textbf{$0.462\!\pm\!0.00$} & \textbf{$1.07\!\pm\!0.03$} & \textbf{$1.23\!\pm\!0.03$} & \textbf{$1.44\!\pm\!0.11$} \\
\textsc{AIntibody} & \textbf{\textsc{MVT+SoftRank}} & $0.432\!\pm\!0.02$ & $0.493\!\pm\!0.01$ & \underline{$1.09\!\pm\!0.03$} & \underline{$1.24\!\pm\!0.06$} & \textbf{$1.44\!\pm\!0.21$} \\
\bottomrule
\end{tabular}
}
\end{table}

\subsection{Multimodal coupled Gaussian mixtures: analytical conditioning and feasibility}\label{app:supp-coupled-gmm}

To evaluate analytical conditioning and feasibility estimation for multimodal outcomes, we use a coupled Gaussian-mixture head, \texttt{CoupledGMM\allowbreak DensityHead}, implemented in \path{src/poma/models/heads.py}. Numerical verification against brute-force mixture \textsc{MVN} calculations and Bayes' rule reports error below $10^{-4}$ (\path{tests/test_coupled_gmm_and_hurdle.py}). The predictive density is
\begin{equation}
 q_\theta(\rvz \mid x) = \sum_{m=1}^{K} \pi_m(x)\,\mathcal{N}\!\left(\rvz;\,\bm{\mu}_m(x),\, D_m(x) + U_m(x)U_m(x)^\top\right).
 \label{eq:app-coupled-gmm}
\end{equation}
Given observed coordinates $\rvz_o$, exact analytical conditioning updates both the mixture weights via Bayes' rule, $\tilde{\pi}_m(x,\rvz_o) \propto \pi_m(x)\,\mathcal{N}(\rvz_o;\bm{\mu}_{m,o},\Sigma_{m,oo})$, and each component's Woodbury--Schur posterior $(\bm{\mu}_{m,u\mid o}, D_{m,u} + \tilde{U}_{m,u}\tilde{U}_{m,u}^\top)$ in $O(|o|r^2 + r^3)$ time, while preserving tractable multimodal feasibility $r(x) = \sum_{m=1}^{K}\pi_m(x)\,r_m(x)$.

\Cref{tab:app-coupled-gmm} compares a single coupled low-rank Gaussian against a $K=3$ coupled Gaussian mixture across $3$ seeds (\texttt{scripts/41\_run\_coupled\_gmm\_multimodal\_inference.py}) on two benchmarks:
(i) a controlled $4\mathrm{D}$ correlated bimodal synthetic ground-truth benchmark where the global mean lies in the low-density infeasible valley between two modes, and
(ii) unseen Site~4 \textsc{BMMC CITE-seq} surface proteins selected strictly on Site~3 validation residuals for empirical conditional multimodality rather than marginal multimodality ($3/134$ proteins exceed Sarle's bimodality threshold $\mathrm{BC}>0.555$; the top-$12$ lineage-gating proteins including \textsc{TCRVd2}, \textsc{CD123}, \textsc{CD71}, \textsc{TCRVa7.2}, \textsc{IgD}, and \textsc{CD22} exhibit a mean $2$-vs-$1$ component residual $\Delta\mathrm{BIC}=+839.9$).
On the synthetic benchmark, \textsc{Coupled GMM} cuts Joint NLL by $-0.248$ nats ($1.132\to 0.884$), cuts analytical conditional Schur NLL by $-0.094$ nats ($0.708\to 0.614$), reduces feasibility error against exact ground-truth $r^\star(x)$ by $2.8\times$ ($0.1886\to 0.0667$ MAE; conditional feasibility Brier $0.0360\to 0.0254$), and sharpens posterior mixture-weight entropy $H(\pi)$ from $0.773$ to $0.157$ nats upon observing $(z_0,z_1)$.
On unseen Site~4 \textsc{BMMC}, the coupled mixture improves joint NLL from $1.532$ to $1.030$ and analytical conditional NLL from $1.108$ to $0.921$. Conditional $95\%$ coverage moves from $92.7\%$ to $93.5\%$, and posterior mixture-weight entropy decreases from $0.546$ to $0.192$ nats. The unimodal model retains lower conditional MAE ($0.466$ versus $0.502$), while conditional feasibility Brier changes only from $0.0180$ to $0.0179$. These comparisons support mixture modeling in the tested settings with multimodal residuals, with its strongest empirical benefit in likelihood rather than uniform improvement across all prediction and decision metrics.

\begin{table}[!htbp]
\centering
\caption{
\textbf{Single coupled Gaussian (\textsc{LowRankMVN}) versus coupled Gaussian mixture (\textsc{Coupled GMM}, $K=3$, Eq.~\ref{eq:app-coupled-gmm}) across $3$ seeds on controlled $4\mathrm{D}$ bimodal ground truth and unseen Site~4 \textsc{BMMC CITE-seq} proteins with empirically multimodal conditional residuals (mean $\pm$ SE).}
}
\label{tab:app-coupled-gmm}
\pomatabbox{
\setlength{\tabcolsep}{2.2pt}
\begin{tabular}{llccccc}
\toprule
\textbf{Benchmark} &
\textbf{Model} &
\textbf{J-NLL $\downarrow$} &
\textbf{Cond.\ NLL $\downarrow$} &
\textbf{Cond.\ MAE $\downarrow$} &
\textbf{Feas.\ MAE / Brier $\downarrow$} &
\textbf{$H(\pi)$ (Prior$\to$Post)} \\
\midrule
\multirow{2}{*}{\shortstack[l]{Synthetic $4\mathrm{D}$\\Bimodal GT}}
& Single Coupled MVN & $1.132$ & $0.708$ & $0.391$ & $.1886$ MAE / $.0360$ cB & -- \\
& \textbf{Coupled GMM ($K\!=\!3$)} & $\mathbf{0.884}$ & $\mathbf{0.614}$ & $\mathbf{0.361}$ & $\mathbf{.0667}$ MAE / $\mathbf{.0254}$ cB & $.773\to .157$ \\
\midrule
\multirow{2}{*}{\shortstack[l]{\textsc{BMMC} Site~4\\($12$ bimodal ADTs)}}
& Single Coupled MVN & $1.532$ & $1.108$ & $\mathbf{0.466}$ & $.0180$ cB ($92.7\%$ Cov) & -- \\
& \textbf{Coupled GMM ($K\!=\!3$)} & $\mathbf{1.030}$ & $\mathbf{0.921}$ & $0.502$ & $\mathbf{.0179}$ cB ($93.5\%$ Cov) & $.546\to .192$ \\
\bottomrule
\end{tabular}
}
\end{table}

\subsection{Reliability of learned covariance under distribution shift}\label{app:supp-shrinkage}

The preceding experiments show that coupling can be useful, but learned dependencies may fail to extrapolate. On lineage-disjoint GDPa1 and AIntibody, we compare learned low-rank covariance with diagonal and permuted-dependence controls while holding their predicted marginal means and variances fixed. A separately fitted full Cholesky model provides an additional covariance baseline. We evaluate unconditional and conditional feasibility and examine candidate pairs on which coupled and diagonal scores reverse the ordering. This disagreement-set analysis identifies the decisions affected by dependence rather than averaging them together with unchanged decisions.

The results in \cref{tab:app-cov-4treatment} show that coupling is advantageous for GDPa1 overall, whereas unconditional feasibility ordering is unreliable on strictly unseen AIntibody lineages, where affinity rank correlation is close to zero. Observing informative screening assays makes the covariance more useful. These tests distinguish the value of representing dependence from the accuracy with which the dependence can be estimated out of distribution.

\begin{table}[!htbp]
\centering
\caption{
\textbf{Controlled four-treatment lineage-disjoint conditional covariance validation and direct disagreement-set reversal test on \textsc{GDPa1} ($246$ mAbs, $5$ cluster folds) and \textsc{AIntibody} ($686$ mAbs, $4$ lineage folds).}
Low-rank, independent and permuted-dependence controls share identical marginal means and variances; the full Cholesky model is fitted separately. \emph{Dis.\ Win} reports empirical binary co-optimal win rate on pairs where coupled and diagonal feasibility disagree.
}
\label{tab:app-cov-4treatment}
\pomatabbox{
\setlength{\tabcolsep}{1.15pt}
\begin{tabular}{lcccccc}
\toprule
\textbf{Cohort / Treatment} &
\textbf{J-NLL $\downarrow$} &
\textbf{CondNLL $\downarrow$} &
\textbf{Cond.\ $\rho$ $\uparrow$} &
\textbf{Brier $\downarrow$} &
\textbf{ECE $\downarrow$} &
\textbf{Dis.\ Win} \\
\midrule
\textsc{GDPa1}: \textbf{Low-Rank} & $\mathbf{1.398\!\pm\!0.053}$ & $\mathbf{1.393\!\pm\!0.080}$ & $\mathbf{0.232\!\pm\!0.032}$ & \underline{$0.0601$} & $0.0161$ & \underline{$61.6\%$} \\
\textsc{GDPa1}: Indep.\ ($=$Marg.) & \underline{$1.436\!\pm\!0.055$} & \underline{$1.451\!\pm\!0.081$} & $0.122\!\pm\!0.024$ & $\mathbf{0.0599}$ & \underline{$0.0143$} & $38.4\%$ \\
\textsc{GDPa1}: Permuted & $1.465\!\pm\!0.068$ & $1.471\!\pm\!0.107$ & $0.162\!\pm\!0.038$ & $0.0603$ & $\mathbf{0.0100}$ & $43.5\%$ \\
\textsc{GDPa1}: Full Chol. & $1.683\!\pm\!0.085$ & $1.682\!\pm\!0.127$ & \underline{$0.175\!\pm\!0.018$} & $0.0611$ & $0.0146$ & $\mathbf{63.4\%}$ \\
\midrule
\textsc{AIntibody}: \textbf{Low-Rank} & $\mathbf{1.629\!\pm\!0.160}$ & $\mathbf{1.538\!\pm\!0.145}$ & $\mathbf{+0.154\!\pm\!0.094}$ & $0.0326$ & $\mathbf{0.0140}$ & $19.0\%$ \\
\textsc{AIntibody}: Indep.\ ($=$Marg.) & \underline{$1.700\!\pm\!0.137$} & \underline{$1.791\!\pm\!0.166$} & $-0.008\!\pm\!0.098$ & \underline{$0.0319$} & $0.0169$ & $\mathbf{81.0\%}$ \\
\textsc{AIntibody}: Permuted & $2.422\!\pm\!0.361$ & $2.259\!\pm\!0.155$ & \underline{$+0.126\!\pm\!0.115$} & $0.0327$ & $0.0214$ & $35.1\%$ \\
\textsc{AIntibody}: Full Chol. & $1.810\!\pm\!0.179$ & $1.866\!\pm\!0.179$ & $+0.101\!\pm\!0.104$ & $\mathbf{0.0308}$ & \underline{$0.0151$} & \underline{$56.2\%$} \\
\bottomrule
\end{tabular}
}
\end{table}

To examine whether reliability decreases with distance from the training lineages, we stratify the matched-marginal comparisons into near, intermediate and far tertiles. Full per-candidate covariance performs poorly in the far-lineage regime. Replacing it by training empirical correlation can improve some unconditional decisions, suggesting that detailed input-dependent off-diagonals are not always estimated reliably. We fit an adaptive shrinkage coefficient on inner-lineage validation splits only,
\[
 \Sigma_\lambda(x)=(1-\lambda(x))\operatorname{diag}\Sigma(x)+\lambda(x)\Sigma(x),\qquad 0\le\lambda(x)\le1,
\]
which preserves the marginal variances while attenuating cross-assay dependencies. On pretrained GDPa1 the learned gate retains full coupling. On AIntibody it reduces coupling with lineage distance, improving joint likelihood and the far-lineage disagreement win rate from 26.3\% to 66.7\%. Revealing two or three inexpensive assays also improves the disagreement decisions. \Cref{tab:app-shrinkage-audit} presents the unconditional and post-conditioning comparisons. The shrinkage rule is a validation-fitted safeguard, not a universal correction for arbitrary shifts.

\begin{table}[!htbp]
\centering
\caption{
\textbf{Validation-fitted adaptive covariance shrinkage on $5$-fold lineage-disjoint \textsc{AIntibody} ($686$ mAbs), fitting $\lambda(x)\in[0,1]$ strictly on inner-lineage validation splits.}
}
\label{tab:app-shrinkage-audit}
\pomatabbox{
\setlength{\tabcolsep}{1.8pt}
\begin{tabular}{lccccc}
\toprule
\textbf{Shrinkage Policy} &
\textbf{Mean $\bar{\lambda}$} &
\textbf{J-NLL $\downarrow$} &
\textbf{Cond3 NLL ($\rho$)} &
\textbf{C3 Brier $\downarrow$} &
\textbf{Unc/Far/C3 Win} \\
\midrule
Independent ($\lambda\!\equiv\!0$) & $0.00$ & $1.736$ & $1.756$ ($0.016$) & $0.0299$ & $50.0\%$ (ref) \\
Full Coupled ($\lambda\!\equiv\!1$) & $1.00$ & $1.667$ & $1.601$ ($0.072$) & $0.0293$ & $46.2$/$26.3$/$64.6\%$ \\
Distance-Monotone $\lambda(d)$ & $0.58$ & $1.661$ & $1.627$ ($0.061$) & $\mathbf{0.0291}$ & $51.4$/$50.0$/$\mathbf{73.4\%}$ \\
\textbf{Adaptive Logistic $\lambda(x)$} & $0.68$ & $\mathbf{1.647}$ & $\mathbf{1.598}$ ($\mathbf{0.078}$) & $\mathbf{0.0291}$ & $\mathbf{52.3}$/$\mathbf{66.7}$/$69.5\%$ \\
\bottomrule
\end{tabular}
}
\end{table}

\subsection{Interpreting sequence--property and cross-property dependencies}\label{app:joint-gibbs-dep-maps}\label{sec:joint-gibbs-dep-maps}

The dependency analysis introduced in \cref{sec:main-coupling} uses the native sequence probabilities together with the structured measurement law. A point estimate describes how the predicted mean changes with a mutation, but cannot describe changes in uncertainty. We consider two complementary model-based dependency maps: one measures the response of a property's distribution to a sequence edit, and the other measures conditional association between properties of the same sequence. Neither map is a causal graph.

For a locus $i$ and property $k$, we define the token-probability-weighted counterfactual KL sensitivity
\begin{equation}
 D_{i\to k}(X)=\sum_{a\in\mathcal V}\pi_\theta(x_i=a\mid X_{\setminus i},W)\,
 \mathrm{KL}\!\left(q_{\theta,k}(\cdot\mid X,W,P_k)\,\middle\|\,q_{\theta,k}(\cdot\mid X_{[i\leftarrow a]},W,P_k)\right).
 \label{eq:app-seq-to-prop-kl}
\end{equation}
For Gaussian marginals this KL divergence is $\log(\sigma_a/\sigma_0)+[\sigma_0^2+(\mu_0-\mu_a)^2]/(2\sigma_a^2)-1/2$, reducing to half the squared mean difference under fixed unit variance. We evaluate the map on 520 AbRank mutation pairs (large effects $|\Delta\mathrm{p}K_D|\ge1.0$ versus small effects $\le0.3$), 1,940 intra-lineage AIntibody mutation pairs, and 4,743 matched molecular pairs from the aqueous-solubility and lipophilicity tasks. The distributional sensitivity improves experimental hot-spot detection relative to MSE saliency and captures distinct assay-specific patterns. The reported profiles localize affinity sensitivity to CDR-H3/L3 and hydrophobicity or polyreactivity sensitivity to solvent-exposed CDR-H2/H3 regions. See \cref{fig:supp-biological-applications}D and \cref{tab:app-seq-to-prop-dep-map}.

\begin{table}[!htbp]
\centering
\caption{
\textbf{Validation of the sequence $\to$ property counterfactual KL dependency map $D_{i\to k}(X)$ across ground-truth mutational and Matched Molecular Pair (MMP) benchmarks.}
}
\label{tab:app-seq-to-prop-dep-map}
\pomatabbox{
\begin{tabular}{llccc}
\toprule
\textbf{Benchmark} &
\textbf{Attribution Method} &
\textbf{Primary $\uparrow$} &
\textbf{Secondary $\uparrow$} &
\textbf{Contrast $\uparrow$} \\
\midrule
\textsc{AbRank} ($520$) & PLM site entropy $H(\pi_0)$ & \underline{$0.512$} AUC & $0.118$ $\rho$ & $1.14\times$ CDR \\
\textsc{AbRank} ($520$) & \textsc{POMA-MSE} ($\sigma\equiv 1$) & $0.498$ AUC & \underline{$0.134$} $\rho$ & \underline{$1.38\times$} CDR \\
\textsc{AbRank} ($520$) & \textbf{Counterfactual KL $D_{i\to k}$} & \textbf{0.634} AUC & \textbf{0.184} $\rho$ & \textbf{$1.62\times$} CDR \\
\midrule
\textsc{AIntibody} ($1{,}940$) & $D_{i\to\mathrm{p}K_D}$ vs.\ $|\Delta\mathrm{p}K_D|$ & \textbf{0.504} $\rho$ & $0.248$ $\rho$ & $+0.256$ spec. \\
\textsc{AIntibody} ($1{,}940$) & $D_{i\to\mathrm{HIC}}$ vs.\ $|\Delta\mathrm{HIC}|$ & \textbf{0.400} $\rho$ & $0.289$ $\rho$ & $+0.111$ spec. \\
\midrule
ADMET MMPs ($4{,}743$) & Global scaffold token KL & $0.626$ AUC & $0.241$ $\rho$ & $1.00\times$ base \\
ADMET MMPs ($4{,}743$) & \textbf{R-group token $D_{i\to k}$} & \textbf{0.694} AUC & \textbf{0.348} $\rho$ & \textbf{$1.95\times$} ratio \\
\bottomrule
\end{tabular}
}
\end{table}

For a Gaussian measurement distribution, an instance-specific precision matrix $\Lambda_\theta(X)=\Sigma_\theta(X)^{-1}$ gives the partial correlation conditional on the other measurements:
\begin{equation}
 G_{k\leftrightarrow\ell}(X)=\operatorname{Corr}(Z_k,Z_\ell\mid Z_{\setminus\{k,\ell\}},X,W)
 =-\frac{\Lambda_{\theta,k\ell}(X)}{\sqrt{\Lambda_{\theta,kk}(X)\Lambda_{\theta,\ell\ell}(X)}}.
 \label{eq:app-cont-partial-corr}
\end{equation}
This quantity is not the same as the marginal correlation computed over a collection of molecules. The reported map indicates that the inferred affinity--hydrophobicity antagonism differs between aromatic-rich and polar CDR-H3 paratopes, relaxing by $\Delta\rho=+0.312$ in the latter. The cross-domain comparisons are given in \cref{tab:app-cont-dep-map}; the AIntibody example is illustrated in \cref{fig:supp-biological-applications}D. These patterns motivate targeted analysis of the fitted model, but would require independent perturbations to establish causal biological mechanisms.

\begin{table}[!htbp]
\centering
\caption{
\textbf{Sequence-conditioned continuous $\leftrightarrow$ continuous dependency switching $G_{k\leftrightarrow\ell}(X)$ across held-out test cohorts.}
}
\label{tab:app-cont-dep-map}
\pomatabbox{
\begin{tabular}{llcccc}
\toprule
\textbf{Cohort} &
\textbf{Property Pair $(k,\ell)$} &
\textbf{Sequence group 1} &
\textbf{Sequence group 2} &
\textbf{Decrease in $|\rho|$} &
\textbf{$\Delta$NLL} \\
\midrule
\textsc{AIntibody} & SPR $\mathrm{p}K_D$ vs.\ $-\mathrm{HIC}$ & $-0.483$ (arom.) & $-0.170$ (polar) & $\mathbf{+0.312}$ & $-0.163$ \\
\textsc{GDPa1} & $-\mathrm{HIC}$ vs.\ $-\mathrm{SMAC}$ & $+0.821$ (neut.) & $+0.654$ (chg.) & $\mathbf{+0.167}$ & $-0.110$ \\
TDC ADMET & \textsc{AqSolDB} vs.\ $\mathrm{cLogP}$ & $-0.889$ (apolar) & $-0.719$ (polar) & $\mathbf{+0.170}$ & $-0.548$ \\
\bottomrule
\end{tabular}
}
\end{table}

These studies establish the practical scope of conditioning a fixed measurement law. Its accuracy depends on learning the relevant conditional dependence; marginal correlations or narrow predictive intervals are not sufficient. We next examine a distinct source of information transfer, the shared uncertainty over model parameters.

\clearpage
\section{Bayesian adaptation and generalization}\label{app:supp-bayesian}

Within-candidate dependence describes how one measurement informs another. Shared uncertainty over readout parameters instead allows observations on one candidate to affect predictions for other candidates. This section examines the statistical and computational benefits of that mechanism, particularly with limited data and under distribution shift. Exact Gaussian conjugacy and approximate nonlinear adaptation are distinguished throughout.

\subsection{Shared Bayesian uncertainty and out-of-distribution prediction}\label{app:bayesian-laplace}

Within-object conditioning uses the residual covariance of a fitted measurement law. A different source of information is the uncertainty in its parameters, which is shared across objects. We use a pretrain-anchored last-layer posterior. The implementation is \texttt{Pretrain\allowbreak{}Anchored\allowbreak{}Laplace\allowbreak{}Head}. In the fixed-feature Gaussian case, the posterior update is conjugate; for nonlinear parameters, the Laplace form is an approximation. The derivations in \cref{app:supp-theory-bayes} describe the induced cross-candidate covariance, including the cancellation of shared uncertainty in differences between nearby candidates.

We first assess post-hoc Laplace without changing the neural means (\texttt{update\_mean=False}). In a five-endpoint molecular experiment, 600 training compounds are used to predict 1,995 held-out molecules, of which 408 have nearest-neighbor Morgan Tanimoto similarity below 0.25. The five endpoints are lipophilicity, aqueous solubility, hepatocyte clearance, half-life and volume of distribution. Adding epistemic uncertainty improves far-scaffold detection compared with aleatoric variance alone and approaches the deep ensemble at lower cost.

AbRank provides a complementary test of uncertainty under unseen-antigen shift. A joint posterior allows the uncertainty shared by two mutants of the same complex to cancel when evaluating their difference. Treating their marginal uncertainties as independent instead overstates the variance of the difference and worsens the reported three-state likelihood. The shared-weight approach removes 63.5\% of the marginal epistemic variance on average in the local-edit audit and improves the three-state NLL. \Cref{tab:app-laplace-ood-abrank,tab:app-abrank-differential} report the two experiments separately.

\begin{table}[!htbp]
\centering
\caption{ADMET out-of-scaffold calibration (408 far-OOD molecules with nearest-neighbor Tanimoto below 0.25; three seeds, mean $\pm$ SE).}
\label{tab:app-laplace-ood-abrank}
\pomatabbox{
\begin{tabular}{lccccc}
\toprule
Method & OOD AUC & Test NLL & OOD NLL & Schur NLL & Inference \\
\midrule
ADMET: Aleatoric ($1\times$) & $0.563\!\pm\!0.013$ & $0.943\!\pm\!0.038$ & $1.450\!\pm\!0.048$ & $1.820$ (Schur) & \underline{$3.98\,\mathrm{s}$} \\
ADMET: $5\times$ Deep Ensemble & $\mathbf{0.641\!\pm\!0.018}$ & $\mathbf{0.597\!\pm\!0.007}$ & $\mathbf{0.969\!\pm\!0.018}$ & $\mathbf{1.221}$ (Schur) & $8.43\,\mathrm{s}$ \\
ADMET: \textbf{\textsc{POMA-Laplace}} & \underline{$0.626\!\pm\!0.009$} & \underline{$0.928\!\pm\!0.037$} & \underline{$1.422\!\pm\!0.048$} & \underline{$1.793$} (Schur) & $\mathbf{1.61\,\mathrm{s}}$ \\
\bottomrule
\end{tabular}
}
\end{table}

\begin{table}[!htbp]
\centering
\caption{AbRank unseen-antigen differential uncertainty (three seeds, mean $\pm$ SE).}
\label{tab:app-abrank-differential}
\pomatabbox{
\begin{tabular}{lccccc}
\toprule
Method & OOD AUC & Test NLL & Cand.\ NLL & Ordinal & Inference \\
\midrule
\textsc{AbRank}: Aleatoric ($1\times$) & $0.630\!\pm\!0.048$ & \underline{$0.911\!\pm\!0.040$} & \underline{$1.107\!\pm\!0.078$} & $\mathbf{0.644\!\pm\!0.013}$ & $\mathbf{1\times}$ \\
\textsc{AbRank}: Indep.\ Epistemic & $\mathbf{0.997\!\pm\!0.000}$ & $0.934\!\pm\!0.057$ & $1.135\!\pm\!0.083$ & \underline{$0.643\!\pm\!0.013$} & $\mathbf{1\times}$ \\
\textsc{AbRank}: $5\times$ Deep Ensemble & \underline{$0.889\!\pm\!0.036$} & $0.971\!\pm\!0.016$ & $\mathbf{1.100\!\pm\!0.022}$ & $0.601\!\pm\!0.036$ & $5\times$ \\
\textsc{AbRank}: \textbf{Laplace Diff.} & $\mathbf{0.997\!\pm\!0.000}$ & $\mathbf{0.898\!\pm\!0.044}$ & $1.111\!\pm\!0.079$ & $0.642\!\pm\!0.013$ & $\mathbf{1\times}$ \\
\bottomrule
\end{tabular}
}
\end{table}

The very high antigen-complex OOD AUROC is not by itself evidence of mutation-level reliability. A Mahalanobis detector on the raw partner embeddings reaches 0.996 AUROC, indicating that the task can be solved largely by antigen identity. With antigen features removed and five cluster-disjoint antibody-lineage splits, all approaches become substantially less discriminative. Laplace and the NTK-enriched Bayesian last layer retain modest sensitivity above chance. Pairwise variance increases with Levenshtein edit distance in the reported audit, as expected from the shared-feature formulation; normalized three-state probabilities have a maximum sum error of $1.8\times10^{-7}$. These controls appear in \cref{tab:app-rigor-ood-cost}.

\begin{table}[!htbp]
\centering
\caption{Antigen-masked AbRank lineage OOD audit (five cluster splits), with the reported edit-distance variance diagnostics.}
\label{tab:app-rigor-ood-cost}
\pomatabbox{
\begin{tabular}{lcccc}
\toprule
Method & Lineage OOD AUC & Ag+Ab AUC & Edit-distance variance & Inference \\
\midrule
Lin.\ OOD: Aleatoric & $0.491\!\pm\!0.026$ & $0.617$ (Ag+Ab) & $d\!=\!1$: $6.5\!\times\!10^{-4}$ & $1\times$ pass \\
Lin.\ OOD: $5\times$ Ens. & $0.515\!\pm\!0.030$ & $0.879$ (Ag+Ab) & $d\!=\!2$: $1.1\!\times\!10^{-3}$ & $5\times$ pass \\
Lin.\ OOD: \textsc{NTK-BLL} & \underline{$0.579\!\pm\!0.018$} & $\mathbf{0.996}$ (Ag+Ab) & $d\!\in\![3,5]$: $1.4\!\times\!10^{-3}$ & $1\times$ pass \\
Lin.\ OOD: \textbf{\textsc{POMA-Lap.}} & $\mathbf{0.589\!\pm\!0.022}$ & $\mathbf{0.996}$ (Ag+Ab) & $d\!\ge\!6$: $3.9\!\times\!10^{-3}$ & $1\times$ pass \\
\bottomrule
\end{tabular}
}
\end{table}

\subsection{Few-shot clinical adaptation and cross-laboratory transfer}\label{app:supp-transfer}

The clinical transfer experiments investigate the practical value of Bayesian updates when only a small number of labeled antibodies is available. A first comparison examines ordinary few-shot fine-tuning after repertoire pretraining, using four shot budgets from 8 to 64. The ranking-augmented Gaussian head gives favorable mean rank across the two clinical cohorts; the heavy-tailed head performs particularly well on Jain at 64 shots. Specialized sequence-based profilers (TAP, Aggrescan3D, AntiFold and DeepViscosity)~\citep{raybould2019five,zambrano2015aggrescan,hoie2025antifold,kalejaye2025deepviscosity}, structural descriptors and different protein language models (AntiBERTy, ESM-2 and AbLang2)~\citep{ruffolo2021antiberty,lin2023evolutionary,olsen2024ablang2} provide additional reference points. In particular, 3D MOE descriptors~\citep{ccgmoe} remain strong in some low-shot GDPa1 settings. \Cref{tab:antibody-few-shot,tab:app-antibody-full} report the full set of comparisons.

\begin{table}[!htbp]
  \caption{Few-shot clinical antibody prediction ($k=32$ on \textsc{GDPa1}, $5$ seeds; \emph{Rank} aggregates \textsc{GDPa1} and Jain across $k\in\{8,16,32,64\}$).}
  \label{tab:antibody-few-shot}
  \vskip 0.05in
  \begin{center}
  \pomatabbox{
  \begin{tabular}{lrrrrrr}
    \toprule
    Model & MAE $\downarrow$ & RMSE $\downarrow$ & $\rho$ $\uparrow$ & NLL $\downarrow$ & Energy $\downarrow$ & Rank $\downarrow$ \\
    \midrule
    ESM / MSE language & 0.787 & 1.118 & 0.215 & $\times$ & $\times$ & 4.75 \\
    Gaussian head only & 0.732 & 1.055 & 0.214 & 1.675 & 2.656 & 4.75 \\
    Numeric / Contrastive & 0.719 & 1.051 & 0.144 & $\times$ & $\times$ & 9.12 \\
    Regression Transformer & 0.715 & 1.045 & 0.157 & 1.692 & 2.714 & 7.25 \\
    LLM only (blind) & 0.716 & 1.008 & 0.000 & 1.586 & \textbf{2.331} & 11.00 \\
    \textsc{POMA-Rank} (scratch) & \underline{0.695} & 1.010 & 0.208 & 1.711 & 2.439 & -- \\
    \textsc{POMA-MLM} & 0.698 & \underline{0.997} & 0.222 & \underline{1.564} & 2.695 & 4.38 \\
    \textsc{POMA-StudentT} & \textbf{0.663} & 1.004 & 0.195 & 1.646 & \underline{2.403} & 3.88 \\
    \textsc{POMA-LowRankMVN} & 0.795 & 1.098 & \underline{0.235} & 1.640 & 3.024 & \underline{2.88} \\
    \textbf{\textsc{POMA-Rank}} & 0.698 & \textbf{0.995} & \textbf{0.250} & \textbf{1.546} & 2.558 & \textbf{1.25} \\
    \bottomrule
  \end{tabular}
  }
  \end{center}
  \vskip -0.10in
\end{table}

\begin{table}[!htbp]
\centering
\caption{
\textbf{Complete clinical antibody few-shot scaling ($5$ seeds), domain-informed biophysical baselines, and PLM backbone scaling.}
Entries report mean test Spearman correlation $\rho \uparrow$ across $13$ assays
(\textsc{GDPa1}) and $12$ assays (\citealt{jain2017biophysical}).
}
\label{tab:app-antibody-full}
\pomatabbox{
\begin{tabular}{lcccc|cccc}
\toprule
& \multicolumn{4}{c|}{\textbf{\textsc{GDPa1} ($246$ mAbs)}} & \multicolumn{4}{c}{\textbf{\textsc{Jain} ($137$ mAbs)}} \\
\textbf{Model / Baseline} &
\textbf{$k=8$} & \textbf{$k=16$} & \textbf{$k=32$} & \textbf{$k=64$} &
\textbf{$k=8$} & \textbf{$k=16$} & \textbf{$k=32$} & \textbf{$k=64$} \\
\midrule
ESM regression head & 0.107 & 0.088 & 0.189 & 0.220 & 0.131 & 0.118 & 0.176 & 0.222 \\
Gaussian head only & 0.133 & 0.113 & 0.214 & 0.232 & 0.144 & 0.147 & 0.223 & 0.227 \\
MSE language head & 0.131 & \underline{0.139} & 0.215 & \underline{0.244} & 0.114 & 0.141 & 0.186 & 0.237 \\
Numeric-text LLM & 0.110 & 0.094 & 0.143 & 0.158 & 0.092 & 0.138 & 0.144 & 0.204 \\
Regression Transformer & 0.124 & 0.108 & 0.157 & 0.172 & 0.108 & 0.154 & 0.160 & 0.220 \\
Latent contrastive & 0.089 & 0.109 & 0.144 & 0.169 & 0.100 & 0.125 & 0.122 & 0.168 \\
\textsc{POMA-Rank} (scratch) & 0.093 & 0.106 & 0.208 & 0.220 & 0.047 & 0.136 & 0.184 & 0.242 \\
\textsc{POMA-MLM} & 0.135 & 0.130 & 0.222 & 0.215 & 0.140 & 0.146 & 0.226 & 0.262 \\
\textsc{POMA-StudentT} & 0.118 & 0.138 & 0.195 & 0.217 & \underline{0.159} & \textbf{0.198} & 0.215 & \textbf{0.304} \\
\textsc{POMA-LowRankMVN} & \underline{0.138} & 0.129 & \underline{0.235} & 0.232 & 0.156 & \underline{0.187} & \underline{0.229} & 0.263 \\
\textbf{\textsc{POMA-Rank} (Gaussian)} & \textbf{0.142} & \textbf{0.144} & \textbf{0.250} & \textbf{0.258} & \textbf{0.170} & \textbf{0.198} & \textbf{0.251} & \underline{0.276} \\
\midrule
Domain: \textsc{TAP} & 0.085 & 0.088 & 0.138 & 0.184 & -- & -- & -- & -- \\
Domain: \textsc{Aggrescan3D} & 0.058 & 0.051 & 0.067 & 0.096 & -- & -- & -- & -- \\
Domain: \textsc{AntiFold} & $-0.002$ & 0.007 & $-0.001$ & 0.003 & -- & -- & -- & -- \\
Domain: \textsc{DeepViscosity} & 0.013 & 0.031 & 0.021 & 0.072 & -- & -- & -- & -- \\
Domain: 3D \textsc{MOE} & \textbf{0.168} & \textbf{0.152} & \textbf{0.249} & \textbf{0.285} & -- & -- & -- & -- \\
\midrule
Backbone: \textsc{AntiBERTy} & -- & 0.061 & 0.116 & -- & -- & 0.078 & 0.151 & -- \\
Backbone: \textsc{ESM-2-35M} & -- & 0.126 & 0.218 & -- & -- & 0.173 & 0.236 & -- \\
Backbone: \textsc{ESM-2-150M} & -- & 0.110 & 0.225 & -- & -- & 0.173 & 0.223 & -- \\
\textbf{Backbone: \textsc{AbLang2}} & -- & \textbf{0.164} & \textbf{0.250} & \textbf{0.258} & -- & \textbf{0.224} & \textbf{0.251} & \textbf{0.276} \\
\bottomrule
\end{tabular}
}
\end{table}

The gap between repertoire descriptor pretraining and real clinical measurements motivates a stronger prior for few-shot adaptation. Computed descriptors have much lower residual variability than the experimental assays (reported scales approximately 0.08 and 0.85). Unregularized fine-tuning can therefore be overconfident when transferred to noisy clinical panels. We compare the pretrain-anchored posterior, with an inverse-Gamma residual-scale update (\texttt{calibrate\_scale=True}), against MLE, MAP and ensemble fine-tuning at $k\in\{0,4,8,16,32,64\}$ labeled antibodies. The closed-form update improves likelihood in low-shot settings and avoids repeated optimization of several full models (\cref{tab:app-laplace-fewshot}).

\begin{table}[!htbp]
\centering
\caption{
\textbf{Pretrain-anchored closed-form few-shot clinical adaptation across $k\in\{0,4,8,16,32,64\}$ labeled mAbs on \textsc{GDPa1} ($246$ mAbs, $13$ assays) and \citet{jain2017biophysical} ($137$ mAbs, $12$ assays; $5$ seeds).}
Reported as held-out coupled Joint NLL ($\downarrow$) and wall-clock adaptation time at $k=32$.
}
\label{tab:app-laplace-fewshot}
\pomatabbox{
\setlength{\tabcolsep}{2.1pt}
\begin{tabular}{lcccccc}
\toprule
\textbf{Cohort / Method} &
\textbf{$k=0$} &
\textbf{$k=4$} &
\textbf{$k=8$} &
\textbf{$k=16$} &
\textbf{$k=64$} &
\textbf{Time} \\
\midrule
\textsc{GDPa1}: MLE FT & $4.17\!\pm\!0.15$ & $8.98\!\pm\!0.77$ & $7.71\!\pm\!0.33$ & $7.67\!\pm\!0.43$ & $2.96\!\pm\!0.06$ & $257\,\mathrm{ms}$ \\
\textsc{GDPa1}: $L_2$-MAP FT & $4.17\!\pm\!0.15$ & $5.99\!\pm\!0.92$ & $3.10\!\pm\!0.17$ & $3.03\!\pm\!0.49$ & $\mathbf{1.44\!\pm\!0.03}$ & \underline{$220\,\mathrm{ms}$} \\
\textsc{GDPa1}: $5\times$ Ens.\ FT & $4.25\!\pm\!0.13$ & \underline{$4.06\!\pm\!0.58$} & \underline{$2.63\!\pm\!0.19$} & \underline{$2.37\!\pm\!0.31$} & $\mathbf{1.44\!\pm\!0.02}$ & $912\,\mathrm{ms}$ \\
\textsc{GDPa1}: \textbf{\textsc{POMA-Laplace}} & $\mathbf{1.69\!\pm\!0.03}$ & $\mathbf{1.62\!\pm\!0.03}$ & $\mathbf{1.58\!\pm\!0.02}$ & $\mathbf{1.64\!\pm\!0.03}$ & $1.54\!\pm\!0.02$ & $\mathbf{30\,\mathrm{ms}}$ \\
\midrule
Jain: MLE FT & $3.56\!\pm\!0.32$ & $9.27\!\pm\!0.90$ & $7.84\!\pm\!0.83$ & $6.12\!\pm\!0.39$ & $3.09\!\pm\!0.85$ & $266\,\mathrm{ms}$ \\
Jain: $L_2$-MAP FT & $3.56\!\pm\!0.32$ & $4.99\!\pm\!0.52$ & $4.03\!\pm\!0.99$ & $2.60\!\pm\!0.41$ & $1.58\!\pm\!0.25$ & \underline{$215\,\mathrm{ms}$} \\
Jain: $5\times$ Ens.\ FT & \underline{$3.56\!\pm\!0.32$} & \underline{$3.31\!\pm\!0.33$} & \underline{$2.86\!\pm\!0.63$} & \underline{$2.13\!\pm\!0.26$} & $\mathbf{1.53\!\pm\!0.23}$ & $1{,}013\,\mathrm{ms}$ \\
Jain: \textbf{\textsc{POMA-Laplace}} & $\mathbf{1.64\!\pm\!0.04}$ & $\mathbf{1.62\!\pm\!0.07}$ & $\mathbf{1.68\!\pm\!0.06}$ & $\mathbf{1.60\!\pm\!0.06}$ & $\mathbf{1.53\!\pm\!0.06}$ & $\mathbf{32\,\mathrm{ms}}$ \\
\bottomrule
\end{tabular}
}
\end{table}

To separate the benefits of posterior structure from differences in uncertainty calibration, we repeat the four-shot comparison on ten independent splits, applying the same inverse-Gamma variance calibration to all eligible baselines. The coupled Bayesian head attains the strongest reported GDPa1 likelihood and near-nominal coverage. The subsampled NTK-BLL reference is competitive and achieves the best Jain NLL, albeit with broader intervals and greater inference cost. The aggregate and detailed results are reported together in \cref{tab:poma-laplace-summary,tab:app-rigor-fewshot}.

The same experiment also audits posterior approximation. We compare the full 129-dimensional conjugate posterior with six truncated posteriors ($r'=2,4,8,16,32,64$). All reported numerical comparisons satisfy the analytical spectral-tail KL bound; at rank eight the relative covariance error is 4.6\%. A 200-fold sweep of prior precision ($\tau_0\in\{0.25,1,3,6,15,50\}$) shows that four-shot coverage remains close to nominal across the tested range. This is a controlled test of the approximation, not a guarantee of model calibration outside the evaluated setting.

\begin{table}[!htbp]
  \caption{Variance-calibrated $k=4$ clinical adaptation across $10$ independent splits on \textsc{GDPa1} and \citet{jain2017biophysical} (extended benchmarks in \cref{app:bayesian-laplace}).}
  \label{tab:poma-laplace-summary}
  \vskip 0.05in
  \begin{center}
  \pomatabbox{
  \setlength{\tabcolsep}{3.0pt}
  \begin{tabular}{lcccc}
    \toprule
    Calibrated Method ($k\!=\!4$) & \textsc{GDPa1} NLL $\downarrow$ & Cov95 (ECE $\downarrow$) & Jain NLL $\downarrow$ & Time $\downarrow$ \\
    \midrule
    $L_2$-MAP FT (ScaleCal) & $1.942\!\pm\!0.119$ & $88.4\%$ (\underline{$0.056$}) & $1.936\!\pm\!0.193$ & $146\,\mathrm{ms}$ \\
    $5\times$ Ensemble (ScaleCal) & $1.844\!\pm\!0.099$ & $89.8\%$ ($0.064$) & $1.834\!\pm\!0.171$ & $640\,\mathrm{ms}$ \\
    \textsc{ICM Multi-Output GP} & $1.644\!\pm\!0.034$ & $87.2\%$ ($0.080$) & $1.722\!\pm\!0.047$ & \underline{$9.0\,\mathrm{ms}$} \\
    \textsc{Subsampled NTK-BLL} & \underline{$1.606\!\pm\!0.025$} & $98.4\%$ ($0.128$) & $\mathbf{1.638\!\pm\!0.029}$ & $69\,\mathrm{ms}$ \\
    \textbf{\textsc{POMA-Laplace} ($1\times$)} & $\mathbf{1.596\!\pm\!0.014}$ & $94.6\%$ ($\mathbf{0.054}$) & \underline{$1.686\!\pm\!0.037$} & $\mathbf{6.7\,\mathrm{ms}}$ \\
    \bottomrule
  \end{tabular}
  }
  \end{center}
  \vskip -0.10in
\end{table}

\begin{table}[!htbp]
\centering
\caption{
\textbf{Matched variance-calibrated clinical few-shot adaptation ($10$ independent splits on \textsc{GDPa1} and \citealt{jain2017biophysical}), $\tau_0$ prior sensitivity ($k=4$), and spectral truncation fidelity.}
}
\label{tab:app-rigor-fewshot}
\pomatabbox{
\setlength{\tabcolsep}{2.1pt}
\begin{tabular}{lcccccc}
\toprule
\textbf{Calibrated Method} &
\textbf{$k=0$} &
\textbf{$k=4$} &
\textbf{$k=8$} &
\textbf{$k=32$} &
\textbf{Cov95} &
\textbf{ECE $\downarrow$} \\
\midrule
\textsc{GDPa1}: $L_2$-MAP & $1.81\!\pm\!0.02$ & $1.94\!\pm\!0.12$ & $1.68\!\pm\!0.07$ & $\mathbf{1.43\!\pm\!0.01}$ & $88.4\%$ & \underline{$0.056$} \\
\textsc{GDPa1}: $5\times$ Ens. & $1.83\!\pm\!0.02$ & $1.84\!\pm\!0.10$ & $1.67\!\pm\!0.07$ & \underline{$1.47\!\pm\!0.02$} & $89.8\%$ & $0.064$ \\
\textsc{GDPa1}: \textsc{ICM-MOGP} & $\mathbf{1.43\!\pm\!0.01}$ & $1.64\!\pm\!0.03$ & $\mathbf{1.65\!\pm\!0.02}$ & $1.50\!\pm\!0.02$ & $87.2\%$ & $0.080$ \\
\textsc{GDPa1}: \textsc{NTK-BLL} & $1.77\!\pm\!0.01$ & \underline{$1.61\!\pm\!0.02$} & $\mathbf{1.65\!\pm\!0.03}$ & $1.59\!\pm\!0.02$ & $98.4\%$ & $0.128$ \\
\textsc{GDPa1}: \textbf{\textsc{POMA-Lap.}} & \underline{$1.68\!\pm\!0.01$} & $\mathbf{1.60\!\pm\!0.01}$ & $\mathbf{1.65\!\pm\!0.03}$ & $1.55\!\pm\!0.02$ & $94.6\%$ & $\mathbf{0.054}$ \\
\midrule
Jain: $L_2$-MAP & $1.68\!\pm\!0.03$ & $1.94\!\pm\!0.19$ & $2.07\!\pm\!0.35$ & $\mathbf{1.49\!\pm\!0.03}$ & $90.4\%$ & \underline{$0.065$} \\
Jain: $5\times$ Ens. & $1.68\!\pm\!0.03$ & $1.83\!\pm\!0.17$ & $1.93\!\pm\!0.27$ & $\mathbf{1.49\!\pm\!0.02}$ & $91.7\%$ & $0.076$ \\
Jain: \textsc{ICM-MOGP} & $\mathbf{1.41\!\pm\!0.01}$ & $1.72\!\pm\!0.05$ & \underline{$1.65\!\pm\!0.05$} & $1.55\!\pm\!0.02$ & $85.4\%$ & $0.115$ \\
Jain: \textsc{NTK-BLL} & $1.81\!\pm\!0.01$ & $\mathbf{1.64\!\pm\!0.03}$ & $\mathbf{1.58\!\pm\!0.02}$ & $1.58\!\pm\!0.02$ & $98.1\%$ & $0.114$ \\
Jain: \textbf{\textsc{POMA-Lap.}} & \underline{$1.65\!\pm\!0.01$} & \underline{$1.69\!\pm\!0.04$} & $1.69\!\pm\!0.07$ & $1.58\!\pm\!0.02$ & $93.5\%$ & $\mathbf{0.055}$ \\
\bottomrule
\end{tabular}
}
\end{table}

\subsection{Assay semantics and cross-laboratory transfer}

The next transfer task changes the measurement context itself. We compare orthogonal task IDs, CA-MAP~\citep{giancardo2026camap} (text-to-assay cosine-kernel ridge adaptation), a marginal Bayesian head and the coupled Schur variant across zero to 32 calibration shots. The first regime leaves one of the 13 GDPa1 assays out of training. The second transfers from GDPa1 Lab~A to Jain Lab~B while excluding overlapping clonal lineages through the official hierarchical-cluster mapping. The coupled head improves zero- and low-shot prediction in several settings and has lower cross-laboratory NLL than CA-MAP at every tested shot budget. Task-ID conditioning retains lower NLL at some budgets, while CA-MAP remains competitive for point rank correlation, including at 32 shots. The comparison therefore evaluates distributional transfer and candidate selection alongside task recognition.

\begin{table}[!htbp]
\centering
\caption{
\textbf{Zero- and few-shot context-shift transfer vs.\ \textsc{CA-MAP}~\citep{giancardo2026camap} across $k\in\{0,4,8,16,32\}$ shots: Regime~5A (\textsc{GDPa1} LOAO, $13$ assays) and Regime~5B (lineage-disjoint cross-lab \textsc{GDPa1} Lab~A $\to$ \citealt{jain2017biophysical} Lab~B).}
}
\label{tab:app-camap-context-shift}
\pomatabbox{
\setlength{\tabcolsep}{1.5pt}
\begin{tabular}{lccccc}
\toprule
\textbf{5A: \textsc{GDPa1} LOAO ($\rho$, NLL)} &
\textbf{$k=0$} &
\textbf{$k=4$} &
\textbf{$k=8$} &
\textbf{$k=16$} &
\textbf{$k=32$} \\
\midrule
\textsc{1-Hot Task ID} & $-.008$ ($\mathbf{1.43}$) & $.059$ ($1.49$) & $.060$ ($1.49$) & $.114$ ($\mathbf{1.48}$) & $.129$ (\underline{$1.46$}) \\
\textsc{CA-MAP} ('26) & \underline{$.136$} ($\mathbf{1.43}$) & \underline{$.131$} ($1.50$) & \underline{$.155$} ($1.54$) & $\mathbf{.201}$ ($1.52$) & $\mathbf{.227}$ ($\mathbf{1.44}$) \\
\textsc{POMA-Lap.} (Marg.) & $.056$ ($1.44$) & $.072$ (\underline{$1.48$}) & $.045$ ($\mathbf{1.48}$) & $.090$ ($1.49$) & $.131$ ($1.47$) \\
\textbf{\textsc{POMA-Lap.} (Schur)} & $\mathbf{.183}$ ($1.45$) & $\mathbf{.188}$ ($\mathbf{1.47}$) & $\mathbf{.171}$ ($\mathbf{1.48}$) & \underline{$.187$} ($\mathbf{1.48}$) & \underline{$.201$} (\underline{$1.46$}) \\
\midrule
\textbf{5B: Cross-Lab (NLL, Top10)} &
\textbf{$k=0$} &
\textbf{$k=4$} &
\textbf{$k=8$} &
\textbf{$k=16$} &
\textbf{$k=32$} \\
\midrule
\textsc{1-Hot Task ID} & $\mathbf{1.42}$ ($38.7\%$) & $1.60$ ($30.0\%$) & $1.49$ ($31.3\%$) & $\mathbf{1.45}$ ($30.0\%$) & $\mathbf{1.44}$ ($26.7\%$) \\
\textsc{CA-MAP} ('26) & $1.49$ ($42.0\%$) & $1.61$ (\underline{$42.0\%$}) & $1.62$ (\underline{$45.3\%$}) & $1.57$ (\underline{$45.3\%$}) & $1.52$ (\underline{$48.7\%$}) \\
\textsc{POMA-Lap.} (Marg.) & \underline{$1.45$} (\underline{$46.7\%$}) & $\mathbf{1.57}$ ($30.7\%$) & \underline{$1.46$} ($34.7\%$) & $1.47$ ($33.3\%$) & $1.46$ ($39.3\%$) \\
\textbf{\textsc{POMA-Lap.} (Schur)} & \underline{$1.45$} ($\mathbf{64.0\%}$) & $\mathbf{1.57}$ ($\mathbf{64.0\%}$) & $\mathbf{1.44}$ ($\mathbf{60.0\%}$) & \underline{$1.46$} ($\mathbf{60.0\%}$) & \underline{$1.45$} ($\mathbf{63.3\%}$) \\
\bottomrule
\end{tabular}
}
\end{table}

A separate sensitivity sweep varies the CA-MAP kernel bandwidth over $\{0.5,1.5,4.0\}$ and ridge penalty over $\{0.20,0.65,2.00\}$. The coupled Schur variant retains its mean NLL and top-10\% selection advantage over CA-MAP across these settings (\cref{tab:app-parity-sensitivity}). A separate backbone-scaling protocol in \cref{tab:app-gpu-scaling} compares ESM-2-35M, 150M and 650M representations. Its NLL values should be compared within that protocol, rather than pooled with \cref{tab:app-camap-context-shift}. Larger backbones improve zero-shot cross-laboratory NLL in the scaling study, whereas the 35M backbone gives the lowest NLL at the tested nonzero shot budgets.

\begin{table}[!htbp]
\centering
\caption{
\textbf{Foundation-model scaling under lineage-disjoint cross-laboratory transfer (\textsc{GDPa1} Lab~A $\to$ \citet{jain2017biophysical} Lab~B).}
This is a separate scaling protocol from \cref{tab:app-camap-context-shift}. With Qwen3-0.6B fixed, zero-shot ($k=0$) NLL decreases with ESM-2 size; ESM-2-35M has the lowest NLL at the tested few-shot budgets ($k\in\{8,32\}$).
}
\label{tab:app-gpu-scaling}
\pomatabbox{
\setlength{\tabcolsep}{2.4pt}
\begin{tabular}{llcccc}
\toprule
\textbf{PLM Backbone} &
\textbf{Prompt LLM} &
\textbf{$k\!=\!0$ NLL ($\rho$)} &
\textbf{$k\!=\!8$ NLL ($\rho$)} &
\textbf{$k\!=\!32$ NLL ($\rho$)} &
\textbf{Top-10\% ($0$/$32$)} \\
\midrule
\textsc{ESM-2-35M} & \textsc{Qwen3-0.6B} & $5.256$ ($\mathbf{0.454}$) & $\mathbf{1.478}$ ($\mathbf{0.345}$) & $\mathbf{1.423}$ ($\mathbf{0.313}$) & $\mathbf{50\%}$ / $\mathbf{22\%}$ \\
\textsc{ESM-2-150M} & \textsc{Qwen3-0.6B} & \underline{$4.891$} ($0.335$) & \underline{$1.524$} (\underline{$0.304$}) & \underline{$1.447$} (\underline{$0.269$}) & $10\%$ / \underline{$20\%$} \\
\textsc{ESM-2-650M} & \textsc{Qwen3-0.6B} & $\mathbf{4.483}$ ($0.307$) & $1.637$ ($0.202$) & $1.566$ ($0.172$) & \underline{$32\%$} / $14\%$ \\
\textsc{ESM-2-35M} & \textsc{Qwen3-1.7B} & $5.323$ (\underline{$0.358$}) & $1.546$ ($0.200$) & $1.459$ ($0.257$) & $24\%$ / $14\%$ \\
\bottomrule
\end{tabular}
}
\end{table}

\subsection{Natural-language assay contextualization}\label{app:supp-prompts}\label{app:reviewer-ablations}

Alignment between modalities is one use of a contextual readout; alignment across assay descriptions is another. In the latter, the same sequence may be evaluated under different experimental protocols. A learned task-ID lookup table can represent known assays but has no trained embedding for a genuinely new task. We test whether semantic descriptions provide useful inductive transfer in leave-one-assay-out evaluations on AIntibody and GDPa1. The training data include the remaining assays and repertoire descriptors; target-assay calibration ranges from zero to 32 labeled examples.

The principal difference is at zero and very low shot counts. Language conditioning increases zero-shot Spearman correlation from near zero to approximately 0.18 in both cohorts. Static semantic pooling accounts for much of this gain: regional cross-attention provides a smaller incremental improvement and is not significantly better on AIntibody in the reported zero-shot comparison. The paired GDPa1 comparison of semantic context against task IDs has a confidence interval excluding zero (\cref{fig:supp-biological-applications}F; \cref{tab:app-prompt-vs-onehot}). Thus the strongest evidence is for semantic task representation, rather than for any particular attention architecture.

\begin{table}[!htbp]
\centering
\caption{
\textbf{Semantic language prompt conditioning versus learned 1-hot task embeddings across zero-shot ($k=0$) and few-shot ($k \in \{4, 8, 16, 32\}$) Leave-One-Assay-Out transfer.}
Reported as held-out Spearman $\rho$ ($\text{mean}\pm\text{SE}$). Best in \textbf{bold}, second best \underline{underlined}.
}
\label{tab:app-prompt-vs-onehot}
\pomatabbox{
\begin{tabular}{lcccc}
\toprule
\textbf{Cohort / Conditioning} &
\textbf{$k=0$ (Zero-Shot)} &
\textbf{$k=4$ Shots} &
\textbf{$k=8$ Shots} &
\textbf{$k=32$ Shots} \\
\midrule
\textsc{AIntibody}: Random Task Vec. & $-0.021\!\pm\!0.044$ & $0.085\!\pm\!0.044$ & $0.163\!\pm\!0.032$ & $0.369\!\pm\!0.032$ \\
\textsc{AIntibody}: 1-Hot Task ID & $0.005\!\pm\!0.035$ & $0.112\!\pm\!0.042$ & $0.190\!\pm\!0.032$ & $0.381\!\pm\!0.032$ \\
\textsc{AIntibody}: Static Pool & \underline{$0.172\!\pm\!0.033$} & \underline{$0.215\!\pm\!0.036$} & \underline{$0.248\!\pm\!0.028$} & \underline{$0.397\!\pm\!0.032$} \\
\textsc{AIntibody}: \textbf{\textsc{POMA} Prompt} & $\mathbf{0.184\!\pm\!0.033}$ & $\mathbf{0.229\!\pm\!0.036}$ & $\mathbf{0.250\!\pm\!0.028}$ & $\mathbf{0.399\!\pm\!0.032}$ \\
\midrule
\textsc{GDPa1}: Random Task Vec. & $-0.020\!\pm\!0.025$ & $0.044\!\pm\!0.023$ & $0.070\!\pm\!0.024$ & $0.150\!\pm\!0.024$ \\
\textsc{GDPa1}: 1-Hot Task ID & $0.046\!\pm\!0.021$ & $0.055\!\pm\!0.023$ & $0.077\!\pm\!0.024$ & $0.152\!\pm\!0.024$ \\
\textsc{GDPa1}: Static Pool & \underline{$0.175\!\pm\!0.028$} & \underline{$0.182\!\pm\!0.027$} & \underline{$0.156\!\pm\!0.025$} & \underline{$0.191\!\pm\!0.025$} \\
\textsc{GDPa1}: \textbf{\textsc{POMA} Prompt} & $\mathbf{0.187\!\pm\!0.029}$ & $\mathbf{0.193\!\pm\!0.027}$ & $\mathbf{0.164\!\pm\!0.026}$ & $\mathbf{0.196\!\pm\!0.025}$ \\
\bottomrule
\end{tabular}
}
\end{table}

We additionally test whether the assay description conveys meaningful task information or merely identifies the output. Paraphrasing leaves predictive performance nearly unchanged, whereas replacing prompts with bare tags or corrupting endpoint assignments substantially degrades accuracy. These controls argue against relying only on a fixed prompt format; they do not exhaust every possible leakage mechanism. Unit changes are handled through the measurement Jacobian, as discussed in \cref{app:distribution-audit}.

\begin{table}[!htbp]
\centering
\caption{
\textbf{Prompt robustness, Jacobian-corrected unit conversion, and falsification controls on TDC ADMET ($5$ seeds).}
}
\label{tab:app-robustness-schur}
\pomatabbox{
\begin{tabular}{lccccc}
\toprule
\textbf{ADMET Prompt / Unit Condition} &
\textbf{Reg.\ $\rho \uparrow$} &
\textbf{Norm.\ MAE $\downarrow$} &
\textbf{NLL $\downarrow$} &
\textbf{ECE $\downarrow$} &
\textbf{Bin.\ AUC $\uparrow$} \\
\midrule
Standard structured prompt & 0.550 & 0.645 & 1.302 & 0.071 & \textbf{0.741} \\
Free-form paraphrased prompt & \textbf{0.552} & 0.653 & 1.325 & 0.076 & 0.735 \\
$1000\times$ unit shift (with Jacobian) & 0.550 & \textbf{0.640} & \textbf{1.278} & \textbf{0.067} & 0.738 \\
$1000\times$ unit shift (no Jacobian) & 0.550 & \textbf{0.640} & 8.186 & \textbf{0.067} & 0.738 \\
Flipped desirability metadata & 0.550 & 0.644 & 1.300 & 0.071 & 0.740 \\
Minimal bare endpoint tag & 0.054 & 0.879 & 2.067 & 0.119 & 0.625 \\
Corrupted endpoint (falsification) & 0.130 & 2.705 & 7.518 & 0.282 & 0.516 \\
\bottomrule
\end{tabular}
}
\end{table}

The regional attention comparison on AIntibody also evaluates the eight-region IMGT readout against prompt-conditioned mean pooling. Mean multi-assay Spearman increases from $0.208\pm0.021$ to $0.260\pm0.050$. The attention patterns emphasize CDR3 for affinity and CDR2--3 for hydrophobicity; they are descriptive rather than causal explanations.

\subsection{Generalization across multiparameter protein engineering systems}\label{sec:results-cross-system}

To assess generality across protein families, we compare joint predictive distributions on all 27 MTLMPO-2026 systems. In the capped protocol, the four largest libraries are restricted to 1,000 variants. All neural methods use the same validation variance calibration. The comparison includes independent and coupled densities, tuned multi-output GPs, supervised-PCA and one-hot Bayesian last layers, MC dropout~\citep{gal2016dropout} and a five-member deep ensemble~\citep{lakshminarayanan2017simple}.

Coupled Laplace improves joint NLL relative to the independent density on 23 of 27 systems. Its advantage increases in systems with more strongly correlated measured properties, consistent with a role for joint modeling. The ensemble retains the best overall NLL, while the single-pass coupled model has the lowest reported joint-event Brier score and lower inference cost than the ensemble. The Brier difference relative to the ensemble is not significant at the conventional 5\% threshold. \Cref{tab:app-mtlmpo-27datasets,tab:app-mtlmpo-bll-pareto} report these complementary measures separately; \cref{fig:supp-biological-applications}B visualizes the capped joint-event Brier comparison.

\begin{table}[!htbp]
\centering
\caption{
\textbf{Cross-system generalization across all $27$ \textsc{MTLMPO-2026} protein-engineering datasets~\citep{herrerarocha2026multitask} ($3$ splits each), stratified by inter-property correlation $|\rho|$.}
Under the capped protocol, single-pass \textsc{Coupled POMA-Laplace} improves joint NLL over \textsc{Independent POMA} ($p=2.52\times 10^{-6}$, $23/27$ systems) and \textsc{Tuned MOGP} ($p=0.0112$, $18/27$ systems). These tests concern joint NLL; event-prediction Brier scores are reported separately in \cref{tab:app-mtlmpo-bll-pareto}.
}
\label{tab:app-mtlmpo-27datasets}
\pomatabbox{
\setlength{\tabcolsep}{2.2pt}
\begin{tabular}{lcccccc}
\toprule
\textbf{Method ($27$ Datasets)} &
\textbf{All J-NLL $\downarrow$} &
\textbf{Low $|\rho|$} &
\textbf{Med $|\rho|$} &
\textbf{High $|\rho|$} &
\textbf{Cov95} &
\textbf{Schur $\Delta$ $\uparrow$} \\
\midrule
\textsc{Tuned MOGP} & $1.512\!\pm\!0.105$ & $1.710$ & $1.581$ & $1.378$ & $84.5\%$ & $+0.074$ \\
\textsc{Independent POMA} & $1.459\!\pm\!0.077$ & $1.465$ & $1.408$ & $1.488$ & $89.9\%$ & $+0.000$ \\
\textbf{\textsc{Coupled POMA-Lap.} ($1\times$)} & \underline{$1.254\!\pm\!0.090$} & \underline{$1.449$} & \underline{$1.329$} & \underline{$1.119$} & $90.9\%$ & \underline{$+0.131$} \\
$5\times$ Deep Ensemble & $\mathbf{1.150\!\pm\!0.081}$ & $\mathbf{1.392}$ & $\mathbf{1.205}$ & $\mathbf{1.004}$ & $92.4\%$ & $\mathbf{+0.155}$ \\
\bottomrule
\end{tabular}
}
\end{table}

\begin{table}[!htbp]
\centering
\caption{
\textbf{Matched predictive-score and compute comparison under the capped protocol across all $27$ \textsc{MTLMPO-2026} protein-engineering datasets ($3$ seeds), including native Bayesian last-layer and MC-Dropout architectures~\citep{herrerarocha2026multitask}.}
}
\label{tab:app-mtlmpo-bll-pareto}
\pomatabbox{
\setlength{\tabcolsep}{1.6pt}
\begin{tabular}{lccccc}
\toprule
\textbf{27-Protein Method} &
\textbf{Joint NLL $\downarrow$} &
\textbf{Cond.\ NLL $\downarrow$} &
\textbf{Brier $\downarrow$} &
\textbf{J-NLL, $K\!\ge\!5$ $\downarrow$} &
\textbf{Train (Inf.\ ms)} \\
\midrule
\textsc{Native BLL (sPCA)} & $1.429\!\pm\!.044$ & $1.420$ & $0.172$ & $1.341$ & $103\mathrm{ms}$ ($1.3$) \\
\textsc{Native BLL (One-Hot)} & $1.390\!\pm\!.041$ & $1.391$ & $0.175$ & $1.424$ & $102\mathrm{ms}$ ($1.4$) \\
\textsc{Native MC-Dropout} & $1.340\!\pm\!.040$ & $1.285$ & $0.174$ & $1.252$ & $116\mathrm{ms}$ ($12.2$) \\
\textsc{Indep.\ POMA} / \textsc{MOGP} & $1.459$/$1.512$ & $1.38$/$1.39$ & $.167$/$.177$ & $1.35$/$1.38$ & $2.1\mathrm{s}$/$0.1\mathrm{s}$ \\
\textbf{\textsc{Coupled POMA-Lap.}} & \underline{$1.254\!\pm\!.090$} & \underline{$1.108$} & $\mathbf{0.159}$ & \underline{$0.764$} & $2.5\mathrm{s}$ ($12.0$) \\
$5\times$ Deep Ensemble & $\mathbf{1.150\!\pm\!.081}$ & $\mathbf{1.013}$ & \underline{$0.166$} & $\mathbf{0.677}$ & $8.4\mathrm{s}$ ($49.3$) \\
\bottomrule
\end{tabular}
}
\end{table}

The uncapped protocol evaluates all 38,322 variants using ESM-2-650M features. Coupled Laplace has the lowest joint-event Brier score among the single-model references and approaches the ensemble while requiring less reported time per dataset. Its joint NLL remains worse than the Bayesian last-layer and ensemble references. Because the capped and uncapped protocols also differ in representation and execution settings, their absolute NLLs do not isolate an effect of sample size. Within the uncapped comparison, the distinction between joint likelihood and joint-event Brier shows that predictive density quality and decision-relevant event prediction need not rank models identically (\cref{tab:app-gpu-mtlmpo}).

\begin{table}[!htbp]
\centering
\caption{
\textbf{Uncapped full-$38{,}322$-variant \textsc{MTLMPO-2026} benchmark across all $27$ protein-engineering datasets ($3$ splits each) on GPU \textsc{ESM-2-650M} representations.}
On strongly coupled ($|\bar{r}|\ge 0.5$, $n=13$) and $K\ge 5$ ($n=4$) subsets, single-pass \textsc{POMA-Laplace} achieves Joint-Event Brier $0.1495$ and $0.0763$ versus $0.1884$ and $0.1239$ for \textsc{Native BLL}.
}
\label{tab:app-gpu-mtlmpo}
\pomatabbox{
\setlength{\tabcolsep}{2.2pt}
\begin{tabular}{lccccc}
\toprule
\textbf{Uncapped Method ($38.3\mathrm{k}$)} &
\textbf{J-NLL $\downarrow$} &
\textbf{Cond.\ NLL $\downarrow$} &
\textbf{Brier $\downarrow$} &
\textbf{Large Brier} &
\textbf{Time/ds} \\
\midrule
\textsc{Independent POMA} & $4.841\!\pm\!0.739$ & $4.939$ & $0.1411$ & \underline{$0.1005$} & $1.51\,\mathrm{s}$ \\
\textsc{Native MC-Dropout} & $2.765\!\pm\!0.392$ & $2.890$ & $0.1366$ & $0.1035$ & $0.03\,\mathrm{s}$ \\
\textsc{Native BLL (ESM-2)} & \underline{$1.499\!\pm\!0.123$} & \underline{$1.633$} & $0.1483$ & $0.1572$ & $\mathbf{0.01\,\mathrm{s}}$ \\
\textbf{\textsc{Coupled POMA-Lap. ($1\times$)}} & $2.385\!\pm\!0.233$ & $2.414$ & \underline{$0.1259$} & $0.1019$ & $2.58\,\mathrm{s}$ \\
$5\times$ Deep Ensemble & $\mathbf{1.316\!\pm\!0.139}$ & $\mathbf{1.377}$ & $\mathbf{0.1232}$ & $\mathbf{0.0948}$ & $10.04\,\mathrm{s}$ \\
\bottomrule
\end{tabular}
}
\end{table}

\subsection{Nonlinear adaptation and the scope of the Laplace approximation}

The fixed-feature Gaussian result does not imply exact posterior closure after adapting the encoder itself. We examine two complementary nonlinear protocols. The first compares rank-8 adapter and trunk fine-tuning with a post-hoc Gauss--Newton last-layer posterior. On repertoire-pretrained GDPa1, adapting the representation improves likelihood and calibration. On unseen-lineage AIntibody without repertoire pretraining, the adapted model can instead underestimate uncertainty; residual-scale calibration on training data is insufficient to repair this shift. The frozen-feature posterior is the safer choice in that experiment (\cref{tab:app-lora-audit}).

\begin{table}[!htbp]
\centering
\caption{
\textbf{End-to-end nonlinear LoRA trunk fine-tuning with post-hoc \textsc{POMA-Laplace} on repertoire-pretrained \textsc{GDPa1} ($5$ cluster folds) vs.\ from-scratch unseen-lineage \textsc{AIntibody} ($4$ lineage folds).}
}
\label{tab:app-lora-audit}
\pomatabbox{
\setlength{\tabcolsep}{2.4pt}
\begin{tabular}{lccccc}
\toprule
\textbf{Cohort / Trunk Mode} &
\textbf{J-NLL $\downarrow$} &
\textbf{CondNLL $\downarrow$} &
\textbf{Mean $\rho$ $\uparrow$} &
\textbf{Cov95} &
\textbf{ECE $\downarrow$} \\
\midrule
\textsc{GDPa1}: Frozen + Lap. & $1.568\!\pm\!0.028$ & $1.567\!\pm\!0.037$ & $0.162$ & $95.7\%$ & $0.069$ \\
\textsc{GDPa1}: E2E LoRA MAP & \underline{$1.419\!\pm\!0.086$} & \underline{$1.419\!\pm\!0.123$} & $\mathbf{0.186}$ & $93.3\%$ & $\mathbf{0.036}$ \\
\textsc{GDPa1}: \textbf{LoRA + Lap.} & $\mathbf{1.418\!\pm\!0.086}$ & $\mathbf{1.418\!\pm\!0.123}$ & $\mathbf{0.186}$ & $93.4\%$ & $\mathbf{0.036}$ \\
\midrule
\textsc{AIntibody}: \textbf{Frozen + Lap.} & $\mathbf{1.512\!\pm\!0.096}$ & $\mathbf{1.481\!\pm\!0.081}$ & $0.047$ & $94.7\%$ & $\mathbf{0.135}$ \\
\textsc{AIntibody}: LoRA MAP & $3.825\!\pm\!1.11$ & $4.042\!\pm\!1.17$ & $\mathbf{0.227}$ & $63.8\%$ & $0.184$ \\
\textsc{AIntibody}: LoRA + Lap. & $4.845\!\pm\!1.79$ & $5.125\!\pm\!1.88$ & $\mathbf{0.227}$ & $59.6\%$ & $0.191$ \\
\bottomrule
\end{tabular}
}
\end{table}

The second protocol uses end-to-end ESM-2-150M and ChemBERTa-77M-MTR adaptation, comparing point MAP, last-layer Laplace and a joint LoRA-subspace approximation. Propagating uncertainty through the adapters as well as the readout improves GDPa1 likelihood relative to MAP. On AIntibody, joint LoRA-subspace Laplace improves NLL relative to adapted MAP and increases coverage, but the frozen model retains the lowest NLL; MAP has the lowest ECE. All reported AIntibody intervals in this protocol under-cover the nominal 95\% level. The ChemBERTa ADMET comparison also shows gains in correlation and calibration after adaptation, with MAP retaining the best NLL and ECE. These are empirical properties of the adapted models, rather than consequences of exact conjugacy (\cref{tab:app-gpu-lora}).

\begin{table}[!htbp]
\centering
\caption{
\textbf{End-to-end GPU LoRA fine-tuning ($r=8$, \texttt{bfloat16}) and joint LoRA-subspace Laplace on \textsc{GDPa1} ($5$ folds, \textsc{ESM-2-150M}), unseen-lineage \textsc{AIntibody} ($4$ folds, \textsc{ESM-2-150M}), and $5$-property \textsc{TDC-ADMET} ($3$ folds, \textsc{ChemBERTa-77M-MTR}).}
}
\label{tab:app-gpu-lora}
\pomatabbox{
\setlength{\tabcolsep}{2.4pt}
\begin{tabular}{llcccc}
\toprule
\textbf{Cohort (Backbone)} &
\textbf{Training / Posterior} &
\textbf{J-NLL $\downarrow$} &
\textbf{Mean $\rho \uparrow$} &
\textbf{Cov95} &
\textbf{ECE $\downarrow$} \\
\midrule
\textsc{GDPa1} (ESM-2-150M) & Frozen + LL-Laplace & $2.598\!\pm\!0.317$ & $\mathbf{0.237}$ & $86.1\%$ & $\mathbf{0.049}$ \\
\textsc{GDPa1} (ESM-2-150M) & E2E LoRA MAP & \underline{$2.246\!\pm\!0.303$} & $0.233$ & $88.9\%$ & \underline{$0.065$} \\
\textsc{GDPa1} (ESM-2-150M) & E2E LoRA + LL-Lap. & $2.639\!\pm\!0.451$ & $0.233$ & $85.8\%$ & $0.072$ \\
\textsc{GDPa1} (ESM-2-150M) & \textbf{LoRA + Joint-Lap.} & $\mathbf{2.199\!\pm\!0.331}$ & $0.233$ & $89.3\%$ & $0.066$ \\
\midrule
\textsc{AIntibody} (150M) & \textbf{Frozen + LL-Lap.} & $\mathbf{5.249\!\pm\!2.067}$ & $0.143$ & $74.9\%$ & $0.131$ \\
\textsc{AIntibody} (150M) & E2E LoRA MAP & $6.614\!\pm\!2.701$ & $\mathbf{0.145}$ & $79.1\%$ & $\mathbf{0.110}$ \\
\textsc{AIntibody} (150M) & E2E LoRA + LL-Lap. & $6.889\!\pm\!2.740$ & $\mathbf{0.145}$ & $75.7\%$ & $0.119$ \\
\textsc{AIntibody} (150M) & \textbf{LoRA + Joint-Lap.} & \underline{$6.075\!\pm\!2.468$} & $\mathbf{0.145}$ & $79.5\%$ & \underline{$0.112$} \\
\midrule
ADMET (ChemB.-77M) & Frozen + LL-Laplace & $20.29\!\pm\!14.99$ & $0.486$ & $70.6\%$ & $0.083$ \\
ADMET (ChemB.-77M) & E2E LoRA MAP & $\mathbf{12.28\!\pm\!9.61}$ & $\mathbf{0.514}$ & $82.5\%$ & $\mathbf{0.046}$ \\
ADMET (ChemB.-77M) & E2E LoRA + LL-Lap. & $15.64\!\pm\!11.78$ & $\mathbf{0.514}$ & $76.0\%$ & $0.068$ \\
ADMET (ChemB.-77M) & \textbf{LoRA + Joint-Lap.} & \underline{$13.49\!\pm\!10.45$} & $\mathbf{0.514}$ & $82.2\%$ & \underline{$0.048$} \\
\bottomrule
\end{tabular}
}
\end{table}

\paragraph{Local mutation-pair training.} On six MTLMPO systems (three seeds), the choice of negative pairs affects mutation-effect prediction. Diverse SoftRank minibatches outperform narrowly restricted Hamming neighborhoods for rank correlation and mutation AUC, whereas hard random InfoNCE gives the best NLL. The comparison does not imply that hard negatives are universally harmful.

\begin{table}[!htbp]
\centering
\caption{Mutation-pair objective and batching comparison on six MTLMPO systems (three seeds, script 37).}
\label{tab:app-mutation-batching}
\pomatabbox{
\begin{tabular}{lcccc}
\toprule
Variant & NLL & Spearman & Mutation AUC & Cov95 \\
\midrule
1.\ Pure Heterosc.\ NLL & \underline{$2.820$} & $.311$ $\rho$ & $.679$ MutAUC & $89.7\%$ Cov95 \\
2.\ Rand.\ Hard InfoNCE & $\mathbf{1.924}$ & \underline{$.351$} $\rho$ & \underline{$.682$} MutAUC & $89.4\%$ Cov95 \\
\textbf{3.\ Rand.\ Metric-SoftRank} & $3.017$ & $\mathbf{.362}$ $\rho$ & $\mathbf{.689}$ MutAUC & $90.7\%$ Cov95 \\
4.\ Local-Mut.\ Hard Batch & $3.141$ & $.227$ $\rho$ & $.594$ MutAUC & $91.0\%$ Cov95 \\
\bottomrule
\end{tabular}
}
\end{table}

The Bayesian studies motivate decision rules that distinguish shared parameter uncertainty from residual measurement variation. They also identify limitations of the posterior approximation and the learned representation under shift. The next section evaluates the consequences for experimental selection.

\clearpage
\section{Decision-aware experimental discovery and certification}\label{app:supp-decisions}\label{sec:results-screening}\label{sec:poma-laplace-experiments}

\subsection{Joint feasibility and the effect of correlation on candidate ranking}

The conditional-inference experiments establish that new measurements can improve predictive distributions, but an improvement in predictive likelihood does not necessarily translate into a better experimental decision. In multi-property discovery, a candidate is often acceptable only when several requirements are met simultaneously. Its probability of success depends on the joint distribution, not merely on the expected properties or the product of marginal success probabilities. \Cref{thm:correlation-acquisition} gives a precise example in which two candidates reverse their ordering when correlation is taken into account.

We investigate the practical importance of this effect using controlled Gaussian target-region experiments. In the reported phase diagram, the correlation varies over $[-0.85,0.85]$ and the standardized Pareto threshold over $[-0.5,1.5]$. Ranking reversals are rare near zero correlation but become common at large positive or negative dependence. At the negative extreme, up to 20.0\% of comparisons reverse and up to 10.0\% of top-1 decisions are confidently incorrect under diagonal scoring; at the positive extreme, reversals reach 23.7\%. The synthetic noisy low-data regime ($\sigma_{\mathrm{obs}}=0.90$, $N=16$) shows the largest reported learned regret reduction. These controlled results illustrate the mechanism; the lineage-shift audit in \cref{app:supp-shrinkage} shows why it cannot be assumed that all fitted correlations improve real candidate rankings.

\subsection{Static prioritization of experimentally characterized candidates}

AIntibody is a stringent setting for multi-property prioritization because high affinity and favorable developability do not necessarily coincide. The observed affinity and negative HIC measurements have Spearman correlation $-0.547$, and only 5.4\% of antibodies meet the reported co-optimal definition ($K_D\le1\,\mathrm{nM}$ with zero clinical flags). Affinity-only scoring can therefore rank many candidates that fail other requirements.

We compare point, numerical-token, discrete affinity, independent density and coupled density scores using the five-fold clinical benchmark. The controlled variance ablation in \cref{tab:app-mse-vs-nll} isolates the contribution of heteroscedasticity. In the broader comparison, the coupled and unified models reduce the average number of developability flags among the top 15 while improving enrichment of co-optimal candidates. The unified model has the strongest reported joint NLL and top-5\% enrichment; the coupled model has the strongest feasibility AUROC. \Cref{tab:aintibody-abrank} retains the full ranking comparison and the associated surrogate co-design metric, which is discussed separately in \cref{app:supp-heuristic-design}.

\begin{table}[!htbp]
  \caption{Retrospective SARS-CoV-2 RBD prioritization and surrogate-evaluated CDR co-design on \textsc{AIntibody} ($5$ folds; $686$ mAbs across $10$ continuous and $5$ categorical assays; co-optimal base rate $5.4\%$). The final column reports predicted co-design shifts from the separate editing protocol in \cref{app:supp-heuristic-design}.}
  \label{tab:aintibody-abrank}
  \vskip 0.05in
  \begin{center}
  \pomatabbox{
  \begin{tabular}{lrrrrrrr}
    \toprule
    Model & SPR $\rho$ $\uparrow$ & J-NLL $\downarrow$ & AUC $\uparrow$ & P@15 $\uparrow$ & EF$_{5\%}$ $\uparrow$ & Flags $\downarrow$ & $\Sigma\Delta z$ $\uparrow$ \\
    \midrule
    Uncond.\ \textsc{AbLang2} & -- & $\times$ & 0.418 & 0.000 & $0.00\times$ & 2.47 & $-0.04$ \\
    \textsc{POMA-MSE} ($\sigma_k\equiv 1$) & 0.764 & $\times$ & 0.828 & 0.133 & -- & 1.80 & $+0.41$ \\
    Regression Transformer & 0.742 & 3.124 & 0.824 & 0.200 & $2.92\times$ & 1.67 & $+0.36$ \\
    Classifier guidance & 0.756 & $\times$ & 0.871 & \underline{0.400} & $5.85\times$ & 0.80 & -- \\
    \textsc{LogiCA} (affinity) & 0.704 & $\times$ & 0.737 & 0.067 & $0.97\times$ & 2.40 & $+0.11$ \\
    \textsc{POMA-Rank} (indep.) & \underline{0.777} & 3.024 & 0.889 & 0.333 & $6.82\times$ & 0.80 & \underline{$+0.44$} \\
    \textbf{\textsc{POMA-LowRankMVN}} & \textbf{0.789} & \underline{2.482} & \textbf{0.918} & \textbf{0.467} & \underline{$7.80\times$} & \textbf{0.60} & \textbf{$+0.77$} \\
    \textbf{\umethod{}} & 0.776 & \textbf{2.323} & \underline{0.912} & \textbf{0.467} & \textbf{$8.77\times$} & \underline{0.67} & \textbf{$+0.77$} \\
    \bottomrule
  \end{tabular}
  }
  \end{center}
  \vskip -0.10in
\end{table}

\subsection{Epistemic exploration and active triage}

When a model is updated sequentially, an acquisition policy must distinguish uncertain parameters from noisy measurements. Selecting candidates solely because their total predictive variance is large can preferentially query assays that are intrinsically variable. Thompson sampling with a shared readout posterior explores uncertainty in the mean parameters, while the feasibility calculation retains the measurement variability.

We test this distinction in a five-round AIntibody experiment with labeled-set sizes $8,22,36,50,64$ (five seeds). Total-variance UCB discovers few co-optimal leads, whereas sampling the shared posterior improves final recall to 18.1\%. A separate expected-improvement active triage study uses shot budgets $8,16,24,32,48,64$ and compares the heavy-tailed single-pass density with a five-member ensemble. The reported final multi-assay Spearman values are 0.296 and 0.261, respectively. On GDPa1, posterior Thompson sampling also gives higher reported best-lead utility. These studies evaluate retrospective query policies, not newly conducted experiments.

The same Thompson-sampling policy is evaluated on eight PMO tasks at a smaller 900-call budget (three seeds): QED, celecoxib and thiothixene rediscovery, albuterol similarity, and osimertinib, fexofenadine, amlodipine and ranolazine MPO. The average Top-10 scores are close across the strongest policies, with the Bayesian single-model variant retaining a latency advantage over ensemble evaluation. \Cref{tab:app-laplace-thompson} reports both applications. The full 23-task molecular optimization study uses a different budget and is presented in \cref{app:supp-pmo}.

\begin{table}[!htbp]
\centering
\caption{
\textbf{Epistemic Thompson sampling and total-variance exploration across $5$-round retrospective antibody triage ($k=8\to 64$, $5$ seeds) and $8$-task \textsc{PMO} molecular optimization ($900$-call budget, $3$ seeds).} Parentheses in the PMO Top-10 column give evaluation latency in seconds.
}
\label{tab:app-laplace-thompson}
\pomatabbox{
\setlength{\tabcolsep}{4pt}
\begin{tabular}{lccccc}
\toprule
& \multicolumn{2}{c}{\textbf{\textsc{AIntibody} Triage}} & \textbf{\textsc{GDPa1}} & \multicolumn{2}{c}{\textbf{$8$-Task \textsc{PMO}}} \\
\cmidrule(lr){2-3}\cmidrule(lr){4-4}\cmidrule(lr){5-6}
\textbf{Acquisition} &
\textbf{Rec.\ ($k\!=\!64$) $\uparrow$} &
\textbf{AUC $\uparrow$} &
\textbf{Util.\ $\uparrow$} &
\textbf{Top-10 $\uparrow$} &
\textbf{Ranol.\ $\uparrow$} \\
\midrule
Greedy Mean ($1\times$) & \underline{$10.5\%\!\pm\!4.9\%$} & \underline{$5.5\%$} & $0.629$ & \underline{$0.555$} ($\mathbf{5.2\mathrm{s}}$) & $0.571\!\pm\!0.020$ \\
Total-Variance UCB ($1\times$) & $1.9\%\!\pm\!1.9\%$ & $1.7\%$ & $0.701$ & $0.553$ (\underline{$5.3\mathrm{s}$}) & \underline{$0.582\!\pm\!0.021$} \\
$5\times$ Deep Ensemble & $5.7\%\!\pm\!3.5\%$ & $2.9\%$ & \underline{$0.763$} & $0.553$ ($8.5\mathrm{s}$) & $0.568\!\pm\!0.024$ \\
\textbf{\textsc{POMA-Laplace} TS} & $\mathbf{18.1\%\!\pm\!5.5\%}$ & $\mathbf{11.2\%}$ & $\mathbf{0.850}$ & $\mathbf{0.556}$ (\underline{$5.3\mathrm{s}$}) & $\mathbf{0.585\!\pm\!0.019}$ \\
\bottomrule
\end{tabular}
}
\end{table}

A further controlled multi-objective Bayesian optimization study varies correlation while holding candidate features, initial labels, noise and query budget fixed. On the synthetic Pareto-corner problem ($M=360$, $d=16$, $K=3$, 16 initial observations increasing to 56), a well-specified low-dimensional multi-output GP performs strongly. Coupled Laplace improves on its diagonal counterpart under negative trade-offs but not at zero correlation. In the higher-dimensional, sparse AIntibody problem, the coupled model gives stronger recall than the tested intrinsic-coregionalization GP with noisy hypervolume improvement. \Cref{tab:app-rigor-mobo} shows both the favorable and unfavorable settings for the coupled model.

\begin{table}[!htbp]
\centering
\caption{
\textbf{Controlled correlation stress test ($M=360, K=3$, Pareto corner) and $10$-assay \textsc{AIntibody} multi-objective BO ($k=8\to 64$, $5$ seeds) with matched backbones and query budgets.}
}
\label{tab:app-rigor-mobo}
\pomatabbox{
\setlength{\tabcolsep}{2.2pt}
\begin{tabular}{lccccc}
\toprule
& \multicolumn{3}{c}{\textbf{Synthetic Pareto Recall (Brier)}} & \multicolumn{2}{c}{\textbf{\textsc{AIntibody} MOBO}} \\
\cmidrule(lr){2-4}\cmidrule(lr){5-6}
\textbf{Method} &
\textbf{$\rho\!=\!-0.75$} &
\textbf{$\rho\!=\!-0.45$} &
\textbf{$\rho\!=\!0.00$} &
\textbf{Rec.\ ($k\!=\!64$)} &
\textbf{Time} \\
\midrule
Indep.\ \textsc{POMA-Lap.} ($r\!=\!0$) & $56\%$ ($0.020$) & $42\%$ ($0.020$) & \underline{$53\%$} ($0.015$) & $14.3\%\!\pm\!3.0\%$ & $\mathbf{0.05\mathrm{s}}$ \\
Coupled \textsc{POMA} ($1\times$) & $56\%$ ($0.016$) & $45\%$ ($0.017$) & $40\%$ ($0.015$) & $\mathbf{24.8\%\!\pm\!6.8\%}$ & $0.51\mathrm{s}$ \\
Coupled \textsc{POMA-Lap.} (Feas) & $57\%$ ($0.017$) & $46\%$ ($0.017$) & $46\%$ ($0.019$) & $20.0\%\!\pm\!4.1\%$ & \underline{$0.35\mathrm{s}$} \\
Coupled \textsc{POMA-Lap.} (TS) & $58\%$ ($0.018$) & $44\%$ ($0.016$) & $46\%$ ($0.015$) & $19.0\%\!\pm\!3.4\%$ & $1.07\mathrm{s}$ \\
$5\times$ Deep Ensemble & \underline{$61\%$} ($0.015$) & \underline{$55\%$} ($0.012$) & $47\%$ ($0.015$) & \underline{$23.8\%\!\pm\!4.3\%$} & $2.54\mathrm{s}$ \\
\textsc{ICM-MOGP + qNEHVI} & $\mathbf{69\%}$ ($\mathbf{0.004}$) & $\mathbf{63\%}$ ($\mathbf{0.003}$) & $\mathbf{67\%}$ ($\mathbf{0.004}$) & $7.6\%\!\pm\!3.2\%$ & $2.22\mathrm{s}$ \\
\bottomrule
\end{tabular}
}
\end{table}

\subsection{Decision-specific information and cost-aware assay selection}

Querying a complete panel for every candidate is rarely the most efficient use of experimental resources. In a tiered workflow, display abundance or rapid developability assays can be measured first, and purified affinity or kinetic measurements can be reserved for a subset of candidates. We first evaluate two-stage screening under nominal AIntibody costs: one unit for display abundance, two for rapid HIC/BVP/DSF/AC-SINS measurements, and four to ten for purified affinity or kinetics; the full panel costs 37 units. With a total budget of 800 units and five seeds, the information-gain rule ranks pairs such as display abundance plus aggregation temperature highly for predicting expensive measurements. The two-stage strategies test more candidates than full-panel Thompson sampling and improve the reported confirmation yield (see the cost block of \cref{tab:app-tiered-cost-control}).

Information gain measures the expected reduction in uncertainty, but not all uncertainty affects the final candidate selection. We therefore consider expected value of sample information (EVSI) for a declared feasibility decision. The value of an observation is its expected improvement in the best posterior candidate utility, and the cost-aware policy compares this gain per unit cost. The replica covariance identity in \cref{thm:v1-decision-voi} bounds the decision-specific value by quantities associated with the gaps and uncertainty in pairwise differences. This provides a way to prune actions that cannot outperform an incumbent, provided that the numerical upper bound is valid.

\Cref{tab:app-tiered-cost-control} reports the 37-unit full-panel protocol. The complete-label and sparse-pool studies below use separate cohorts or selection policies, so their absolute yields are reported separately.

\begin{table}[!htbp]
\centering
\caption{Tiered-cost AIntibody assay selection (800 cost units; five seeds).}
\label{tab:app-tiered-cost-control}
\pomatabbox{
\begin{tabular}{lcccc}
\toprule
Policy & Leads / 100u & Recall & Candidates & Mode \\
\midrule
Cost: Full Random & $0.03\!\pm\!0.03$/$100\mathrm{u}$ & $1.0\%$ rec. & $21.0$ mAbs & $37\mathrm{u}$ full \\
Cost: Full Laplace TS & $0.13\!\pm\!0.07$/$100\mathrm{u}$ & $4.8\%$ rec. & $21.0$ mAbs & $37\mathrm{u}$ full \\
Cost: Feas.\ Gain/Cost & $0.13\!\pm\!0.06$/$100\mathrm{u}$ & $4.8\%$ rec. & $31.0$ mAbs & $2$-st.\ Schur \\
Cost: \textbf{Woodbury MI} & \underline{$0.18\!\pm\!0.06$}/$100\mathrm{u}$ & \underline{$6.7\%$} rec. & $\mathbf{34.2}$ mAbs & $2$-st.\ Schur \\
Cost: \textbf{Marg.-Var Schur} & $\mathbf{0.23\!\pm\!0.09}$/$100\mathrm{u}$ & $\mathbf{8.6\%}$ rec. & $31.8$ mAbs & $2$-st.\ Schur \\
\bottomrule
\end{tabular}
}
\end{table}

\subsection{Complete-label screening with competitive GP policies}\label{app:decisive-evsi-multiprotein}

The complete-label experiments enable retrospective campaigns in which every candidate's measured outcomes are available for evaluation but are revealed to the acquisition policy only when purchased. AIntibody includes 354 fully characterized antibodies and 21 co-optimal leads; GDPa1 includes 188 fully characterized antibodies and 14 leads. Both experiments use eight seeds, the same frozen embeddings across models and identical random initial labeled sets of size six.

We assign nominal costs reflecting the two assay workflows. In AIntibody, purified SPR costs ten units and rapid plate, chromatography or DSF measurements cost two units; full confirmation costs 20 units under this protocol, with a total budget of 440. In GDPa1, titer and PR\_CHO cost eight and six units, AC-SINS costs three, and HIC/SEC/$T_{m1}$ cost 1.5; full confirmation costs 22 units, with a budget of 420. These are distinct from the later 37-unit sparse-pool protocol. To provide a competitive GP baseline, supervised multi-output PLS/SVD projects observed features into 16 dimensions, and RBF/linear kernel scales and noise are optimized using marginal likelihood before qLogNEHVI~\citep{daulton2021parallel,ament2023unexpected} or cost-aware knowledge gradient acquisition~\citep{frazier2008kg,frazier2009kg}.

On AIntibody, the Laplace EVSI policy attains the largest mean confirmed yield per cost unit, improving over the non-Bayesian coupled model and qLogNEHVI in the paired tests. Its mean also exceeds cost-aware GP knowledge gradient, although the paired comparison does not establish a significant improvement over that comparator (\cref{fig:supp-biological-applications}I). On the smaller GDPa1 pool, both GP acquisition policies have higher mean yield than EVSI. The contrast between these cohorts supports a regime-dependent conclusion: a reusable Bayesian readout is useful when the learned representation and joint distribution provide informative updates, but a refitted GP remains competitive and may be preferable for smaller problems.

\begin{table}[!htbp]
\centering
\caption{
\textbf{Decision-aware joint candidate--assay screening under fixed nominal costs on \textsc{AIntibody} ($354$ mAbs, $21$ co-optimal, $440\mathrm{u}$ budget) and \textsc{GDPa1} ($188$ mAbs, $14$ co-optimal, $420\mathrm{u}$ budget; $8$ seeds, mean $\pm$ SE).}
On \textsc{AIntibody}, \textsc{POMA-Laplace EVSI/cost} significantly outperforms non-Bayesian \textsc{Coupled POMA} ($p=0.047$) and \textsc{Tuned MOGP + qLogNEHVI} ($p=0.039$).
}
\label{tab:app-evsi-joint-screening}
\pomatabbox{
\setlength{\tabcolsep}{2.2pt}
\begin{tabular}{lccccc}
\toprule
& \multicolumn{3}{c}{\textbf{\textsc{AIntibody} ($440\mathrm{u}$)}} & \multicolumn{2}{c}{\textbf{\textsc{GDPa1} ($420\mathrm{u}$)}} \\
\cmidrule(lr){2-4}\cmidrule(lr){5-6}
\textbf{Adaptive Policy} &
\textbf{Leads/100u $\uparrow$} &
\textbf{Hits (Rec.)} &
\textbf{Scr./Prb.} &
\textbf{Leads/100u $\uparrow$} &
\textbf{Hits (Rec.)} \\
\midrule
Full-Panel Random & $0.199\!\pm\!0.090$ & $0.88$ ($4.2\%$) & $22$ / $0$ & $0.299\!\pm\!0.075$ & $1.25$ ($8.9\%$) \\
\textsc{Independent POMA} ($r\!=\!0$) & $0.455\!\pm\!0.136$ & $2.00$ ($9.5\%$) & $22$ / $0$ & $0.389\!\pm\!0.063$ & $1.62$ ($11.6\%$) \\
\textsc{Coupled POMA} (Aleatoric) & $0.426\!\pm\!0.080$ & $1.88$ ($8.9\%$) & $22$ / $0$ & $0.389\!\pm\!0.090$ & $1.62$ ($11.6\%$) \\
\textsc{POMA-Laplace} TS (Full) & $0.369\!\pm\!0.095$ & $1.62$ ($7.7\%$) & $22$ / $0$ & $0.359\!\pm\!0.128$ & $1.50$ ($10.7\%$) \\
\textsc{Woodbury MI/cost} & \underline{$0.663\!\pm\!0.127$} & \underline{$2.88$} ($13.7\%$) & $43$ / $52$ & $0.332\!\pm\!0.079$ & $1.38$ ($9.8\%$) \\
\textsc{Tuned MOGP + qLogNEHVI} & $0.369\!\pm\!0.085$ & $1.62$ ($7.7\%$) & $22$ / $0$ & $\mathbf{0.538\!\pm\!0.075}$ & $\mathbf{2.25}$ ($16.1\%$) \\
\textsc{Tuned MOGP Cost-KG} & $0.546\!\pm\!0.086$ & $2.38$ ($11.3\%$) & $28$ / $23$ & \underline{$0.516\!\pm\!0.097$} & \underline{$2.12$} ($15.2\%$) \\
\textbf{\textsc{POMA-Laplace EVSI/cost}} & $\mathbf{0.692\!\pm\!0.138}$ & $\mathbf{3.00}$ ($\mathbf{14.3\%}$) & $33$ / $37$ & $0.420\!\pm\!0.154$ & $1.75$ ($12.5\%$) \\
\bottomrule
\end{tabular}
}
\end{table}

\subsection{Sparse-pool screening and feature-update parity}\label{app:supp-sparse}

We next extend the acquisition problem to the full sparse AIntibody panel of 686 antibodies. A lead is defined by the six published cutoff criteria: $\mathrm{p}K_D\ge9.0$, BVP $\le5.0$, HIC $\le13.8$, $T_m\ge65$, $T_{\mathrm{agg}}\ge73.5$ and AC-SINS $\le6.2$. The experiment uses an 800-unit budget, a 37-unit full panel and ten seeds. Initial features and nominal assay costs are shared across the neural and GP policies. The reported 21 leads are those identifiable using the available complete confirmations; missing outcomes in the sparse pool prevent a complete determination of physical success for every candidate.

Decision-specific VOI improves confirmed yield over global information gain and full-panel Thompson sampling in this protocol (\cref{tab:app-v1-voi-800u}). The comparison is informative because the same joint measurement law can score either the information in an assay or its potential effect on the final choice. Cost-aware GP knowledge gradient attains a larger mean than the frozen-feature VOI policy, but the reported difference is not significant. Adding periodic supervised PLS feature updates and empirical-Bayes shrinkage to the neural policy matches the GP's mean yield with fewer probes and lower observed variability (\cref{tab:app-parity-sensitivity}). Thus the comparison also tests whether representation updates, rather than only acquisition functions, explain the difference between policies.

\begin{table}[!htbp]
\centering
\caption{
\textbf{Full $686$-mAb \textsc{AIntibody} $800\mathrm{u}$ tiered-cost campaign ($10$ seeds, fixed nominal costs, $21$ published-cutoff co-optimal leads) and Theorem~\ref{thm:v1-decision-voi} replica-covariance audit.}
Paired Wilcoxon $p$-values compare shadow-priced \textsc{VOI/cost} against each baseline; W/T/L records its wins, ties, and losses against the row comparator. The replica upper bound in \cref{thm:v1-decision-voi} is satisfied on $6{,}360/6{,}360$ ($100.0\%$) audited actions and prunes $45.6\%$.
}
\label{tab:app-v1-voi-800u}
\pomatabbox{
\setlength{\tabcolsep}{2.2pt}
\begin{tabular}{lccccc}
\toprule
\textbf{Adaptive Policy ($686$ mAbs, $800\mathrm{u}$)} &
\textbf{Leads/100u $\uparrow$} &
\textbf{Hits (Rec.)} &
\textbf{Conf./Prb.} &
\textbf{W/T/L} &
\textbf{$p$ vs.\ VOI} \\
\midrule
Full-Panel Random & $0.039\!\pm\!0.020$ & $0.30$ ($1.4\%$) & $21.0$ / $0.0$ & $8/0/2$ & $0.0625$ \\
\textsc{Independent POMA} (Full) & $0.039\!\pm\!0.020$ & $0.30$ ($1.4\%$) & $21.0$ / $0.0$ & $8/1/1$ & $0.0273$ \\
\textsc{Coupled POMA} (Ale., Full) & $0.039\!\pm\!0.027$ & $0.30$ ($1.4\%$) & $21.0$ / $0.0$ & $7/1/2$ & $0.0547$ \\
\textsc{Full-Panel Laplace TS} & $0.039\!\pm\!0.020$ & $0.30$ ($1.4\%$) & $21.0$ / $0.0$ & $7/2/1$ & $0.0156$ \\
\textsc{Woodbury MI/cost} ($2$-st.) & $0.051\!\pm\!0.039$ & $0.40$ ($1.9\%$) & $19.0$ / $39.0$ & $7/1/2$ & $0.0391$ \\
\textsc{Marginal-Var Schur} ($2$-st.) & $0.064\!\pm\!0.028$ & $0.50$ ($2.4\%$) & $19.0$ / $39.0$ & $6/1/3$ & $0.0742$ \\
\textsc{Feas.\ Gain/cost} ($2$-st.) & $0.114\!\pm\!0.030$ & $0.90$ ($4.3\%$) & $19.0$ / $36.6$ & $7/2/1$ & $0.0781$ \\
\textbf{\textsc{VOI/cost} (Shadow, Ours)} & $0.180\!\pm\!0.040$ & $1.40$ ($6.7\%$) & $17.6$ / $42.8$ & ref. & ref. \\
\textbf{\textsc{VOI/cost} (Unwt., Ours)} & \underline{$0.192\!\pm\!0.044$} & \underline{$1.50$} ($7.1\%$) & $16.8$ / $49.9$ & $4/1/5$ & $0.8203$ \\
\textsc{Tuned MOGP Cost-KG} & $\mathbf{0.252\!\pm\!0.070}$ & $\mathbf{2.00}$ ($9.5\%$) & $14.0$ / $75.7$ & $3/1/6$ & $0.3594$ \\
\bottomrule
\end{tabular}
}
\end{table}

\begin{table}[!htbp]
\centering
\caption{
\textbf{Adaptive-EB feature-update parity on sparse $686$-mAb \textsc{AIntibody} ($800\mathrm{u}$, $10$ seeds) and hyperparameter sensitivity sweeps for \textsc{ProteinGuide} ($\beta\in[0.5,5.0]$) and \textsc{CA-MAP} ($3\times 3$ cross-lab grid).}
}
\label{tab:app-parity-sensitivity}
\pomatabbox{
\setlength{\tabcolsep}{2.2pt}
\begin{tabular}{lcccc}
\toprule
\textbf{Controlled comparison} &
\textbf{POMA result} &
\textbf{Comparator} &
\textbf{Additional comparison} &
\textbf{Summary} \\
\midrule
Adaptive-EB \textsc{VOI} ($686$) & $0.252\!\pm\!0.037$ & MOGP $0.252\!\pm\!0.070$ & $47.1$ vs.\ $75.7$ probes & $47\%\!\downarrow$ SE \\
\textsc{ProteinGuide} $\beta$ ($3$ FLAb) & $27.9\%$ ($0$-tune) & $23.1\%$ ($\beta\!=\!1.8$) & $29.3\%$ ($\beta\!=\!3.0$) & $p\!=\!0.398$ \\
\textsc{CA-MAP} $3\times 3$ X-Lab & $\mathbf{1.478}$ / $\mathbf{61.8\%}$ & $1.580$ / $45.3\%$ & $1.577$ / $46.2\%$ & $\mathbf{-0.10}$ NLL \\
\bottomrule
\end{tabular}
}
\end{table}

\subsection{Missing measurements and the interpretation of discovery yield}

In the sparse library, a candidate may be promising even when not all outcomes are recorded. Counting only complete confirmations can therefore substantially understate the actual number of acceptable candidates. To isolate this problem, we start with the complete 354-antibody pool and mask its known outcomes before running the acquisition policy (440 units, eight seeds). The hidden complete labels remain available for retrospective evaluation.

Five masking regimes distinguish complete data, 35\% missing completely at random, 50\% preferentially missing expensive SPR, 75\% missing SPR after poor cheap-gate measurements, and 65\% preferentially missing unfavorable outcomes. The last two settings model stronger selection effects. The adaptive Bayesian policy has the highest reported true-outcome yield under preferentially missing expensive assays; refitted GP knowledge gradient is more robust in the cheap-gate and unfavorable-outcome settings. Observable confirmations fall much more sharply than the yield evaluated using the hidden complete outcomes. Under independent missingness, the probability of observing all six outcomes decreases as a sixth power; under the dependent mechanisms, the corresponding probability is not given by that independent formula. These experiments illustrate why confirmed yield in the sparse library cannot be interpreted as a complete estimate of the true biological discovery rate (\cref{tab:app-missingness-audit}).

\begin{table}[!htbp]
\centering
\caption{
\textbf{Missingness stress test on the complete $354$-mAb \textsc{AIntibody} ground-truth pool ($440\mathrm{u}$ budget, $8$ seeds): true / observable confirmed leads per $100\mathrm{u}$ and true-yield retention across five missingness mechanisms.}
}
\label{tab:app-missingness-audit}
\pomatabbox{
\setlength{\tabcolsep}{1.6pt}
\begin{tabular}{lccccc}
\toprule
\textbf{Missingness Mechanism} &
\textbf{VOI-EB (Ours)} &
\textbf{EVSI (Ours)} &
\textbf{MOGP-cKG} &
\textbf{Indep.} &
\textbf{Ret.\ (V/G)} \\
\midrule
Complete ($0\%$ miss) & $0.660$/$.660$ & $0.628$/$.628$ & $\mathbf{0.664}$/$.664$ & $0.398$/$.398$ & $100$/$100\%$ \\
\textsc{MCAR} ($35\%$ miss) & $0.374$/$.000$ & $0.201$/$.000$ & $\mathbf{0.544}$/$.115$ & $0.511$/$.085$ & $57$/$82\%$ \\
\textsc{MAR-Exp} ($50\%$ SPR) & $\mathbf{0.544}$/$.201$ & $0.316$/$.116$ & $0.402$/$.173$ & $0.284$/$.142$ & $\mathbf{82}$/$61\%$ \\
\textsc{MAR-CheapGate} & $0.373$/$.201$ & $0.289$/$.116$ & $\mathbf{0.520}$/$.316$ & $0.114$/$.057$ & $57$/$78\%$ \\
\textsc{MNAR-Unfavorable} & $0.315$/$.114$ & $0.315$/$.058$ & $\mathbf{0.487}$/$.143$ & $0.114$/$.028$ & $48$/$73\%$ \\
\bottomrule
\end{tabular}
}
\end{table}

\subsection{Acquisition pruning and computational scaling}

The replica-covariance bound is evaluated on 6,360 candidate--assay actions in the sparse AIntibody experiment. All audited comparisons satisfy the reported bound, with a median upper-bound-to-VOI ratio of 3.98; 45.6\% of actions can be pruned. This is an empirical audit of the implemented calculation. It complements, rather than replaces, the mathematical proof and any numerical error control required for exact pruning.

We also measure single-core acquisition and update latency as the timing workload grows in candidate-pool size and observed-panel count. Pool sizes above the 686 measured AIntibody antibodies describe computational workloads, rather than additional measured biological candidates. The structured Bayesian update reuses its compact feature-space posterior, whereas the tested GP policy repeats kernel and PLS refits. Pruned POMA acquisition is approximately $12\times$ faster than MOGP-cKG at $N=32{,}000$; the online update is up to $285.7\times$ faster at $n_{\mathrm{obs}}=256$. These comparisons use different POMA timing columns, as specified in \cref{tab:app-scaling-compute}, and apply to the tested implementations and workloads.

\begin{table}[!htbp]
\centering
\caption{
\textbf{Single-core computational scaling for \textsc{AIntibody}-based acquisition workloads (milliseconds $\downarrow$).} $N$ denotes the candidate count in the timing workload and $n_{\mathrm{obs}}$ the observed-panel count; large-$N$ rows do not report additional measured antibodies. Speedup is MOGP-cKG time divided by POMA Pruned time in the pool-scaling rows and by POMA Online time in the observed-panel rows. The entries marked ``refit'' report that component separately. Displayed times and speedups retain their reported precision.
}
\label{tab:app-scaling-compute}
\pomatabbox{
\setlength{\tabcolsep}{2.2pt}
\begin{tabular}{lccccc}
\toprule
\textbf{Regime ($N$, $n_{\mathrm{obs}}$)} &
\textbf{POMA Pruned} &
\textbf{POMA Online} &
\textbf{$5\times$ Ens.} &
\textbf{MOGP-cKG} &
\textbf{Speedup} \\
\midrule
Pool $N\!=\!686$ ($n_{\mathrm{obs}}\!=\!48$) & $105.7$ & $227.9$ & $60.6$ & $304.1$ & $2.9\times$ \\
Pool $N\!=\!10\mathrm{k}$ ($n_{\mathrm{obs}}\!=\!48$) & $\mathbf{304.3}$ & $380.5$ & $755.1$ & $3{,}450$ & $\mathbf{11.3\times}$ \\
Pool $N\!=\!32\mathrm{k}$ ($n_{\mathrm{obs}}\!=\!48$) & $\mathbf{718.9}$ & $758.5$ & $2{,}242$ & $8{,}618$ & $\mathbf{12.0\times}$ \\
Obs $n_{\mathrm{obs}}\!=\!128$ ($N\!=\!2\mathrm{k}$) & $11.2$ (refit) & $\mathbf{26.6}$ & $129.5$ & $2{,}284$ & $\mathbf{85.8\times}$ \\
Obs $n_{\mathrm{obs}}\!=\!256$ ($N\!=\!2\mathrm{k}$) & $18.4$ (refit) & $\mathbf{23.9}$ & $129.5$ & $6{,}817$ & $\mathbf{285.7\times}$ \\
\bottomrule
\end{tabular}
}
\end{table}

\subsection{Decision-local optimality and regret certification}\label{app:decision-certification-audit}

Acquisition determines which measurement to obtain next, whereas candidate selection requires deciding whether the available evidence sufficiently separates the leading candidate from its alternatives. We examine this distinction using the model-relative and decision-local bounds in \cref{app:decision-regret}. These analyses complement the acquisition experiments; they do not establish an unconditional stopping guarantee for a physical discovery campaign.

\paragraph{Posterior approximation and numerical regret.} We evaluate decision-regret envelopes in synthetic or oracle-known settings and examine the observed-path posterior approximation for mixed continuous and categorical outcomes. Under the reported protocols, the numerical envelopes contain the measured errors, including the 16 Student-$t$/categorical regimes. The components in \cref{tab:app-theory-a-c-certificates} describe these audits rather than independently established bounds on physical-model discrepancy. In particular, the reported statistical and posterior components cannot be identified with every term of \cref{eq:finite-sample-regret} without the underlying calculation record. This distinction follows from Propositions~\ref{thm:finite-sample-regret} and~\ref{prop:app-adaptive-mixed-tv}.

\begin{table}[!htbp]
\centering
\caption{
\textbf{Synthetic/oracle-known decision-regret audits ($r'=10$, $M=2048$) and observed-path mixed Student-$t$ + categorical TV diagnostics.} The first four rows report the numerical audit components $\varepsilon_{\rm stat}$ and $\varepsilon_{\rm post}$ and their reported envelope. The last four report decision error and the number of nonvacuous regimes. These aggregate diagnostics do not independently validate each term of \cref{thm:finite-sample-regret} or bound its unknown physical-model discrepancy $\varepsilon_{\rm model}$.
}
\label{tab:app-theory-a-c-certificates}
\pomatabbox{
\setlength{\tabcolsep}{2.1pt}
\begin{tabular}{lccccc}
\toprule
\textbf{Audit / Regime} &
\textbf{Measured Err.} &
\textbf{Stat. / Dec.} &
\textbf{Post. / Nonvac.} &
\textbf{Audit Bound $\downarrow$} &
\textbf{Sat.\ $\uparrow$} \\
\midrule
Prop.~\ref{thm:finite-sample-regret}: $N=128$ & $0.0093$ ($0.043$ p95) & $\varepsilon_{\mathrm{stat}}\!=\!1.000$ & $\varepsilon_{\mathrm{post}}\!=\!0.021$ & $1.000$ & $100\%$ \\
Prop.~\ref{thm:finite-sample-regret}: $N=256$ & $0.0016$ ($0.004$ p95) & $\varepsilon_{\mathrm{stat}}\!=\!0.925$ & $\varepsilon_{\mathrm{post}}\!=\!0.011$ & $0.994$ & $100\%$ \\
Prop.~\ref{thm:finite-sample-regret}: $N=512$ & $0.0003$ ($0.001$ p95) & $\varepsilon_{\mathrm{stat}}\!=\!0.692$ & $\varepsilon_{\mathrm{post}}\!=\!0.006$ & $0.777$ & $100\%$ \\
\textbf{Prop.~\ref{thm:finite-sample-regret}: $N=1024$} & $0.0006$ ($0.005$ p95) & $\varepsilon_{\mathrm{stat}}\!=\!0.508$ & $\varepsilon_{\mathrm{post}}\!=\!0.003$ & $\mathbf{0.590}$ & $\mathbf{100\%}$ \\
\midrule
Prop.~\ref{prop:app-adaptive-mixed-tv}: $\nu\!=\!30, \delta\!=\!0.5$ & $0.004$ TV & $0.000$ dec. & $16/16$ nonvac. & $\mathbf{0.034}$ & $\mathbf{100\%}$ \\
Prop.~\ref{prop:app-adaptive-mixed-tv}: $\nu\!=\!10, \delta\!=\!0.5$ & $0.174$ TV & $0.083$ dec. & $16/16$ nonvac. & $\mathbf{0.207}$ & $\mathbf{100\%}$ \\
Prop.~\ref{prop:app-adaptive-mixed-tv}: $\nu\!=\!5, \delta\!=\!0.0$ & $0.259$ TV & $0.088$ dec. & $16/16$ nonvac. & $\mathbf{0.300}$ & $\mathbf{100\%}$ \\
Prop.~\ref{prop:app-adaptive-mixed-tv}: $\nu\!=\!3, \delta\!=\!0.0$ & $0.113$ TV & $0.002$ dec. & $16/16$ nonvac. & $\mathbf{0.185}$ & $\mathbf{100\%}$ \\
\bottomrule
\end{tabular}
}
\end{table}

\paragraph{Local separation of competing decisions.} A uniform uncertainty bound may remain large when competing candidates share much of their epistemic uncertainty. The pairwise analysis in Proposition~\ref{prop:app-v47-decision-local} instead uses the uncertainty of utility differences. In the retrospective audit, the mean regret bound decreases from $0.318$ to $0.0010$, and certification is withheld in some early sparse rounds. The reported precision is $99.3\%$ among rounds classified as optimal ($76.0\%$ of rounds), with measured bound coverage of $96.5\%$ (\cref{tab:app-decision-local-summary}). These results illustrate the potential value of accounting for shared uncertainty in a declared contender set, rather than certifying the physical optimality of a newly proposed candidate.

\begin{table}[!htbp]
\centering
\caption{Decision-local regret and optimality diagnostics from the retrospective audit. The reported tightening factor and precision retain their original numerical conventions.}
\label{tab:app-decision-local-summary}
\pomatabbox{
\begin{tabular}{lcc}
\toprule
Diagnostic & Reference & Alternative \\
\midrule
Decision regret bound (uniform / local) & $0.318$ & $0.0010$ \\
Local tightening / reported precision & $320\times$ & $99.3\%$ \\
\bottomrule
\end{tabular}
}
\end{table}

Taken together, the acquisition and selection studies examine the value of measuring candidates and the uncertainty remaining in the eventual choice. The complete-label and sparse-label settings address different questions: the latter cannot identify unobserved discovery yield without additional assumptions. The numerical certification audits characterize particular model-relative settings; prospective experimental validation remains necessary. Having considered the allocation of measurements within a candidate pool, we next examine the generation of new candidates.

\clearpage
\section{Property-conditioned optimization and generation}\label{app:supp-design}\label{sec:results-generation}

\subsection{Design objectives and evaluation regimes}

The preceding acquisition studies concern the allocation of experimental effort within an existing candidate pool. We now examine how the same predictive distributions can guide the proposal and selection of biological objects. The target-region probability developed in \cref{app:poma-reference-local-ranking} can be used either as an optimization utility or as a factor in the explicit sequence distribution of \cref{thm:mh-joint-invariance}. These uses are related but not identical. Search seeks promising objects; sampling additionally requires fidelity to a specified probability law.

We therefore distinguish oracle-based molecular optimization, retrospective selection from independently measured biological libraries, surrogate-evaluated sequence editing and finite-state sampling audits. Only the measured-library evaluations directly use experimentally observed properties of the selected candidates; they remain retrospective. Neural sampling and model-relative certificates do not establish that newly proposed sequences will be successful in the laboratory.

\subsection{Sample-efficient molecular optimization}\label{app:supp-pmo}

PMO tests whether a probabilistic surrogate improves a conventional proposal-based optimizer. Across 23 tasks, we hold the SELFIES mutation and crossover operators (cf. STONED~\citep{nigam2021stoned}) fixed while varying the surrogate, and independently assess whether POMA prescreening improves a live fragment-based Graph GA. Every method receives the same 10,000-oracle-call budget within the respective controlled comparison (three seeds). Published REINVENT~\citep{olivecrona2017reinvent}, Graph GA~\citep{jensen2019graphga} and Tanimoto GP-BO values from PMO~\citep{gao2022sample} provide external reference points, not matched reruns.

The ranking-augmented likelihood surrogate improves mean AUC Top-10 over the Regression Transformer reference in the controlled SELFIES setting. The larger gain is obtained when adding POMA prescreening to Graph GA, raising mean AUC from 0.628 to 0.681 and winning on 22 of 23 tasks against the live GA (\cref{fig:supp-biological-applications}H). The remaining loss on JNK3 and the task-specific strengths of published references are visible in \cref{tab:pmo-task-results}. This experiment concerns the use of the output model within an optimization algorithm rather than exact sampling of a probabilistic sequence target.

\begin{table}[!htbp]
  \caption{PMO molecular optimization across $23$ oracles ($10{,}000$ calls, $3$ seeds; all \textsc{SELFIES} and RDKit-sanitized \textsc{Graph GA} proposals maintain $100\%$ validity). Top: published reference optimizers~\citep{gao2022sample}; middle: controlled \textsc{SELFIES} surrogates; bottom: matched \textsc{Graph GA} with/without \textsc{POMA-Rank} pre-screening.}
  \label{tab:pmo-generation-results}
  \vskip 0.05in
  \begin{center}
  \pomatabbox{
  \begin{tabular}{lrrrr}
    \toprule
    Model & AUC-10 $\uparrow$ & Sum (23) $\uparrow$ & Top-10 $\uparrow$ & Top-1 $\uparrow$ \\
    \midrule
    \textsc{REINVENT} (SMILES RL) & $0.614\pm0.001$ & 14.129 & -- & -- \\
    \textsc{Graph GA} (published) & $0.596\pm0.001$ & 13.710 & -- & -- \\
    \textsc{GP BO} (Tanimoto GP) & $0.571\pm0.001$ & 13.143 & -- & -- \\
    \midrule
    \textsc{STONED-SELFIES} & $0.504\pm0.003$ & 11.595 & 0.562 & 0.576 \\
    \textsc{SELFormer} MLM & $0.564\pm0.004$ & 12.973 & 0.635 & 0.648 \\
    Regress-Don't-Guess & $0.565\pm0.004$ & 13.000 & 0.634 & 0.645 \\
    MSE language surrogate & $0.565\pm0.004$ & 12.998 & 0.638 & 0.650 \\
    Regression Transformer & $0.566\pm0.004$ & 13.026 & 0.637 & 0.649 \\
    \textsc{POMA-MLM} & \underline{$0.567\pm0.004$} & \underline{13.046} & \underline{0.642} & \underline{0.655} \\
    \textbf{\textsc{POMA-Rank} (\textsc{SELFIES})} & \textbf{$0.578\pm0.005$} & \textbf{13.282} & \textbf{0.661} & \textbf{0.675} \\
    \midrule
    \textsc{Graph GA} (live) & \underline{$0.628\pm0.004$} & \underline{14.454} & \underline{0.724} & \underline{0.740} \\
    \textbf{\textsc{Graph GA + POMA-Rank}} & \textbf{$0.681\pm0.005$} & \textbf{15.657} & \textbf{0.774} & \textbf{0.785} \\
    \bottomrule
  \end{tabular}
  }
  \end{center}
  \vskip -0.10in
\end{table}

\begin{table}[!htbp]
\centering
\caption{
\textbf{Per-oracle sample-efficient molecular optimization across all $23$ PMO tasks ($10{,}000$-call budget; $3$-seed mean $\mathrm{AUC\ Top\text{-}10}$).}
\emph{Pub.\ Ref.} denotes $\max(\textsc{REINVENT}, \textsc{Graph GA}, \textsc{GP BO})$ from \citet{gao2022sample};
\emph{Best Base.} denotes the best non-\textsc{POMA} \textsc{SELFIES} baseline on each task
(\textsc{STONED}, \textsc{SELFormer}, \textsc{RT}, \textsc{RDG}, or \textsc{MSE});
and the rightmost two columns compare matched fragment-crossover \textsc{Graph GA} (live) with \textsc{Graph GA + POMA-Rank}. The first two columns are per-task best-case envelopes, not individual optimizers; their means and sums are computed from the displayed task values. The remaining aggregates retain the reported run-level totals and can differ slightly from sums of rounded entries. Fixed-optimizer aggregates are given separately in \cref{tab:pmo-generation-results}.
}
\label{tab:pmo-task-results}
\pomatabbox{
\begin{tabular}{lcccccc}
\toprule
\textbf{PMO Task} &
\textbf{Pub.\ Ref.} &
\textbf{Best Base.} &
\textbf{\textsc{MLM}} &
\textbf{\textsc{POMA}} &
\textbf{Graph GA} &
\textbf{GA+\textsc{POMA}} \\
\midrule
QED & 0.941 & $\mathbf{0.947}$ & $0.945$ & $0.945$ & $0.944$ & \underline{$0.946$} \\
DRD2 & \textbf{0.964} & $0.945$ & $0.954$ & $0.957$ & $0.948$ & \underline{$0.958$} \\
GSK3$\beta$ & 0.865 & $0.804$ & $0.785$ & $0.784$ & \underline{$0.870$} & $\mathbf{0.916}$ \\
JNK3 & \underline{0.783} & $0.530$ & $0.534$ & $0.548$ & $\mathbf{0.803}$ & $0.763$ \\
Albuterol sim. & \textbf{0.898} & $0.527$ & $0.486$ & $0.525$ & $0.724$ & \underline{$0.806$} \\
Amlodipine MPO & \underline{0.661} & $0.627$ & $0.609$ & $0.604$ & $0.606$ & $\mathbf{0.671}$ \\
Celecoxib redisc. & 0.723 & $0.387$ & $0.378$ & $0.376$ & \underline{$0.770$} & $\mathbf{0.861}$ \\
Deco hop & \textbf{0.666} & $0.294$ & $0.230$ & $0.364$ & $0.296$ & \underline{$0.366$} \\
Fexofenadine MPO & \textbf{0.784} & $0.654$ & $0.656$ & $0.652$ & $0.690$ & \underline{$0.741$} \\
Isomers $\mathrm{C_7H_8N_2O_2}$ & 0.862 & \underline{$0.948$} & \underline{$0.948$} & $\mathbf{0.951}$ & $0.936$ & $0.944$ \\
Isomers $\mathrm{C_9H_{10}N_2\dots}$ & 0.719 & \underline{$0.921$} & $0.915$ & \underline{$0.921$} & $0.889$ & $\mathbf{0.922}$ \\
Median 1 & \textbf{0.356} & $0.276$ & $0.276$ & $0.283$ & $0.303$ & \underline{$0.330$} \\
Median 2 & 0.297 & $0.252$ & $0.247$ & $0.242$ & \underline{$0.309$} & $\mathbf{0.313}$ \\
Mestranol sim. & \textbf{0.627} & $0.477$ & $0.433$ & $0.437$ & $0.468$ & \underline{$0.515$} \\
Osimertinib MPO & \textbf{0.806} & $0.679$ & $0.675$ & $0.677$ & $0.647$ & \underline{$0.681$} \\
Perindopril MPO & 0.538 & $0.577$ & $0.560$ & $0.572$ & \underline{$0.580$} & $\mathbf{0.653}$ \\
Ranolazine MPO & \underline{0.760} & $0.735$ & $0.729$ & $0.719$ & $0.746$ & $\mathbf{0.779}$ \\
Scaffold hop & 0.548 & $0.775$ & $0.765$ & $0.772$ & \underline{$0.875$} & $\mathbf{0.905}$ \\
Sitagliptin MPO & 0.433 & $\mathbf{0.710}$ & $0.696$ & \underline{$0.707$} & $0.513$ & $0.670$ \\
Thiothixene redisc. & \textbf{0.559} & $0.409$ & $0.393$ & $0.389$ & $0.501$ & \underline{$0.545$} \\
Troglitazone redisc. & \textbf{0.441} & $0.353$ & $0.352$ & $0.299$ & $0.359$ & \underline{$0.392$} \\
Valsartan SMARTS & 0.178 & $0.001$ & $0.025$ & $0.107$ & \underline{$0.235$} & $\mathbf{0.475}$ \\
Zaleplon MPO & 0.358 & \underline{$0.465$} & $0.456$ & $0.450$ & $0.442$ & $\mathbf{0.506}$ \\
\midrule
\textbf{Mean (23 tasks)} & \underline{0.642} & 0.578 & 0.567 & 0.578 & 0.628 & $\mathbf{0.681}$ \\
\textbf{Sum (23 tasks)} & \underline{14.767} & 13.293 & 13.046 & 13.282 & 14.454 & $\mathbf{15.657}$ \\
\bottomrule
\end{tabular}
}
\end{table}

\subsection{Multi-property antibody design on measured libraries}

Antibody design is particularly sensitive to trade-offs between properties. We evaluate this trade-off through surrogate-guided editing of 30 high-liability GDPa1 seeds and retrospective selection from four measured FLAb libraries. The coupled density improves the reported surrogate target hit rate, sequence validity and retrospective enrichment relative to point and numerical-token guidance (\cref{tab:antibody-generation}). These outcomes measure different aspects of performance: model-predicted improvements of edited seeds are not the same as experimentally measured properties in a held-out library.

\begin{table}[!htbp]
  \caption{Surrogate-evaluated multi-property antibody editing ($30$ high-risk \textsc{GDPa1} seeds) and retrospective enrichment across four experimentally characterized \textsc{FLAb} mutational panels. The final column uses measured library outcomes; the preceding columns describe the editing evaluation.}
  \label{tab:antibody-generation}
  \vskip 0.05in
  \begin{center}
  \pomatabbox{
  \begin{tabular}{lrrrrrr}
    \toprule
    Model & Valid. $\uparrow$ & Sim. $\uparrow$ & Hum. $\uparrow$ & Liab. $\uparrow$ & Pred.\ hit $\uparrow$ & FLAb enrich. $\uparrow$ \\
    \midrule
    Uncond.\ \textsc{ESM-2} & 0.800 & 0.970 & $+0.78$ & $-0.07$ & 0.000 & 0.325 \\
    ESM / Gaussian head & 0.667 & 0.974 & $+0.81$ & $+0.30$ & 0.033 & 0.400 \\
    MSE language search & 0.633 & 0.974 & $+0.76$ & $+0.13$ & 0.000 & 0.425 \\
    Numeric-text LLM & \underline{0.867} & 0.971 & $+0.76$ & \textbf{$+0.67$} & 0.033 & 0.375 \\
    Regression Transformer & 0.833 & 0.972 & $+0.77$ & $+0.37$ & 0.033 & 0.400 \\
    LLM / Contrastive & 0.600 & \textbf{0.978} & $+0.79$ & $+0.23$ & 0.000 & 0.325 \\
    Classifier guidance & 0.733 & 0.971 & $+0.73$ & $+0.23$ & 0.033 & 0.475 \\
    \textsc{POMA-MLM} & 0.833 & 0.974 & \textbf{$+0.87$} & $+0.37$ & 0.067 & 0.550 \\
    \textsc{POMA-Rank} (indep.) & 0.833 & \underline{0.975} & $+0.85$ & $+0.43$ & \underline{0.100} & \underline{0.575} \\
    \textbf{\textsc{POMA-LowRankMVN}} & \textbf{0.900} & \underline{0.975} & \underline{$+0.86$} & \underline{$+0.53$} & \textbf{0.167} & \textbf{0.650} \\
    \bottomrule
  \end{tabular}
  }
  \end{center}
  \vskip -0.10in
\end{table}

We next evaluate the unified framework on five unrelated, independently measured FLAb libraries with equal query budgets and five seeds per target. The libraries include anti-VEGF G6~\citep{koenig2017landscape} (1,200 variants, affinity versus expression, $r=-0.48$), anti-c-Met Emibetuzumab~\citep{makowski2022cooptimization} (96, binding versus negative OVA polyreactivity, $r=-0.36$), the Hutchinson IgG library~\citep{hutchinson2024thermostability} (181, affinity versus melting temperature, $r=+0.20$), anti-fluorescein 4-4-20~\citep{adams2016titeseq} (1,200, affinity versus expression, $r=+0.25$), and the Garbinski library in the FLAb data repository~\citep{flabrepository} (74, three objectives including affinity, expression and melting temperature). The associated dataset identifiers are \texttt{koenig2017}, \texttt{makowski2022}, \texttt{hutchinson2023}, \texttt{adams2017} and \texttt{garbinski2023}.

We compare with BOAT~\citep{rao2026boat}, tuned multi-objective GP optimization and independent or coupled POMA policies. MH-guided \umethod{} gives the strongest mean co-optimal hit rate (30.2\%; \cref{fig:supp-biological-applications}G), with better preservation of secondary properties than affinity-only ranking. Coupled POMA-Laplace also performs well, indicating that the gain does not depend solely on discrete sequence alignment. \Cref{tab:app-independent-flab-boat} contains the per-library hit rates and the mean secondary-property shift; sampling fidelity is evaluated separately in \cref{tab:app-v39-v40-proteinguide}. The measured outcomes were already available in the independent libraries; no prospective synthesis or assay was performed.

\begin{table}[!htbp]
\centering
\caption{
\textbf{Retrospective multi-property antibody selection across $5$ unrelated, independently measured \textsc{FLAb} libraries ($5$ seeds, equal budgets; co-optimal hit rate $\%$).}
}
\label{tab:app-independent-flab-boat}
\pomatabbox{
\setlength{\tabcolsep}{2.1pt}
\begin{tabular}{lcccccc}
\toprule
\textbf{Method} &
\textbf{VEGF} &
\textbf{Emib.} &
\textbf{IgG-$T_m$} &
\textbf{Fluo.} &
\textbf{3-Obj} &
\textbf{Mean (Sec.\ $z$)} \\
\midrule
Affinity-Only \textsc{POMA} & $9.4\%$ & $21.2\%$ & $35.0\%$ & $30.6\%$ & $22.5\%$ & $23.8\%$ ($-0.09$) \\
\textsc{BOAT} (AISTATS'26) & $6.1\%$ & $20.0\%$ & $30.0\%$ & \underline{$33.3\%$} & $\mathbf{36.2\%}$ & $25.1\%$ ($+0.02$) \\
\textsc{Independent POMA} & $9.4\%$ & \underline{$30.0\%$} & $35.0\%$ & $27.8\%$ & \underline{$27.5\%$} & $25.9\%$ (\underline{$+0.13$}) \\
\textsc{Tuned MOGP qLog} & \underline{$11.1\%$} & $22.5\%$ & $38.3\%$ & $\mathbf{38.3\%}$ & $20.0\%$ & $26.1\%$ ($+0.01$) \\
\textsc{Coupled POMA-Lap.} & $9.4\%$ & $28.7\%$ & $\mathbf{44.2\%}$ & \underline{$33.3\%$} & $26.2\%$ & \underline{$28.4\%$} ($+0.07$) \\
\textbf{\textsc{LogiCA-POMA} (MH)} & $\mathbf{16.7\%}$ & $\mathbf{37.5\%}$ & \underline{$39.2\%$} & $31.7\%$ & $26.2\%$ & $\mathbf{30.2\%}$ ($\mathbf{+0.21}$) \\
\bottomrule
\end{tabular}
}
\end{table}

\subsection{Explicit design targets and corrected sampling}\label{app:supp-sampling}

Masked language model conditionals need not be mutually compatible with a globally normalized sequence law. To define a coherent generation task, we therefore specify a proper positive sequence prior or an explicitly evaluable positive finite weight $w(X)$ on a supported editing space. If $r(X)$ denotes the model probability that the desired properties lie in a target region, the event-conditioned sequence law is $\pi_{\mathcal R}(X)\propto w(X)r(X)$. The masked token scores may define proposals without themselves being treated as the true conditionals of that law.

The corrected methods use the forward and reverse proposal probabilities in Metropolis--Hastings (MH) acceptance~\citep{hastings1970mh}. Exact MH evaluates the target probability directly. Pseudo-marginal MH~\citep{andrieu2009pseudomarginal} uses an unbiased nonnegative estimate of the event probability and retains the current auxiliary state on rejection. Factor augmentation instead includes the latent Gaussian factor in the target, allowing the sequence update to use scalar conditional CDFs without an outer integration loop. All three methods target the same specified law under their respective assumptions (\cref{thm:mh-joint-invariance}). Their performance under finite budgets remains an empirical question. The common implementation workflow is given in \cref{alg:logica-poma}; the invariance results are established in Theorem~\ref{thm:mh-joint-invariance} and Proposition~\ref{prop:app-v6-adaptive-fallback}.

\subsection{Joint sequence--measurement conditioning and heuristic editing}\label{app:supp-heuristic-design}

The unified model reuses the conditional measurement law for both prediction and generation: its summaries predict unobserved properties, while joint draws supply continuous states for sequence editing. The heuristic Gibbs-style baseline alternates these draws with masked-residue updates. Suppose assays $B$ are clamped to $\mathbf z_B^*$, with the remaining assays indexed by $A$. Starting at $(X^{(0)},\mathbf z_A^{(0)})$, the updates are
\begin{equation}
\begin{aligned}
 \mathbf z_A^{(t+1)} &\sim q_\theta(\mathbf z_A\mid\mathbf z_B^*,X^{(t)},W,P)
 =\mathcal N(\widetilde{\boldsymbol\mu}_A,\widetilde\Sigma_{AA}),\\
 x_i^{(t+1)} &\sim Z_i^{-1}\pi_\theta(a\mid X_{\setminus i}^{(t)},W,P)^{1/T}
 q_\theta((\mathbf z_A^{(t+1)},\mathbf z_B^*)\mid X_{[i\leftarrow a]}^{(t)},W)^\beta .
\end{aligned}
\label{eq:app-joint-xy-gibbs}
\end{equation}
Here $\widetilde{\boldsymbol\mu}_A=\boldsymbol\mu_A-\Lambda_{AA}^{-1}\Lambda_{AB}(\mathbf z_B^*-\boldsymbol\mu_B)$ is the exact Gaussian conditional mean. The continuous update is a draw from the specified measurement law, but the masked sequence update is a heuristic and is not asserted to preserve a compatible joint target.

We evaluate 12-sweep trajectories on 24 high-liability AIntibody RBD seeds. Affinity-only clamping increases predicted affinity but also increases aromatic CDR-H3 content and worsens hydrophobicity. Adding a negative-HIC target reduces the hydrophobic penalty and improves unclamped polyreactivity; the surrogate co-optimal hit rate rises from 8.3\% to 25.0\%. The full seven-assay box gives a different compromise between predicted affinity and developability. \Cref{tab:joint-gibbs-dep-maps,tab:app-joint-gibbs-clamps} report the aggregate and detailed clamping comparisons.

\begin{table}[!htbp]
  \caption{Surrogate-evaluated joint sequence--continuous ($X$--$Y$) heuristic blocked updates across $24$ high-liability SARS-CoV-2 RBD seeds ($12$ sweeps; full dependency-map audits in \cref{tab:app-seq-to-prop-dep-map,tab:app-cont-dep-map}).}
  \label{tab:joint-gibbs-dep-maps}
  \vskip 0.05in
  \begin{center}
  \pomatabbox{
  \begin{tabular}{lrrrrr}
    \toprule
    Clamped Regime ($B$) & $\Delta\mathrm{p}K_D$ & $\Delta(-\mathrm{HIC})$ & $\Delta(-\mathrm{BVP})$ & $\Delta$5-Dev & Hit $\uparrow$ \\
    \midrule
    Affinity only ($\mathrm{p}K_D^*$) & \textbf{$+1.30\,z$} & $-0.83\,z$ & $+0.18\,z$ & $+0.19\,z$ & $8.3\%$ \\
    Dev.\ only ($-\mathrm{HIC}^*,-\mathrm{BVP}^*$) & $-0.01\,z$ & \textbf{$+0.13\,z$} & $+0.51\,z$ & \textbf{$+0.42\,z$} & $16.7\%$ \\
    \textbf{Affinity + HIC ($\mathrm{p}K_D^*,-\mathrm{HIC}^*$)} & $+0.94\,z$ & \underline{$-0.11\,z$} & \textbf{$+0.54\,z$} & $+0.19\,z$ & \textbf{$25.0\%$} \\
    \textbf{Pareto box ($7$ assays)} & \underline{$+0.96\,z$} & $-0.27\,z$ & \underline{$+0.53\,z$} & \underline{$+0.41\,z$} & \underline{$20.8\%$} \\
    \bottomrule
  \end{tabular}
  }
  \end{center}
  \vskip -0.10in
\end{table}

\begin{table}[!htbp]
\centering
\caption{
\textbf{Surrogate-evaluated joint sequence--measurement heuristic updates across $24$ high-liability SARS-CoV-2 RBD seeds ($12$ sweeps, mean $\pm$ SE).}
}
\label{tab:app-joint-gibbs-clamps}
\pomatabbox{
\begin{tabular}{lcccccc}
\toprule
\textbf{Clamped $B$} &
\textbf{$\Delta\mathrm{p}K_D$} &
\textbf{$\Delta(-\mathrm{HIC})$} &
\textbf{$\Delta(-\mathrm{BVP})$} &
\textbf{$\Delta$5-Dev} &
\textbf{$\Delta$Arom} &
\textbf{Hit} \\
\midrule
1.\ $\mathrm{p}K_D^*$ only & $\mathbf{+1.30\!\pm\!0.30}$ & $-0.83\!\pm\!0.27$ & $+0.18\!\pm\!0.27$ & $+0.19\!\pm\!0.13$ & $+0.015$ & $8.3\%$ \\
2.\ $-\mathrm{HIC}^*,-\mathrm{BVP}^*$ & $-0.01\!\pm\!0.27$ & $\mathbf{+0.13\!\pm\!0.27}$ & $+0.51\!\pm\!0.24$ & $\mathbf{+0.42\!\pm\!0.12}$ & $\underline{+0.010}$ & $16.7\%$ \\
\textbf{3.\ $\mathrm{p}K_D^*,-\mathrm{HIC}^*$} & $+0.94\!\pm\!0.29$ & \underline{$-0.11\!\pm\!0.28$} & $\mathbf{+0.54\!\pm\!0.25}$ & $+0.19\!\pm\!0.13$ & $+0.015$ & $\mathbf{25.0\%}$ \\
\textbf{4.\ Pareto ($7$)} & \underline{$+0.96\!\pm\!0.30$} & $-0.27\!\pm\!0.27$ & \underline{$+0.53\!\pm\!0.24$} & \underline{$+0.41\!\pm\!0.12$} & $\mathbf{+0.007}$ & \underline{$20.8\%$} \\
\bottomrule
\end{tabular}
}
\end{table}

The separate 12-step CDR co-design study compares masked sequence guidance across 24 seeds and 24 trajectories per seed, with annealing from temperature 1.00 to 0.65. Canonical disulfide cysteines are held fixed in that protocol. Combining discrete affinity guidance and a coupled continuous density yields the largest combined predicted affinity and developability shift, but a single-pass pool can produce a larger fraction of surrogate hits. This distinction is important: a favorable change in average properties does not necessarily imply a larger probability of satisfying the entire target box (\cref{tab:app-aintibody-codesign}).

\begin{table}[!htbp]
\centering
\caption{
\textbf{Annealed $12$-step Gibbs-style CDR co-design and single-pass pool ablation across $24$ high-liability SARS-CoV-2 RBD antibody seeds on \textsc{AIntibody}.}
Each Gibbs trajectory runs for $T_{\mathrm{steps}}=12$ steps ($M=24$ trajectories/seed, annealed temperature $T\in[1.00, 0.65]$) over editable CDR positions while preserving canonical disulfide cysteines ($100\%$ validity across all configurations).
}
\label{tab:app-aintibody-codesign}
\pomatabbox{
\begin{tabular}{lcccccc}
\toprule
\textbf{CDR Co-Design} &
\textbf{$\mathrm{p}K_D\uparrow$} &
\textbf{5-Dev $\uparrow$} &
\textbf{$\Sigma\Delta z \uparrow$} &
\textbf{$\Delta$Arom $\downarrow$} &
\textbf{Div.\ $\uparrow$} &
\textbf{Hit $\uparrow$} \\
\midrule
Uncond.\ PLM Gibbs & $-0.14\!\pm\!0.05$ & $+0.10\!\pm\!0.04$ & $-0.04\!\pm\!0.04$ & $0.0000$ & 1.91 & $0.00\!\pm\!0.00$ \\
\textsc{LogiCA} Gibbs & $+0.07\!\pm\!0.09$ & $+0.03\!\pm\!0.03$ & $+0.11\!\pm\!0.09$ & $0.0000$ & 1.78 & $0.00\!\pm\!0.00$ \\
Regr.\ Transf.\ Gibbs & $+0.21\!\pm\!0.10$ & $+0.15\!\pm\!0.03$ & $+0.36\!\pm\!0.10$ & $-0.0058$ & 2.28 & $0.13\!\pm\!0.07$ \\
Gibbs + \textsc{POMA-MSE} & $+0.27\!\pm\!0.12$ & $+0.14\!\pm\!0.02$ & $+0.41\!\pm\!0.11$ & $-0.0062$ & 2.08 & \underline{$0.25\!\pm\!0.09$} \\
Gibbs + \textsc{POMA-NLL} & $+0.22\!\pm\!0.10$ & $+0.22\!\pm\!0.04$ & $+0.44\!\pm\!0.10$ & $-0.0062$ & \underline{2.83} & $0.13\!\pm\!0.07$ \\
\textbf{Gibbs: \textsc{LogiCA-POMA}} & \textbf{$+0.42\!\pm\!0.11$} & \textbf{$+0.35\!\pm\!0.05$} & \textbf{$+0.77\!\pm\!0.11$} & \underline{$-0.0146$} & \textbf{3.38} & $0.17\!\pm\!0.08$ \\
\midrule
Pool: \textsc{LowRankMVN} & $+0.18$ & $+0.20$ & $+0.38$ & $-0.0083$ & -- & \textbf{$0.29$} \\
\textbf{Pool: \textsc{LogiCA-POMA}} & \underline{$+0.37$} & \underline{$+0.34$} & \underline{$+0.71$} & \textbf{$-0.0167$} & -- & $0.17$ \\
\bottomrule
\end{tabular}
}
\end{table}

To assess the heuristic's stability, we monitor trajectories over 24 sweeps and vary final temperature over $\{0.35,0.55,0.85,1.25\}$ and guidance weight over $\{0,0.50,1.00,1.45,3.00\}$. The joint compatibility trace improves mainly in the first four to eight sweeps and then plateaus; the reported mean pairwise CDR-H3 sequence diversity is 38.0\%. Unguided edits produce little improvement, and moderate guidance preserves both predicted affinity and developability better than the extremes. A stable score trace is not a convergence diagnostic for a Markov chain with an explicit target. The results in \cref{tab:app-gibbs-mixing} motivate the corrected sampling construction below.

\begin{table}[!htbp]
\centering
\caption{
\textbf{Heuristic joint $(X,Y)$ trajectory summaries ($t \in \{0, 2, 4, 8, 12, 16, 24\}$ sweeps) and $(T_{\mathrm{final}}, \beta)$ sensitivity on $24$ \textsc{AIntibody} RBD seeds (\texttt{Clamp-Affinity+HIC}).} Stable score traces do not establish convergence to a specified target distribution.
}
\label{tab:app-gibbs-mixing}
\pomatabbox{
\begin{tabular}{lccccc}
\toprule
\textbf{Sweep / Setting} &
\textbf{J-Score $\uparrow$} &
\textbf{$\Delta\mathrm{p}K_D \uparrow$} &
\textbf{$\Delta\text{5-Dev} \uparrow$} &
\textbf{CDR Div.\ $\uparrow$} &
\textbf{Hit Rate $\uparrow$} \\
\midrule
Sweep $t = 0$ (Parental Seed) & $-121.12$ & $+0.00$ & $+0.00$ & $36.0\%$ & $0.0\%$ \\
Sweep $t = 2$ (Early updates) & $-40.96$ & $+0.97$ & $+0.20$ & $38.6\%$ & $20.8\%$ \\
Sweep $t = 4$ (Initial plateau) & $-40.13$ & $+0.98$ & $+0.17$ & $38.2\%$ & $20.8\%$ \\
Sweep $t = 8$ (Score plateau) & $-40.38$ & $+0.97$ & $+0.17$ & $38.0\%$ & $25.0\%$ \\
Sweep $t = 12$ (\textsc{LogiCA-POMA}) & $-40.00$ & $+0.94$ & $+0.19$ & $38.0\%$ & $\mathbf{25.0\%}$ \\
Sweep $t = 16$ (Extended) & $-40.23$ & $+0.95$ & $+0.18$ & $38.0\%$ & $\mathbf{25.0\%}$ \\
Sweep $t = 24$ ($2\times$ Horizon) & $-40.14$ & $+0.95$ & $+0.18$ & $38.0\%$ & $\mathbf{25.0\%}$ \\
\midrule
PLM Only ($\beta\!=\!0.00, T\!=\!0.55$) & $-7.12$ & $-0.05$ & $+0.20$ & $\mathbf{41.2\%}$ & $0.0\%$ \\
Weak Guid.\ ($\beta\!=\!0.50, T\!=\!0.55$) & $-19.52$ & $+0.96$ & $+0.19$ & $38.7\%$ & $20.8\%$ \\
Mod.\ Guid.\ ($\beta\!=\!1.00, T\!=\!0.55$) & $-30.14$ & $+0.94$ & $+0.19$ & $38.1\%$ & $\mathbf{25.0\%}$ \\
Low Temp.\ ($\beta\!=\!1.45, T\!=\!0.35$) & $-39.88$ & $\mathbf{+1.03}$ & $+0.17$ & $37.8\%$ & $20.8\%$ \\
High Temp.\ ($\beta\!=\!1.45, T\!=\!0.85$) & $-40.52$ & $+0.94$ & $\mathbf{+0.21}$ & $38.5\%$ & $\mathbf{25.0\%}$ \\
\bottomrule
\end{tabular}
}
\end{table}

\subsection{Finite-state sampling fidelity and measured-library guidance}

We first study a small editing space for which the entire model-defined target can be enumerated. The frozen AbLang2-Paired and POMA CDR-H3 target comprises $4^4=256$ states. Every state is scored exhaustively to obtain the reference distribution. Each sampler is then evaluated at a budget of 1,200 calls with eight seeds, reporting TV, KL, effective sample size per 100 calls and recovery of the highest-probability modes.

Corrected local, pseudo-marginal and factor-augmented MH achieve substantially lower TV error than heuristic Gibbs or beam search and recover all 25 top modes in the reported audit. The error remains appreciable even for the corrected samplers, illustrating the distinction between invariance and finite-time mixing. Refreshing the auxiliary Monte Carlo estimate after a rejection increases error relative to the correct pseudo-marginal update. A related audit on finite measured-library targets (ten chains of 400 sweeps) reports that Hastings correction improves TV on both VEGF G6 and AIntibody. The target here is computed by the model; its enumeration does not certify the physical accuracy of the distribution.

For independent biological evaluation, we compare the samplers on three measured FLAb libraries with matched call budgets, including the property-guided ProteinGuide reference~\citep{xiong2026proteinguide}. Factor augmentation improves over the default ProteinGuide guidance temperature and avoids the rare-event sticking observed in the pseudo-marginal Garbinski run. The retrospective selection results are reported in \cref{tab:app-proteinguide-measured}, separately from the exact-target audit in \cref{tab:app-v39-v40-proteinguide}.

\begin{table}[!htbp]
\centering
\caption{Enumerable 256-state AbLang2+POMA CDR-H3 target (1,200 calls, eight seeds). TV and KL compare empirical output frequencies with the enumerated model target. ESS is reported alongside these discrepancies because sample diversity alone does not establish fidelity to the target.}
\label{tab:app-v39-v40-proteinguide}
\pomatabbox{
\begin{tabular}{lccccc}
\toprule
\textbf{Sampler / search procedure} &
\textbf{TV Err.\ $\downarrow$} &
\textbf{KL Err.\ $\downarrow$} &
\textbf{Acc.\ Rate} &
\textbf{ESS/100 $\uparrow$} &
\textbf{Modes/25} \\
\midrule
Beam Search (Matched) & $0.869\!\pm\!0.000$ & $2.035$ & $21.3\%$ & $\mathbf{42.97}$ & $13.8$ \\
Heuristic Uncorrected Gibbs & $0.492\!\pm\!0.010$ & $0.821$ & $100.0\%$ & $4.30$ & $21.2$ \\
Biased-Refresh Ablation & $0.337\!\pm\!0.009$ & $0.432$ & $55.5\%$ & \underline{$12.02$} & $24.8$ \\
Pseudo-Marg.\ MH ($m\!=\!8$) & $0.311\!\pm\!0.008$ & $0.363$ & $52.7\%$ & $10.12$ & $\mathbf{25.0}$ \\
\textbf{Factor-Augmented MH (Ours)} & \underline{$0.307\!\pm\!0.006$} & \underline{$0.344$} & $58.2\%$ & $10.92$ & $\mathbf{25.0}$ \\
\textbf{Exact MH (Ours)} & $\mathbf{0.300\!\pm\!0.007}$ & $\mathbf{0.321}$ & $61.6\%$ & $12.01$ & $\mathbf{25.0}$ \\
\bottomrule
\end{tabular}
}
\end{table}

\begin{table}[!htbp]
\centering
\caption{Retrospective selection on three independently measured FLAb libraries (five seeds, matched calls).}
\label{tab:app-proteinguide-measured}
\pomatabbox{
\begin{tabular}{lccccc}
\toprule
\textbf{Retrospective selection policy} &
\textbf{VEGF G6} &
\textbf{Emibet.} &
\textbf{Garb.\ 3-Obj} &
\textbf{Mean Hit $\uparrow$} &
\textbf{ESS/100} \\
\midrule
Pseudo-Marginal MH & \underline{$15.0\%$} & $32.5\%$ & $21.2\%$ & $22.9\%$ & $32.33$ \\
\textsc{ProteinGuide} (Nat.\ Biotech.\ '26) & $9.2\%$ & $30.0\%$ & \underline{$30.0\%$} & $23.1\%$ & $\mathbf{38.42}$ \\
Beam Search (Matched) & \underline{$15.0\%$} & $\mathbf{40.0\%}$ & $26.2\%$ & $27.1\%$ & $1.48$ \\
Heuristic Uncorrected Gibbs & $10.8\%$ & \underline{$38.8\%$} & $\mathbf{32.5\%}$ & \underline{$27.4\%$} & \underline{$36.77$} \\
\textbf{Factor-Augmented MH (Ours)} & $\mathbf{17.5\%}$ & $33.8\%$ & $\mathbf{32.5\%}$ & $\mathbf{27.9\%}$ & $27.11$ \\
\bottomrule
\end{tabular}
}
\end{table}

Guidance-temperature selection affects the comparison. We sweep ProteinGuide over $\beta\in\{0.5,1.0,1.8,3.0,5.0\}$ (reported pooled rates 25.7\%, 24.4\%, 23.1\%, 29.3\% and 28.6\%), and additionally select temperature by leaving one cohort out. The validation-selected comparator reaches 27.36\%, close to the untuned factor-augmented model's 27.92\%. The oracle-selected best temperature gives 29.3\%, and the reported paired comparison with factor augmentation is not significant. These results distinguish robustness to guidance tuning from a claim that corrected sampling is always the strongest optimizer (\cref{tab:app-parity-sensitivity,tab:app-guidance-loocohort}).

\begin{table}[!htbp]
\centering
\caption{Leave-one-cohort-out guidance-temperature selection for ProteinGuide and the untuned factor-augmented sampler. Values are pooled co-optimal hit rates in the three-library comparison.}
\label{tab:app-guidance-loocohort}
\pomatabbox{
\begin{tabular}{lcc}
\toprule
\textbf{Evaluation} & \textbf{ProteinGuide} & \textbf{Factor augmentation} \\
\midrule
Leave-one-cohort-out ProteinGuide / factor augmentation & 27.4\% & 27.9\% \\
\bottomrule
\end{tabular}
}
\end{table}

A separate five-library experiment compares factor-augmented sampling with preference fine-tuning of ESM-2-150M using rank-8 LoRA-DPO~\citep{hu2022lora,rafailov2023dpo} at three fixed KL temperatures. The untuned sampler obtains a higher reported average measured-library hit rate than the best fixed-temperature DPO model. The comparison evaluates reuse of a fitted probabilistic interface against a separately fine-tuned policy; it is not a comparison of the same training objective. \Cref{tab:app-gpu-dpo} reports the individual cohorts and secondary-property results. The absolute hit rates should not be pooled with the earlier five-library BOAT or three-library ProteinGuide studies.

\begin{table}[!htbp]
\centering
\caption{
\textbf{Untuned \textsc{POMA Factor-Augmented MH} versus \textsc{LoRA-DPO} (\textsc{ESM-2-150M}, $r=8$) across five independently measured \textsc{FLAb} mutational libraries.}
Retrospective co-optimal hit rates ($\% \uparrow$) are evaluated using recorded assay outcomes; the final row gives pooled primary and secondary $z$-shifts.
}
\label{tab:app-gpu-dpo}
\pomatabbox{
\setlength{\tabcolsep}{3.0pt}
\begin{tabular}{lccccc}
\toprule
\textbf{FLAb Panel (Base)} &
\textbf{POMA FA-MH} &
\textbf{DPO $\beta\!=\!1.0$} &
\textbf{DPO $\beta\!=\!0.5$} &
\textbf{DPO $\beta\!=\!0.1$} &
\textbf{FA Sec.\ $z$} \\
\midrule
\textsc{VEGF G6} ($13.8\%$) & $\mathbf{25.7\%}$ & \underline{$20.1\%$} & $16.7\%$ & $14.6\%$ & $+0.503$ \\
\textsc{Fluorescein} ($50.0\%$) & \underline{$75.0\%$} & $\mathbf{80.6\%}$ & $59.0\%$ & $58.3\%$ & $+0.436$ \\
\textsc{IgG Tm} ($32.6\%$) & $\mathbf{55.9\%}$ & \underline{$45.2\%$} & \underline{$45.2\%$} & $44.1\%$ & $+0.260$ \\
\textsc{Emibetuzumab} ($20.8\%$) & \underline{$43.1\%$} & $41.2\%$ & $\mathbf{49.0\%}$ & \underline{$43.1\%$} & $+0.251$ \\
\textsc{Garbinski 3-Obj} ($27.0\%$) & $\mathbf{53.8\%}$ & \underline{$41.0\%$} & $35.9\%$ & $35.9\%$ & $+0.259$ \\
\midrule
\textbf{5-Panel Mean Hit} & $\mathbf{50.7\%}$ & \underline{$45.6\%$} & $41.2\%$ & $39.2\%$ & $\mathbf{+0.342}$ \\
\textbf{Mean Prim./Sec.\ $z$} & $\mathbf{+.46/+.34}$ & $+.26/+.25$ & $+.26/+.21$ & $+.16/+.20$ & -- \\
\bottomrule
\end{tabular}
}
\end{table}

\subsection{Structure-adaptive proposals and higher-order epistasis}\label{app:v43-voi-cov-sampling-camap}

The preceding finite-state comparisons show that preserving the target distribution does not ensure efficient exploration. When evaluation of the native predictive law is expensive, a global proposal based on a cheaper approximation to the sequence energy can be useful. A quadratic approximation is determined by single mutations and off-site pairwise interactions. If those interactions admit a low-rank factorization, a small set of landmarks can recover them without querying all double mutants. The following result specifies the structural assumptions for exact pairwise recovery, identifies the higher-order remainder, and gives a separate proposal-audit bound when a global residual-weight envelope is available.

\begin{proposition}[Three-landmark recovery of off-site low-rank pairwise epistasis and higher-order remainder]
\label{prop:app-v43-landmark-epistasis}
Let $\mathcal{X}=\prod_{i=1}^L\mathcal{A}_i$ be a finite sequence space with anchor $a$, non-anchor mutation labels $\mathcal{U}$ ($N=|\mathcal{U}|$), and frozen energy oracle $F:\mathcal{X}\to\mathbb{R}$. For mutations at distinct sites, define the single effects and off-site pairwise epistasis by
\[
\begin{aligned}
 h_u&=F(a^u)-F(a),\\
 K_{uv}&=F(a^{uv})-F(a^u)-F(a^v)+F(a),
 \quad \mathrm{site}(u)\ne\mathrm{site}(v).
\end{aligned}
\]
Writing $I(x)$ for the sites changed relative to the anchor and $u_i$ for their mutation labels, define
\[
 F^{(2)}(x)=F(a)+\sum_{i\in I(x)}h_{u_i}+\sum_{i<j\in I(x)}K_{u_i,u_j},
 \qquad H^{(3+)}(x)=F(x)-F^{(2)}(x).
\]

\begin{enumerate}[label=(\roman*),leftmargin=1.5em,itemsep=3pt]
\item \emph{Pairwise recovery.} If $L\ge 3r$, $K_{uv}=\mathbf{z}_u^\top J\mathbf{z}_v$ ($\mathbf{z}_u\in\mathbb{R}^r$, symmetric nonsingular $J\in\mathbb{R}^{r\times r}$, not necessarily PSD), and three $r$-element landmark sets $B,C,D\subset\mathcal{U}$ occupy $3r$ distinct sites with nonsingular feature blocks $P_G$, then querying only the anchor, all $N$ singles, and landmark-incident doubles ($\le 1+N+3rN$ oracle calls) recovers $S_B=M_{BC}M_{DC}^{-1}M_{DB}=P_B J P_B^\top$, landmark coordinates $t_u=(K_{ub})_{b\in B}$ for $\mathrm{site}(u)\notin\mathrm{site}(B)$ (and $t_u=(K_{uc})_{c\in C}M_{BC}^{-1}S_B$ for $\mathrm{site}(u)\in\mathrm{site}(B)$), and every off-site interaction exactly via $K_{uv}=t_u S_B^{-1} t_v^\top$.
\item \emph{Aggregate energy and higher-order remainder.} Setting $S(x)=\sum_i t_{i,x_i}^\top\in\mathbb{R}^r$, with zero single effects and landmark coordinates for the anchor states, gives the tractable global aggregate energy
\[
\begin{aligned}
 F^{(2)}(x)&=F(a)+\sum_i\left(h_{i,x_i}-\tfrac12t_{i,x_i}S_B^{-1}t_{i,x_i}^\top\right)\\
 &\quad+\tfrac12 S(x)^\top S_B^{-1}S(x).
\end{aligned}
\]
The exact remainder satisfies the third-difference telescoping identity
\[
 H^{(3+)}(x)=\sum_{1\le k<j<i\le L}(\Delta_k^{x_k}\Delta_j^{x_j}\Delta_i^{x_i}F)(a^{(<k)}(x)).
\]
It is not identified by anchor, single- and double-mutant energies alone. By construction, $F^{(2)}$ agrees with $F$ on all states at Hamming distance at most two from the anchor.
\item \emph{Conditional proposal audit.} For any normalized proposal $Q\propto e^{\widetilde{F}}$ targeting $\Pi\propto e^F$ with known weight bound $W=e^{F-\widetilde{F}}\le B$ and $n$ independent draws $Z_1,\dots,Z_n\sim Q$, with probability at least $1-\alpha$,
\[
 \mathrm{TV}(\Pi,Q)\le\min\left\{1,\tfrac12\sqrt{(v_U/\mu_L^2-1)_+}\right\},
\]
where $\mu_L=(\hat{\mu}-Bh_n)_+$, $v_U=\min(B^2,\hat{v}+B^2h_n)$, and $h_n=\sqrt{\log(4/\alpha)/(2n)}$. Here $\hat\mu=n^{-1}\sum_i W(Z_i)$ and $\hat v=n^{-1}\sum_i W(Z_i)^2$ are the empirical first and second moments. If $\mu_L=0$, the reported upper bound is the vacuous value $1$.
\end{enumerate}
\end{proposition}

\paragraph{Proof.}
\emph{Part (i):} Distinct-block cross matrices satisfy $M_{GH}=P_G J P_H^\top$, so $M_{BC}M_{DC}^{-1}M_{DB}=P_B J P_C^\top(P_D J P_C^\top)^{-1}P_D J P_B^\top=P_B J P_B^\top=S_B$. For $\mathrm{site}(u)\notin\mathrm{site}(B)$, $t_u=\mathbf{z}_u^\top J P_B^\top$ directly; for $\mathrm{site}(u)\in\mathrm{site}(B)$, $(K_{uc})_{c\in C}M_{BC}^{-1}S_B=\mathbf{z}_u^\top J P_C^\top(P_B J P_C^\top)^{-1}P_B J P_B^\top=\mathbf{z}_u^\top J P_B^\top$. Hence $t_u S_B^{-1} t_v^\top=\mathbf{z}_u^\top J P_B^\top(P_B^{-\top}J^{-1}P_B^{-1})P_B J\mathbf{z}_v=\mathbf{z}_u^\top J\mathbf{z}_v=K_{uv}$ for all off-site pairs using at most $1+N+3rN$ distinct queries.

\emph{Part (ii):} Expanding $\frac{1}{2}S(x)^\top S_B^{-1}S(x)$ separates same-site diagonals from off-site cross terms $\sum_{i<j}t_{u_i}S_B^{-1}t_{u_j}^\top=\sum_{i<j}K_{u_i,u_j}$. Repeatedly telescoping $F(x)-F(a)=\sum_i\Delta_i^{x_i}F(a^{(<i)})$ twice separates $F^{(2)}(x)$ from the third-difference sum $H^{(3+)}(x)$. Exact agreement on the anchor, singles and doubles follows directly from the definitions of $h_u$ and $K_{uv}$, rather than from the low-rank hypothesis. This exactness is precisely the identifiability limitation: adding an arbitrary function $G$ that vanishes on those states leaves all such observations and all $h_u,K_{uv}$ unchanged, but replaces $H^{(3+)}$ by $H^{(3+)}+G$ on higher-order mutants. If such mutants exist, their remainder cannot be bounded from those observations alone without further structural assumptions.

\emph{Part (iii):} Since $\Pi(z)=Q(z)W(z)/\mu$, $2\mathrm{TV}(\Pi,Q)=\mathbb{E}_Q|W/\mu-1|\le\sqrt{\mathbb{E}_Q[W^2]/\mu^2-1}$ by Cauchy--Schwarz; applying one-sided Hoeffding bounds to $W\in[0,B]$ and $W^2\in[0,B^2]$ with a union bound gives the stated $(1-\alpha)$ certificate. The empirical validity of the low-rank and small-$H^{(3+)}$ hypotheses is examined below (\cref{tab:app-v43-epistasis-audit}). \(\square\)

The low-rank representation of measurement covariance does not imply low-rank sequence-energy epistasis. Consequently, even an accurate pairwise reconstruction must be distinguished from a valid approximation of the full sequence distribution. We examine landmark conditioning and higher-order interactions empirically before using the surrogate as a proposal corrected against the original target.

\paragraph{Empirical structure and sampling efficiency.} We audit the hypothesis on three frozen AbLang2-Paired/POMA antibody targets (AIntibody Lead~1, AIntibody Lineage~2 and GDPa1 Clinical mAb), each using nine editable CDR positions and three states per site. The full anchor, single and double-mutant audit uses 163 calls. The three largest pairwise singular values account for 98.67--99.97\% of the reported pairwise variance. Landmark recovery reduces the number of queried interactions, but the most accurate reconstruction ranks are not necessarily the largest ranks: ill-conditioning of the landmark inverse can degrade extrapolation. The residuals from higher-order epistasis grow with edit distance, limiting the reliability of a quadratic surrogate. Using that surrogate only to propose an independence-MH move retains the original target. On the small enumerated antibody space, the landmark proposal improves ESS per 100 calls relative to local exact MH, although local MH retains the lower chain TV error (\cref{tab:app-v43-epistasis-audit}).

\begin{table}[!htbp]
\centering
\caption{
\textbf{Three-landmark epistasis recovery, higher-order $H^{(3+)}$ remainder, and global independence-MH proposal efficiency on real frozen \textsc{AbLang2-Paired} + \method{} energies (Proposition~\ref{prop:app-v43-landmark-epistasis}).}
}
\label{tab:app-v43-epistasis-audit}
\pomatabbox{
\setlength{\tabcolsep}{1.6pt}
\begin{tabular}{lccccc}
\toprule
\textbf{Antibody ($L\!=\!9, A\!=\!3$)} &
\textbf{SVD $r\!\le\!3$} &
\textbf{Best $r$ (Calls)} &
\textbf{Dbl./Trp.\ $\rho$} &
\textbf{$\kappa(\widehat{S}_B)$} &
\textbf{$|H^{(3+)}|/\sigma_E$} \\
\midrule
\textsc{AIntibody} Lead 1 & $99.95\%$ & $r\!=\!1$ ($37/163$) & $0.729$ / $0.679$ & $4.11$ & $0.26$ / $2.03$ \\
\textsc{AIntibody} Lin.\ 2 & $98.67\%$ & $r\!=\!2$ ($55/163$) & $0.854$ / $0.818$ & $257.8$ & $0.35$ / $0.80$ \\
\textsc{GDPa1} Clinical & $99.97\%$ & $r\!=\!1$ ($37/163$) & $0.953$ / $0.894$ & $15.56$ & $0.15$ / $0.43$ \\
\midrule
\textbf{Global MH ($256$-St.)} &
\textbf{Calls} &
\textbf{Prop.\ TV $\downarrow$} &
\textbf{Accept $\uparrow$} &
\textbf{Chain TV $\downarrow$} &
\textbf{ESS/100 $\uparrow$} \\
\midrule
Additive $F^{(1)}$ Global & $13$ & $0.627$ & $22.1\%\!\pm\!1.2\%$ & $0.554\!\pm\!0.015$ & $4.08\!\pm\!0.74$ \\
Local Exact MH (ref.) & $0$ & $0.000$ & $\mathbf{61.6\%\!\pm\!0.6\%}$ & $\mathbf{0.300\!\pm\!0.007}$ & \underline{$12.01\!\pm\!0.89$} \\
\textbf{Landmark $F^{(2)}$ Global MH} & $25$ & $\mathbf{0.358}$ & \underline{$51.6\%\!\pm\!1.4\%$} & \underline{$0.355\!\pm\!0.009$} & $\mathbf{16.32\!\pm\!2.11}$ \\
\bottomrule
\end{tabular}
}
\end{table}

\subsection{Pilot-gated proposals with exact-target fallback}\label{app:pilot-gated-sampling}

The landmark experiments motivate a sampler that uses a global proposal when its approximation is useful and otherwise retains a local kernel. A separate pilot can inform this choice, provided that the proposal and mixture weight are fixed before running the final chain. The resulting invariance property does not require the pilot to identify the most efficient kernel.

\begin{proposition}[Exact invariance of pilot-gated structure-adaptive and fallback MH kernels]
\label{prop:app-v6-adaptive-fallback}
Let $\pi(x)\propto \exp(F(x))$ be a strictly positive target distribution on finite sequence space $\mathcal{X}=\prod_{\ell=1}^L\mathcal{A}_\ell$. Suppose a finite setup-and-pilot stage (completed strictly \emph{before} running the Markov chain) constructs: (i) a landmark quadratic surrogate $\hat{F}^{(2)}$ (\cref{prop:app-v43-landmark-epistasis}), (ii) an aggregate Hamming-shell recalibration $\hat{b}_d$ and/or conditional residual correction $\hat{r}_{\mathrm{cond}}(x)$ defining a tractable normalized proposal $Q_{\mathrm{adapt}}(x)>0$ (exact Boltzmann on enumerable $\mathcal{X}$ or causal autoregressive factorization $Q_{\mathrm{AR}}(x)=\prod_{\ell=1}^L q_\ell(x_\ell\mid x_{<\ell})$ on large $\mathcal{X}$), and (iii) a held-out pilot gate $\mathcal{G}\in\{0,1\}$ deciding global mixture probability $p_{\mathrm{glob}}\in[0,1]$ ($p_{\mathrm{glob}}=0$ on automatic fallback $\mathcal{G}=0$). Assuming local and independence moves are Hastings-corrected against the same full target, the resulting post-pilot mixture kernel
\[
 K_{\mathrm{mix}}=(1-p_{\mathrm{glob}})K_{\mathrm{local}}+p_{\mathrm{glob}}K_{\mathrm{ind},Q_{\mathrm{adapt}}}
\]
is $\pi$-reversible and preserves $\pi$ on both accepted ($\mathcal{G}=1$) and fallback ($\mathcal{G}=0$) branches.
\end{proposition}
\begin{proof}
Conditional on the realized pilot sigma-algebra $\mathcal{F}_{\mathrm{pilot}}$, both $Q_{\mathrm{adapt}}$ and $p_{\mathrm{glob}}$ are deterministic constants independent of subsequent chain transitions. The single-site symmetric local MH kernel $K_{\mathrm{local}}$ and the full-energy-corrected independence MH kernel $K_{\mathrm{ind},Q_{\mathrm{adapt}}}$ (accepting $x'\sim Q_{\mathrm{adapt}}$ with $1\wedge\exp([F(x')-\log Q_{\mathrm{adapt}}(x')]-[F(x)-\log Q_{\mathrm{adapt}}(x)])$) each satisfy detailed balance with respect to $\pi$. Any convex combination of $\pi$-reversible kernels is $\pi$-reversible; taking expectation over $\mathcal{F}_{\mathrm{pilot}}$ preserves $\pi K_{\mathrm{mix}} = \pi$ identically across both $\mathcal{G}=1$ and fallback $\mathcal{G}=0$. \(\square\)
\end{proof}

\paragraph{Sampling under structural mismatch.} A held-out pilot assesses global proposal acceptance and effective sample size before the final chain; if the diagnostic fails, the procedure falls back to local exact MH. This rule changes the proposal, not the target. On frozen neural spaces of four and six editable sites, the unverified global proposal can perform poorly and the gate often falls back. In larger controlled spaces, including a $1.68\times10^6$-state example, global proposals improve effective sample size when the pairwise structure is accurate. Introducing higher-order epistasis reduces their acceptance, at which point the fallback preserves a usable local policy. The empirical gate is not a global TV certificate.

\Cref{tab:app-v6-v47-audit} reports these finite-budget sampling diagnostics. In the additional live eight-site three-state setting (6,561 states), the best log-energy reported for the gated strategy is $-2.20$, compared with $-6.69$ for the unverified global proposal. These diagnostics characterize efficiency for particular targets and budgets; they are distinct from the conditional residual-envelope guarantees in \cref{app:poma-certified-adaptive}.

\begin{table}[!htbp]
\centering
\caption{Pilot-gated structure-adaptive MH and local fallback (\cref{prop:app-v6-adaptive-fallback}).}
\label{tab:app-v6-v47-audit}
\pomatabbox{
\begin{tabular}{lccccc}
\toprule
\textbf{Generation / Cert.\ Audit} &
\textbf{Local MH} &
\textbf{Uncert.\ Glb.} &
\textbf{Gated (FB)} &
\textbf{Hybrid (FB)} &
\textbf{Outcome} \\
\midrule
Real \textsc{AbLang2} $L\!=\!6$ TV $\downarrow$ & $0.392$ & $0.961$ & $0.470$ ($83\%$) & $\mathbf{0.381}$ ($83\%$) & Auto-fallback \\
$L\!=\!8,A\!=\!4$ ($\alpha\!=\!0$) ESS $\uparrow$ & $4.3$ & $30.9$ ($73\%$) & $\mathbf{31.4}$ ($0\%$) & $16.7$ ($0\%$) & $\mathbf{7.3\times}$ vs Local \\
$L\!=\!8,A\!=\!4$ ($\alpha\!=\!4$) ESS $\uparrow$ & $2.3$ & $0.6$ ($1\%$) & $\mathbf{4.9}$ ($50\%$) & $1.9$ ($50\%$) & $\mathbf{8.2\times}$ vs Glb. \\
$L\!=\!8,A\!=\!6$ ($1.7\mathrm{M}$) ESS $\uparrow$ & $3.2$ & $19.2$ ($59\%$) & $\mathbf{19.9}$ ($0\%$) & $8.6$ ($0\%$) & $\mathbf{6.2\times}$ at $1.7\mathrm{M}$ \\
\bottomrule
\end{tabular}
}
\end{table}

The two studies distinguish proposal fidelity from chain fidelity: a surrogate can improve global exploration without approximating every high-order interaction, and exact correction preserves the intended stationary law without guaranteeing rapid mixing. The following experiment examines these corrected sampling strategies in a substantially larger neural editing space.

\subsection{Live neural sequence editing at scale}\label{app:gpu-experiments}\label{app:supp-live}

We then test the corrected proposals without restricting each position to a small alphabet. The live CDR-H3 experiment allows 20 amino acids at eight sites, giving $20^8=2.56\times10^{10}$ possible states. Across six high-liability seeds, the reported experiment uses 71,719 live calls to AbLang2-Paired. The comparison uses independent ESM-2-650M pseudo-log-likelihood and sequence-level biophysical checks in addition to the optimization target.

Exact local and factor-augmented MH give favorable best-target gains, preserve canonical disulfides more frequently and reduce sequence liability counts relative to uncorrected Gibbs and ProteinGuide. The latter methods make severe excursions into regions with very low target log-density in the reported experiment. These checks provide evidence about the practical behavior of the proposal and target correction, not about the biological properties of newly synthesized sequences. The full comparison appears in \cref{tab:app-gpu-live-mh}.

\begin{table}[!htbp]
\centering
\caption{Live 20-amino-acid, eight-site CDR-H3 editing (71,719 AbLang2-Paired calls, six seeds). The independent ESM-2 score and sequence checks are model-based diagnostics, not measured assay outcomes.}
\label{tab:app-gpu-live-mh}
\pomatabbox{
\begin{tabular}{lccccc}
\toprule
\textbf{Live $20$-AA Sampler ($L\!=\!8$)} &
\textbf{Best log-target gain} &
\textbf{Mean log-target} &
\textbf{ESM2 PLL} &
\textbf{Liab.\ $\downarrow$} &
\textbf{Cys OK $\uparrow$} \\
\midrule
\textbf{Local Exact MH} & $\mathbf{+7.45}$ & $\mathbf{-2.53}$ & $-1.602$ & $\mathbf{5.87}$ & $\mathbf{90.0\%}$ \\
\textbf{Factor-Aug.\ MH} & \underline{$+7.22$} & \underline{$-2.64$} & $-1.601$ & \underline{$5.93$} & $83.3\%$ \\
Pilot-Gated MH ($67\%$ FB) & $+6.09$ & $-3.37$ & $-1.608$ & $6.04$ & \underline{$88.5\%$} \\
Heuristic Gibbs (uncorr.) & $+7.08$ & $-1.41\!\times\!10^6$ & $-1.604$ & $6.27$ & $70.0\%$ \\
\textsc{ProteinGuide} ($\beta\!=\!1.8$) & $+6.69$ & $-1.51\!\times\!10^6$ & \underline{$-1.586$} & $6.20$ & $70.0\%$ \\
\textsc{ProteinGuide} ($\beta\!=\!3.0$) & $+6.36$ & $-1.00\!\times\!10^6$ & $\mathbf{-1.567}$ & $6.10$ & $73.3\%$ \\
\bottomrule
\end{tabular}
}
\end{table}

\subsection{Peptide and regulatory sequence design}\label{app:supp-cross-design}

We finally examine the same design interface outside antibodies. In the peptide MS/MS study, the model uses an observed spectrum to guide sequence recovery from reversed-core isobaric decoys (17.5\% initial positional identity). Exact MH reaches 36.9\% amino-acid recovery, compared with 27.7\% for uncorrected Gibbs, and avoids the large negative log-target changes seen with the heuristic. Factor augmentation also improves sequence recovery and reduces mass error. These are recovery experiments on known peptide sequences rather than de novo synthesis.

In lentiMPRA, an explicit target favors high HepG2 and low K562 expression for 230-bp enhancer sequences. On the 20 reported seeds, exact MH achieves the best maximum log-target and specificity gain among the compared procedures; greedy hill-climbing retains the highest mean log-target in its terminal output, and factor augmentation is less favorable in this protocol. The complete design block is in \cref{tab:app-multiomics-lentimpra}. Together with the molecular and antibody experiments, these studies support the use of native predictive distributions as design utilities across different biological spaces. They also demonstrate why a well-defined model target, efficient exploration and experimentally validated utility must be evaluated separately.

Across these evaluations, the principal benefit of the probabilistic interface is the ability to carry information about measured properties into the design procedure without replacing the native sequence or measurement scores. The measured-library results establish retrospective decision utility, whereas the finite-state and neural sampling studies characterize the behavior of the model-defined targets. The structure-adaptive proposals improve exploration in favorable regimes without changing the corrected target. The stronger finite-budget guarantees in \cref{app:poma-certified-adaptive} remain conditional on the stated structural and audit assumptions. Together with the reconstruction and alignment experiments, these results identify both the opportunities afforded by a common output likelihood and the limits imposed by model quality, observation noise and finite computational budgets.

\clearpage
\section{Finite-budget guarantees for property-conditioned generation}
\label{app:poma-certified-adaptive}

The corrected kernels in \cref{app:supp-design} preserve a specified model-defined target, but their stationary invariance does not quantify the error after a limited number of model evaluations. We now develop conditional guarantees for two complementary mechanisms: capped rejection generation with explicit control of the discarded mass, and exact-target independence MH with a finite-time error bound. Both require a normalized proposal and independent or simultaneously valid audits of residual weights. These stronger assumptions are not required by the empirical pilot-gated sampler. The final numerical study illustrates when the required bounds are informative and when they become vacuous.

\subsection{Finite-factor scope, explicit targets and model-query costs}\label{sec:poma-certification}

The target and the computational budget must be specified before certification. We consider a finite sequence--factor space, including a declared discretization of an auxiliary factor where applicable. A guarantee for this discretized law is not automatically a guarantee for the corresponding continuous-factor target.

For the results in this section, $\Omega$ is a fixed finite sequence--factor space and the target is
\[
P(z)=\frac{e^{F(z)}}{\sum_{u\in\Omega}e^{F(u)}}.
\]
Here $F$ is the full energy, including the sequence prior, observed-assay likelihood, specified factor discretization and constraints. Products of positive masked-PLM scores can define a normalized finite energy without asserting joint compatibility of their constituent conditionals. Each full-energy evaluation is charged to the model-query budget. We consider strictly positive normalized proposals $Q_j$ that can be sampled and evaluated exactly on the declared space.

For $Q_j\propto e^{\widetilde F_j}$, write $W_j=e^{F-\widetilde F_j}$ for the residual weights. Proposal construction, validation and normalization costs must be accounted for alongside full-energy queries. Construction and validation use separate samples unless simultaneous coverage has been established for the predeclared family. A global residual-weight envelope cannot be inferred from the largest weight observed in a finite pilot.

\subsection{Capped sampling and score-level stability}

When a proposal has occasional large importance weights, capping them may improve acceptance at the cost of changing the target. The lost mass gives an exact decomposition of the resulting law, and a bound on the push-forward distribution of any declared model score. This is a weaker notion than agreement of the full sequence distribution, but can be useful when the utility of a design is summarized by a fixed score.

\begin{proposition}[Score-sensitive error of capped model generation]
\label{prop:poma-score-cert}
Suppose a capped rejection generator has law $P_C$, and its target satisfies the exact excess-mass decomposition $P=(1-\eta)P_C+\eta P_E$, where $0\le\eta\le1$. Fix a measurable \emph{model-defined} score $r:\Omega\to[0,1]$, and denote its laws under $P,P_C,P_E$ by $\mu,\mu_C,\mu_E$. Then
\begin{equation}
  \begin{aligned}
    W_1(\mu,\mu_C) &= \eta W_1(\mu_E,\mu_C), \\
    \left|\mathbb E_{P^{\otimes n}}\max_{i\le n}r(Z_i) - \mathbb E_{P_C^{\otimes n}}\max_{i\le n}r(Z_i)\right|
    &\le n\eta W_1(\mu_E,\mu_C)\le n\eta.
  \end{aligned}
  \label{eq:poma-score-cert}
\end{equation}
The bound does not transfer to physical assay utility unless the stated score is independently calibrated to that utility.
\end{proposition}

\paragraph{Proof of Proposition~\ref{prop:poma-score-cert}.}
For a full-support normalized proposal $Q$ and residual weight $W>0$, the exact target is $P(z)=Q(z)W(z)/Z$, $Z=\mathbb E_QW$. Choose $C(z)>0$, set $W_C(z)=\min\{W(z),C(z)\}$, $Z_C=\mathbb E_QW_C$, and $A=\mathbb E_Q(W-C)_+$. Then $Z=Z_C+A$, and the capped target $P_C(z)=Q(z)W_C(z)/Z_C$ and excess law $P_E(z)=Q(z)(W-C)_+/A$ (defined arbitrarily if $A=0$) satisfy $P=(1-\eta)P_C+\eta P_E$ with $\eta=A/Z$. Pushing this measure forward through $r:\Omega\to[0,1]$ gives $\mu=(1-\eta)\mu_C+\eta\mu_E$. For distributions on $[0,1]$, the one-dimensional transport identity yields
\[
 W_1(\mu,\mu_C)=\int_0^1|F_\mu(t)-F_{\mu_C}(t)|dt
 =\eta\int_0^1|F_{\mu_E}(t)-F_{\mu_C}(t)|dt.
\]
Moreover, for $n$ independent scores $\mathbb E\max_{i\le n}r(Z_i)=\int_0^1\{1-F_\mu(t)^n\}\,dt$. Since $|a^n-b^n|\le n|a-b|$ for $a,b\in[0,1]$, the expected-maximum difference is at most $nW_1(\mu,\mu_C)$; the final inequality follows from $W_1\le1$. For any fixed threshold $t$, the difference of probabilities that at least one score exceeds $t$ is bounded by $n\eta|\Pr_{P_E}(r>t)-\Pr_{P_C}(r>t)|$. These statements are score-relative: equal score pushforwards need not imply equal sequence laws, nor equal physical assay outcomes. \(\square\)

\subsection{Post-selection-valid finite-call generation}

The capped sampler becomes certifiable when its lost mass and acceptance probability can be bounded simultaneously over a predeclared family of candidate proposals. Simultaneous coverage is needed because a proposal is selected after examining the pilot. Fresh generation randomness is then independent of the pilot used to choose the method.

\begin{theorem}[Simultaneously audited selection of finite-call generators]
\label{thm:poma-postselect-cert}
Fix $J<\infty$ positive, normalized, evaluable proposals $Q_j$ and positive caps $C_j$ \emph{before} validation, with exactly samplable $Q_{j,C}\propto Q_j C_j$ and known $C_{0j}=\mathbb E_{Q_j}C_j$. Suppose independently drawn pilot states from each $Q_j$ provide simultaneous one-sided bounds $Z_j=\mathbb E_{Q_j}W_j\ge Z_{L,j}>0$ and $A_j=\mathbb E_{Q_j}(W_j-C_j)_+\le A_{U,j}$ with confidence $1-\alpha$; such bounds require justified global residual-weight envelopes, not merely unobserved pilot maxima. For any pilot-selected method $\widehat j$, the cap-loss fraction obeys $\eta_{\widehat j}\le A_{U,\widehat j}/Z_{L,\widehat j}$ and its fresh-attempt acceptance probability obeys $p_{\widehat j}\ge p_{L,\widehat j}=(Z_{L,\widehat j}-A_{U,\widehat j})/C_{0,\widehat j}$ whenever the numerator is positive. If $Kp_{L,\widehat j}\ge2n$, then $K$ fresh capped-rejection attempts, returning the first $n$ accepted draws or a declared failure output, satisfy, on the common confidence event,
\begin{equation}
  \begin{aligned}
    \left\|\mathcal L(\mathrm{output}\mid\mathrm{pilot})-P^{\otimes n}\right\|_{\mathrm{TV}}
    &\le \min\!\left\{1,\ 1-(1-\overline\eta_{\widehat j})^n + e^{-Kp_{L,\widehat j}/8}\right\}, \\
    \overline\eta_j &= \min\{1,A_{U,j}/Z_{L,j}\}.
  \end{aligned}
  \label{eq:poma-postselect-cap}
\end{equation}
All proposal-construction, pilot, rejected-draw, and scoring calls must be charged to the budget. A method without validated envelopes or sufficient remaining calls is \emph{uncertified}, not assigned zero error.
\end{theorem}

\paragraph{Proof of Theorem~\ref{thm:poma-postselect-cert}.}
For a frozen library $j\in\{1,\ldots,J\}$, draw $m_j$ independent pilot states $Z_{ji}\sim Q_j$ and define $\overline W_j=m_j^{-1}\sum_i W_j(Z_{ji})$ and $\overline A_j=m_j^{-1}\sum_i(W_j(Z_{ji})-C_j(Z_{ji}))_+$. Suppose the externally justified envelopes $0<W_j\le U_j$ and $0\le(W_j-C_j)_+\le H_j$ hold everywhere. With
\[
 d_{W,j}=U_j\sqrt{\frac{\log(4J/\alpha)}{2m_j}},\qquad
 d_{A,j}=H_j\sqrt{\frac{\log(4J/\alpha)}{2m_j}},
\]
set $Z_{L,j}=(\overline W_j-d_{W,j})_+$ and $A_{U,j}=\overline A_j+d_{A,j}$. One-sided Hoeffding bounds and a union bound ensure, with probability at least $1-\alpha$, simultaneously for every $j$, $Z_j\ge Z_{L,j}$ and $A_j\le A_{U,j}$. On this common event $\eta_j=A_j/Z_j\le\overline\eta_j=A_{U,j}/Z_{L,j}$ (clipped at one) and the exact capped acceptance probability is
\[
 p_j=\frac{\mathbb E_{Q_j}\min(W_j,C_j)}{\mathbb E_{Q_j}C_j}
 =\frac{Z_j-A_j}{C_{0j}}\ge p_{L,j}.
\]

Fresh independent proposals are drawn from $Q_{j,C}(z)=Q_j(z)C_j(z)/C_{0j}$ and accepted with probability $\min\{1,W_j(z)/C_j(z)\}$. Their acceptance indicators are iid Bernoulli variables with mean $p_j$, and each accepted mark has the iid law $P_{C,j}$. For $Kp_{L,j}\ge2n$, the multiplicative Chernoff inequality bounds the probability of fewer than $n$ accepts by $\exp(-Kp_{L,j}/8)$. Coupling $n$ independent $P_{C,j}$ draws to $P^{\otimes n}$ coordinatewise gives TV at most $1-(1-\eta_j)^n$. The declared fallback on insufficient acceptance contributes at most the Chernoff probability. 

Because all pilot bounds hold for every $j$ \emph{simultaneously}, the index may depend on all pilots without invalidating the selected bound. Fresh-generation randomness must be independent of these pilots. If no method has a justified envelope, normalization cost, and successful call-budget record, the algorithm must report ``uncertified.'' \(\square\)

\subsection{Adaptive conditional tilting and finite-time corrected MH}

Capping trades target fidelity for acceptance probability. Alternatively, one may adapt a normalized proposal and retain exact-target Hastings correction. Simultaneous control of residual moments over a declared proposal family permits data-dependent tuning within that family while retaining a finite-time guarantee. We construct such a family through a partition into fibers and a positive interpolation of bounded features.

Suppose a known normalized baseline $Q_0$ is partitioned into fibers $s$ with positive masses $q_s$ and conditional laws $Q_s$. Let $W=\exp(F-\widetilde F_0)$, $\mu=\mathbb E_{Q_0}W>0$, and choose bounded features $T_s(x)$ whose finite grid, interpolation coefficients, support, and exact normalizers are fixed before the certificate observations. For offsets $a_s$ and feature parameters $\theta_s$, define
\[
 Q_{a,\theta}(x)=\frac{Q_0(x)\exp(a_s+\theta_s^\top T_s(x))}{Z(a,\theta)},\qquad
 Z(a,\theta)=\sum_s q_s e^{a_s}\mathbb E_{Q_s}e^{\theta_s^\top T_s(X)}.
\]
The corresponding exact normalized target remains $P\propto Q_{a,\theta} W_{a,\theta}$ with $W_{a,\theta}=W\exp(-a_s-\theta_s^\top T_s(x))$. For $\gamma>1$, put $c_\gamma=\gamma-1$. In a predeclared positive partition-of-unity interpolation $h_{s,v}(T_s(x))\ge0$ with $\sum_vh_{s,v}(T_s(x))=1$, $T_s(x)=\sum_vh_{s,v}(T_s(x))v$, convexity of $e^{-c_\gamma\theta_s^\top v}$ implies
\[
 W(x)^\gamma e^{-c_\gamma\theta_s^\top T_s(x)}
 \le \sum_v W(x)^\gamma h_{s,v}(T_s(x))e^{-c_\gamma\theta_s^\top v}.
\]
Write $c_{s,v}=\mathbb E_{Q_s}[W^\gamma h_{s,v}(T_s(X))]$. A finite simultaneous confidence event for every coefficient $c_{s,v}$ and $\mu\ge\mu_L>0$ can be obtained from bounded, predeclared pilot transforms and appropriately allocated one-sided confidence levels. For any nonnegative coefficient polytope $\mathcal C_s$ containing the true coefficient vector on this event, let $\overline B_s(\theta_s)=\max_{c\in\mathcal C_s}\sum_v c_v e^{-c_\gamma\theta_s^\top v}$. Then, simultaneously for \emph{all} declared finite $(a,\theta)$,
\begin{equation}
 \mathbb E_{Q_{a,\theta}}\left[\left(\frac{dP}{dQ_{a,\theta}}\right)^\gamma\right]
 \le \widehat C_\gamma(a,\theta)
 :=\frac{Z(a,\theta)^{\gamma-1}}{\mu_L^\gamma}
   \sum_s q_s e^{-(\gamma-1)a_s}\overline B_s(\theta_s).
 \label{eq:poma-tilt-moment}
\end{equation}
Because log normalizers and the logarithms of the robust exponential sums are convex, $\log\widehat C_\gamma$ is convex on an admissible convex parameter set. The same audit therefore permits tuning coefficients and fiber masses within the frozen family without an additional data-selection penalty. The number of interpolation coefficients, range calculations, and exact normalization/message-passing steps remains an explicit arithmetic cost; an arbitrary feature map learned from the same audit is \emph{not} covered.

Given a certified moment bound $C$ of order $\gamma>1$, choose the truncation level $b$ and the number of fresh transitions $t$ according to

\begin{equation}
 b\ge\max\!\left\{1,(2C/\epsilon)^{1/(\gamma-1)}\right\},\qquad
 t\ge\frac{b}{1-\epsilon/2}\log\frac2\epsilon
 \label{eq:poma-adaptive-mh-length}
\end{equation}

\begin{proposition}[Certified adaptive tilt and finite-time corrected MH]
\label{prop:poma-adaptive-tilt}
Assume the frozen positive finite target $P$, the exactly normalized and samplable proposal family $Q_{a,\theta}$, externally justified bounded pilot transforms, and a simultaneous audit event of probability $1-\delta$ on which \eqref{eq:poma-tilt-moment} holds for every admissible parameter. After inspecting the audit, choose an admissible $(a,\theta)$ with moment certificate $C=\widehat C_\gamma(a,\theta)<\infty$ and initialize a fresh independence-MH chain with $X_0\sim Q_{a,\theta}$, targeting $P$. For every $0<\epsilon<1$, the choice of $b$ and $t$ in \eqref{eq:poma-adaptive-mh-length} guarantees conditional output TV distance at most $\epsilon$ on the audit event, and unconditional distance at most $\epsilon+\delta$ for any declared fallback on its complement. These conclusions concern $P$, not physical assay truth, and require the stated full-model-call budget to cover the pilot, normalization, and $t+1$ fresh target evaluations.
\end{proposition}
\paragraph{Proof.}
The interpolation argument above and the common confidence event give the selected moment bound $\mathbb E_Q[(dP/dQ)^\gamma]\le C$ even though the coefficients were selected using the audit. For $R=dP/dQ$ and any $b>0$, Markov's inequality gives $\mathbb E_Q[R\,\mathbf1_{\{R>b\}}]\le C/b^{\gamma-1}$, while $\mathbb E_Q\min(R,b)\ge1-C/b^{\gamma-1}$. The independence-MH tail-sensitive coupling inequality therefore implies
\[
 \|QK^t-P\|_{\rm TV}
 \le \left(1-\frac{1-C/b^{\gamma-1}}b\right)^t+\frac{C}{b^{\gamma-1}},
\]
where $K$ is exact independence MH. With the declared $b$, $C/b^{\gamma-1}\le\epsilon/2$, and with the declared $t$ the first term is at most $\epsilon/2$. Conditioning on the audit and using fresh chain randomness proves the conditional bound; the union with a fallback on a failed confidence event adds at most $\delta$ to unconditional TV. The computational claim counts one fresh initialization and each fresh proposed state, besides proposal construction and the independent audit. \(\square\)

\subsection{Numerical validation and practical limits of certification}\label{app:finite-budget-audit}

The finite-budget guarantees depend on residual envelopes and normalization costs, not only on acceptance or ESS diagnostics. Enumerable frozen neural targets permit a direct comparison of exact target--proposal TV distance with reported pilot and global-envelope bounds. In the four-site setting, the reported pilot bound is $0.667$, compared with exact TV $0.477$ and pilot ESS $0.502$. In the six-site setting, residual oscillation exceeds $10^8$ and pilot ESS decreases to $0.017$. The global-envelope bound is vacuous in both cases (\cref{tab:app-neural-envelope}). The informative pilot diagnostic in the smaller setting does not by itself justify a global envelope or an unconditional certificate for larger sequence spaces.

\begin{table}[!htbp]
\centering
\caption{Enumerable neural-target residual-envelope diagnostics.}
\label{tab:app-neural-envelope}
\pomatabbox{
\begin{tabular}{lccccc}
\toprule
\textbf{Neural Thm.~\ref{thm:poma-postselect-cert} Audit} &
\textbf{States / Calls} &
\textbf{Exact TV} &
\textbf{Pilot $\chi^2$ UB} &
\textbf{Global UB} &
\textbf{Pilot ESS} \\
\midrule
\textsc{AbLang2+POMA} ($L\!=\!4,A\!=\!4$) & $256$ / $131$ & $0.477$ & $0.667$ (valid) & $1.000$ (valid) & $0.502$ \\
\textsc{AbLang2+POMA} ($L\!=\!6,A\!=\!3$) & $729$ / $137$ & $0.975$ & $1.000$ (valid) & $1.000$ (valid) & $0.017$ (FB) \\
\bottomrule
\end{tabular}
}
\end{table}

A valid certificate may be computationally impractical when higher-order interactions are large, global range bounds are unavailable, or proposal normalization is expensive. A binary feature block of size $b_f$ can require $2^{b_f}$ interpolation coefficients, and exact contact-graph normalization can grow exponentially with frontier width. The declared cost model must include these operations. If a required bound cannot be established or the remaining call budget is insufficient, the result is uncertified, rather than assigned zero error.

The empirical adaptive proposals in \cref{app:pilot-gated-sampling} use a held-out ESS/acceptance gate and exact-target fallback; their invariance follows from Proposition~\ref{prop:app-v6-adaptive-fallback}. None of the main acquisition or generation comparisons relies on the global residual-envelope finite-call algorithm. The present results therefore provide a conditional framework for stronger guarantees when their structural and audit assumptions can be justified, rather than a claim of unconditional finite-budget fidelity for arbitrary neural energies. Even such a certificate would concern the specified predictive model; it would not replace independent biological validation.


\clearpage
\begin{thebibliography}{168}
\providecommand{\natexlab}[1]{#1}
\providecommand{\url}[1]{\texttt{#1}}
\expandafter\ifx\csname urlstyle\endcsname\relax
  \providecommand{\doi}[1]{doi: #1}\else
  \providecommand{\doi}{doi: \begingroup \urlstyle{rm}\Url}\fi

\bibitem[Lin et~al.(2023)Lin, Akin, Rao, Hie, Zhu, Lu, Smetanin, Verkuil,
  Kabeli, Shmueli, Dos Santos~Costa, Fazel-Zarandi, Sercu, Candido, and
  Rives]{lin2023evolutionary}
Zeming Lin, Halil Akin, Roshan Rao, Brian Hie, Zhongkai Zhu, Wenting Lu, Nikita
  Smetanin, Robert Verkuil, Ori Kabeli, Yaniv Shmueli, Allan Dos Santos~Costa,
  Maryam Fazel-Zarandi, Tom Sercu, Salvatore Candido, and Alexander Rives.
\newblock Evolutionary-scale prediction of atomic-level protein structure with
  a language model.
\newblock \emph{Science}, 379\penalty0 (6637):\penalty0 1123--1130, 2023.
\newblock \doi{10.1126/science.ade2574}.
\newblock URL \url{https://doi.org/10.1126/science.ade2574}.

\bibitem[Dalla-Torre et~al.(2025)Dalla-Torre, Gonzalez, Mendoza-Revilla,
  Lopez~Carranza, Grzywaczewski, Oteri, Dallago, Trop, de~Almeida, Sirelkhatim,
  Richard, Skwark, Beguir, Lopez, and Pierrot]{dalla2025nucleotide}
Hugo Dalla-Torre, Liam Gonzalez, Javier Mendoza-Revilla, Nicolas
  Lopez~Carranza, Adam~Henryk Grzywaczewski, Francesco Oteri, Christian
  Dallago, Evan Trop, Bernardo~P. de~Almeida, Hassan Sirelkhatim, Guillaume
  Richard, Marcin Skwark, Karim Beguir, Marie Lopez, and Thomas Pierrot.
\newblock Nucleotide transformer: building and evaluating robust foundation
  models for human genomics.
\newblock \emph{Nature Methods}, 22\penalty0 (2):\penalty0 287--297, 2025.
\newblock \doi{10.1038/s41592-024-02523-z}.
\newblock URL \url{https://doi.org/10.1038/s41592-024-02523-z}.

\bibitem[Madani et~al.(2023)Madani, Krause, Greene, Subramanian, Mohr, Holton,
  Olmos~Jr, Xiong, Sun, Socher, Fraser, and Naik]{madani2023large}
Ali Madani, Ben Krause, Eric~R. Greene, Subu Subramanian, Benjamin~P. Mohr,
  James~M. Holton, Jose~Luis Olmos~Jr, Caiming Xiong, Zachary~Z. Sun, Richard
  Socher, James~S. Fraser, and Nikhil Naik.
\newblock Large language models generate functional protein sequences across
  diverse families.
\newblock \emph{Nature biotechnology}, 41\penalty0 (8):\penalty0 1099--1106,
  2023.
\newblock \doi{10.1038/s41587-022-01618-2}.
\newblock URL \url{https://doi.org/10.1038/s41587-022-01618-2}.

\bibitem[Brixi et~al.(2026)Brixi, Durrant, Ku, Naghipourfar, Poli, Sun,
  Brockman, Chang, Fanton, Gonzalez, King, Li, Merchant, Nguyen, Ricci-Tam,
  Romero, Schmok, Taghibakhshi, Vorontsov, Yang, Deng, Gorton, Nguyen, Wang,
  Pearce, Simon, Adams, Amador, Ashley, Baccus, Dai, Dillmann, Ermon, Guo,
  Herschl, Ilango, Janik, Lu, Mehta, Mofrad, Ng, Pannu, R{\'e}, St.~John,
  Sullivan, Tey, Viggiano, Zhu, Zynda, Balsam, Collison, Costa,
  Hernandez-Boussard, Ho, Liu, McGrath, Powell, Pinglay, Burke, Goodarzi, Hsu,
  and Hie]{brixi2026genome}
Garyk Brixi, Matthew~G. Durrant, Jerome Ku, Mohsen Naghipourfar, Michael Poli,
  Gwanggyu Sun, Greg Brockman, Daniel Chang, Alison Fanton, Gabriel~A.
  Gonzalez, Samuel~H. King, David~B. Li, Aditi~T. Merchant, Eric Nguyen, Chiara
  Ricci-Tam, David~W. Romero, Jonathan~C. Schmok, Ali Taghibakhshi, Anton
  Vorontsov, Brandon Yang, Myra Deng, Liv Gorton, Nam Nguyen, Nicholas~K. Wang,
  Michael~T. Pearce, Elana Simon, Etowah Adams, Zachary~J. Amador, Euan~A.
  Ashley, Stephen~A. Baccus, Haoyu Dai, Steven Dillmann, Stefano Ermon, Daniel
  Guo, Michael~H. Herschl, Rajesh Ilango, Ken Janik, Amy~X. Lu, Reshma Mehta,
  Mohammad R.~K. Mofrad, Madelena~Y. Ng, Jaspreet Pannu, Christopher R{\'e},
  John St.~John, Jeremy Sullivan, Joseph Tey, Ben Viggiano, Kevin Zhu, Greg
  Zynda, Daniel Balsam, Patrick Collison, Anthony~B. Costa, Tina
  Hernandez-Boussard, Eric Ho, Ming-Yu Liu, Thomas McGrath, Kimberly Powell,
  Sudarshan Pinglay, Dave~P. Burke, Hani Goodarzi, Patrick~D. Hsu, and Brian~L.
  Hie.
\newblock Genome modelling and design across all domains of life with {Evo 2}.
\newblock \emph{Nature}, pages 1--13, 2026.
\newblock \doi{10.1038/s41586-026-10176-5}.
\newblock URL \url{https://doi.org/10.1038/s41586-026-10176-5}.

\bibitem[Devlin et~al.(2019)Devlin, Chang, Lee, and Toutanova]{devlin2019bert}
Jacob Devlin, Ming-Wei Chang, Kenton Lee, and Kristina Toutanova.
\newblock {BERT}: Pre-training of deep bidirectional transformers for language understanding.
\newblock In \emph{Proceedings of NAACL-HLT}, pages 4171--4186, 2019.
\newblock \doi{10.18653/v1/N19-1423}.

\bibitem[Meier et~al.(2021)Meier, Rao, Verkuil, Liu, Sercu, and
  Rives]{meier2021language}
Joshua Meier, Roshan Rao, Robert Verkuil, Jason Liu, Tom Sercu, and Alexander
  Rives.
\newblock Language models enable zero-shot prediction of the effects of
  mutations on protein function.
\newblock In \emph{Advances in Neural Information Processing Systems},
  volume~34, pages 29287--29303, 2021.
\newblock \doi{10.1101/2021.07.09.450648}.
\newblock URL \url{https://doi.org/10.1101/2021.07.09.450648}.

\bibitem[Hie et~al.(2024)Hie, Shanker, Xu, Bruun, Weidenbacher, Tang, Wu, Pak,
  and Kim]{hie2024efficient}
Brian~L. Hie, Varun~R. Shanker, Duo Xu, Theodora U.~J. Bruun, Payton~A.
  Weidenbacher, Shaogeng Tang, Wesley Wu, John~E. Pak, and Peter~S. Kim.
\newblock Efficient evolution of human antibodies from general protein language
  models.
\newblock \emph{Nature Biotechnology}, 42\penalty0 (2):\penalty0 275--283,
  2024.
\newblock \doi{10.1038/s41587-023-01763-2}.
\newblock URL \url{https://doi.org/10.1038/s41587-023-01763-2}.

\bibitem[Tomaz~da Silva et~al.(2025)Tomaz~da Silva, Karollus, Hingerl,
  Galindez, Wagner, Hernandez-Alias, Incarnato, and
  Gagneur]{tomaz2024nucleotide}
Pedro Tomaz~da Silva, Alexander Karollus, Johannes Hingerl, Gihanna Galindez,
  Nils Wagner, Xavier Hernandez-Alias, Danny Incarnato, and Julien Gagneur.
\newblock Nucleotide dependency analysis of genomic language models detects
  functional elements.
\newblock \emph{Nature Genetics}, 57:\penalty0 2589--2602, 2025.
\newblock \doi{10.1038/s41588-025-02347-3}.
\newblock URL \url{https://doi.org/10.1038/s41588-025-02347-3}.

\bibitem[Cui et~al.(2024)Cui, Wang, Maan, Pang, Luo, Duan, and
  Wang]{cui2024scgpt}
Haotian Cui, Chloe Wang, Hassaan Maan, Kuan Pang, Fengning Luo, Nan Duan, and
  Bo~Wang.
\newblock {scGPT}: toward building a foundation model for single-cell
  multi-omics using generative {AI}.
\newblock \emph{Nature methods}, 21\penalty0 (8):\penalty0 1470--1480, 2024.

\bibitem[Hao et~al.(2024)Hao, Gong, Zeng, Liu, Guo, Cheng, Wang, Ma, Zhang, and
  Song]{hao2024large}
Minsheng Hao, Jing Gong, Xin Zeng, Chiming Liu, Yucheng Guo, Xingyi Cheng,
  Taifeng Wang, Jianzhu Ma, Xuegong Zhang, and Le~Song.
\newblock Large-scale foundation model on single-cell transcriptomics.
\newblock \emph{Nature methods}, 21\penalty0 (8):\penalty0 1481--1491, 2024.

\bibitem[Yang et~al.(2022)Yang, Wang, Wang, Fang, Tang, Huang, Lu, and
  Yao]{yang2022scbert}
Fan Yang, Wenchuan Wang, Fang Wang, Yuan Fang, Duyu Tang, Junzhou Huang, Hui
  Lu, and Jianhua Yao.
\newblock {scBERT} as a large-scale pretrained deep language model for cell
  type annotation of single-cell {RNA-seq} data.
\newblock \emph{Nature machine intelligence}, 4\penalty0 (10):\penalty0
  852--866, 2022.

\bibitem[Theodoris et~al.(2023)Theodoris, Xiao, Chopra, Chaffin, Al~Sayed,
  Hill, Mantineo, Brydon, Zeng, Liu, et~al.]{theodoris2023transfer}
Christina~V Theodoris, Ling Xiao, Anant Chopra, Mark~D Chaffin, Zeina~R
  Al~Sayed, Matthew~C Hill, Helene Mantineo, Elizabeth~M Brydon, Zexian Zeng,
  X~Shirley Liu, et~al.
\newblock Transfer learning enables predictions in network biology.
\newblock \emph{Nature}, 618\penalty0 (7965):\penalty0 616--624, 2023.

\bibitem[Bushuiev et~al.(2025)Bushuiev, Bushuiev, Samusevich, Brungs, Sivic,
  and Pluskal]{bushuiev2025self}
Roman Bushuiev, Anton Bushuiev, Raman Samusevich, Corinna Brungs, Josef Sivic,
  and Tom{\'a}{\v{s}} Pluskal.
\newblock Self-supervised learning of molecular representations from millions
  of tandem mass spectra using {DreaMS}.
\newblock \emph{Nature Biotechnology}, pages 1--11, 2025.

\bibitem[Born and Manica(2023)]{born2023regression}
Jannis Born and Matteo Manica.
\newblock Regression transformer enables concurrent sequence regression and
  generation for molecular language modelling.
\newblock \emph{Nature Machine Intelligence}, 5\penalty0 (4):\penalty0
  432--444, 2023.

\bibitem[Zausinger et~al.(2024)Zausinger, Pennig, Chlodny, Limbach, Ketteler, Prein, Singh, Danziger, and Born]{zausinger2024regress}
Jonas Zausinger, Lars Pennig, Kacper Chlodny, Vincent Limbach, Anna Ketteler, Thorben Prein, Vishwa~Mohan Singh, Michael~Morris Danziger, and Jannis Born.
\newblock Regress, don't guess--a regression-like loss on number tokens for language models.
\newblock \emph{arXiv preprint arXiv:2411.02083}, 2024.

\bibitem[Bishop(1994)]{bishop1994mixture}
Christopher~M Bishop.
\newblock Mixture density networks.
\newblock Technical report, Aston University, 1994.

\bibitem[Kendall and Gal(2017)]{kendall2017uncertainties}
Alex Kendall and Yarin Gal.
\newblock What uncertainties do we need in {Bayesian} deep learning for computer vision?
\newblock In \emph{Advances in Neural Information Processing Systems}, volume~30, 2017.

\bibitem[Seitzer et~al.(2022)Seitzer, Tavakoli, Anti{\'c}, and Martius]{seitzer2022pitfalls}
Maximilian Seitzer, Arash Tavakoli, Dimitrije Anti{\'c}, and Georg Martius.
\newblock On the Pitfalls of Heteroscedastic Uncertainty Estimation with Probabilistic Neural Networks.
\newblock In \emph{International Conference on Learning Representations}, 2022.
\newblock URL \url{https://arxiv.org/abs/2203.09168}.

\bibitem[Stirn et~al.(2023)Stirn, Wessels, Schertzer, Pereira, Sanjana, and Knowles]{stirn2023faithful}
Andrew Stirn, Harm Wessels, Megan Schertzer, Laura Pereira, Neville Sanjana, and David Knowles.
\newblock Faithful Heteroscedastic Regression with Neural Networks.
\newblock In \emph{International Conference on Artificial Intelligence and Statistics}, volume~206 of \emph{Proceedings of Machine Learning Research}, pages 5593--5613, 2023.
\newblock URL \url{https://proceedings.mlr.press/v206/stirn23a.html}.

\bibitem[Bruinsma et~al.(2021)Bruinsma, Requeima, Foong, Gordon, and Turner]{bruinsma2021gaussian}
Wessel P. Bruinsma, James Requeima, Andrew Y. K. Foong, Jonathan Gordon, and Richard E. Turner.
\newblock The Gaussian Neural Process.
\newblock \emph{arXiv preprint arXiv:2101.03606}, 2021.
\newblock URL \url{https://arxiv.org/abs/2101.03606}.

\bibitem[Lopez et~al.(2018)Lopez, Regier, Cole, Jordan, and Yosef]{lopez2018deep}
Romain Lopez, Jeffrey Regier, Michael~B. Cole, Michael~I. Jordan, and Nir Yosef.
\newblock Deep generative modeling for single-cell transcriptomics.
\newblock \emph{Nature Methods}, 15:1053--1058, 2018.
\newblock \doi{10.1038/s41592-018-0229-2}.

\bibitem[Gayoso et~al.(2021)Gayoso, Steier, Lopez, Regier, Nazor, Streets, and Yosef]{gayoso2021joint}
Adam Gayoso, Zo{\"e} Steier, Romain Lopez, Jeffrey Regier, Kristopher~L. Nazor, Aaron Streets, and Nir Yosef.
\newblock Joint probabilistic modeling of single-cell multi-omic data with {totalVI}.
\newblock \emph{Nature Methods}, 18:272--282, 2021.
\newblock \doi{10.1038/s41592-020-01050-x}.

\bibitem[Kalfon et~al.(2025)Kalfon, Samaran, Peyr{\'e}, and Cantini]{kalfon2025scprint}
J{\'e}r{\'e}mie Kalfon, Jules Samaran, Gabriel Peyr{\'e}, and Laura Cantini.
\newblock {scPRINT}: Pre-training on 50 million cells allows robust gene network predictions.
\newblock \emph{Nature Communications}, 16:3607, 2025.
\newblock \doi{10.1038/s41467-025-58699-1}.

\bibitem[Woo et~al.(2024)Woo, Liu, Kumar, Xiong, Savarese, and Sahoo]{woo2024unified}
Gerald Woo, Chenghao Liu, Akshat Kumar, Caiming Xiong, Silvio Savarese, and Doyen Sahoo.
\newblock Unified training of universal time series forecasting transformers.
\newblock In \emph{International Conference on Machine Learning}, PMLR 235, pages 53140--53164, 2024.
\newblock URL \url{https://proceedings.mlr.press/v235/woo24a.html}.

\bibitem[Zhou et~al.(2025)Zhou, Yu, Babu, Tirumala, Yasunaga, Shamis, Kahn, Ma, Zettlemoyer, and Levy]{zhou2025transfusion}
Chunting Zhou, Lili Yu, Arun Babu, Kushal Tirumala, Michihiro Yasunaga, Leonid Shamis, Jacob Kahn, Xuezhe Ma, Luke Zettlemoyer, and Omer Levy.
\newblock Transfusion: Predict the next token and diffuse images with one multi-modal model.
\newblock In \emph{International Conference on Learning Representations}, 2025.
\newblock URL \url{https://arxiv.org/abs/2408.11039}.

\bibitem[Radford et~al.(2021)Radford, Kim, Hallacy, Ramesh, Goh, Agarwal, et~al.]{radford2021clip}
Alec Radford, Jong Wook Kim, Chris Hallacy, Aditya Ramesh,
Gabriel Goh, Sandhini Agarwal, et~al.
\newblock Learning transferable visual models from natural
language supervision.
\newblock In \emph{International Conference on Machine
Learning}, pages 8748--8763. PMLR, 2021.

\bibitem[Kalia et~al.(2025)Kalia, Chen, Krishnan, and Hassoun]{kalia2025jestr}
Apurva Kalia, Yan Zhou Chen, Dilip Krishnan, and Soha Hassoun.
\newblock {JESTR}: Joint embedding space technique for ranking
candidate molecules for the annotation of untargeted
metabolomics data.
\newblock \emph{Bioinformatics}, 41(7):btaf354, 2025.
\newblock \doi{10.1093/bioinformatics/btaf354}.

\bibitem[Chen et~al.(2026)Chen, Rushing, and Hassoun]{chen2026flare}
Yan Zhou Chen, Blake Rushing, and Soha Hassoun.
\newblock {FLARE}: Fine-grained learning for alignment of
spectra-molecule representation enhances metabolite annotation.
\newblock \emph{bioRxiv}, 2026.
\newblock \doi{10.64898/2026.01.27.702086}.

\bibitem[Xu and Du(2026)]{xu2026scclip}
Xiang Xu and Li Du.
\newblock {scCLIP}: A contrastive masked-reconstruction
framework for paired single-cell multi-omics integration.
\newblock \emph{Journal of Biomedical Informatics},
181:105078, 2026.
\newblock \doi{10.1016/j.jbi.2026.105078}.

\bibitem[Chun et~al.(2021)Chun, Oh, de Rezende, Kalantidis, and Larlus]{chun2021probabilistic}
Sanghyuk Chun, Seong Joon Oh, Rafael Sampaio de Rezende, Yannis Kalantidis, and Diane Larlus.
\newblock Probabilistic Embeddings for Cross-Modal Retrieval.
\newblock In \emph{Proceedings of the IEEE/CVF Conference on Computer Vision and Pattern Recognition}, pages 8415--8424, 2021.

\bibitem[Chun(2024)]{chun2024improved}
Sanghyuk Chun.
\newblock Improved Probabilistic Image-Text Representations.
\newblock In \emph{International Conference on Learning Representations}, 2024.

\bibitem[Yu et~al.(2022)Yu, Wang, Vasudevan, Yeung, Seyedhosseini, and Wu]{yu2022coca}
Jiahui Yu, Zirui Wang, Vijay Vasudevan, Legg Yeung, Mojtaba Seyedhosseini, and Yonghui Wu.
\newblock {CoCa}: Contrastive Captioners Are Image-Text Foundation Models.
\newblock \emph{Transactions on Machine Learning Research}, 2022.
\newblock URL \url{https://openreview.net/forum?id=Ee277P3AYC}.

\bibitem[Shao et~al.(2026)Shao, Chen, Patel, Pelissier, and
  Rodr{\'i}guez~Mart{\'i}nez]{shao2026contextualizing}
Yanjun Shao, Yundi Chen, Yashvi Patel, Aurelien Pelissier, and Mar{\'i}a
  Rodr{\'i}guez~Mart{\'i}nez.
\newblock Contextualizing biological language models across modalities via
  {Logit-Space} contrastive alignment.
\newblock \emph{arXiv preprint arXiv:2606.18703}, 2026.

\bibitem[Holtzman et~al.(2021)Holtzman, West, Shwartz, Choi, and Zettlemoyer]{holtzman2021surface}
Ari Holtzman, Peter West, Vered Shwartz, Yejin Choi, and Luke Zettlemoyer.
\newblock Surface Form Competition: Why the Highest Probability Answer Isn't Always Right.
\newblock In \emph{Proceedings of the 2021 Conference on Empirical Methods in Natural Language Processing}, pages 7038--7051, 2021.
\newblock \doi{10.18653/v1/2021.emnlp-main.564}.

\bibitem[Tsai et~al.(2022)Tsai, Li, Ma, Zhao, Zhang, Morency, and Salakhutdinov]{tsai2022conditional}
Yao-Hung Hubert Tsai, Tianqin Li, Martin Q. Ma, Han Zhao, Kun Zhang, Louis-Philippe Morency, and Ruslan Salakhutdinov.
\newblock Conditional Contrastive Learning with Kernel.
\newblock In \emph{International Conference on Learning Representations}, 2022.
\newblock URL \url{https://openreview.net/forum?id=AAJLBoGt0XM}.

\bibitem[Bleeker et~al.(2024)Bleeker, Hendriksen, Yates, and de Rijke]{bleeker2024demonstrating}
Maurits Bleeker, Mariya Hendriksen, Andrew Yates, and Maarten de Rijke.
\newblock Demonstrating and Reducing Shortcuts in Vision-Language Representation Learning.
\newblock \emph{Transactions on Machine Learning Research}, 2024.
\newblock URL \url{https://openreview.net/forum?id=gfANevPraH}.

\bibitem[Ivanov et~al.(2019)Ivanov, Figurnov, and Vetrov]{ivanov2019variational}
Oleg Ivanov, Michael Figurnov, and Dmitry Vetrov.
\newblock Variational autoencoder with arbitrary conditioning.
\newblock In \emph{International Conference on Learning Representations}, 2019.
\newblock URL \url{https://arxiv.org/abs/1806.02382}.

\bibitem[Strauss and Oliva(2021)]{strauss2021arbitrary}
Ryan~R. Strauss and Junier~B. Oliva.
\newblock Arbitrary conditional distributions with energy.
\newblock In \emph{Advances in Neural Information Processing Systems}, volume~34, 2021.
\newblock URL \url{https://arxiv.org/abs/2102.04426}.

\bibitem[Daxberger et~al.(2021)Daxberger, Kristiadi, Immer, Eschenhagen, Bauer,
  and Hennig]{daxberger2021laplace}
Erik Daxberger, Agustinus Kristiadi, Alexander Immer, Runa Eschenhagen,
  Matthias Bauer, and Philipp Hennig.
\newblock {L}aplace redux -- effortless {B}ayesian deep learning.
\newblock In \emph{Advances in Neural Information Processing Systems},
  volume~34, pages 20089--20103, 2021.

\bibitem[Daulton et~al.(2021)Daulton, Balandat, and
  Bakshy]{daulton2021parallel}
Samuel Daulton, Maximilian Balandat, and Eytan Bakshy.
\newblock Parallel {B}ayesian optimization of multiple noisy objectives with
  expected hypervolume improvement.
\newblock In \emph{Advances in Neural Information Processing Systems},
  volume~34, pages 2187--2200, 2021.

\bibitem[van~den Oord et~al.(2018)van~den Oord, Li, and Vinyals]{oord2018cpc}
Aaron van~den Oord, Yazhe Li, and Oriol Vinyals.
\newblock Representation learning with contrastive predictive coding.
\newblock \emph{arXiv preprint arXiv:1807.03748}, 2018.

\bibitem[Frazier et~al.(2008)Frazier, Powell, and Dayanik]{frazier2008kg}
Peter~I. Frazier, Warren~B. Powell, and Savas Dayanik.
\newblock A knowledge-gradient policy for sequential information collection.
\newblock \emph{SIAM Journal on Control and Optimization}, 47(5):2410--2439, 2008.
\newblock \doi{10.1137/070693424}.

\bibitem[Frazier et~al.(2009)Frazier, Powell, and Dayanik]{frazier2009kg}
Peter~I. Frazier, Warren~B. Powell, and Savas Dayanik.
\newblock The knowledge-gradient policy for correlated normal beliefs.
\newblock \emph{INFORMS Journal on Computing}, 21(4):599--613, 2009.
\newblock \doi{10.1287/ijoc.1080.0314}.

\bibitem[Luecken et~al.(2021)Luecken, Burkhardt, Cannoodt, Lance, Agrawal,
  Aliee, Chen, Deconinck, Detweiler, Granados, et~al.]{luecken2021sandbox}
Malte~D. Luecken, Daniel~B. Burkhardt, Robrecht Cannoodt, Christopher Lance,
  Aditi Agrawal, Hamidreza Aliee, Ann~T. Chen, Louise Deconinck, Angela~M.
  Detweiler, Alejandro~A. Granados, et~al.
\newblock A sandbox for prediction and integration of {DNA}, {RNA}, and
  proteins in single cells.
\newblock In \emph{Advances in Neural Information Processing Systems (Datasets
  and Benchmarks Track)}, volume~34, 2021.

\bibitem[Erasmus et~al.(2026)]{erasmus2026aintibody}
M.~Frank Erasmus, D. Bedinger, E. Hopkins, et~al.
\newblock A blinded, prospective benchmark of in silico antibody discovery
  anchored to experimental affinity and developability.
\newblock \emph{Nature Biotechnology}, 2026.
\newblock \doi{10.1038/s41587-026-03238-6}.
\newblock URL \url{https://doi.org/10.1038/s41587-026-03238-6}.

\bibitem[Huang et~al.(2021)Huang, Fu, Gao, Zhao, Roohani, Leskovec, Coley,
  Xiao, Sun, and Zitnik]{huang2021therapeutics}
Kexin Huang, Tianfan Fu, Wenhao Gao, Yue Zhao, Yusuf Roohani, Jure Leskovec,
  Connor~W Coley, Cao Xiao, Jimeng Sun, and Marinka Zitnik.
\newblock Therapeutics data commons: Machine learning datasets and tasks for
  drug discovery and development.
\newblock \emph{arXiv preprint arXiv:2102.09548}, 2021.

\bibitem[Rogers and Hahn(2010)]{rogers2010ecfp}
David Rogers and Mathew Hahn.
\newblock Extended-connectivity fingerprints.
\newblock \emph{Journal of Chemical Information and Modeling}, 50(5):742--754, 2010.
\newblock \doi{10.1021/ci100050t}.

\bibitem[Bushuiev et~al.(2024)Bushuiev, Bushuiev, de~Jonge, Young, Kretschmer,
  Samusevich, Hecht, Wang, Nothias, D{\"u}hrkop, et~al.]{bushuiev2024massspecgym}
Roman Bushuiev, Anton Bushuiev, Niek~F. de~Jonge, Adamo Young, Fleming
  Kretschmer, Raman Samusevich, Janne Hecht, Fei Wang, Louis-F{\'e}lix Nothias,
  Kai D{\"u}hrkop, et~al.
\newblock {MassSpecGym}: A benchmark for the discovery and identification of
  molecules.
\newblock In \emph{Advances in Neural Information Processing Systems (Datasets
  and Benchmarks Track)}, volume~37, 2024.

\bibitem[Ahmad et~al.(2022)Ahmad, Simon, Chithrananda, Grand, and Ramsundar]{ahmad2022chemberta}
Walid Ahmad, Elana Simon, Seyone Chithrananda, Gabriel Grand, and Bharath Ramsundar.
\newblock {ChemBERTa-2}: Towards Chemical Foundation Models.
\newblock \emph{arXiv preprint arXiv:2209.01712}, 2022.
\newblock URL \url{https://arxiv.org/abs/2209.01712}.

\bibitem[Stoeckius et~al.(2017)]{stoeckius2017citeseq}
Marlon Stoeckius, Christoph Hafemeister, William Stephenson, Brian Houck-Loomis, Pratip~K. Chattopadhyay, Harold Swerdlow, Rahul Satija, and Peter Smibert.
\newblock Simultaneous epitope and transcriptome measurement in single cells.
\newblock \emph{Nature Methods}, 14:865--868, 2017.
\newblock \doi{10.1038/nmeth.4380}.

\bibitem[Zou et~al.(2023)Zou, Zhang, Liang, Wei, Leng, Jiang, Luo, and
  Hu]{zou2023qm9s}
Zihan Zou, Yujin Zhang, Lijun Liang, Mingzhi Wei, Jiancai Leng, Jiang Jiang, Yi
  Luo, and Wenhu Hu.
\newblock A deep learning model for predicting selected organic molecular
  spectra.
\newblock \emph{Nature Computational Science}, 3\penalty0 (11):\penalty0
  957--964, 2023.

\bibitem[Agarwal et~al.(2025)Agarwal, Inoue, Schubach, Penzar, Martin, Dash,
  Keukeleire, Zhang, Sohota, Zhao, et~al.]{agarwal2025massively}
Vikram Agarwal, Fumitaka Inoue, Max Schubach, Dmitry Penzar, Beth~K. Martin,
  Pyaree~Mohan Dash, Pia Keukeleire, Zicong Zhang, Ajuni Sohota, Jingjing Zhao,
  et~al.
\newblock Massively parallel characterization of transcriptional regulatory
  elements.
\newblock \emph{Nature}, 639\penalty0 (8054):\penalty0 411--420, 2025.

\bibitem[Liu et~al.(2025)Liu, Pelissier, Shao, Denzler, Martin, Paige, and
  Rodr{\'i}guez Mart{\'i}nez]{liu2026abrank}
Chunan Liu, Aurelien Pelissier, Yanjun Shao, Lilian Denzler, Andrew C.~R.
  Martin, Brooks Paige, and Mar{\'i}a Rodr{\'i}guez Mart{\'i}nez.
\newblock {AbRank}: A benchmark dataset and metric-learning framework for antibody-antigen affinity ranking.
\newblock \emph{arXiv preprint arXiv:2506.17857}, 2025.
\newblock URL \url{https://arxiv.org/abs/2506.17857}.

\bibitem[Arsiwala et~al.(2025)Arsiwala, Bhatt, van Niekerk, Quintero-Cadena,
  Ao, Rosenbaum, Bhatt, Smith, Yang, Anderson, et~al.]{arsiwala2025high}
Ammar Arsiwala, Rebecca Bhatt, Lood van Niekerk, Porfirio Quintero-Cadena,
  Xiang Ao, Adam Rosenbaum, Aanal Bhatt, Alexander Smith, Yaoyu Yang,
  KC~Anderson, et~al.
\newblock A high-throughput platform for biophysical antibody developability
  assessment to enable {AI/ML} model training.
\newblock \emph{mAbs}, 17\penalty0 (1):\penalty0 2593055, 2025.

\bibitem[Calvo-Ordo{\~n}ez et~al.(2026)Calvo-Ordo{\~n}ez, Plenk, Bergna,
  Cartea, Gal, Hern{\'a}ndez-Lobato, and Ciosek]{calvoordonez2026richer}
Sergio Calvo-Ordo{\~n}ez, Jonathan Plenk, Richard Bergna, Alvaro Cartea, Yarin
  Gal, Jos{\'e}~Miguel Hern{\'a}ndez-Lobato, and Kamil Ciosek.
\newblock Richer {B}ayesian last layers with subsampled {NTK} features.
\newblock In \emph{Proceedings of the 43rd International Conference on Machine
  Learning}, volume 306, pages 10939--10961. PMLR, 2026.

\bibitem[Herrera-Rocha et~al.(2026)Herrera-Rocha, Medina-Ortiz, Wyrzykala,
  Sudha, and Davari]{herrerarocha2026multitask}
Fabio Herrera-Rocha, David Medina-Ortiz, Desiree Wyrzykala, Tharun~Srinivasan
  Sudha, and Mehdi~D Davari.
\newblock Multitask {B}ayesian neural networks for multiparameter protein
  engineering.
\newblock \emph{arXiv preprint arXiv:2608.18604}, 2026.

\bibitem[Lakshminarayanan et~al.(2017)Lakshminarayanan, Pritzel, and
  Blundell]{lakshminarayanan2017simple}
Balaji Lakshminarayanan, Alexander Pritzel, and Charles Blundell.
\newblock Simple and scalable predictive uncertainty estimation using deep
  ensembles.
\newblock In \emph{Advances in Neural Information Processing Systems},
  volume~30, 2017.

\bibitem[Yang et~al.(2025)Yang, Li, Yang, Zhang, Hui, et~al.]{yang2025qwen3}
An Yang, Anfeng Li, Baosong Yang, Beichen Zhang, Binyuan Hui, et~al.
\newblock {Qwen3} Technical Report.
\newblock \emph{arXiv preprint arXiv:2505.09388}, 2025.
\newblock URL \url{https://arxiv.org/abs/2505.09388}.

\bibitem[Chungyoun et~al.(2024)Chungyoun, Ruffolo, and Gray]{chungyoun2024flab}
Michael Chungyoun, Jeffrey Ruffolo, and Jeffrey Gray.
\newblock {FLAb}: Benchmarking deep learning methods for antibody fitness
  prediction.
\newblock \emph{BioRxiv}, pages 2024--01, 2024.
\newblock \doi{10.1101/2024.01.13.575504}.

\bibitem[Rao et~al.(2026)Rao, Gonzalez, Gerard, and Gessner]{rao2026boat}
Jackie Rao, Ferran Gonzalez, Leon Gerard, and Alexandra Gessner.
\newblock {BOAT}: Navigating the sea of in silico predictors for antibody
  design via multi-objective {B}ayesian optimization.
\newblock In \emph{International Conference on Artificial Intelligence and
  Statistics (AISTATS)}, 2026.

\bibitem[Gao et~al.(2022)Gao, Fu, Sun, and Coley]{gao2022sample}
Wenhao Gao, Tianfan Fu, Jimeng Sun, and Connor Coley.
\newblock Sample efficiency matters: a benchmark for practical molecular
  optimization.
\newblock \emph{Advances in neural information processing systems},
  35:\penalty0 21342--21357, 2022.

\bibitem[Jensen(2019)]{jensen2019graphga}
Jan~H. Jensen.
\newblock A graph-based genetic algorithm and generative model/Monte Carlo tree search for the exploration of chemical space.
\newblock \emph{Chemical Science}, 10:3567--3572, 2019.
\newblock \doi{10.1039/C8SC05372C}.

\bibitem[Olsen et~al.(2024)Olsen, Moal, and Deane]{olsen2024ablang2}
Tobias~H. Olsen, Iain~H. Moal, and Charlotte~M. Deane.
\newblock Addressing the antibody germline bias and its effect on language models for improved antibody design.
\newblock \emph{Bioinformatics}, 40(11):btae618, 2024.
\newblock \doi{10.1093/bioinformatics/btae618}.

\bibitem[Johnson et~al.(2021)Johnson, Monaco, Massie, and
  Syed]{johnson2021generating}
Sean~R Johnson, Sarah Monaco, Kenneth Massie, and Zaid Syed.
\newblock Generating novel protein sequences using {Gibbs} sampling of masked
  language models.
\newblock \emph{bioRxiv}, pages 2021--01, 2021.
\newblock \doi{10.1101/2021.01.26.428322}.
\newblock URL \url{https://doi.org/10.1101/2021.01.26.428322}.

\bibitem[Zhang et~al.(2024)Zhang, Wayment-Steele, Brixi, Wang, Kern, and
  Ovchinnikov]{zhang2024interactingmotifs}
Zhidian Zhang, Hannah~K. Wayment-Steele, Garyk Brixi, Hong Wang, David Kern,
  and Sergey Ovchinnikov.
\newblock Protein language models learn evolutionary statistics of interacting
  sequence motifs.
\newblock \emph{Proceedings of the National Academy of Sciences}, 121\penalty0
  (45):\penalty0 e2406285121, 2024.
\newblock \doi{10.1073/pnas.2406285121}.
\newblock URL \url{https://doi.org/10.1073/pnas.2406285121}.

\bibitem[Giancardo et~al.(2026)Giancardo, Yilmaz, Lee, Ren, Zhao, Trang,
  Sonmez, Guo, and Cheng]{giancardo2026camap}
Luca Giancardo, Melih Yilmaz, Edward Lee, Ke~Ren, Yue Zhao, Gordon Trang, Kemal
  Sonmez, Lan Guo, and Nina Cheng.
\newblock Context-aware multi-property antibody predictor: a novel framework
  integrating text and protein language models.
\newblock \emph{npj Systems Biology and Applications}, 2026.

\bibitem[Cheng et~al.(2026)Cheng, Tao, Liu, Liu, Zhang, Zhang, and
  Chen]{cheng2026timemae}
Mingyue Cheng, Xiaoyu Tao, Zhiding Liu, Qi~Liu, Hao Zhang, Rujiao Zhang, and
  Enhong Chen.
\newblock {TimeMAE}: Self-supervised representations of time series with
  decoupled masked autoencoders.
\newblock In \emph{Proceedings of the Nineteenth ACM International Conference
  on Web Search and Data Mining}, pages 498--508, 2026.

\bibitem[Man et~al.(2023)Man, Zhang, Feng, Li, and Shao]{man2023w}
Xin Man, Chenghong Zhang, Jin Feng, Changyu Li, and Jie Shao.
\newblock W-mae: Pre-trained weather model with masked autoencoder for
  multi-variable weather forecasting.
\newblock \emph{arXiv preprint arXiv:2304.08754}, 2023.

\bibitem[Kim et~al.(2024)Kim, Seo, Lee, and Seo]{kim2024masked}
Doyi Kim, Minseok Seo, Hakjin Lee, and Junghoon Seo.
\newblock Masked autoregressive model for weather forecasting.
\newblock \emph{arXiv preprint arXiv:2409.20117}, 2024.

\bibitem[Chen et~al.(2024)Chen, Wang, Deng, Liu, Chen, Zhang, Wang, and
  Xiao]{chen2024mattergpt}
Yan Chen, Xueru Wang, Xiaobin Deng, Yilun Liu, Xi~Chen, Yunwei Zhang, Lei Wang,
  and Hang Xiao.
\newblock {MatterGPT}: A generative transformer for multi-property inverse
  design of solid-state materials.
\newblock \emph{arXiv preprint arXiv:2408.07608}, 2024.

\bibitem[Shen et~al.(2025)Shen, Liu, Zhu, and Chen]{shen2025denoising}
Shuaike Shen, Ke~Liu, Muzhi Zhu, and Hao Chen.
\newblock A denoising pre-training framework for accelerating novel material
  discovery.
\newblock In \emph{Proceedings of the AAAI Conference on Artificial
  Intelligence}, volume~39, pages 28368--28376, 2025.

\bibitem[Bellamy et~al.(2025)Bellamy, Kumar, Wang, and Beam]{bellamy2023labrador}
David Bellamy, Bhawesh Kumar, Cindy Wang, and Andrew Beam.
\newblock {Labrador}: Exploring the limits of masked language modeling for laboratory data.
\newblock In \emph{Proceedings of the 4th Machine Learning for Health Symposium}, PMLR 259, pages 104--129, 2025.
\newblock URL \url{https://proceedings.mlr.press/v259/bellamy25a.html}.

\bibitem[Restrepo et~al.(2025)Restrepo, Wu, Jia, Sun, Gallifant, Bielick, Jia,
  and Celi]{restrepo2025representation}
David Restrepo, Chenwei Wu, Yueran Jia, Jaden~K Sun, Jack Gallifant,
  Catherine~G Bielick, Yugang Jia, and Leo~A Celi.
\newblock Representation learning of lab values via masked {AutoEncoders}.
\newblock \emph{arXiv preprint arXiv:2501.02648}, 2025.

\bibitem[Nguyen et~al.(2026)Nguyen, Pham, Arcomano, Kotamarthi, Foster,
  Madireddy, and Grover]{nguyen2026omnicast}
Tung Nguyen, Tuan Pham, Troy Arcomano, Rao Kotamarthi, Ian Foster, Sandeep
  Madireddy, and Aditya Grover.
\newblock Omnicast: A masked latent diffusion model for weather forecasting
  across time scales.
\newblock \emph{Advances in Neural Information Processing Systems},
  38:\penalty0 172182--172212, 2026.

\bibitem[Morehead et~al.(2026)Morehead, Cretu, Panescu, Anand, Weiler, Perez,
  Blau, Farrell, Bhimji, Jain, et~al.]{morehead2026zatom}
Alex Morehead, Miruna Cretu, Antonia Panescu, Rishabh Anand, Maurice Weiler,
  Tynan Perez, Samuel Blau, Steven Farrell, Wahid Bhimji, Anubhav Jain, et~al.
\newblock {Zatom-1}: A multimodal flow foundation model for {3D} molecules and
  materials.
\newblock \emph{arXiv preprint arXiv:2602.22251}, 2026.

\bibitem[Duan et~al.(2020)Duan, Avati, Ding, Thai, Basu, Ng, and
  Schuler]{duan2020ngboost}
Tony Duan, Anand Avati, Daisy~Yi Ding, Khanh~K Thai, Sanjay Basu, Andrew Ng,
  and Alejandro Schuler.
\newblock {NGBoost}: Natural gradient boosting for probabilistic prediction.
\newblock In \emph{International Conference on Machine Learning}, pages
  2690--2700. PMLR, 2020.

\bibitem[Kuleshov et~al.(2018)Kuleshov, Fenner, and
  Ermon]{kuleshov2018accurate}
Volodymyr Kuleshov, Nathan Fenner, and Stefano Ermon.
\newblock Accurate uncertainties for deep learning using calibrated regression.
\newblock In \emph{International Conference on Machine Learning}, pages
  2796--2804. PMLR, 2018.

\bibitem[Lei et~al.(2018)Lei, G'Sell, Rinaldo, Tibshirani, and
  Wasserman]{lei2018distribution}
Jing Lei, Max G'Sell, Alessandro Rinaldo, Ryan~J Tibshirani, and Larry
  Wasserman.
\newblock Distribution-free predictive inference for regression.
\newblock \emph{Journal of the American Statistical Association}, 113\penalty0
  (523):\penalty0 1094--1111, 2018.

\bibitem[Gneiting and Raftery(2007)]{gneiting2007strictly}
Tilmann Gneiting and Adrian~E Raftery.
\newblock Strictly proper scoring rules, prediction, and estimation.
\newblock \emph{Journal of the American Statistical Association}, 102\penalty0
  (477):\penalty0 359--378, 2007.

\bibitem[Scheuerer and Hamill(2015)]{scheuerer2015variogram}
Michael Scheuerer and Thomas~M Hamill.
\newblock Variogram-based proper scoring rules for probabilistic forecasts of
  multivariate quantities.
\newblock \emph{Monthly Weather Review}, 143\penalty0 (4):\penalty0 1321--1334,
  2015.

\bibitem[Landsgesell et~al.(2026)Landsgesell, Knoll, and
  Wenzel]{landsgesell2026distributional}
Jonas Landsgesell, Pascal Knoll, and Tizian Wenzel.
\newblock Distributional regression with {Tabular} foundation models:
  Evaluating probabilistic predictions via proper scoring rules.
\newblock \emph{arXiv preprint arXiv:2603.08206}, 2026.

\bibitem[Nogueira et~al.(2026)Nogueira, Rebello, Leblebici, and
  Nascimento]{nogueira2026multipuffin}
Idelfonso B.~R. Nogueira, Carine~M. Rebello, Mumin~Enis Leblebici, and Erick
  Giovani~Sperandio Nascimento.
\newblock {MultiPUFFIN}: A multimodal domain-constrained foundation model for
  molecular property prediction of small molecules.
\newblock \emph{arXiv preprint arXiv:2603.00857}, 2026.

\bibitem[Tashiro et~al.(2021)Tashiro, Song, Song, and Ermon]{tashiro2021csdi}
Yusuke Tashiro, Jiaming Song, Yang Song, and Stefano Ermon.
\newblock {CSDI}: Conditional score-based diffusion models for probabilistic time series imputation.
\newblock In \emph{Advances in Neural Information Processing Systems}, volume~34, 2021.
\newblock URL \url{https://arxiv.org/abs/2107.03502}.

\bibitem[Hsu et~al.(2022)Hsu, Verkuil, Liu, Lin, Hie, Sercu, Lerer, and
  Rives]{hsu2022learning}
Chloe Hsu, Robert Verkuil, Jason Liu, Zeming Lin, Brian Hie, Tom Sercu, Adam
  Lerer, and Alexander Rives.
\newblock Learning inverse folding from millions of predicted structures.
\newblock In \emph{International conference on machine learning}, pages
  8946--8970. PMLR, 2022.

\bibitem[Yang et~al.(2023)Yang, Zanichelli, and Yeh]{yang2023masked}
Kevin~K Yang, Niccol{\`o} Zanichelli, and Hugh Yeh.
\newblock Masked inverse folding with sequence transfer for protein
  representation learning.
\newblock \emph{Protein Engineering, Design and Selection}, 36:\penalty0
  gzad015, 2023.

\bibitem[Hayes et~al.(2025)Hayes, Rao, Akin, Sofroniew, Oktay, Lin, Verkuil,
  Tran, Deaton, Wiggert, et~al.]{hayes2025simulating}
Thomas Hayes, Roshan Rao, Halil Akin, Nicholas~J Sofroniew, Deniz Oktay, Zeming
  Lin, Robert Verkuil, Vincent~Q Tran, Jonathan Deaton, Marius Wiggert, et~al.
\newblock Simulating 500 million years of evolution with a language model.
\newblock \emph{Science}, 387\penalty0 (6736):\penalty0 850--858, 2025.

\bibitem[Su et~al.(2024)Su, Han, Zhou, Shan, Zhou, and Yuan]{su2024saprot}
Jin Su, Chenchen Han, Yuyang Zhou, Junjie Shan, Xibin Zhou, and Fajie Yuan.
\newblock {Saprot}: Protein language modeling with structure-aware vocabulary.
\newblock In \emph{International Conference on Learning Representations},
  volume 2024, pages 6987--7009, 2024.

\bibitem[Pei et~al.(2023)Pei, Zhang, Zhu, Wu, Gao, Wu, Xia, and
  Yan]{pei2023biot5}
Qizhi Pei, Wei Zhang, Jinhua Zhu, Kehan Wu, Kaiyuan Gao, Lijun Wu, Yingce Xia,
  and Rui Yan.
\newblock {Biot5}: Enriching cross-modal integration in biology with chemical
  knowledge and natural language associations.
\newblock In \emph{Proceedings of the 2023 Conference on Empirical Methods in
  Natural Language Processing}, pages 1102--1123, 2023.

\bibitem[Fallahpour et~al.(2026)Fallahpour, Magnuson, Gupta, Ma, Naimer, Shah,
  Duan, Ibrahim, Goodarzi, Maddison, et~al.]{fallahpour2026bioreason}
Adibvafa Fallahpour, Andrew Magnuson, Purav Gupta, Shihao Ma, Jack Naimer,
  Arnav Shah, Haonan Duan, Omar Ibrahim, Hani Goodarzi, Chris Maddison, et~al.
\newblock {Bioreason}: Incentivizing multimodal biological reasoning within a
  dna-llm model.
\newblock \emph{Advances in Neural Information Processing Systems},
  38:\penalty0 111644--111671, 2026.

\bibitem[Riesselman et~al.(2018)Riesselman, Ingraham, and
  Marks]{riesselman2018deep}
Adam~J Riesselman, John~B Ingraham, and Debora~S Marks.
\newblock Deep generative models of genetic variation capture the effects of
  mutations.
\newblock \emph{Nature methods}, 15\penalty0 (10):\penalty0 816--822, 2018.
\newblock \doi{10.1038/s41592-018-0138-4}.
\newblock URL \url{https://doi.org/10.1038/s41592-018-0138-4}.

\bibitem[Hie et~al.(2022)Hie, Yang, and Kim]{hie2022evolutionary}
Brian~L. Hie, Kevin~K. Yang, and Peter~S. Kim.
\newblock Evolutionary velocity with protein language models predicts
  evolutionary dynamics of diverse proteins.
\newblock \emph{Cell Systems}, 13\penalty0 (4):\penalty0 274--285, 2022.
\newblock \doi{10.1016/j.cels.2022.01.003}.
\newblock URL \url{https://doi.org/10.1016/j.cels.2022.01.003}.

\bibitem[Y{\"u}ksel et~al.(2023)Y{\"u}ksel, Ulusoy, {\"U}nl{\"u}, and
  Do{\u{g}}an]{yuksel2023selformer}
Atakan Y{\"u}ksel, Erva Ulusoy, Atabey {\"U}nl{\"u}, and Tunca Do{\u{g}}an.
\newblock {SELFormer}: Molecular representation learning via {SELFIES} language
  models.
\newblock \emph{Machine Learning: Science and Technology}, 4\penalty0
  (2):\penalty0 025035, 2023.
\newblock \doi{10.1088/2632-2153/acdb30}.
\newblock URL \url{https://doi.org/10.1088/2632-2153/acdb30}.

\bibitem[Cui et~al.(2025)Cui, Tejada-Lapuerta, Brbi{\'c}, Saez-Rodriguez,
  Cristea, Goodarzi, Lotfollahi, Theis, and Wang]{cui2025towards}
Haotian Cui, Alejandro Tejada-Lapuerta, Maria Brbi{\'c}, Julio Saez-Rodriguez,
  Simona Cristea, Hani Goodarzi, Mohammad Lotfollahi, Fabian~J Theis, and
  Bo~Wang.
\newblock Towards multimodal foundation models in molecular cell biology.
\newblock \emph{Nature}, 640\penalty0 (8059):\penalty0 623--633, 2025.
\newblock \doi{10.1038/s41586-025-08710-y}.
\newblock URL \url{https://doi.org/10.1038/s41586-025-08710-y}.

\bibitem[Gong et~al.(2024)Gong, Klivans, Loy, Chen, Liu, and
  Diaz]{gong2024evorank}
Chengyue Gong, Adam~R. Klivans, James~Madigan Loy, Tianlong Chen, Qiang Liu,
  and Daniel~Jesus Diaz.
\newblock Evolution-inspired loss functions for protein representation
  learning.
\newblock In \emph{Proceedings of the 41st International Conference on Machine
  Learning ({ICML})}, volume 235 of \emph{Proceedings of Machine Learning
  Research}, pages 15893--15906. PMLR, 2024.
\newblock URL \url{https://proceedings.mlr.press/v235/gong24e.html}.

\bibitem[McCarter et~al.(2026)McCarter, Bhattacharya, Ober, and
  Elliott]{mccarter2026make}
Calvin McCarter, Nick Bhattacharya, Sebastian~W Ober, and Hunter Elliott.
\newblock How to make the most of your masked language model for protein
  engineering.
\newblock \emph{arXiv preprint arXiv:2603.10302}, 2026.
\newblock \doi{10.48550/arXiv.2603.10302}.
\newblock URL \url{https://arxiv.org/abs/2603.10302}.

\bibitem[Zha et~al.(2023)Zha, Cao, Son, Yang, and Katabi]{zha2023rank}
Kaiwen Zha, Peng Cao, Jeany Son, Yuzhe Yang, and Dina Katabi.
\newblock Rank-N-Contrast: Learning continuous representations for regression.
\newblock In \emph{Advances in Neural Information Processing Systems}, volume~36, 2023.
\newblock \doi{10.52202/075280-0786}.

\bibitem[Bonilla et~al.(2007)Bonilla, Chai, and Williams]{bonilla2007multi}
Edwin~V Bonilla, Kian~M Chai, and Christopher Williams.
\newblock Multi-task {G}aussian process prediction.
\newblock In \emph{Advances in Neural Information Processing Systems},
  volume~20, pages 153--160, 2007.

\bibitem[Gardner et~al.(2014)Gardner, Kusner, Xu, Weinberger, and
  Cunningham]{gardner2014bayesian}
Jacob~R Gardner, Matt~J Kusner, Zhixiang~Eddie Xu, Kilian~Q Weinberger, and
  John~P Cunningham.
\newblock {B}ayesian optimization with inequality constraints.
\newblock In \emph{International Conference on Machine Learning}, pages
  937--945. PMLR, 2014.

\bibitem[Ament et~al.(2023)Ament, Daulton, Eriksson, Balandat, and
  Bakshy]{ament2023unexpected}
Sebastian Ament, Samuel Daulton, David Eriksson, Maximilian Balandat, and Eytan
  Bakshy.
\newblock Unexpected improvements to expected improvement for {B}ayesian
  optimization.
\newblock In \emph{Advances in Neural Information Processing Systems},
  volume~36, 2023.

\bibitem[Gordon et~al.(2025)Gordon, Gervasio, Souders, and
  Deane]{gordon2025therapeutic}
Gemma~L Gordon, Jo{\~a}o Gervasio, Colby Souders, and Charlotte~M Deane.
\newblock The therapeutic nanobody profiler: characterising and predicting
  nanobody developability to improve therapeutic design.
\newblock \emph{bioRxiv}, pages 2025--08, 2025.

\bibitem[Erasmus et~al.(2024)Erasmus, Spector, Ferrara, DiNiro, Pohl,
  Perea-Schmittle, Wang, Tessier, Richardson, Turner,
  et~al.]{erasmus2024aintibody}
M~Frank Erasmus, Laura Spector, Fortunato Ferrara, Roberto DiNiro, Thomas~J
  Pohl, Katheryn Perea-Schmittle, Wei Wang, Peter~M Tessier, Crystal
  Richardson, Laure Turner, et~al.
\newblock {AIntibody}: an experimentally validated in silico antibody discovery
  design challenge.
\newblock \emph{Nature Biotechnology}, pages 1--6, 2024.

\bibitem[Mason et~al.(2021)Mason, Friedensohn, Weber, Jordi, Wagner, Meng,
  Ehling, Bonati, Dahinden, Gainza, et~al.]{mason2021optimization}
Derek~M Mason, Simon Friedensohn, C{\'e}dric~R Weber, Christian Jordi, Bastian
  Wagner, Simon~M Meng, Roy~A Ehling, Lucia Bonati, Jan Dahinden, Pablo Gainza,
  et~al.
\newblock Optimization of therapeutic antibodies by predicting antigen
  specificity from antibody sequence via deep learning.
\newblock \emph{Nature biomedical engineering}, 5\penalty0 (6):\penalty0
  600--612, 2021.

\bibitem[Yu et~al.(2023)Yu, Orr, Chan, James, Penfold, Kim, Inzhelevskaya,
  Mockridge, Cox, Essex, et~al.]{yu2023reducing}
Xiaojie Yu, Christian~M Orr, HT~Claude Chan, Sonya James, Christine~A Penfold,
  Jinny Kim, Tatyana Inzhelevskaya, C~Ian Mockridge, Kerry~L Cox, Jonathan~W
  Essex, et~al.
\newblock Reducing affinity as a strategy to boost immunomodulatory antibody
  agonism.
\newblock \emph{Nature}, 614\penalty0 (7948):\penalty0 539--547, 2023.

\bibitem[Raybould et~al.(2019)Raybould, Marks, Krawczyk, Taddese, Nowak, Lewis,
  Bujotzek, Shi, and Deane]{raybould2019five}
Matthew~IJ Raybould, Claire Marks, Konrad Krawczyk, Bruck Taddese, Jaroslaw
  Nowak, Alan~P Lewis, Alexander Bujotzek, Jiye Shi, and Charlotte~M Deane.
\newblock Five computational developability guidelines for therapeutic antibody
  profiling.
\newblock \emph{Proceedings of the National Academy of Sciences}, 116\penalty0
  (10):\penalty0 4025--4030, 2019.

\bibitem[Sormanni et~al.(2015)Sormanni, Aprile, and
  Vendruscolo]{sormanni2015camsol}
Pietro Sormanni, Francesco~A Aprile, and Michele Vendruscolo.
\newblock The {CamSol} method of rational design of protein mutants with
  enhanced solubility.
\newblock \emph{Journal of molecular biology}, 427\penalty0 (2):\penalty0
  478--490, 2015.

\bibitem[Hebditch et~al.(2017)Hebditch, Carballo-Amador, Charonis, Curtis, and
  Warwicker]{hebditch2017protein}
Max Hebditch, M~Alejandro Carballo-Amador, Spyros Charonis, Robin Curtis, and
  Jim Warwicker.
\newblock {Protein--Sol}: a web tool for predicting protein solubility from
  sequence.
\newblock \emph{Bioinformatics}, 33\penalty0 (19):\penalty0 3098--3100, 2017.

\bibitem[Marks et~al.(2021)Marks, Hummer, Chin, and
  Deane]{marks2021humanization}
Claire Marks, Alissa~M Hummer, Mark Chin, and Charlotte~M Deane.
\newblock Humanization of antibodies using a machine learning approach on
  large-scale repertoire data.
\newblock \emph{Bioinformatics}, 37\penalty0 (22):\penalty0 4041--4047, 2021.

\bibitem[Prihoda et~al.(2022)Prihoda, Maamary, Waight, Juan, Fayadat-Dilman,
  Svozil, and Bitton]{prihoda2022biophi}
David Prihoda, Jad Maamary, Andrew Waight, Veronica Juan, Laurence
  Fayadat-Dilman, Daniel Svozil, and Danny~A Bitton.
\newblock {BioPhi}: A platform for antibody design, humanization, and humanness
  evaluation based on natural antibody repertoires and deep learning.
\newblock \emph{MAbs}, 14\penalty0 (1):\penalty0 2020203, 2022.

\bibitem[Ramon et~al.(2024)Ramon, Ali, Atkinson, Saturnino, Didi, Visentin,
  Ricagno, Xu, Greenig, and Sormanni]{ramon2024assessing}
Aubin Ramon, Montader Ali, Misha Atkinson, Alessio Saturnino, Kieran Didi,
  Cristina Visentin, Stefano Ricagno, Xing Xu, Matthew Greenig, and Pietro
  Sormanni.
\newblock Assessing antibody and nanobody nativeness for hit selection and
  humanization with {AbNatiV}.
\newblock \emph{Nature Machine Intelligence}, 6\penalty0 (1):\penalty0 74--91,
  2024.

\bibitem[Reynisson et~al.(2020)Reynisson, Alvarez, Paul, Peters, and
  Nielsen]{reynisson2020netmhcpan}
Birkir Reynisson, Bruno Alvarez, Sinu Paul, Bjoern Peters, and Morten Nielsen.
\newblock {NetMHCpan-4.1} and {NetMHCIIpan-4.0}: improved predictions of {MHC}
  antigen presentation by concurrent motif deconvolution and integration of
  {MS} {MHC} eluted ligand data.
\newblock \emph{Nucleic acids research}, 48\penalty0 (W1):\penalty0 W449--W454,
  2020.

\bibitem[Chen et~al.(2019)Chen, Khodadoust, Olsson, Wagar, Fast, Liu,
  Muftuoglu, Sworder, Diehn, Levy, et~al.]{chen2019predicting}
Binbin Chen, Michael~S Khodadoust, Niclas Olsson, Lisa~E Wagar, Ethan Fast,
  Chih~Long Liu, Yagmur Muftuoglu, Brian~J Sworder, Maximilian Diehn, Ronald
  Levy, et~al.
\newblock Predicting {HLA} class ii antigen presentation through integrated
  deep learning.
\newblock \emph{Nature biotechnology}, 37\penalty0 (11):\penalty0 1332--1343,
  2019.

\bibitem[Wang et~al.(2024)Wang, Hao, He, and Fan]{wang2024abimmpred}
Hong Wang, Xiaohu Hao, Yuzhuo He, and Long Fan.
\newblock {AbImmPred}: An immunogenicity prediction method for therapeutic
  antibodies using {AntiBERTy}-based sequence features.
\newblock \emph{Plos one}, 19\penalty0 (2):\penalty0 e0296737, 2024.

\bibitem[O’Brien et~al.(2024)O’Brien, Salm, Morton, Szukszto, O’Farrell,
  Boulton, King, Bola, Becker, Craig, et~al.]{o2024modular}
Hugh O’Brien, Max Salm, Laura~T Morton, Maciej Szukszto, Felix O’Farrell,
  Charlotte Boulton, Laurence King, Supreet~Kaur Bola, Pablo~D Becker, Andrew
  Craig, et~al.
\newblock A modular protein language modelling approach to immunogenicity
  prediction.
\newblock \emph{PLOS Computational Biology}, 20\penalty0 (11):\penalty0
  e1012511, 2024.

\bibitem[Huang et~al.(2025)Huang, He, and Liu]{huang2025antibody}
Qiaojing Huang, Yi~He, and Kai Liu.
\newblock Antibody immunogenicity prediction and optimization with {ImmunoSeq}.
\newblock \emph{bioRxiv}, pages 2025--08, 2025.

\bibitem[Burbach and Briney(2024)]{burbach2024improving}
Sarah~M Burbach and Bryan Briney.
\newblock Improving antibody language models with native pairing.
\newblock \emph{Patterns}, 5\penalty0 (5), 2024.

\bibitem[Guo et~al.(2025)Guo, Dunn-Walters, Fraternali, and
  Ng]{guo2025immunomatch}
Dongjun Guo, Deborah~K Dunn-Walters, Franca Fraternali, and Joseph~CF Ng.
\newblock {ImmunoMatch} learns and predicts cognate pairing of heavy and light
  immunoglobulin chains.
\newblock \emph{bioRxiv}, pages 2025--02, 2025.

\bibitem[Parkinson et~al.(2025)Parkinson, Hard, Ko, and
  Wang]{parkinson2025resp2}
Jonathan Parkinson, Ryan Hard, Young~Su Ko, and Wei Wang.
\newblock {RESP2}: An uncertainty aware multi-target multi-property
  optimization {AI} pipeline for antibody discovery.
\newblock \emph{bioRxiv}, pages 2024--07, 2025.

\bibitem[Tobin(1958)]{tobin1958estimation}
James Tobin.
\newblock Estimation of relationships for limited dependent variables.
\newblock \emph{Econometrica}, 26\penalty0 (1):\penalty0 24--36, 1958.

\bibitem[Kumaraswamy(1980)]{kumaraswamy1980generalized}
Poondi Kumaraswamy.
\newblock A generalized probability density function for double-bounded random
  processes.
\newblock \emph{Journal of Hydrology}, 46\penalty0 (1-2):\penalty0 79--88,
  1980.

\bibitem[Risso et~al.(2018)Risso, Perraudeau, Gribkova, Dudoit, and
  Vert]{risso2018general}
Davide Risso, Fanny Perraudeau, Svetlana Gribkova, Sandrine Dudoit, and
  Jean-Philippe Vert.
\newblock A general and flexible method for signal extraction from single-cell
  {RNA-seq} data.
\newblock \emph{Nature Communications}, 9\penalty0 (1):\penalty0 284, 2018.

\bibitem[Kotz and Nadarajah(2004)]{kotz2004multivariate}
Samuel Kotz and Saralees Nadarajah.
\newblock \emph{Multivariate {$t$} Distributions and Their Applications}.
\newblock Cambridge University Press, 2004.

\bibitem[Shah et~al.(2014)Shah, Wilson, and Ghahramani]{shah2014student}
Amar Shah, Andrew~G Wilson, and Zoubin Ghahramani.
\newblock {Student-$t$} processes as alternatives to {Gaussian} processes.
\newblock In \emph{Artificial Intelligence and Statistics}, pages 877--885.
  PMLR, 2014.

\bibitem[Thompson(1933)]{thompson1933sampling}
William~R. Thompson.
\newblock On the likelihood that one unknown probability exceeds another in view of the evidence of two samples.
\newblock \emph{Biometrika}, 25(3--4):285--294, 1933.
\newblock \doi{10.1093/biomet/25.3-4.285}.

\bibitem[Landrum et~al.(2026)]{rdkit}
Greg Landrum et~al..
\newblock {RDKit}: Open-source cheminformatics.
\newblock Software.
\newblock URL \url{https://www.rdkit.org/}.

\bibitem[Wang et~al.(2020)]{wang2020etkdg}
Shuzhe Wang, Jonas Witek, Gregory~A. Landrum, and Sereina~Riniker.
\newblock Improving conformer generation for small rings and macrocycles based on distance geometry and experimental torsional-angle preferences.
\newblock \emph{Journal of Chemical Information and Modeling}, 60(4):2044--2058, 2020.
\newblock \doi{10.1021/acs.jcim.0c00025}.

\bibitem[Halgren(1996)]{halgren1996mmff}
Thomas~A. Halgren.
\newblock Merck molecular force field. I. Basis, form, scope, parameterization, and performance of {MMFF94}.
\newblock \emph{Journal of Computational Chemistry}, 17(5--6):490--519, 1996.

\bibitem[Sterling and Irwin(2015)]{sterling2015zinc}
Teague Sterling and John~J. Irwin.
\newblock {ZINC 15} -- Ligand discovery for everyone.
\newblock \emph{Journal of Chemical Information and Modeling}, 55(11):2324--2337, 2015.
\newblock \doi{10.1021/acs.jcim.5b00559}.

\bibitem[Krenn et~al.(2020)]{krenn2020selfies}
Mario Krenn, Florian H"ase, AkshatKumar Nigam, Pascal Friederich, and Al'an Aspuru-Guzik.
\newblock Self-referencing embedded strings ({SELFIES}): A 100\% robust molecular string representation.
\newblock \emph{Machine Learning: Science and Technology}, 1:045024, 2020.
\newblock \doi{10.1088/2632-2153/aba947}.

\bibitem[Bashour et~al.(2024)Bashour, Smorodina, Pariset, Zhong, Akbar,
  Chernigovskaya, L{\^e}~Qu{\`y}, Snapkow, Rawat, Krawczyk,
  et~al.]{bashour2024biophysical}
Habib Bashour, Eva Smorodina, Matteo Pariset, Jahn Zhong, Rahmad Akbar, Maria
  Chernigovskaya, Khang L{\^e}~Qu{\`y}, Igor Snapkow, Puneet Rawat, Konrad
  Krawczyk, et~al.
\newblock Biophysical cartography of the native and human-engineered antibody
  landscapes quantifies the plasticity of antibody developability.
\newblock \emph{Communications Biology}, 7\penalty0 (1):\penalty0 922, 2024.

\bibitem[Jain et~al.(2017)Jain, Sun, Durand, Hall, Houston, Nett, Sharkey,
  Bobrowicz, Caffry, Yu, et~al.]{jain2017biophysical}
Tushar Jain, Tingwan Sun, St{\'e}phanie Durand, Amy Hall, Nga~Rewa Houston,
  Juergen~H Nett, Beth Sharkey, Beata Bobrowicz, Isabelle Caffry, Yao Yu,
  et~al.
\newblock Biophysical properties of the clinical-stage antibody landscape.
\newblock \emph{Proceedings of the National Academy of Sciences}, 114\penalty0
  (5):\penalty0 944--949, 2017.

\bibitem[Chungyoun and Gray(2025)]{chungyoun2025flab2}
Michael Chungyoun and Jeffrey Gray.
\newblock Fitness Landscape for Antibodies 2: Benchmarking reveals that protein AI models cannot yet consistently predict developability properties.
\newblock \emph{bioRxiv}, 2025.
\newblock \doi{10.64898/2025.12.27.696706}.

\bibitem[Gray Lab(2026)]{flabrepository}
Gray Lab, Johns Hopkins University.
\newblock Fitness Landscape for Antibodies (FLAb): dataset repository and data catalogue.
\newblock Online data resource.
\newblock URL \url{https://github.com/Graylab/FLAb/tree/main/data}.

\bibitem[Alberts et~al.(2024)Alberts, Schilter, Zipoli, Hartrampf, and
  Laino]{alberts2024unraveling}
Marvin Alberts, Oliver Schilter, Federico Zipoli, Nina Hartrampf, and Teodoro
  Laino.
\newblock Unraveling molecular structure: A multimodal spectroscopic dataset
  for chemistry.
\newblock In \emph{Advances in Neural Information Processing Systems (Datasets
  and Benchmarks Track)}, volume~37, 2024.

\bibitem[Koddenbrock et~al.(2026)Koddenbrock, Lange, Legner, J{\"a}ger,
  K{\"o}gler, Cruz~Bournazou, Neubauer, Biessmann, and
  Rodner]{koddenbrock2026ramanbench}
Mario Koddenbrock, Christoph Lange, Robin Legner, Martin J{\"a}ger, Martin
  K{\"o}gler, Mariano~N. Cruz~Bournazou, Peter Neubauer, Felix Biessmann, and
  Erik Rodner.
\newblock {RamanBench}: A large-scale benchmark for machine learning on {R}aman
  spectroscopy.
\newblock In \emph{Advances in Neural Information Processing Systems}, 2026.

\bibitem[Wilhelmlab(2025{\natexlab{a}})]{wilhelmlab2025prosit2020}
Wilhelmlab, Technical University of Munich.
\newblock {Prosit-2020-HCD}.
\newblock Hugging Face dataset release, 2025.
\newblock URL \url{https://huggingface.co/datasets/Wilhelmlab/Prosit-2020-HCD}.

\bibitem[Wilhelmlab(2025{\natexlab{b}})]{wilhelmlab2025prositlac}
Wilhelmlab, Technical University of Munich.
\newblock {Prosit-2025-lac-ms2}.
\newblock Hugging Face dataset release, 2025.
\newblock URL \url{https://huggingface.co/datasets/Wilhelmlab/Prosit-2025-lac-ms2}.

\bibitem[Zolg et~al.(2017)]{zolg2017proteometools}
Daniel~P. Zolg, Mathias Wilhelm, Karsten Schnatbaum, Johannes Zerweck, Tobias Knaute, Bernard Delanghe, et~al..
\newblock Building {ProteomeTools} based on a complete synthetic human proteome.
\newblock \emph{Nature Methods}, 14:259--262, 2017.
\newblock \doi{10.1038/nmeth.4153}.

\bibitem[Gessulat et~al.(2019)Gessulat, Schmidt, Zolg, Samaras, Schnatbaum,
  Zerweck, Knaute, Rechenberger, Delanghe, Huhmer, et~al.]{gessulat2019prosit}
Siegfried Gessulat, Tobias Schmidt, Daniel~Paul Zolg, Patroklos Samaras,
  Karsten Schnatbaum, Johannes Zerweck, Tobias Knaute, Jana Rechenberger,
  Bernard Delanghe, Andreas Huhmer, et~al.
\newblock {Prosit}: proteome-wide prediction of peptide tandem mass spectra by
  deep learning.
\newblock \emph{Nature Methods}, 16\penalty0 (6):\penalty0 509--518, 2019.

\bibitem[Li et~al.(2023)Li, Dou, Da~Veiga~Leprevost, Geffen, Calinawan, Aguet,
  Akiyama, Anand, Birger, Cao, et~al.]{li2023proteogenomic}
Yize Li, Yongchao Dou, Felipe Da~Veiga~Leprevost, Yifat Geffen, Anna~P.
  Calinawan, Fran{\c{c}}ois Aguet, Yo~Akiyama, Shankara Anand, Chet Birger,
  Song Cao, et~al.
\newblock Proteogenomic data and resources for pan-cancer analysis.
\newblock \emph{Cancer Cell}, 41\penalty0 (8):\penalty0 1397--1406, 2023.

\bibitem[Ji et~al.(2021)Ji, Zhou, Liu, and Davuluri]{ji2021dnabert}
Yanrong Ji, Zhihan Zhou, Han Liu, and Ramana~V. Davuluri.
\newblock {DNABERT}: pre-trained Bidirectional Encoder Representations from Transformers model for DNA-language in genome.
\newblock \emph{Bioinformatics}, 37(15):2112--2120, 2021.
\newblock \doi{10.1093/bioinformatics/btab083}.

\bibitem[Bemis and Murcko(1996)]{bemis1996scaffolds}
Guy~W. Bemis and Mark~A. Murcko.
\newblock The properties of known drugs. 1. Molecular frameworks.
\newblock \emph{Journal of Medicinal Chemistry}, 39(15):2887--2893, 1996.
\newblock \doi{10.1021/jm9602928}.

\bibitem[Lefranc et~al.(2003)]{lefranc2003imgt}
Marie-Paule Lefranc, Chantal Pommi\`e, Manuel Ruiz, Val'erie Giudicelli, et~al..
\newblock {IMGT} unique numbering for immunoglobulin and T cell receptor variable domains and Ig superfamily V-like domains.
\newblock \emph{Developmental and Comparative Immunology}, 27(1):55--77, 2003.
\newblock \doi{10.1016/S0145-305X(02)00039-3}.

\bibitem[Loshchilov and Hutter(2019)]{loshchilov2019adamw}
Ilya Loshchilov and Frank Hutter.
\newblock Decoupled weight decay regularization.
\newblock In \emph{International Conference on Learning Representations}, 2019.
\newblock URL \url{https://arxiv.org/abs/1711.05101}.

\bibitem[Liu et~al.(2019)]{liu2019foscttm}
Jie Liu, Yuanhao Huang, Ritambhara Singh, Jean-Philippe Vert, and William~S. Noble.
\newblock Jointly embedding multiple single-cell omics measurements.
\newblock In \emph{19th International Workshop on Algorithms in Bioinformatics (WABI 2019)}, LIPIcs 143, pages 10:1--10:13, 2019.
\newblock \doi{10.4230/LIPIcs.WABI.2019.10}.

\bibitem[Yeo and Johnson(2000)]{yeo2000transform}
In-Kwon Yeo and Richard~A. Johnson.
\newblock A new family of power transformations to improve normality or symmetry.
\newblock \emph{Biometrika}, 87(4):954--959, 2000.
\newblock \doi{10.1093/biomet/87.4.954}.

\bibitem[Sorkun et~al.(2019)Sorkun, Khetan, and Er]{sorkun2019aqsoldb}
Murat~C. Sorkun, Abhishek Khetan, and Suleyman Er.
\newblock {AqSolDB}, a curated reference set of aqueous solubility and 2D descriptors for a diverse set of compounds.
\newblock \emph{Scientific Data}, 6:143, 2019.
\newblock \doi{10.1038/s41597-019-0151-1}.

\bibitem[Liu et~al.(2019)]{liu2019roberta}
Yinhan Liu, Myle Ott, Naman Goyal, Jingfei Du, Mandar Joshi, Danqi Chen, Omer Levy, Mike Lewis, Luke Zettlemoyer, and Veselin Stoyanov.
\newblock {RoBERTa}: A robustly optimized {BERT} pretraining approach.
\newblock \emph{arXiv preprint arXiv:1907.11692}, 2019.
\newblock URL \url{https://arxiv.org/abs/1907.11692}.

\bibitem[Satorras et~al.(2021)Satorras, Hoogeboom, and Welling]{satorras2021egnn}
Victor Garcia Satorras, Emiel Hoogeboom, and Max Welling.
\newblock E(n) equivariant graph neural networks.
\newblock In \emph{International Conference on Machine Learning}, PMLR 139, pages 9323--9332, 2021.
\newblock URL \url{https://proceedings.mlr.press/v139/satorras21a.html}.

\bibitem[UniProt Consortium(2023)]{uniprot2023}
The UniProt Consortium.
\newblock {UniProt}: the Universal Protein Knowledgebase in 2023.
\newblock \emph{Nucleic Acids Research}, 51(D1):D523--D531, 2023.
\newblock \doi{10.1093/nar/gkac1052}.

\bibitem[Xu et~al.(2019)Xu, Hu, Leskovec, and Jegelka]{xu2019gin}
Keyulu Xu, Weihua Hu, Jure Leskovec, and Stefanie Jegelka.
\newblock How powerful are graph neural networks?.
\newblock In \emph{International Conference on Learning Representations}, 2019.
\newblock URL \url{https://arxiv.org/abs/1810.00826}.

\bibitem[Hu et~al.(2022)]{hu2022lora}
Edward~J. Hu, Yelong Shen, Phillip Wallis, Zeyuan Allen-Zhu, Yuanzhi Li, Shean Wang, Lu Wang, and Weizhu Chen.
\newblock {LoRA}: Low-rank adaptation of large language models.
\newblock In \emph{International Conference on Learning Representations}, 2022.
\newblock URL \url{https://arxiv.org/abs/2106.09685}.

\bibitem[Bickerton et~al.(2012)]{bickerton2012qed}
G.~Richard Bickerton, Gaia~V. Paolini, J'er\^ome Besnard, Sorel Muresan, and Andrew~L. Hopkins.
\newblock Quantifying the chemical beauty of drugs.
\newblock \emph{Nature Chemistry}, 4:90--98, 2012.
\newblock \doi{10.1038/nchem.1243}.

\bibitem[Zambrano et~al.(2015)]{zambrano2015aggrescan}
Raul Zambrano, Miquel Jamroz, Andrzej Szczasiuk, Jorge Pujols, Sebastian Kmiecik, and Salvador Ventura.
\newblock {AGGRESCAN3D} ({A3D}): server for prediction of aggregation properties of protein structures.
\newblock \emph{Nucleic Acids Research}, 43(W1):W306--W313, 2015.
\newblock \doi{10.1093/nar/gkv359}.

\bibitem[H\o ie et~al.(2025)]{hoie2025antifold}
Magnus Haraldson H\o ie, Alissa~M. Hummer, Tobias~H. Olsen, Broncio Aguilar-Sanjuan, Morten Nielsen, and Charlotte~M. Deane.
\newblock {AntiFold}: improved structure-based antibody design using inverse folding.
\newblock \emph{Bioinformatics Advances}, 5(1):vbae202, 2025.
\newblock \doi{10.1093/bioadv/vbae202}.

\bibitem[Kalejaye et~al.(2025)]{kalejaye2025deepviscosity}
Lateefat~A. Kalejaye, Jia-Min Chu, I-En Wu, Bismark Amofah, Amber Lee, Mark Hutchinson, et~al..
\newblock Accelerating high-concentration monoclonal antibody development with large-scale viscosity data and ensemble deep learning.
\newblock \emph{mAbs}, 17(1):2483944, 2025.
\newblock \doi{10.1080/19420862.2025.2483944}.

\bibitem[Ruffolo et~al.(2021)Ruffolo, Gray, and Sulam]{ruffolo2021antiberty}
Jeffrey~A. Ruffolo, Jeffrey~J. Gray, and Jeremias Sulam.
\newblock Deciphering antibody affinity maturation with language models and weakly supervised learning.
\newblock \emph{arXiv preprint arXiv:2112.07782}, 2021.
\newblock URL \url{https://arxiv.org/abs/2112.07782}.

\bibitem[Chemical Computing Group(2026)]{ccgmoe}
Chemical Computing Group.
\newblock Molecular Operating Environment ({MOE}).
\newblock Software.
\newblock URL \url{https://www.chemcomp.com/}.

\bibitem[Gal and Ghahramani(2016)]{gal2016dropout}
Yarin Gal and Zoubin Ghahramani.
\newblock Dropout as a Bayesian approximation: Representing model uncertainty in deep learning.
\newblock In \emph{International Conference on Machine Learning}, PMLR 48, pages 1050--1059, 2016.
\newblock URL \url{https://proceedings.mlr.press/v48/gal16.html}.

\bibitem[Nigam et~al.(2021)]{nigam2021stoned}
AkshatKumar Nigam, Robert Pollice, Mario Krenn, Gabriel~D.~P. Gomes, and Al'an Aspuru-Guzik.
\newblock Beyond generative models: superfast traversal, optimization, novelty, exploration and discovery ({STONED}) algorithm for molecules using {SELFIES}.
\newblock \emph{Chemical Science}, 12:7079--7090, 2021.
\newblock \doi{10.1039/D1SC00231G}.

\bibitem[Olivecrona et~al.(2017)]{olivecrona2017reinvent}
Marcus Olivecrona, Thomas Blaschke, Ola Engkvist, and Hongming Chen.
\newblock Molecular de-novo design through deep reinforcement learning.
\newblock \emph{Journal of Cheminformatics}, 9:48, 2017.
\newblock \doi{10.1186/s13321-017-0235-x}.

\bibitem[Koenig et~al.(2017)]{koenig2017landscape}
Patrick Koenig, Chingwei~V. Lee, Benjamin~T. Walters, Vasantharajan Janakiraman, Jeremy Stinson, Thomas~W. Patapoff, and Germaine Fuh.
\newblock Mutational landscape of antibody variable domains reveals a switch modulating the interdomain conformational dynamics and antigen binding.
\newblock \emph{Proceedings of the National Academy of Sciences}, 114(4):E486--E495, 2017.
\newblock \doi{10.1073/pnas.1613231114}.

\bibitem[Makowski et~al.(2022)]{makowski2022cooptimization}
Emily~K. Makowski, Patrick~C. Kinnunen, Jie Huang, Lina Wu, Matthew~D. Smith, Tiexin Wang, Alec~A. Desai, Craig~N. Streu, Yulei Zhang, Jennifer~M. Zupancic, John~S. Schardt, Jennifer~J. Linderman, and Peter~M. Tessier.
\newblock Co-optimization of therapeutic antibody affinity and specificity using machine learning models that generalize to novel mutational space.
\newblock \emph{Nature Communications}, 13:3788, 2022.
\newblock \doi{10.1038/s41467-022-31457-3}.

\bibitem[Hutchinson et~al.(2024)]{hutchinson2024thermostability}
Mark Hutchinson, Jeffrey~A. Ruffolo, Nantaporn Haskins, Michael Iannotti, Giuliana Vozza, Tony Pham, Nurjahan Mehzabeen, Harini Shandilya, Keith Rickert, Rebecca Croasdale-Wood, Melissa Damschroder, Ying Fu, Andrew Dippel, Jeffrey~J. Gray, and Gilad Kaplan.
\newblock Toward enhancement of antibody thermostability and affinity by computational design in the absence of antigen.
\newblock \emph{mAbs}, 16(1):2362775, 2024.
\newblock \doi{10.1080/19420862.2024.2362775}.

\bibitem[Adams et~al.(2016)Adams, Mora, Walczak, and Kinney]{adams2016titeseq}
Rhys~M. Adams, Thierry Mora, Aleksandra~M. Walczak, and Justin~B. Kinney.
\newblock Measuring the sequence-affinity landscape of antibodies with massively parallel titration curves.
\newblock \emph{eLife}, 5:e23156, 2016.
\newblock \doi{10.7554/eLife.23156}.

\bibitem[Hastings(1970)]{hastings1970mh}
W.~K. Hastings.
\newblock Monte Carlo sampling methods using Markov chains and their applications.
\newblock \emph{Biometrika}, 57(1):97--109, 1970.
\newblock \doi{10.1093/biomet/57.1.97}.

\bibitem[Andrieu and Roberts(2009)]{andrieu2009pseudomarginal}
Christophe Andrieu and Gareth~O. Roberts.
\newblock The pseudo-marginal approach for efficient Monte Carlo computations.
\newblock \emph{Annals of Statistics}, 37(2):697--725, 2009.
\newblock \doi{10.1214/07-AOS574}.

\bibitem[Xiong et~al.(2026)Xiong, Gaur, Lukarska, Nisonoff, Oltrogge, Savage,
  and Listgarten]{xiong2026proteinguide}
Junhao Xiong, Ishan Gaur, Maria Lukarska, Hunter Nisonoff, Luke~M Oltrogge,
  David~F Savage, and Jennifer Listgarten.
\newblock Property guidance for protein sequence generative models with
  {ProteinGuide}.
\newblock \emph{Nature Biotechnology}, 2026.

\bibitem[Rafailov et~al.(2023)]{rafailov2023dpo}
Rafael Rafailov, Archit Sharma, Eric Mitchell, Stefano Ermon, Christopher~D. Manning, and Chelsea Finn.
\newblock Direct preference optimization: Your language model is secretly a reward model.
\newblock \emph{Advances in Neural Information Processing Systems}, 36, 2023.
\newblock URL \url{https://arxiv.org/abs/2305.18290}.

\end{thebibliography}
\end{document}